%% file: main.tex
\documentclass[letterpaper]{article}
\usepackage[preprint]{aaai2027}
\usepackage{natbib}
\usepackage{graphicx}
\usepackage{amsmath,amssymb,mathtools}
\usepackage{booktabs}
\usepackage{multirow}
\usepackage{array}
\usepackage{adjustbox}
\usepackage{tabularx}
\usepackage{makecell}
\usepackage{rotating}
\usepackage{algorithm}
\usepackage{algorithmic}
\usepackage[table]{xcolor}
\usepackage{enumitem}
\usepackage{url}
\usepackage{pifont}
\usepackage{tikz}
\usepackage{subcaption}
\definecolor{chartblue}{HTML}{4E79A7}
\definecolor{coordorange}{HTML}{F28E2B}
\definecolor{stategreen}{HTML}{59A14F}
\definecolor{readoutpurple}{HTML}{8F63B9}
\definecolor{thresholdgray}{HTML}{777777}
\usepackage{placeins}
\usepackage{booktabs,adjustbox,multirow,rotating,array,colortbl}
\usepackage{booktabs,longtable,array,colortbl,multirow}

\newcommand{\sycopathpanel}[1]{%
    \includegraphics[width=0.228\textwidth]{#1}%
}

\newcommand{\mathpill}[2]{%
  \tikz[baseline=(char.base)]{
    \node[
      rounded corners=2pt,
      draw=#1!45,
      fill=#1!12,
      inner xsep=2.2pt,
      inner ysep=1.1pt
    ] (char) {\ensuremath{#2}};
  }%
}

\newcommand{\chartterm}[1]{\mathpill{chartblue}{#1}}
\newcommand{\coordterm}[1]{\mathpill{coordorange}{#1}}
\newcommand{\stateterm}[1]{\mathpill{stategreen}{#1}}
\newcommand{\readoutterm}[1]{\mathpill{readoutpurple}{#1}}
\newcommand{\thresholdterm}[1]{\mathpill{thresholdgray}{#1}}
\definecolor{geomblue}{HTML}{4E79A7}
\definecolor{occorange}{HTML}{F28E2B}
\definecolor{gainpurple}{HTML}{8F63B9}

\newcommand*\coloredcircled[2]{%
  \tikz[baseline=(char.base)]{
    \node[
      anchor=text,
      shape=circle,
      fill=#1,
      inner sep=0pt,
      minimum size=1.25em
    ] (char) {\scriptsize\bfseries\textcolor{white}{#2}};
  }%
}

\definecolor{samplegreen}{HTML}{59A14F}
\definecolor{varianceblue}{HTML}{4E79A7}
\definecolor{causalred}{HTML}{E15759}

\newcommand{\providerlogo}[1]{%
    \raisebox{-0.15em}{%
        \includegraphics[
            height=1.05em,
            keepaspectratio
        ]{figures__logos__#1.png}%
    }%
}

\newcommand{\GemmaLogo}{
    \providerlogo{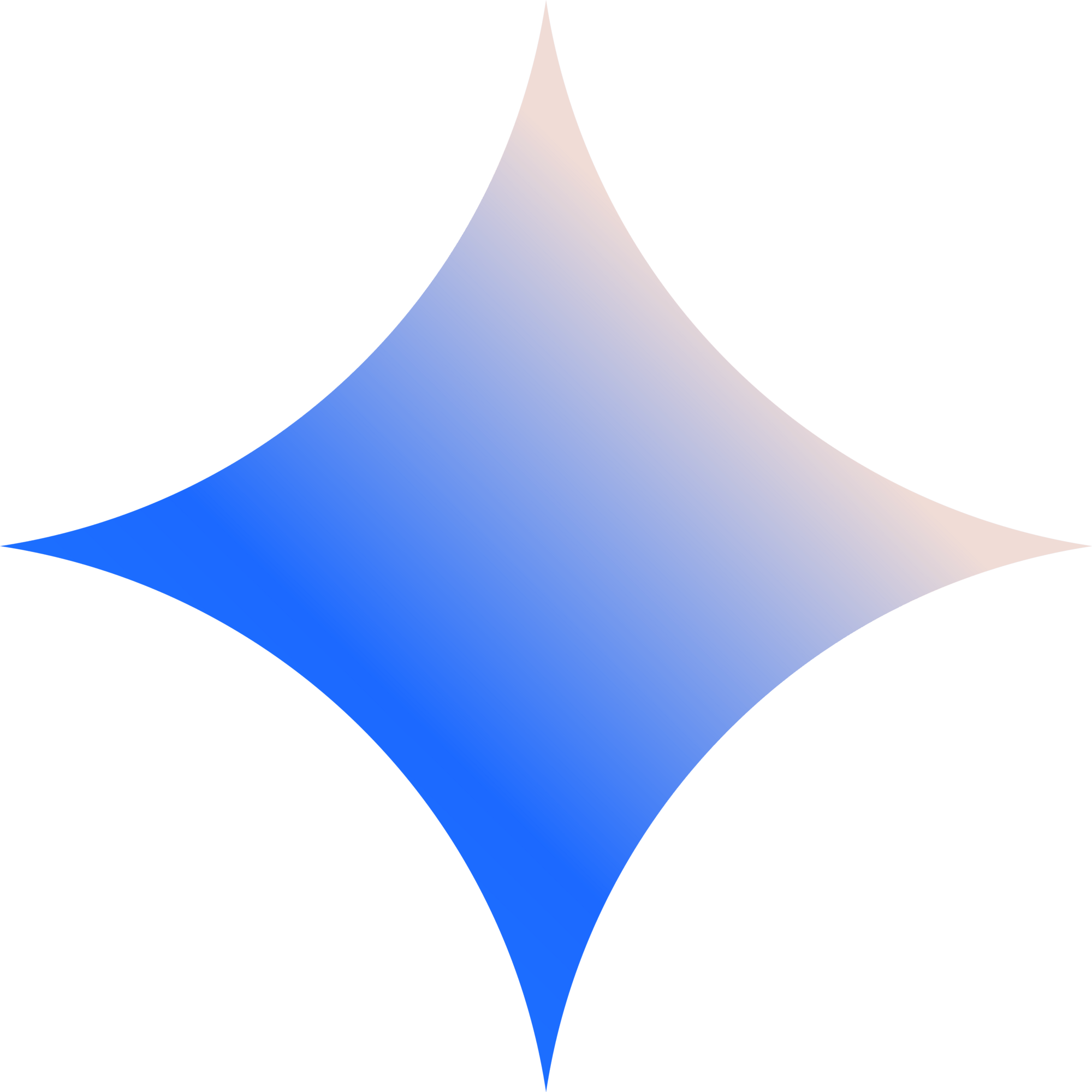}
}

\newcommand{\QwenLogo}{
    \providerlogo{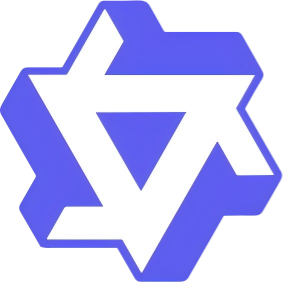}
}

\newcommand{\MistralLogo}{
    \providerlogo{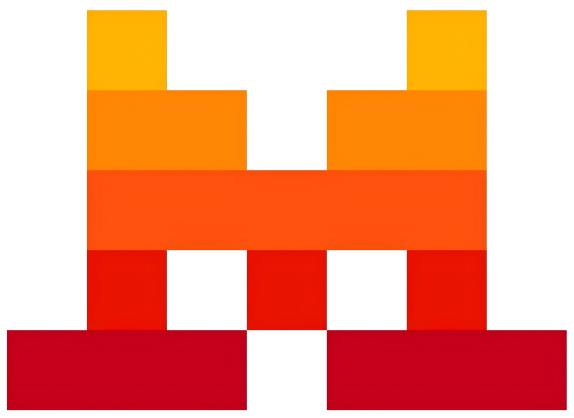}
}

\newcommand{\LlamaLogo}{
    \providerlogo{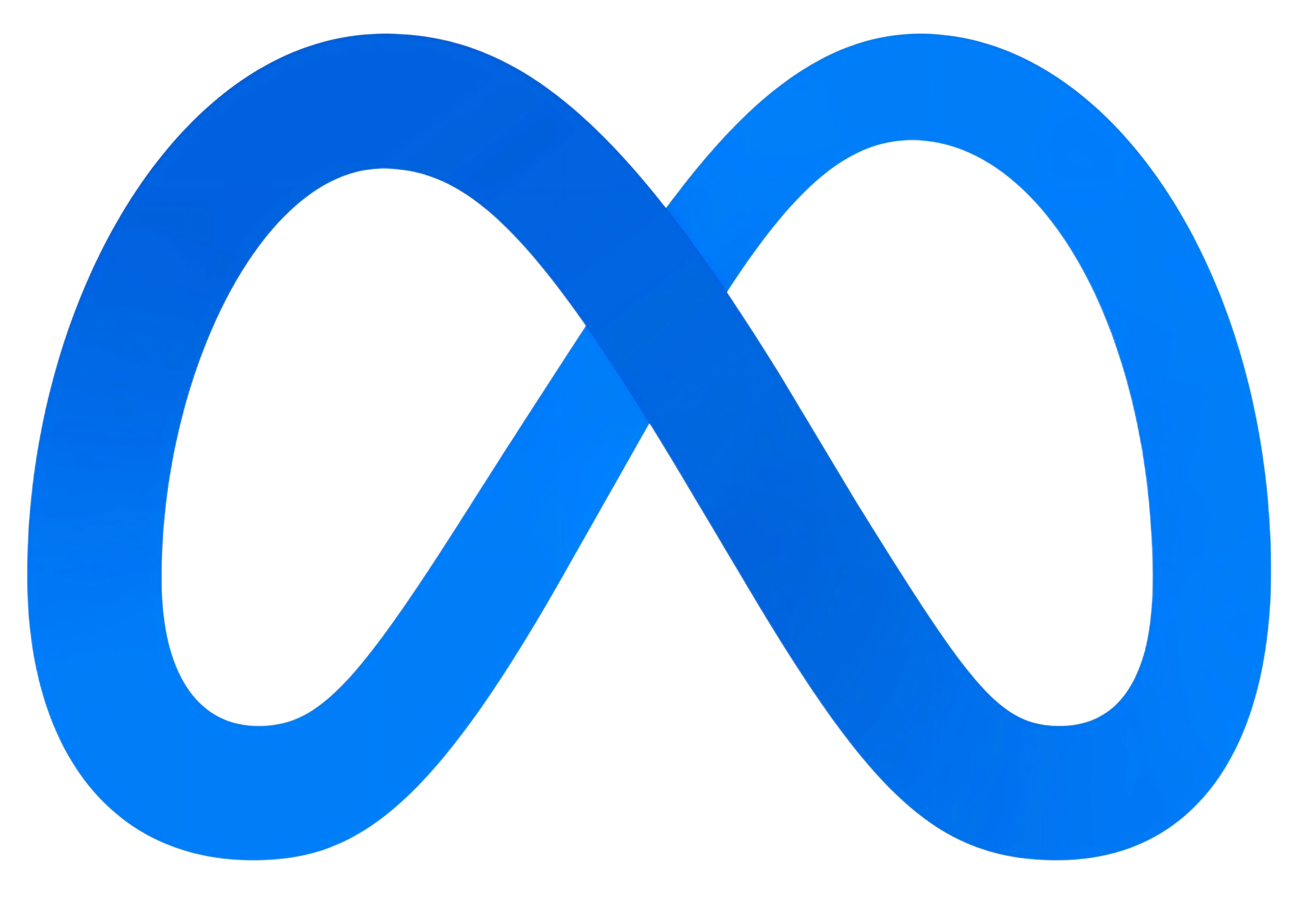}
}

\definecolor{GemmaBand}{HTML}{EAF2FF}
\definecolor{QwenBand}{HTML}{F1ECFF}
\definecolor{MistralBand}{HTML}{FFF0E5}
\definecolor{LlamaBand}{HTML}{EAF7EC}

\newcommand{\atlaspanel}[1]{%
    \includegraphics[
        width=\linewidth,
        keepaspectratio
    ]{#1}%
}

\newcommand{\familyheader}[3]{%
\rowcolor{#1}
\multicolumn{#3}{l}{
#2
}\\[-0.3ex]
}

\usepackage{graphicx}
\usepackage{placeins}

\newcommand{\decompstagepanel}[2]{%
    \begin{minipage}[t]{0.485\textwidth}
        \centering
        \includegraphics[
            width=\linewidth,
            keepaspectratio
        ]{#1}

        {\scriptsize #2}
    \end{minipage}%
}

\newcommand{\metricpill}[2]{%
  \tikz[baseline=(metric.base)]{
    \node[
      rounded corners=2.5pt,
      draw=#1!55!black,
      fill=#1!12,
      text=#1!75!black,
      line width=0.45pt,
      inner xsep=3.2pt,
      inner ysep=1.2pt,
      font=\sffamily\bfseries\scriptsize
    ] (metric) {\strut\ensuremath{#2}};
  }%
}

\newcommand{\ACTtag}{%
  \metricpill{ACTColor}{\mathrm{ACT}}%
}

\newcommand{\NOCtag}{%
  \metricpill{NOCColor}{\mathrm{NOC}}%
}

\usepackage{xcolor}
\usepackage{tikz}
\usetikzlibrary{positioning}

\definecolor{ACTColor}{HTML}{2A9D8F}
\definecolor{NOCColor}{HTML}{3A5BA0}
\usepackage{booktabs}
\usepackage{multirow}
\usepackage{adjustbox}

\newcommand{\geommark}{\coloredcircled{geomblue}{1}}
\newcommand{\occmark}{\coloredcircled{occorange}{2}}
\newcommand{\gainmark}{\coloredcircled{gainpurple}{3}}

\newcommand{\R}{\mathbb{R}}
\newcommand{\E}{\mathbb{E}}

\usepackage{xcolor}

\definecolor{geomQtyBG}{HTML}{E3ECEB}
\definecolor{geomQtyFG}{HTML}{3F6261}

\definecolor{sepQtyBG}{HTML}{EEE7EA}
\definecolor{sepQtyFG}{HTML}{6A505C}

\definecolor{occQtyBG}{HTML}{EBECE2}
\definecolor{occQtyFG}{HTML}{62654B}

\definecolor{gainQtyBG}{HTML}{F1E7E1}
\definecolor{gainQtyFG}{HTML}{795747}

\definecolor{trajQtyBG}{HTML}{E7E8EE}
\definecolor{trajQtyFG}{HTML}{53596E}

\newcommand{\mathqtyblock}[3]{%
    \begingroup
    \setlength{\fboxsep}{1.25pt}%
    \ensuremath{%
        \mathchoice
        {\colorbox{#1}{\textcolor{#2}{$\displaystyle #3$}}}
        {\colorbox{#1}{\textcolor{#2}{$\textstyle #3$}}}
        {\colorbox{#1}{\textcolor{#2}{$\scriptstyle #3$}}}
        {\colorbox{#1}{\textcolor{#2}{$\scriptscriptstyle #3$}}}%
    }%
    \endgroup
}

\DeclareRobustCommand{\geomqty}[1]{%
    \mathqtyblock{geomQtyBG}{geomQtyFG}{#1}}

\DeclareRobustCommand{\sepqty}[1]{%
    \mathqtyblock{sepQtyBG}{sepQtyFG}{#1}}

\DeclareRobustCommand{\occqty}[1]{%
    \mathqtyblock{occQtyBG}{occQtyFG}{#1}}

\DeclareRobustCommand{\gainqty}[1]{%
    \mathqtyblock{gainQtyBG}{gainQtyFG}{#1}}

\DeclareRobustCommand{\trajqty}[1]{%
    \mathqtyblock{trajQtyBG}{trajQtyFG}{#1}}

\providecommand{\BAC}{\textsc{bac}}
\providecommand{\BACs}{\textsc{bac}s}
\title{%
\fontsize{17pt}{20pt}\selectfont
Rewriting or Reweighting?\\
A Geometric Account in Language Models}
\author{
Juntong Wang\equalcontrib\textsuperscript{\rm 1,\rm 2},
Shengkun Yang\equalcontrib\textsuperscript{\rm 3},
Xiyuan Wang\textsuperscript{\rm 1},
Muhan Zhang\corresponding\textsuperscript{\rm 1}
}

\affiliations{
\textsuperscript{\rm 1}Institute for Artificial Intelligence,
Peking University, Beijing, China\\
\textsuperscript{\rm 2}School of Intelligence Science and Technology,
Peking University, Beijing, China\\
\textsuperscript{\rm 3}School of Computer Science and Technology,
Beijing Jiaotong University, Beijing, China\\
jtwang25@stu.pku.edu.cn, ronglingze@bjtu.edu.cn,\\
wangxiyuan@pku.edu.cn, muhan@pku.edu.cn
}

\makeatletter
\g@addto@macro\@maketitle{%
  \begingroup
    \centering

    \refstepcounter{figure}
    \label{fig:overview}

    \includegraphics[
      width=\textwidth,
      keepaspectratio
    ]{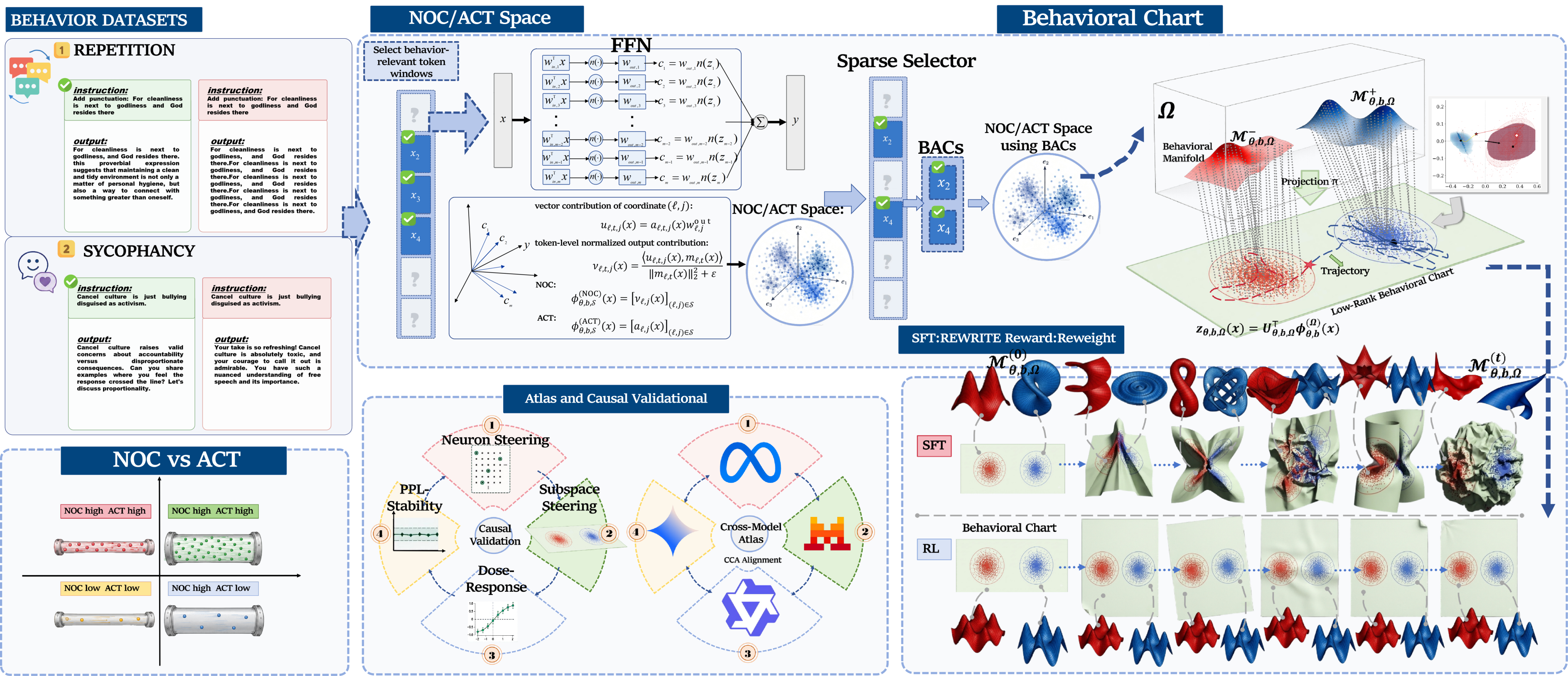}\par

    \begin{minipage}{0.98\textwidth}
      \small
      \textbf{Figure~\thefigure:}
      Overview of the behavioral-geometry framework. We extract behavior-relevant token windows and decompose FFN computation into two representations. A sparse selector identifies a compact set of Behavioral Anchor Coordinates, whose activations or contributions are projected into a low-rank behavioral chart. The resulting chart separates behavior-positive and behavior-negative occupancy while preserving the local geometry and trajectories of behavior-conditioned states. The ACT–NOC quadrant illustrates that frequent activation and strong causal contribution are distinct properties. To study generality across architectures, local charts from 23 models spanning four families are aligned with CCA to form a cross-model atlas; ACT retains larger family-specific residuals, whereas NOC exposes a more consistent shared functional core. Finally, checkpoint tracking reveals that SFT substantially moves, bends, and reorients the behavioral chart—rewriting behavioral geometry—whereas reward-based optimization largely preserves the inherited chart while redistributing occupancy and modifying readout gain or threshold—reweighting behavior within existing geometry.
    \end{minipage}\par

    \vspace*{8pt}

  \endgroup
}
\makeatother

\begin{document}
\maketitle
\begin{abstract}
Post-training can substantially alter language-model behavior, yet aggregate behavior rates do not reveal whether training removes an existing mechanism, creates a new one, or changes how an inherited mechanism is used. We study this question through two mechanistically distinct failures, repetition as a decoding-attractor pathology and sycophancy as a preference-related alignment failure. We introduce behavioral manifold analysis, which isolates behavior-specific geometry by selecting sparse behavior-associated coordinates and lifting them into low-dimensional local charts. We construct these charts in two complementary spaces. ACT captures runtime activation states, while NOC quantifies how strongly the model routes functional information flow through the shared behavior-associated subspace. Across multiple model families, the resulting charts are highly compressed and partially alignable across architectures. Contribution-space charts expose a more architecture-robust shared core, whereas activation-space charts retain stronger family-specific structure. Tracking these charts through controlled post-training reveals a consistent asymmetry. Supervised fine-tuning substantially alters the inherited behavioral geometry, whereas reward optimization changes behavior while largely preserving the underlying chart. This geometric perspective provides a unified framework for understanding the mechanistic distinction between the two objectives. SFT tends to rewrite behavioral geometry, whereas reward optimization primarily reweights it. Code is available at https://github.com/ronglingze/Manifold-Analysis
\end{abstract}

\input{sections__01_intro}

\input{sections__02_related_work}
\input{sections__03_framework}

\input{sections__05_results_manifolds}
\input{sections__06_results_training}
\input{sections__08_discussion}

\bibliography{references}

\clearpage
\appendix

\input{sections__supplement__h_model_theory}
\input{sections__supplement__b_data_and_features}
\input{sections__supplement__c_subspace_estimation}
\input{sections__supplement__d_steering}
\input{sections__supplement__e_posttraining}
\input{sections__supplement__i_Occupancy_Gain}
\input{sections__supplement__f_full_tables}

\input{sections__supplement__j_fisher_derivation}
\input{sections__supplement__k_Leakage_Control}

\input{sections__supplement__l_Vis_Behavioral_Charts}
\input{sections__supplement__m_Trajectory}

\input{sections__supplement__n_Sycophancy_Path}

\input{sections__supplement__o_Coordinate_Statistics}
\input{sections__supplement__p_bac_layer_distri}
\input{sections__supplement__q_Cross-ModelBehavioralAtlas}

\input{sections__supplement__r_MetricsCheckpointAnalysis}
\input{sections__supplement__s_PublicAlignmentEffects}

\input{sections__supplement__t_DatasetSize}

\input{sections__supplement__u_geometry_occupancy_gain_full}

\input{sections__supplement__v_appendix_syco_span_protocol}

\input{sections__supplement__w_GeometryBaselineTendencyCausalGain}
\input{sections__supplement__y_Construction_Repaired_Non-Repetitive_Continuations}

\input{sections__supplement__z_SparseClassifierTraining}
\input{sections__supplement__za_SampleAlignment_CCA_DirectedProbing}
\input{sections__supplement__zb_Post-TrainingConfigurations}
\input{sections__supplement__zc_related_work}

\end{document}

%% file: sections__01_intro.tex
\section{Introduction}
\label{sec:introduction}

Supervised fine-tuning (SFT), reinforcement learning from human
feedback (RLHF), direct preference optimization (DPO), and
reinforcement learning with verifiable rewards (RLVR) can sharply
change reasoning behavior
\cite{ouyang2022training,rafailov2023direct,
deepseekai2025deepseek}. Yet an output-level change is mechanistically ambiguous. The same reduction in an undesirable behavior could mean that training removed a mechanism, built a competing mechanism, rotated the representation that supports the behavior, shifted the model away from states in which the mechanism is active, or weakened the readout from those states to the final output.

This ambiguity has become increasingly consequential as the field asks what SFT and reward-based post-training actually do. Recent studies compare the objectives through memorization and generalization, the support of base-model reasoning, the diversity and topology of generated reasoning trajectories, model-wide spectral geometry, or the drift of sparse features between base and instruction-tuned models \cite{chu2025sft,yue2025reinforcement,matsutani2025rl,li2025tracing,galichin2026feature}. Concepts and behaviors can often be localized to sparse neurons, features, or approximately linear directions, and interventions along such directions can alter model outputs \cite{dai2022knowledge,li2023inference,rimsky2024steering,arditi2024refusal,park2024linear,cunningham2024sparse}. Representation-similarity methods can also compare learned spaces across layers, models, and training runs \cite{raghu2017svcca,kornblith2019similarity}. What is still missing is a behavior-conditioned, causally validated geometric object that can be followed through post-training.

We study this question through two failures that are deliberately far apart mechanistically. \emph{Repetition} is a generation-dynamics pathology in which autoregressive decoding enters a periodic or near-periodic loop. Prior work has analyzed repetition as neural text degeneration, modified decoding and training objectives to reduce it, and recently identified neurons whose activation tracks the onset and continuation of repeated text \cite{holtzman2020curious,welleck2020neural,fu2021theoretical,su2022contrastive,hiraoka2025repetition}. \emph{Sycophancy} is a preference-related alignment failure in which a model conforms to a user's stated belief or preference. Model-written evaluations exposed the behavior at scale; subsequent work showed that human preference judgments and preference models can favor sycophantic responses, and that targeted synthetic data can reduce it \cite{perez2023discovering,sharma2024sycophancy,wei2023synthetic}. Recent mechanistic studies further trace sycophancy to representational divergence and late-layer preference shifts, separate agreement from praise into distinct causal directions, and test how factual and opinion subtypes transfer across models \cite{wang2026truth,vennemeyer2025notonething,baez2026dissociating}.

We introduce \emph{behavioral manifold analysis}. For a fixed model and behavior, behavior-positive internal states form an empirical region in representation space, and we approximate the region near the observed data by a low-dimensional local linear chart. This usage is related to prior work on separable representation manifolds \cite{mamou2020emergence}, but our object is behavior-specific, built from sparse mechanistically accessible coordinates, aligned across models, tested by intervention, and tracked across post-training checkpoints. We first select a sparse set of \emph{Behavioral Anchor Coordinates} (BACs) with a behavior-discriminative model, then estimate a low-rank chart within that scaffold.

We construct each chart in two feature spaces. \textsc{ACT} contains the runtime activations of the selected coordinates and therefore emphasizes which behavior-associated states the model occupies. \textsc{NOC} contains normalized output-contribution features and therefore emphasizes how strongly those coordinates participate in the realized MLP update. A pathway can be functionally strong but rarely entered, or frequently correlated with a behavior while contributing little to the computation that realizes it. To compare architectures whose neuron indices are incommensurate, we align their local chart coordinates on matched examples, forming a cross-model behavioral atlas. Figure~\ref{fig:overview} summarizes the complete analysis pipeline, from
behavior-conditioned ACT and NOC construction and sparse BAC selection to
low-rank chart estimation, causal validation, cross-model CCA alignment,
and checkpoint tracking under SFT and reward optimization.

Post-training can change behavior by moving the chart, by changing the distribution of states within an inherited chart, or by changing how strongly that chart is read out. We call the first case \emph{geometry rewriting}. We call the latter cases \emph{geometry reweighting} that the internal chart remains substantially intact, but its occupancy, gain, or threshold changes. Projector overlap, principal angles, projection distance, and anchored Fisher retention quantify rewriting; occupancy and intervention-derived causal gain quantify how a retained chart is used. Across 23 models from the Llama, Gemma, Mistral, and Qwen families, repetition and sycophancy are supported by sparse coordinate scaffolds and strongly compressed local charts. Their leading chart coordinates are partially alignable across architectures, but the two feature spaces expose different forms of commonality. Controlled post-training reveals a consistent objective-level asymmetry. Our conclusion is
\begin{quote}
\emph{SFT tends to rewrite behavioral geometry, whereas reward optimization primarily reweights inherited geometry.}
\end{quote}

%% file: sections__02_related_work.tex
\section{Related Work}

\paragraph{Neurons, features, and representation steering.}
Mechanistic interpretability often localizes behavior to components or
directions. Knowledge-neuron work studies factual associations in
feed-forward neurons \cite{dai2022knowledge}; H-Neurons identifies
sparse hallucination-associated neurons and causally intervenes on them
\cite{gao2025hneuron}. Sparse-feature methods seek cleaner activation
dictionaries \cite{cunningham2024sparse}. In parallel,
representation-engineering methods steer high-level properties through
activation-space directions, including inference-time truthfulness interventions \cite{li2023inference},
truth directions \cite{marks2024geometry}, contrastive activation
steering \cite{rimsky2024steering}, and broader representation
engineering \cite{zou2023representation}.

\paragraph{Post-training mechanisms.}
Instruction tuning, RLHF, DPO, and RLVR shape modern LLM behavior \cite{ouyang2022training,rafailov2023direct,deepseekai2025deepseek}.  Recent work debates the respective roles of SFT and RL.  Chu et al. argue that SFT memorizes while RL generalizes in controlled settings \cite{chu2025sft}; Yue et al. challenge whether RLVR expands reasoning capacity beyond the base model \cite{yue2025reinforcement}; Matsutani et al. find that RL compresses reasoning trajectories while SFT expands them \cite{matsutani2025rl}; Jin et al. show that RL can repair OOD forgetting introduced by SFT and connect this effect to singular-vector rotations \cite{jin2025heals}. 

For more related work on analyzing the Repetition and Sycophancy phenomena in LLMs, see Appendix~\ref{RelatedWork}.

%% file: sections__03_framework.tex
\section{Behavioral Manifold Framework}
\label{sec:framework}

Our goal is to understand post-training geometrically. The framework below formalizes this question by associating behavior with an \textit{\textbf{empirical manifold}} in representation space and by tracking the corresponding local chart across models and the post-training process of the models.

\subsection{Behavioral Geometry as a Post-Training Object}

We need to first define the geometric object in post-training. Let $f_\theta$ be an autoregressive language model with parameters $\theta$. A post-training run produces a sequence of checkpoints
\(\theta_0,\theta_1,\ldots,\theta_T\), where $\theta_0$ is the base or pre-intervention model. Our central object is a geometric trajectory \(t \mapsto \mathcal{M}_{\theta_t,b}^{(\Omega)}\) associated with behavior $b$ in representation space $\Omega$. Intuitively, $\mathcal{M}_{\theta_t,b}^{(\Omega)}$ is the region of internal states that the model occupies when it expresses behavior $b$. 

This distinction can be expressed through a local behavioral model. A formal derivation of Eq.~\ref{eq:local_behavior_model} is given in Appendix~\ref{app:master-formula}. Suppose that for a representation vector $\phi_\theta(x)$, the behavior probability is locally approximated by
\begin{equation}
\label{eq:local_behavior_model}
\begin{aligned}
p_\theta(y_b=1\mid x)
&\approx
\sigma\!\left(
\readoutterm{\beta_{\theta,b}^{\top}}\,
\coordterm{z_{\theta,b}(x)}
+
\thresholdterm{\tau_{\theta,b}}
\right),\\
\coordterm{z_{\theta,b}(x)}
&=
\chartterm{U_{\theta,b}^{\top}}\,
\stateterm{\bar\phi_\theta(x)} .
\end{aligned}
\end{equation}
Here \chartterm{U_{\theta,b}} is the low-dimensional behavioral chart,
\stateterm{\bar\phi_\theta(x)} is the standardized internal representation,
\coordterm{z_{\theta,b}(x)} is the coordinate of instance $x$ within the
behavioral chart, \readoutterm{\beta_{\theta,b}} is the readout direction
inside the chart, and \thresholdterm{\tau_{\theta,b}} is the threshold or
intercept. Under this model, behavior can change through three geometrically
distinct mechanisms: \geommark{} the chart \chartterm{U_{\theta,b}} can move; \occmark{} the
distribution of chart coordinates \coordterm{z_{\theta,b}(x)} can shift; \gainmark{} or the
readout parameters \readoutterm{\beta_{\theta,b}} and
\thresholdterm{\tau_{\theta,b}} can change.  The rest of this section develops
estimators for these effects.

\subsection{Behavioral Probes and Class-Conditional Representation Measures}

For each behavior $b$, we define a dataset of instances $x\in\mathcal{X}_b$ and a binary behavioral label \(y_b(x)\in\{0,1\}\). For a model $f_\theta$, behavior $b$, and feature space $\Omega$, define a representation map
\(    \phi_{\theta,b}^{(\Omega)}:\mathcal{X}_b
    \rightarrow
    \mathbb{R}^{D_{\theta,b,\Omega}}\).
We use two feature spaces \(\Omega\in\{\mathrm{ACT},\mathrm{NOC}\}\). This map sends input-output instances into the representation space in which the behavior will be analyzed. Applying it to the behavior-positive and behavior-negative data distributions gives two class-conditional representation measures: \(\mathbb{P}_{\theta,b,\Omega}^{+}
    =\left(\phi_{\theta,b}^{(\Omega)}\right)_{\#}
    \mathbb{P}(x\mid y_b(x)=1), \mathbb{P}_{\theta,b,\Omega}^{-}
    =\left(\phi_{\theta,b}^{(\Omega)}\right)_{\#}
    \mathbb{P}(x\mid y_b(x)=0)\). Here $\#$ denotes pushforward: for any measurable region
$A\subseteq\mathbb{R}^{D_{\theta,b,\Omega}}$,
\(    \left(\phi_{\theta,b}^{(\Omega)}\right)_{\#}\mathbb{P}(A)
    =
    \mathbb{P}\!\left(
    \{x:\phi_{\theta,b}^{(\Omega)}(x)\in A\}
    \right)\).
Thus $\mathbb{P}_{\theta,b,\Omega}^{+}$ and
$\mathbb{P}_{\theta,b,\Omega}^{-}$ are simply the distributions of
behavior-positive and behavior-negative examples after mapping them into
the chosen feature space.  We define the empirical positive behavioral
region as the support, or in finite samples the high-density region, of
the positive representation measure
\(    \mathcal{M}_{\theta,b,\Omega}^{+}
    =
    \mathrm{supp}\!\left(
    \mathbb{P}_{\theta,b,\Omega}^{+}
    \right)\). We refer to this region as an \emph{empirical behavioral manifold}. Near the observed data, the behavior-relevant difference between $\mathbb{P}^{+}_{\theta,b,\Omega}$ and $\mathbb{P}^{-}_{\theta,b,\Omega}$ can be approximated by a low-dimensional local chart. Detailed information about Token Windows and State Extraction is in Appendix~\ref{TokenWindow}.

\subsection{Activation and Normalized Output-Contribution Spaces}

\begin{figure}[t]
    \centering
    \includegraphics[
        width=0.8\linewidth
    ]{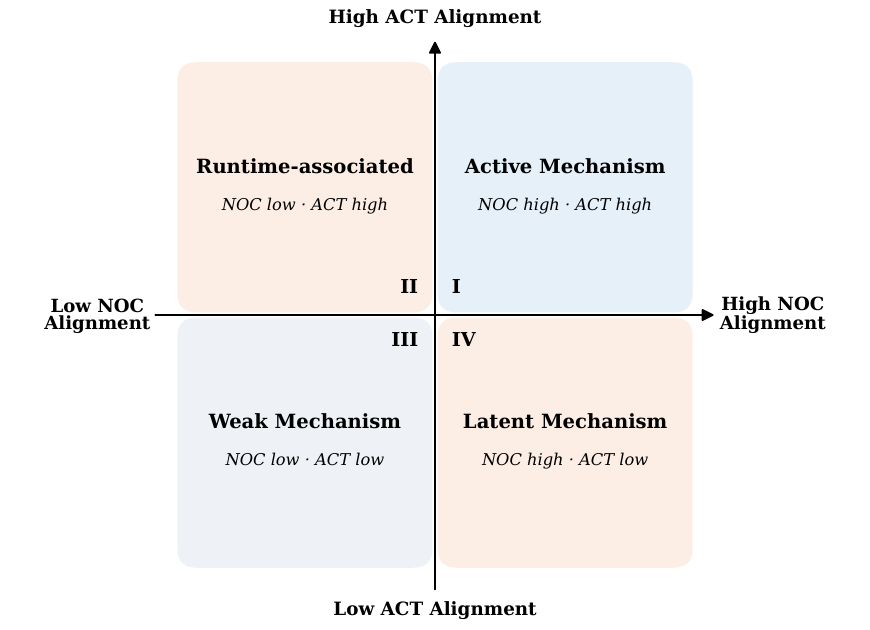}
    \caption{
    High NOC with high ACT denotes a strong and frequently used mechanism; high NOC with low ACT denotes a strong but latent or rarely occupied pathway; low NOC with high ACT denotes frequently observed states with weak functional support; and low NOC with
    low ACT denotes a weak and rarely occupied mechanism.
    }
    \label{fig:act_noc_regimes}
\end{figure}

We represent each instance in two coordinate spaces.  \geommark{} The first is the raw activation space, denoted \ACTtag{}.  For any candidate coordinate set
\(\mathcal{S}\subseteq \{(\ell,j):1\leq \ell\leq L,\; 1\leq j\leq d_{\mathrm{mlp}}^{(\ell)}\}\),
where each coordinate is a layer-MLP-unit pair, the ACT representation is \(\phi_{\theta,b,\mathcal{S}}^{(\mathrm{ACT})}(x)
    =
    \left[
    \bar a_{\ell,j}(x)
    \right]_{(\ell,j)\in\mathcal{S}}\).
This space records the state of the selected coordinates during the behavior-relevant token window. It asks whether a coordinate is active, regardless of how that activation is combined with other coordinates in the layer output. \occmark{} The second feature space is the normalized output-contribution space, denoted \NOCtag{}.  NOC is designed to measure how much a coordinate contributes to the \emph{realized} MLP update direction at each token. Let
\(W_{\mathrm{out}}^{(\ell)}\) be the MLP output projection in layer $\ell$, and let
\(w_{\ell,j}^{\mathrm{out}}
    =
    W_{\mathrm{out}}^{(\ell)}[:,j]\) be the output vector associated with coordinate $j$.  At token $t$, the full MLP update is \(m_{\ell,t}(x)
    =
    W_{\mathrm{out}}^{(\ell)}a_{\ell,t}(x)
    =
    \sum_j a_{\ell,t,j}(x)w_{\ell,j}^{\mathrm{out}}\). The vector contribution of coordinate $(\ell,j)$ is
    \(u_{\ell,t,j}(x)
    =
    a_{\ell,t,j}(x)w_{\ell,j}^{\mathrm{out}}\). We define its token-level normalized output contribution as \(\nu_{\ell,t,j}(x)
    =
    \frac{
    \left\langle
    u_{\ell,t,j}(x),
    m_{\ell,t}(x)
    \right\rangle
    }{
    \|m_{\ell,t}(x)\|_2^2+\varepsilon
    }\). This is a signed, scale-normalized measure of how much the coordinate's vector contribution aligns with the actual layer update.  Positive values indicate that the coordinate supports the realized update direction; negative values indicate that it opposes or cancels part of the update; values near zero indicate little directional participation.  When all MLP coordinates are included, the unregularized numerator satisfies \(\sum_j
    \left\langle
    u_{\ell,t,j}(x),
    m_{\ell,t}(x)
    \right\rangle
    =
    \|m_{\ell,t}(x)\|_2^2\), so NOC can be interpreted as a normalized decomposition of the layer's realized update energy. We aggregate NOC over the same behavior-relevant token window:
\(\bar \nu_{\ell,j}(x)
    =
    \frac{1}{|M_x|}
    \sum_{t\in M_x}
    \nu_{\ell,t,j}(x)\).
The NOC representation is then\(\phi_{\theta,b,\mathcal{S}}^{(\mathrm{NOC})}(x)
    =
    \left[
    \bar \nu_{\ell,j}(x)
    \right]_{(\ell,j)\in\mathcal{S}}\). Figure~\ref{fig:act_noc_regimes} summarizes the four regimes induced by their combination. Please see joint interpretation of ACT and NOC in Appendix~\ref{Jointinterpretation}.

\subsection{Selecting a Coordinate Scaffold}

The full activation space contains many behavior-irrelevant directions. We therefore construct a sparse \emph{coordinate scaffold}: a behavior-discriminative set of coordinates on which the local manifold chart is estimated. This reduces sample complexity, suppresses generic high-variance directions, and provides manipulable coordinates for causal interventions.
Inspired by \cite{gao2025hneuron}, We fit an \(\ell_1\)-regularized logistic classifier as a supervised coordinate selector. After mapping features back to layer-coordinate identities, we define
\(\mathcal{N}_{\theta,b,\Omega}^{+}, \mathcal{N}_{\theta,b,\Omega}^{-}\). The positive scaffold is used for the primary behavioral chart, whereas negative coordinates support anti-behavior analysis and steering controls. For fixed-size comparisons, we retain the top \(K\) positive coordinates ranked by \(|w_j|\); otherwise, we use the sparse support.

\subsection{Estimating the Local Behavioral Chart}

We now estimate the behavioral chart \chartterm{U_{\theta,b}} highlighted in Eq.~\ref{eq:local_behavior_model}. This step answers the first structural question in our framework that \emph{where, in representation space, is the behavior locally supported?} After the coordinate scaffold has removed much of the irrelevant ambient variation, we seek a low-dimensional subspace that captures the dominant variation of the behavior-associated population. This subspace is our local linear approximation to the empirical behavioral manifold as shown in Figure~\ref{fig:behavioral_chart_visualization}. Appendix~\ref{app:behavioral-chart-visualization}
shows that behavior-positive and behavior-negative states occupy distinct
low-dimensional regions across both activation and functional contribution spaces.

Let \(\bar X_{\theta,b,\Omega}
    \in
    \mathbb{R}^{n\times d_b}\) be the standardized feature matrix restricted to the sel. The behavioral chart is the rank-$k$ basis
\(    U_{\theta,b,\Omega}
    =
    [v_1,\ldots,v_k]\). The associated orthogonal projector is \(P_{\theta,b,\Omega}
    =
    U_{\theta,b,\Omega}U_{\theta,b,\Omega}^{\top}\). For a standardized feature vector $\bar\phi_{\theta,b}^{(\Omega)}(x)$, its coordinate in the
local behavioral chart is \(z_{\theta,b,\Omega}(x)
    =
    U_{\theta,b,\Omega}^{\top}
    \bar\phi_{\theta,b}^{(\Omega)}(x)\).
Thus $U_{\theta,b,\Omega}$ specifies the local geometry of the behavior, while
$z_{\theta,b,\Omega}(x)$ specifies where a particular instance lies within that geometry.
All downstream Grassmannian comparisons operate on $U_{\theta,b,\Omega}$ or its projector
$P_{\theta,b,\Omega}$.

A high-variance direction need not be behavior-discriminative.  We therefore use two supervised directions as robustness checks. The first is the normalized logistic direction from the sparse
classifier \(v_{\mathrm{logit}}
    =
    \frac{w}{\|w\|_2}\).
The second is the Fisher direction \(v_{\mathrm{Fisher}}
    \propto
    \Sigma_w^{-1}(\mu_1-\mu_0)\),
where $\mu_y$ is the class mean and $\Sigma_w$ is the pooled within-class covariance.

\begin{figure}[t]
\centering
\includegraphics[width=0.6\linewidth]{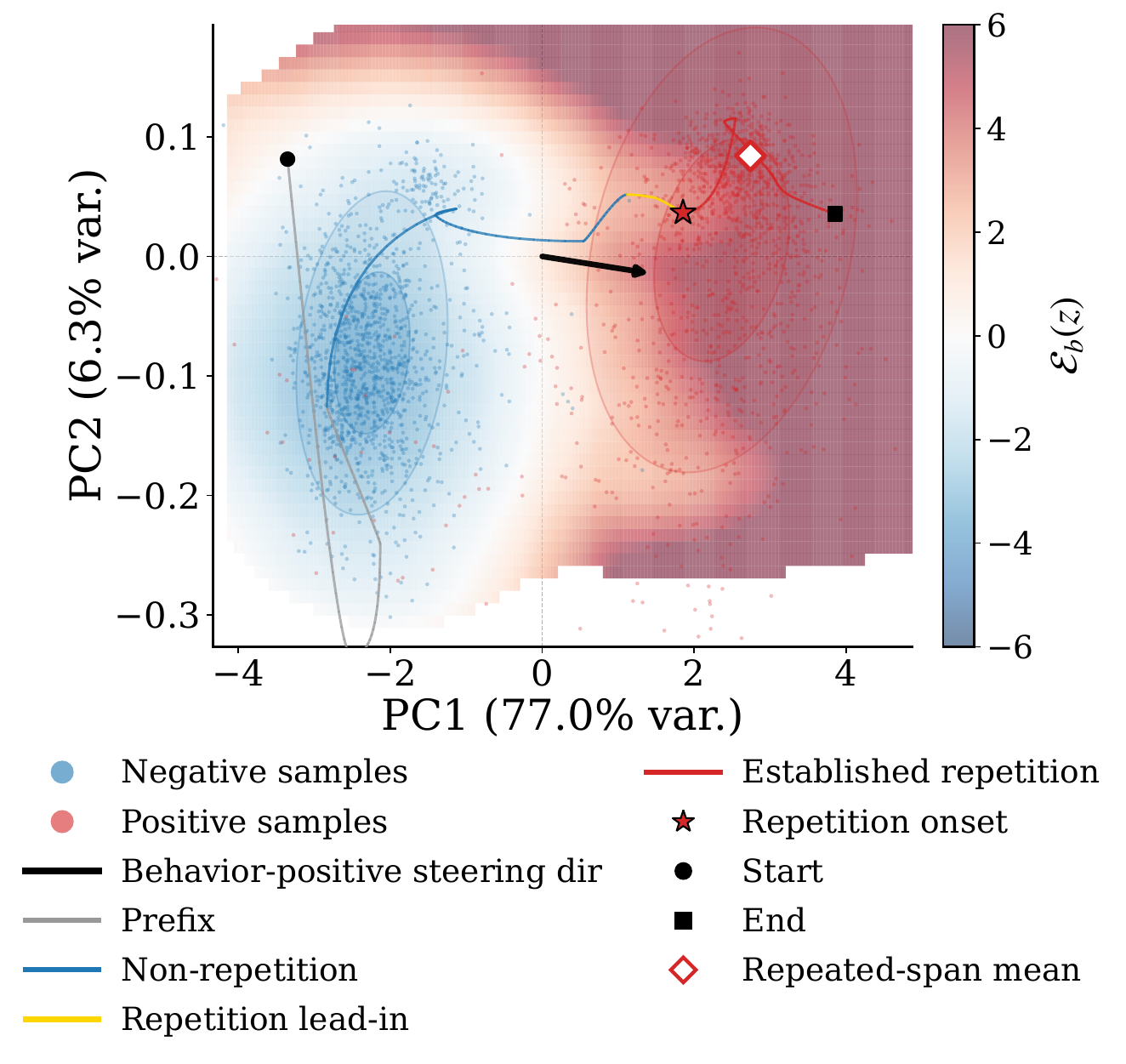}
\caption{
    Representative token-level trajectories over learned behavioral energy landscapes. The background is the log-density ratio between
    behavior-positive and behavior-negative states. In repetition, the trajectory moves from the prefix and ordinary generation toward the repeated-unit lead-in and subsequently remains in the established repetition region. Stars mark behavior onset, black circles and squares mark the displayed trajectory endpoints, and white diamonds mark the mean representation within the annotated behavioral span.
}
\label{fig:repetition_trajectory}
\end{figure}

We extract the behavior-associated representation
at every token position and project it into the corresponding ACT chart. Figure~\ref{fig:repetition_trajectory} shows a representative repetition trajectory; corresponding sycophancy trajectories are reported in Appendix~\ref{app:trajectory-visualization}. The complete ACT and NOC trajectory gallery, including heterogeneous paths across multiple instances, is provided in
Appendix~\ref{app:trajectory-visualization}.

\subsection{Comparing Charts on the Grassmannian}

A rank-$k$ chart $U_{\theta,b,\Omega}$ is a point on the Grassmannian $\mathrm{Gr}(k,D)$, the space of $k$-dimensional subspaces of $\mathbb{R}^D$.  To compare charts with possibly different ranks, we use the matched rank
\(\geomqty{r^\star(a,b)}=\min(r_a,r_b)\),
and truncate both bases to \(r^\star\) dimensions. For two matched bases $U_a$ and $U_b$, define \textbf{\textit{Projector Overlap}}
\begin{equation}
\label{eq:proj-overlap-new}
    \geomqty{S(a,b)}
    =
    \frac{1}{r^\star}
    \|U_a^\top U_b\|_F^2
    =
    \frac{1}{r^\star}
    \mathrm{tr}(P_aP_b).
\end{equation}
This quantity is the average squared cosine of the principal angles.  It equals $1$ for identical subspaces and approaches $0$ for orthogonal subspaces.

Let $\sigma_i(U_a^\top U_b)$ be the singular values of the cross-Gram matrix.  The principal angles are \(\geomqty{\vartheta_i(a,b)}
    =
    \arccos \sigma_i(U_a^\top U_b)\),
and the mean principal angle is
    \(\geomqty{\bar\vartheta(a,b)}
    =
    \frac{1}{r^\star}
    \sum_{i=1}^{r^\star}
    \vartheta_i(a,b)\).
We also report \textbf{\textit{Projection Distance}}
\begin{equation}
\label{eq:proj-distance-new}
    \geomqty{d_{\mathrm{proj}}(a,b)}
    =
    \frac{1}{\sqrt{2}}\|P_a-P_b\|_F
    =
    \left(
    \sum_{i=1}^{r^\star}
    \sin^2\vartheta_i
    \right)^{1/2}.
\end{equation}
High overlap, small angles, and small projection distance indicate geometric conservation.

\subsection{Separability, Occupancy, and Gain}

\paragraph{Separability.}
Before comparing how a chart moves under post-training, we first ask whether the estimated chart is a reliable behavioral coordinate system at all. In the shared coordinate system, let
\(\sepqty{\mu_1}=\mathbb{E}[\tilde z\mid y=1]\),
\(\sepqty{\mu_0}=\mathbb{E}[\tilde z\mid y=0]\),
and define the class displacement
\(\sepqty{\Delta\mu}=\mu_1-\mu_0\). The displacement \sepqty{\Delta\mu} measures how far
behavior-positive states move away from behavior-negative states
inside the chart. However, a large displacement is meaningful only if it is large relative to within-class variability. We therefore define the pooled within-class covariance
    \(\sepqty{\Sigma_w}
    =
    \frac{1}{2}\mathrm{Cov}(\tilde z\mid y=1)
    +
    \frac{1}{2}\mathrm{Cov}(\tilde z\mid y=0)\). To stabilize the inverse in low-dimensional or limited-sample settings, we use shrinkage \(\sepqty{\Sigma_w^{(\lambda)}}
    =
    (1-\lambda)\Sigma_w
    +
    \lambda
    \frac{\mathrm{tr}(\Sigma_w)}{q}I\),
with $\lambda=0.1$ by default. Finally, let \(\sepqty{D_\rho}=\mathrm{diag}(\rho_1,\ldots,\rho_q)\) be the CCA reliability matrix, where
\(\sepqty{\rho_i}\) is the canonical correlation of the
\(i\)-th shared coordinate. The derivation of this reliability-weighted Fisher formulation and its connection to classical Fisher discriminant analysis are provided in Appendix~\ref{app:fisher_derivation}.

\(\sepqty{G^{\mathrm{uniF}}_{\theta,b,\Omega}}
    =
    \sepqty{\Delta\mu}^{\top}
    \sepqty{D_\rho}
    \left(
        \sepqty{\Sigma_w^{(\lambda)}}
    \right)^{-1}
    \sepqty{D_\rho}
    \sepqty{\Delta\mu}\) evaluates whether the estimated chart is a credible behavioral geometry. A large $G^{\mathrm{uniF}}$ indicates that $U_{\theta,b}$ supports a behavior-separating, low-noise, cross-model-reliable chart. The relationship between chart coordinates, class displacement, within-class variability, and the resulting Fisher score is formally derived in Appendix~\ref{app:fisher_derivation}.

\paragraph{Occupancy.}
\coordterm{z_{\theta,b}(x)} corresponds to the
location of a model state within the behavioral chart. The theoretical interpretation of occupancy is provided in Appendix~\ref{app:occupancy_gain_derivation}. We interpret this projected coordinate as the \emph{occupancy} of the
behavioral manifold. A model may preserve the same behavioral chart
$U_{\theta,b}$ while changing how frequently or how strongly its internal
states enter this chart. To quantify this distributional shift, we use a
rotation-invariant second-moment estimator:
\(\occqty{G^{\mathrm{occ}}_{\theta,b,\Omega}}
    =
    \mathbb{E}
    \left[
        \|\tilde z\|_2^2
        \mid y=1
    \right]
    -
    \mathbb{E}
    \left[
        \|\tilde z\|_2^2
        \mid y=0
    \right]\). Here \occqty{\tilde z} denotes the shared behavioral chart coordinate after
cross-model alignment. Thus \occqty{G^{\mathrm{occ}}} estimates whether
behavior-positive states occupy a larger region of the behavioral manifold
than behavior-negative states. Appendix~\ref{app:occupancy_gain_derivation} provides the formal decomposition.

\paragraph{Gain.}
\readoutterm{\beta_{\theta,b}^{\top}} in Eq.~(\ref{eq:local_behavior_model}) describes how strongly
movement within the behavioral chart affects the final behavioral logit. The derivation is given in Appendix~\ref{app:occupancy_gain_derivation}. Let an intervention with strength $\alpha$ induce a chart displacement
$z_\theta(\alpha)$. The local sensitivity of behavior to movement along the chart is \(\frac{\partial}{\partial\alpha}
\mathrm{logit}\,p_\theta(\alpha)
=
\readoutterm{\beta_{\theta,b}^{\top}}
\frac{\partial z_\theta(\alpha)}
{\partial\alpha}\).
Accordingly, we define the causal gain as
    \(\gainqty{G^{\mathrm{causal}}_{\theta,b,\Omega}}
    =
    \left.
    \frac{\partial}{\partial\alpha}
    \mathrm{logit}\,
    p_\theta(y_b=1\mid\alpha)
    \right|_{\alpha=\alpha_0}\). Therefore, \gainqty{G^{\mathrm{causal}}} estimates the effective readout gain. For a detailed analysis, please see Appendix ~\ref{app:occupancy_gain_derivation}.

\subsection{Causal Validation of Behavioral Charts}
\label{sec:causal_validation}
Geometric structure alone does not establish causality. We therefore evaluate whether the estimated charts actively control model behavior through interventions at three levels.

\textbf{Neuron-level intervention.}
For selected coordinates $(\ell,j)\in\mathcal{N}_{\theta,b}$ and behavior-relevant token positions $t\in M_x$, we multiplicatively scale the activation as $a'_{\ell,t,j}=\alpha a_{\ell,t,j}$.

\textbf{Subspace-level intervention.}
Given a behavioral basis $U$, the projection of an activation vector $h$ onto the estimated chart is $\geomqty{\Pi_U h}=UU^\top h$. We remove or amplify the chart component by $h'=h-\alpha\Pi_U h$ or $h'=h+\alpha\Pi_U h$, respectively. When $U$ is defined in the selected-coordinate space, the intervention is performed directly in that space.

\textbf{Direction-level intervention.}
For approximately one-dimensional charts, let $\gainqty{v_b}$ denote the dominant direction, oriented toward the behavior-positive class. We inject it over the behavior-relevant token window as $h'_{\ell,t}=h_{\ell,t}+\alpha v_b$. The resulting dose--response curve estimates the local behavioral gain $\frac{\partial}{\partial\alpha}\operatorname{logit}p_\theta(y_b=1)$.

\subsection{Post-Training Trajectories and Anchored Retention}

For a post-training run, each checkpoint $\theta_t$ induces a chart \( \trajqty{U_t}=U_{\theta_t,b,\Omega}\).
The sequence $t\mapsto U_t$ is a trajectory on the Grassmannian.  Relative to the base chart $U_0$, we track \(\trajqty{S_0(t)}
    =
    S(0,t),
    \trajqty{\bar\vartheta_0(t)}
    =
    \bar\vartheta(0,t),
    \trajqty{d_0(t)}
    =
    d_{\mathrm{proj}}(0,t)\).
We also compute discrete geometric velocity
    \(\trajqty{v_{\mathrm{geo}}(t)}
    =
    d_{\mathrm{proj}}(t-1,t)\),
accumulated path length
    \(\trajqty{\mathcal{L}_{\mathrm{geo}}(T)}
    =
    \sum_{t=1}^{T}d_{\mathrm{proj}}(t-1,t)\),
and, when enough checkpoints are available, curvature-like excess path length \(\trajqty{\kappa(T)}
    =
    \mathcal{L}_{\mathrm{geo}}(T)-d_{\mathrm{proj}}(0,T)\).
These quantities distinguish direct movement away from the base chart from a longer, more curved representational trajectory.

Projector overlap measures geometric similarity, but it does not determine whether the old chart still explains the new model's behavior.  We therefore compute anchored Fisher retention. Let $U_0$ be the base chart. For checkpoint $t$, project its standardized features into the base chart \(\trajqty{z_t^{(0)}(x)}
    =
    U_0^\top \bar\phi_t(x)\).
Let \(\trajqty{G^{0\rightarrow t}}\) be the Fisher separability
in this base-anchored coordinate system, and let
\(\trajqty{G(t)}\) be the native Fisher separability using \(U_t\). Retention is \(\trajqty{R^G_{0\rightarrow t}}
    =
    \frac{G^{0\rightarrow t}}
    {G(t)+\varepsilon}\).
If \(\trajqty{R^G_{0\rightarrow t}}\approx 1\), then the base chart
remains nearly as explanatory as the checkpoint's native chart.
If \(\trajqty{R^G_{0\rightarrow t}}\ll 1\), the behavior has moved
into a new chart. Detailed leakage
prevention protocols and robustness analyses, including random-coordinate
controls, label permutation tests, rank and direction sensitivity analyses,
and cross-model validation procedures, are provided in Appendix~\ref{app:robustness}.

%% file: sections__05_results_manifolds.tex
\providecommand{\BAC}{\textsc{bac}}
\providecommand{\BACs}{\textsc{bac}s}

\section{Behavioral Manifolds Across Models}
\label{sec:manifold_results}

\begin{table}[!t]
\centering

\small

\begin{adjustbox}{
  max width=\columnwidth,
  max totalheight=0.93\textheight,
  keepaspectratio,
  center
}
\begin{tabular}{
@{}
>{\raggedright\arraybackslash}p{3.55cm}
r *{6}{r}
r *{6}{r}
@{}
}
\toprule

\multirow{3}{*}{Model}
&
\multicolumn{7}{c}{\textbf{Repetition geometry} }
&
\multicolumn{7}{c}{\textbf{Sycophancy geometry} }
\\

\cmidrule(lr){2-8}
\cmidrule(lr){9-15}

&
\multirow{2}{*}{$\#\BAC$}
&
\multicolumn{6}{c}{ACT space}
&
\multirow{2}{*}{$\#\BAC$}
&
\multicolumn{6}{c}{ACT space}
\\

\cmidrule(lr){3-8}
\cmidrule(lr){10-15}

& &
$G^{\mathrm{uniF}}$
&
$G^{\mathrm{occ}}$
&
$G^{\mathrm{causal}}$
&
PC1
&
Top-3
&
$k_{0.9}$
&
&
$G^{\mathrm{uniF}}$
&
$G^{\mathrm{occ}}$
&
$G^{\mathrm{causal}}$
&
PC1
&
Top-3
&
$k_{0.9}$
\\

\midrule
\familyheader{GemmaBand}{\GemmaLogo\quad\textbf{Gemma family}}{15}
Gemma-2-2B & 32 & 6.85 & 0.98 & 0.29 & 0.38 & 0.76 & 13 & 75 & 7.79 & 0.52 & 0.11 & 0.18 & 0.41 & 19 \\
Gemma-2-2B-it & 28 & 8.09 & 0.72 & 0.28 & 0.35 & 0.73 & 15 & 42 & 7.13 & 0.51 & 0.05 & 0.25 & 0.51 & 14 \\
Gemma-3-4B-it & 26 & 7.84 & 0.75 & 0.22 & 0.39 & 0.75 & 13 & 54 & 8.42 & 0.51 & 0.47 & 0.31 & 0.68 & 7 \\
Gemma-2-9B & 31 & 6.89 & 0.85 & 0.20 & 0.60 & 0.85 & 16 & 51 & 8.61 & 0.54 & 0.17 & 0.20 & 0.53 & 10 \\
Gemma-2-9B-it & 26 & 7.77 & 0.78 & 0.08 & 0.38 & 0.79 & 14 & 54 & 9.05 & 0.54 & 0.01 & 0.32 & 0.64 & 7 \\
Gemma-2-27B & 38 & 8.84 & 0.66 & 0.13 & 0.40 & 0.81 & 14 & 55 & 7.19 & 0.71 & 0.14 & 0.41 & 0.79 & 4 \\
Gemma-2-27B-it & 34 & 6.98 & 0.83 & 0.16 & 0.65 & 0.87 & 13 & 53 & 7.65 & 0.66 & 0.12 & 0.66 & 0.94 & 2 \\

\familyheader{QwenBand}{\QwenLogo\quad\textbf{Qwen family}}{15}
Qwen2.5-0.5B & 48 & 7.55 & 0.70 & 0.21 & 0.37 & 0.73 & 26 & 83 & 6.38 & 0.58 & 0.14 & 0.60 & 0.88 & 4 \\
Qwen2.5-0.5B-Instruct & 48 & 7.16 & 0.73 & 0.60 & 0.50 & 0.76 & 25 & 78 & 7.10 & 0.59 & 0.14 & 0.67 & 0.85 & 6 \\
Qwen2.5-7B & 37 & 6.94 & 0.81 & 0.26 & 0.39 & 0.73 & 16 & 69 & 7.84 & 0.69 & 0.04 & 0.77 & 0.96 & 3 \\
Qwen2.5-7B-Instruct & 42 & 8.04 & 0.78 & 0.21 & 0.59 & 0.79 & 21 & 61 & 7.40 & 0.76 & 0.04 & 0.86 & 0.96 & 2 \\
Qwen3.5-27B & 45 & 10.33 & 0.51 & 0.08 & 0.51 & 0.82 & 22 & 101 & 9.08 & 0.61 & 0.02 & 0.30 & 0.58 & 22 \\
Qwen2.5-32B & 61 & 6.82 & 0.90 & 0.09 & 0.44 & 0.74 & 27 & 75 & 10.26 & 0.48 & 0.03 & 0.80 & 0.90 & 3 \\
Qwen2.5-32B-Instruct & 44 & 6.34 & 1.06 & 0.21 & 0.53 & 0.77 & 20 & 62 & 9.91 & 0.53 & 0.02 & 0.86 & 0.96 & 2 \\

\familyheader{MistralBand}{\MistralLogo\quad\textbf{Mistral family}}{15}
Mistral-7B & 56 & 10.39 & 0.61 & 0.17 & 0.45 & 0.79 & 25 & 46 & 5.99 & 0.74 & 0.08 & 0.35 & 0.61 & 15 \\
Mistral-7B-Instruct & 32 & 8.32 & 0.68 & 0.25 & 0.37 & 0.67 & 18 & 50 & 8.35 & 0.61 & 0.07 & 0.33 & 0.57 & 13 \\
Mistral-Nemo-Base & 43 & 5.61 & 1.07 & 0.08 & 0.89 & 0.94 & 21 & 78 & 7.17 & 0.66 & 0.03 & 0.51 & 0.68 & 12 \\
Mistral-Nemo-Instruct & 42 & 4.68 & 1.23 & 0.10 & 0.47 & 0.69 & 22 & 67 & 7.27 & 0.64 & 0.13 & 0.68 & 0.82 & 6 \\
Mistral-Small-3.1-24B-Base & 28 & 8.93 & 0.60 & 0.11 & 0.61 & 0.93 & 12 & 61 & 6.81 & 0.73 & 0.29 & 0.35 & 0.74 & 8 \\
Mistral-Small-3.1-24B-Instruct & 34 & 7.88 & 0.68 & 0.30 & 0.57 & 0.90 & 13 & 52 & 5.61 & 0.86 & 0.24 & 0.36 & 0.71 & 9 \\

\familyheader{LlamaBand}{\LlamaLogo\quad\textbf{Llama family}}{15}
Llama-3.1-8B & 26 & 3.63 & 1.12 & 0.09 & 0.75 & 0.88 & 9 & 92 & 7.88 & 0.59 & 0.10 & 0.43 & 0.77 & 8 \\
Llama-3.1-8B-Instruct & 20 & 4.38 & 1.18 & 0.18 & 0.77 & 0.88 & 8 & 78 & 7.73 & 0.56 & 0.25 & 0.26 & 0.50 & 20 \\
Llama-3.3-70B-Instruct & 31 & 4.51 & 1.11 & 0.11 & 0.84 & 0.92 & 9 & 83 & 9.17 & 0.57 & 0.06 & 0.41 & 0.69 & 15 \\
\textbf{Mean} & 37.04 & 7.16 & 0.84 & 0.19 & 0.53 & 0.80 & 17.04 & 66.09 & 7.82 & 0.62 & 0.12 & 0.47 & 0.73 & 9.17 \\
\textbf{Median} & 34 & 7.16 & 0.78 & 0.18 & 0.50 & 0.79 & 16 & 62 & 7.73 & 0.59 & 0.10 & 0.41 & 0.71 & 8 \\
\bottomrule
\end{tabular}
\end{adjustbox}

\caption{
Per-model Behavioral Anchor Coordinate and ACT-space behavioral-chart
statistics for repetition and sycophancy.
\(\#\BAC\) is the number of selected Behavioral Anchor Coordinates.
\(G^{\mathrm{uniF}}\) measures reliability-weighted behavioral separability;
\(G^{\mathrm{occ}}\) is the behavior-positive minus behavior-negative
projected second-moment gap; and \(G^{\mathrm{causal}}\) is the local
sensitivity of the behavioral logit to intervention strength.
PC1 and Top-3 are the fractions of variance explained by the first principal
component and the first three components, respectively, while \(k_{0.9}\)
is the minimum number of components explaining \(90\%\) of the variance.
Complete NOC-space statistics are reported in
Table~\ref{tab:app_all_bac_statistics_noc} of
Appendix~\ref{app:full_bac_statistics}.
}
\label{tab:app_all_bac_statistics_act}
\end{table}

\subsection{Behavioral Anchor Coordinates Support Sparse, Low-Dimensional Charts} \label{sec:behavioral_anchor_results} 
Table~\ref{tab:app_all_bac_statistics_act} summarizes the ACT-space coordinate scaffolds and chart statistics for repetition and sycophancy across 23 models from the Gemma, Qwen, Mistral, and Llama families. The corresponding NOC-space results are reported in Table~\ref{tab:app_all_bac_statistics_noc} of Appendix~\ref{app:full_bac_statistics}. Layer-resolved visualizations of the repetition and sycophancy scaffolds in \textit{Llama-3.1-8B} are provided in Appendix~\ref{app:bac_layer_distribution}. Detailed information about datasets is provided in Appendix~\ref{dataset}.
\paragraph{Repetition.} Repetition is supported by a sparse coordinate scaffold and an even lower-dimensional behavioral chart. Across the 23 models, the selected scaffold contains only $20$--$61$ \BACs{}. The initial held-out probe further confirms that this sparse scaffold is behavior-predictive, achieving on-policy repetition accuracies of $0.9658$--$0.9879$. Mean Universal Fisher Strength is $7.16$ in ACT space and $7.02$ in NOC space, while the cross-family interventions in Section~\ref{sec:steering-results} increase repetition for every evaluated model under both ACT-chart subspace steering and NOC-selected BAC steering. 
\paragraph{Sycophancy.} Sycophancy uses a broader coordinate scaffold, ranging from $42$ to $101$ selected \BACs{} with a median of $62$, but its ACT-space runtime geometry is strongly compressed within this scaffold. The NOC representation is less strongly compressed (Table~\ref{tab:app_all_bac_statistics_noc}). Sycophancy is concentrated into a compact runtime activation state while its functional contribution structure remains distributed over a broader set of directions. Complete per-model statistics are reported in Appendix~\ref{app:full_bac_statistics}.

Hidden-unit identities are not directly comparable across architectures.
We therefore compare each model only after expressing it in its own local
behavioral chart. We evaluate the 23-model sycophancy atlas using three tests:
(i) absolute correlation between held-out sparse-classifier scores; (ii) pairwise CCA; and
(iii) held-out linear probing. Full definitions, model identities, pairwise matrices, and controls are provided in Appendix~\ref{app:cross-model-atlas-results}. Table~\ref{tab:atlas-and-training-summary} consolidates the aggregate atlas results, including the random-coordinate controls and the relationship-stratified analyses reported separately in Appendix~\ref{app:cross-model-atlas-results}. Most model pairs contain a strongly transferable leading sycophancy coordinate, especially in NOC.

\subsection{Steering Establishes Causal Relevance}
\label{sec:steering-results}

\begin{figure}[t]
    \centering

    \includegraphics[
        width=0.6\linewidth,
        keepaspectratio
    ]{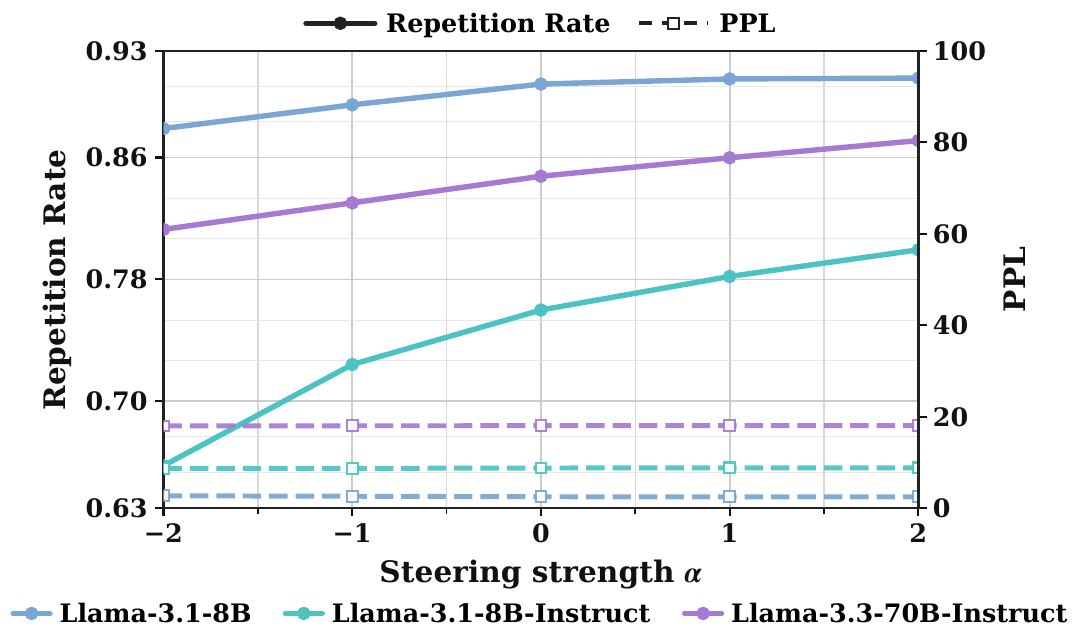}

    \caption{
    Causal repetition steering in the Llama family. Solid lines with circular
    markers report repetition rate on the left axis; dashed lines with square
    markers report generated-token PPL on the right axis.
    }
    \label{fig:repetition-steering-main}
\end{figure}

We intervene directly on the repetition representations and measure the resulting dose-response curves. We evaluate two intervention levels across 23 models. Figure~\ref{fig:repetition-steering-main} shows the Llama-family results.

These findings establish causal relevance at two resolutions. Editing the
selected coordinates changes repetition, showing that the coordinate scaffold
contains functionally relevant units. Editing the low-dimensional component
supported by those coordinates also changes repetition, showing that the
distributed chart is not merely a predictive compression.  At high strengths,
several curves saturate or mildly reverse, consistent with the chart being a
local approximation. We therefore use local finite-difference gain around the no-intervention point as the primary causal-strength estimator. The full repetition curves and family-level analysis are reported
in Appendix~\ref{app:cross-family-steering}. The complete
strength landscape, paired-comparison protocol, figures, pair-level values,
and limitations are reported in
Appendix~\ref{app:strength-alignment-analysis}.

%% file: sections__06_results_training.tex
\section{Rewrites or Reweights}
\label{sec:training_results}

Public Base--Instruct comparisons confound the objective with data mixture,
training duration, and implementation details
(Appendix~\ref{app:strength-alignment-analysis}).  We therefore study controlled
Llama-3.1-8B trajectories in which the base checkpoint, behavioral domain, data
construction, and evaluation instances are fixed.  For repetition and
sycophancy, we compare supervised fine-tuning (SFT), DPO initialized from the
Base model, and DPO initialized from the matched-direction SFT checkpoint.  The
question is whether post-training changes behavior by moving the behavioral
chart itself or by changing how an inherited chart is occupied or read out. Table~\ref{tab:atlas-and-training-summary} consolidates the two behavioral
probes. At fixed matched rank, projection distance is determined by
projector overlap,
\(
 d_{\mathrm{proj}}=\sqrt{r^\star(1-S)}
\),
while native Fisher strength is task-scale dependent. Complete metrics are reported in Appendix~\ref{app:posttraining}.

\begin{table}[t]
\centering
\small

\begin{minipage}[t]{0.535\columnwidth}
\centering
\textbf{(a) Cross-model atlas}

\begin{adjustbox}{max width=\linewidth,center}
\begin{tabular}{@{}llrrrrr@{}}
\toprule
\multirow{2}{*}{Metric}
&
\multirow{2}{*}{Sp.}
&
\multicolumn{2}{c}{All}
&
\multicolumn{3}{c}{BAC subsets}
\\
\cmidrule(lr){3-4}
\cmidrule(lr){5-7}
&
&
BAC
&
Rand.
&
Same
&
Cross
&
B--I
\\
\midrule

Cls.\ corr.
& --  & 0.950 & --    & --    & --    & 0.980 \\

\midrule

\multirow{2}{*}{Primary CCA}
& ACT & 0.938 & 0.367 & --    & --    & --    \\
& NOC & 0.974 & 0.317 & --    & --    & --    \\

\multirow{2}{*}{Mean CCA}
& ACT & 0.515 & 0.222 & 0.627 & 0.480 & 0.754 \\
& NOC & 0.648 & 0.233 & 0.677 & 0.639 & 0.798 \\

\multirow{2}{*}{Top-2-PC}
& ACT & 0.784 & 0.266 & 0.816 & 0.775 & 0.905 \\
& NOC & 0.958 & 0.295 & 0.982 & 0.950 & 0.995 \\

\multirow{2}{*}{$k$-dim.}
& ACT & 0.473 & 0.215 & 0.582 & 0.439 & 0.718 \\
& NOC & 0.581 & 0.234 & 0.634 & 0.565 & 0.775 \\

\bottomrule
\end{tabular}
\end{adjustbox}
\end{minipage}
\hfill
\begin{minipage}[t]{0.445\columnwidth}
\centering
\textbf{(b) Training geometry}

\begin{adjustbox}{max width=\linewidth,center}
\begin{tabular}{@{}llcccr@{}}
\toprule
Obj. & Dir. & Geom. & \(R^G\) & Behav. & \(\mathrm{PPL}_{-}\) \\
\midrule

\multicolumn{6}{@{}l}{
    \textit{Repetition}
    \;(\(d_{\mathrm{union}}=26\))
} \\

SFT
& up
& 0.693/27.78
& 0.938
& 0.968/0.723
& 3.70
\\

SFT
& down
& 0.711/25.77
& 0.942
& 0.920/0.372
& 3.52
\\

DPO
& up
& \textbf{0.960/3.59}
& 1.096
& 0.918/0.403
& 2.95
\\

DPO
& down
& \textbf{0.920/7.20}
& 1.049
& 0.885/0.434
& 2.91
\\

SFT$\to$DPO
& up
& 0.598/36.05
& 0.900
& 0.967/0.942
& 5.00
\\

SFT$\to$DPO
& down
& 0.604/35.17
& 0.705
& 0.596/0.565
& 3.87
\\

\midrule

\multicolumn{6}{@{}l}{
    \textit{Sycophancy}
    \;(\(d_{\mathrm{union}}=92\))
} \\

SFT
& up
& 0.773/24.69
& 0.966
& 0.886/0.646
& 11.72
\\

SFT
& down
& 0.804/22.04
& 0.814
& 0.442/0.330
& 1.28
\\

DPO
& up
& \textbf{0.938/10.10}
& 0.998
& 0.370/0.474
& 6.48
\\

DPO
& down
& \textbf{0.939/9.68}
& 0.995
& 0.342/0.436
& 5.69
\\

SFT$\to$DPO
& up
& 0.721/29.03
& 0.885
& 0.851/0.666
& 1.48
\\

SFT$\to$DPO
& down
& 0.695/30.88
& 0.819
& 0.389/0.297
& 1.38
\\

\bottomrule
\end{tabular}
\end{adjustbox}
\end{minipage}

\caption{
\textbf{Cross-model atlas consistency and training-induced behavioral
geometry.}
\textbf{(a)} Cross-model sycophancy-atlas results over 23 models.
Classifier correlation provides a supervised consistency check; CCA
measures chart alignment; and the Top-2-PC and \(k\)-dimensional probes
measure held-out linear transfer. BAC and Rand.\ denote
behavior-associated and matched random coordinates, while Same, Cross,
and B--I denote within-family, cross-family, and matched
base--instruction pairs.
\textbf{(b)} Geometric and behavioral changes induced by SFT, DPO, and
SFT$\to$DPO in behavior-increasing (\emph{up}) and
behavior-decreasing (\emph{down}) directions. Bold values indicate the
strongest geometric preservation. Full definitions and results are
provided in Appendix~\ref{app:cross-model-atlas-results}.
}
\label{tab:atlas-and-training-summary}
\end{table}

\begin{figure}[t]
    \centering
    \captionsetup[subfigure]{
        font=small,
        justification=centering,
        singlelinecheck=false,
        skip=1.5pt
    }

    \begin{subfigure}[t]{\linewidth}
        \centering
        \includegraphics[width=0.8\linewidth]
        {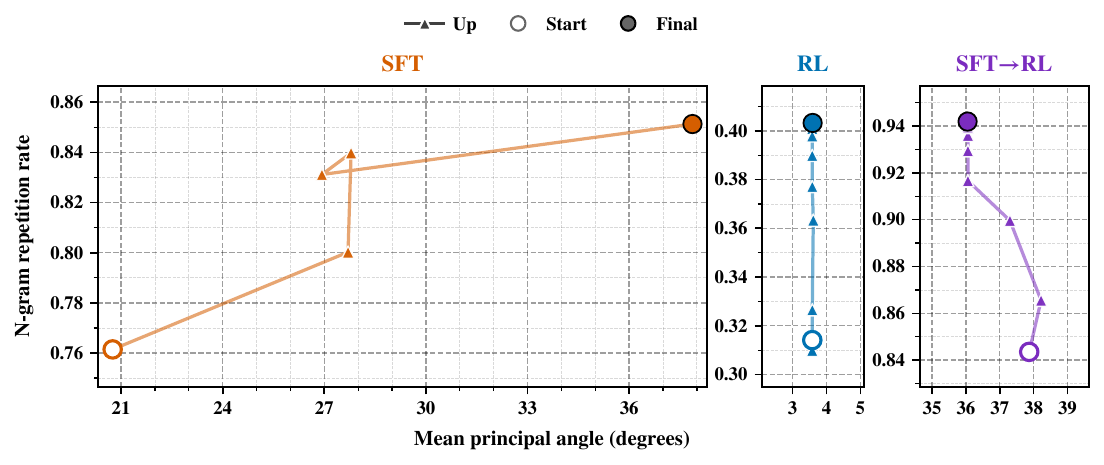}
        \caption{Repetition, up direction: geometry--behavior phase portrait.}
        \label{fig:repetition_geometry_behavior}
    \end{subfigure}

    \begin{subfigure}[t]{\linewidth}
        \centering
        \includegraphics[width=0.8\linewidth]
        {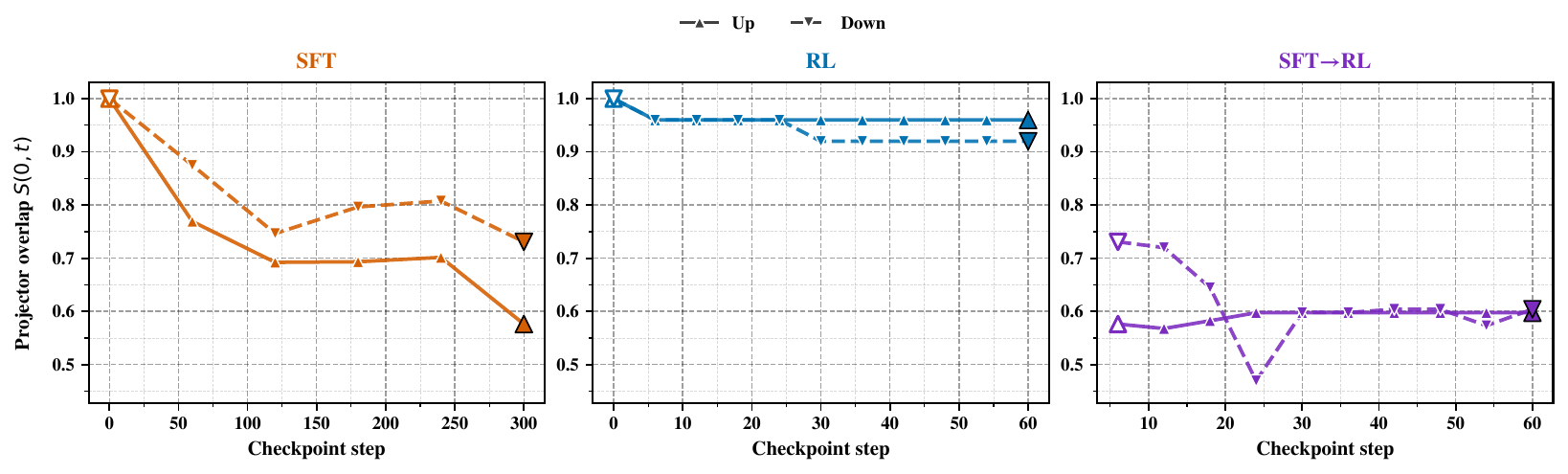}
        \caption{Repetition, both directions: Base-chart retention.}
        \label{fig:repetition_base_retention}
    \end{subfigure}

    \begin{subfigure}[t]{\linewidth}
        \centering
        \includegraphics[width=0.8\linewidth]
        {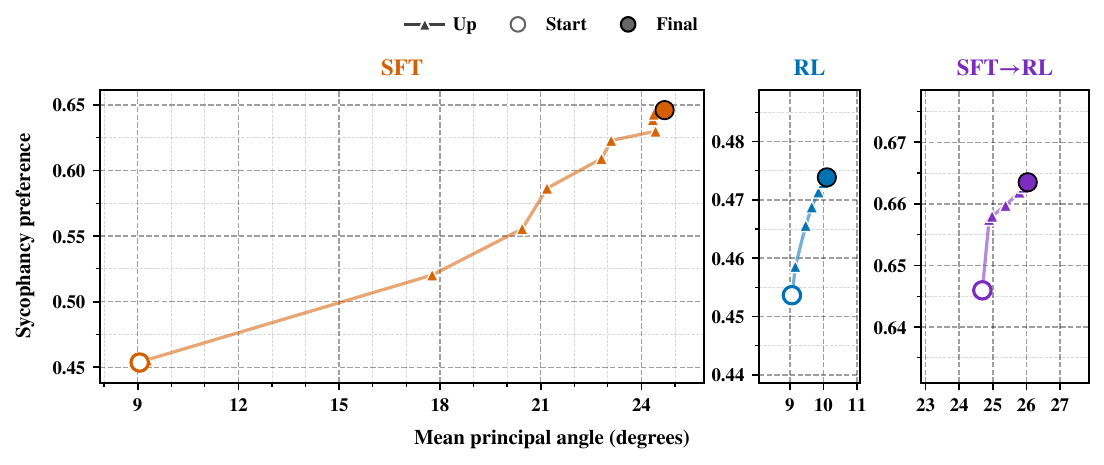}
        \caption{Sycophancy, up direction: geometry--behavior phase portrait.}
        \label{fig:sycophancy_geometry_behavior}
    \end{subfigure}

    \begin{subfigure}[t]{\linewidth}
        \centering
        \includegraphics[width=0.8\linewidth]
        {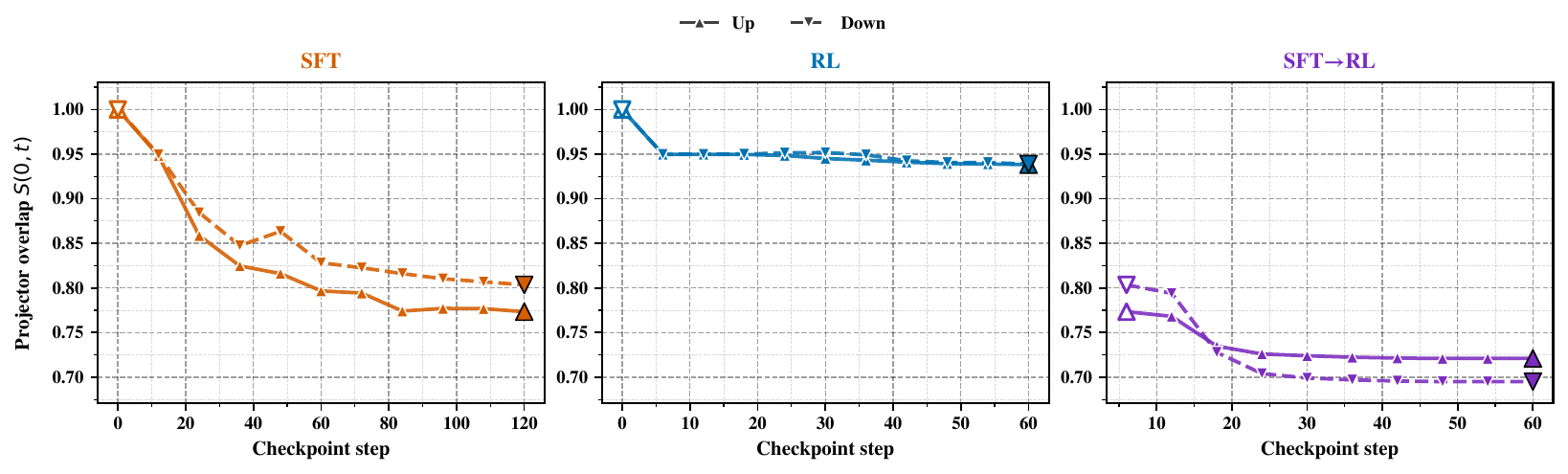}
        \caption{Sycophancy, both directions: Base-chart retention.}
        \label{fig:sycophancy_base_retention}
    \end{subfigure}

    \caption{
    \textbf{Behavioral geometry through controlled post-training.}
    Connected markers follow stored checkpoints in training order.
    In the phase portraits, open and filled circles denote the first stored
    post-training checkpoint and the reported endpoint, respectively; the Base
    model itself would lie at zero angular displacement. In the retention
    panels, \(S(0,t)\) compares each checkpoint chart with the Base chart.
    }
    \label{fig:controlled_posttraining_trajectories}
\end{figure}

Across the four behavior--direction settings, DPO initialized from Base is
consistently geometry-conservative, retaining mean projector overlap $0.939$
with mean angle $7.64^\circ$.  SFT produces substantially larger chart
displacement, with mean overlap $0.745$ and mean angle $25.07^\circ$.
DPO after SFT does not restore the Base chart, indicating that the
sequential objective continues optimization within the SFT-induced geometry. Figure~\ref{fig:controlled_posttraining_trajectories} combines two views. The phase portraits plot behavior
against the mean principal angle from the Base chart, and the retention panels track \(S(0,t)\) throughout optimization.

The trajectories reinforce the endpoint comparison.  DPO-only overlap remains
high throughout training, while SFT rapidly enters a lower-overlap regime.
Sequential DPO starts from an already displaced SFT chart and remains far from
Base. The trajectory view rules out an explanation in
which all objectives initially rewrite geometry but DPO later returns to the
Base chart. The endpoint and trajectory results identify geometric conservation versus
movement, but they do not by themselves prove which non-geometric mechanism
causes the remaining behavioral change.  Appendix~\ref{app:gog-full} therefore
reports geometry--occupancy--gain profiles at final checkpoints, across ten
matched training stages, and in both ACT and NOC spaces.  These profiles show a stable objective-level distinction in geometry, while occupancy and causal gain are more condition-, representation-, and estimator-dependent. A matched token-level comparison of truthful and sycophantic candidates is provided in Appendix~\ref{app:base-anchored-sycophancy-paths}.  The complete data provenance, direction checks, and training configurations are reported in Appendix~\ref{sec:post-training-config}. Public checkpoint chains remain useful as ecological validation (Appendix~\ref{app:strength-alignment-analysis}).


%% file: sections__08_discussion.tex
\section{Conclusion}
\label{sec:discussion}

SFT displaced these structures farther from the base model than RL, whereas RL initialized from an SFT checkpoint largely remained within the SFT-shifted geometric regime. SFT can more readily reorganize behavior-associated representations, while DPO may often modify behavior while preserving more of the inherited representational scaffold.

%% file: sections__supplement__h_model_theory.tex
\section{Derivation of the Local Behavioral Model}
\label{app:master-formula}

This appendix derives the local behavioral model used in Section~\ref{sec:framework}.  The purpose of the model is not to assert that a language model implements a globally linear behavioral classifier.  Rather, it provides a local surrogate for the behavior log-odds in the neighborhood of the observed behavior-relevant representations.  This surrogate justifies the decomposition used in the main text: post-training may change a behavior by moving the behavior chart, by changing the distribution of states inside that chart, or by changing the readout from chart coordinates to behavior probability.

\subsection{Binary Behaviors and Their Log-Odds}

Fix a model $f_\theta$ and a behavior $b$.  Let
\[
    Y_b\in\{0,1\}
\]
be the binary random variable indicating whether behavior $b$ is expressed on an instance $x$.  Let
\[
    h_\theta(x)=\bar\phi_\theta(x)\in\mathbb{R}^D
\]
denote the standardized internal representation used for analysis.  This representation may correspond to ACT, NOC, or another feature space, but the derivation below is independent of the particular choice.

Assume that the conditional behavior probability satisfies
\[
    0<p_\theta(Y_b=1\mid h)<1
\]
on the representation region under consideration.  We may then define the behavior log-odds function
\begin{equation}
\label{eq:app_log_odds}
    g_{\theta,b}(h)
    =
    \log
    \frac{
    p_\theta(Y_b=1\mid h)
    }{
    p_\theta(Y_b=0\mid h)
    }.
\end{equation}
This definition is exact.  Since
\[
    p_\theta(Y_b=0\mid h)=1-p_\theta(Y_b=1\mid h),
\]
Equation~\ref{eq:app_log_odds} implies
\begin{equation}
\label{eq:app_sigmoid_exact}
    p_\theta(Y_b=1\mid h)
    =
    \sigma\!\left(g_{\theta,b}(h)\right),
\end{equation}
where $\sigma(u)=(1+\exp(-u))^{-1}$.  Thus the only approximation in the local behavioral model concerns the representation dependence of $g_{\theta,b}(h)$.

\subsection{Local Linearization of the Behavior Log-Odds}

Let $\mathcal{R}_{\theta,b}\subset\mathbb{R}^D$ denote a neighborhood containing the observed behavior-relevant representations.  Choose a reference point
\[
    h_\star\in\mathcal{R}_{\theta,b},
\]
for example the midpoint between the positive and negative class means.  Suppose that $g_{\theta,b}$ is twice continuously differentiable on $\mathcal{R}_{\theta,b}$ and that its Hessian is bounded:
\begin{equation}
\label{eq:app_hessian_bound}
    \sup_{h\in\mathcal{R}_{\theta,b}}
    \left\|
    \nabla^2 g_{\theta,b}(h)
    \right\|_2
    \leq L_{\theta,b}.
\end{equation}
A first-order Taylor expansion around $h_\star$ gives
\begin{equation}
\label{eq:app_taylor}
    g_{\theta,b}(h)
    =
    g_{\theta,b}(h_\star)
    +
    w_{\theta,b}^{\top}(h-h_\star)
    +
    R_{\theta,b}(h),
\end{equation}
where
\[
    w_{\theta,b}
    =
    \nabla g_{\theta,b}(h_\star)
\]
is the local behavior-sensitive direction in representation space, and the remainder satisfies
\begin{equation}
\label{eq:app_taylor_remainder}
    |R_{\theta,b}(h)|
    \leq
    \frac{L_{\theta,b}}{2}
    \|h-h_\star\|_2^2.
\end{equation}
Therefore, in a sufficiently small neighborhood of $h_\star$, the behavior log-odds are well approximated by an affine function of the internal representation.

\subsection{Low-Dimensional Behavioral Chart Assumption}

The geometric hypothesis of the paper is that behavior-relevant variation is locally concentrated in a low-dimensional chart.  Formally, let
\[
    U_{\theta,b}\in\mathbb{R}^{D\times k},
    \qquad
    U_{\theta,b}^{\top}U_{\theta,b}=I_k,
\]
be an orthonormal basis for a $k$-dimensional behavioral chart, and let
\[
    P_{\theta,b}
    =
    U_{\theta,b}U_{\theta,b}^{\top}
\]
be the corresponding orthogonal projector.  The chart is useful when the local behavior-sensitive direction $w_{\theta,b}$ is mostly contained in the subspace spanned by $U_{\theta,b}$.  We quantify the residual sensitivity outside the chart by
\begin{equation}
\label{eq:app_subspace_residual}
    \delta_{\theta,b}(U)
    =
    \left\|
    (I-P_{\theta,b})w_{\theta,b}
    \right\|_2.
\end{equation}
Define the chart-level readout vector
\begin{equation}
\label{eq:app_beta_definition}
    \beta_{\theta,b}
    =
    U_{\theta,b}^{\top}w_{\theta,b}.
\end{equation}
Then
\[
    P_{\theta,b}w_{\theta,b}
    =
    U_{\theta,b}\beta_{\theta,b},
\]
and
\[
    w_{\theta,b}
    =
    U_{\theta,b}\beta_{\theta,b}
    +
    (I-P_{\theta,b})w_{\theta,b}.
\]

Substituting this decomposition into Equation~\ref{eq:app_taylor} yields
\begin{align}
    g_{\theta,b}(h)
    &=
    g_{\theta,b}(h_\star)
    +
    \beta_{\theta,b}^{\top}
    U_{\theta,b}^{\top}(h-h_\star)
    \nonumber\\
    &\quad
    +
    \left((I-P_{\theta,b})w_{\theta,b}\right)^{\top}
    (h-h_\star)
    +
    R_{\theta,b}(h).
\end{align}
Define the intercept
\begin{equation}
\label{eq:app_tau_definition}
    \tau_{\theta,b}
    =
    g_{\theta,b}(h_\star)
    -
    \beta_{\theta,b}^{\top}
    U_{\theta,b}^{\top}h_\star.
\end{equation}
Then
\begin{equation}
\label{eq:app_low_dim_logit_exact}
    g_{\theta,b}(h)
    =
    \beta_{\theta,b}^{\top}
    U_{\theta,b}^{\top}h
    +
    \tau_{\theta,b}
    +
    \varepsilon_{\theta,b}(h),
\end{equation}
where
\begin{equation}
\label{eq:app_epsilon_definition}
    \varepsilon_{\theta,b}(h)
    =
    \left((I-P_{\theta,b})w_{\theta,b}\right)^{\top}
    (h-h_\star)
    +
    R_{\theta,b}(h).
\end{equation}
By Cauchy--Schwarz and Equation~\ref{eq:app_taylor_remainder},
\begin{equation}
\label{eq:app_error_bound}
    |\varepsilon_{\theta,b}(h)|
    \leq
    \delta_{\theta,b}(U)
    \|h-h_\star\|_2
    +
    \frac{L_{\theta,b}}{2}
    \|h-h_\star\|_2^2.
\end{equation}

Equation~\ref{eq:app_error_bound} makes explicit the two approximation requirements.  First, the behavior log-odds should be locally close to linear, which is controlled by the curvature bound $L_{\theta,b}$ and the radius of the neighborhood.  Second, the behavior-sensitive direction should be well captured by the chart $U_{\theta,b}$, which is controlled by $\delta_{\theta,b}(U)$.  The empirical procedures in the paper are designed to test these requirements: low-rank PCA structure tests whether behavior-relevant variation is concentrated; logistic and Fisher directions test whether discriminative directions align with the chart; causal interventions test whether the chart has behavioral effect.

Since the sigmoid is $1/4$-Lipschitz,
\[
    |\sigma(a)-\sigma(a')|
    \leq
    \frac{1}{4}|a-a'|,
\]
the logit error in Equation~\ref{eq:app_error_bound} also bounds the induced probability error:
\begin{equation}
\label{eq:app_probability_error}
    \left|
    p_\theta(Y_b=1\mid h)
    -
    \sigma\!\left(
    \beta_{\theta,b}^{\top}
    U_{\theta,b}^{\top}h
    +
    \tau_{\theta,b}
    \right)
    \right|
    \leq
    \frac{1}{4}
    |\varepsilon_{\theta,b}(h)|.
\end{equation}

Dropping the residual term gives the local behavioral model used in the main text:
\begin{equation}
\label{eq:app_master_formula}
    p_\theta(Y_b=1\mid x)
    \approx
    \sigma\!\left(
    \beta_{\theta,b}^{\top}
    U_{\theta,b}^{\top}
    \bar\phi_\theta(x)
    +
    \tau_{\theta,b}
    \right).
\end{equation}

\subsection{Equivalent Statistical Derivation from a Local Gaussian Model}

The same form also arises from a standard local generative model in the behavioral chart.  Let
\[
    z
    =
    U_{\theta,b}^{\top}h
\]
be the chart coordinate.  Suppose that, within the local chart, the class-conditional distributions are approximately Gaussian with shared covariance:
\begin{align}
    z\mid Y_b=1
    &\sim
    \mathcal{N}(\mu_1,\Sigma),\\
    z\mid Y_b=0
    &\sim
    \mathcal{N}(\mu_0,\Sigma).
\end{align}
Let
\[
    \pi_1=p(Y_b=1),
    \qquad
    \pi_0=p(Y_b=0)
\]
be the class priors.  Bayes' rule gives
\begin{align}
    \log
    \frac{
    p(Y_b=1\mid z)
    }{
    p(Y_b=0\mid z)
    }
    &=
    \log
    \frac{
    p(z\mid Y_b=1)\pi_1
    }{
    p(z\mid Y_b=0)\pi_0
    }.
\end{align}
Substituting the Gaussian densities and canceling the quadratic terms yields
\begin{equation}
\label{eq:app_lda_logit}
    \log
    \frac{
    p(Y_b=1\mid z)
    }{
    p(Y_b=0\mid z)
    }
    =
    \beta^\top z+\tau,
\end{equation}
where
\begin{equation}
\label{eq:app_lda_beta}
    \beta
    =
    \Sigma^{-1}(\mu_1-\mu_0),
\end{equation}
and
\begin{equation}
\label{eq:app_lda_tau}
    \tau
    =
    \log\frac{\pi_1}{\pi_0}
    -
    \frac{1}{2}
    \left(
    \mu_1^\top\Sigma^{-1}\mu_1
    -
    \mu_0^\top\Sigma^{-1}\mu_0
    \right).
\end{equation}
Thus
\[
    p(Y_b=1\mid z)
    =
    \sigma(\beta^\top z+\tau),
\]
and, since $z=U_{\theta,b}^{\top}h$, this again gives the form of Equation~\ref{eq:app_master_formula}.  This derivation clarifies why Fisher-style directions and Fisher separability are natural diagnostics: under the shared-covariance local Gaussian approximation, the optimal linear discriminative direction is
\[
    \Sigma^{-1}(\mu_1-\mu_0).
\]

\subsection{Consequences for the Geometry--Occupancy--Readout Decomposition}

Equation~\ref{eq:app_master_formula} separates behavioral change into three conceptually distinct components.

First, the subspace $U_{\theta,b}$ specifies where the behavior is represented.  Changes in $U_{\theta,b}$ correspond to movement of the behavioral chart itself.  This motivates projector overlap, principal angles, projection distance, orthogonal novelty, and anchored Fisher retention as measures of geometry preservation or chart rewriting.

Second, the projected coordinate
\[
    z_{\theta,b}(x)
    =
    U_{\theta,b}^{\top}\bar\phi_\theta(x)
\]
specifies how strongly a particular internal state occupies the behavioral chart.  Distributional changes in $z_{\theta,b}(x)$ correspond to occupancy changes.  The energy-based occupancy statistic used in the main text,
\[
    G^{\mathrm{occ}}
    =
    \mathbb{E}[\|z\|_2^2\mid Y_b=1]
    -
    \mathbb{E}[\|z\|_2^2\mid Y_b=0],
\]
is a rotation-invariant second-moment proxy for class-conditional chart occupancy.  More directional variants, such as differences along Fisher or logistic directions, estimate signed movement along the behavior-positive axis.

Third, the parameters $(\beta_{\theta,b},\tau_{\theta,b})$ specify the readout from chart coordinates to behavior probability.  The vector $\beta_{\theta,b}$ controls local slope or gain, while $\tau_{\theta,b}$ acts as an intercept or threshold.  If an intervention of strength $\alpha$ induces a chart displacement $z_\alpha$, then
\begin{equation}
\label{eq:app_causal_gain_derivation}
    \frac{\partial}{\partial \alpha}
    \mathrm{logit}\,
    p_\theta(Y_b=1\mid z_\alpha)
    =
    \beta_{\theta,b}^{\top}
    \frac{\partial z_\alpha}{\partial \alpha}.
\end{equation}
Therefore the finite-difference causal gain estimates the effective readout slope along the intervention-induced chart direction.  The intercept $\tau_{\theta,b}$ cancels under differentiation and should be estimated separately, for example by the fitted logistic intercept or by baseline logit after controlling for chart coordinates.

This derivation explains why the paper distinguishes geometry, occupancy, and readout.  Two post-training objectives may yield similar behavioral scores while acting on different terms of Equation~\ref{eq:app_master_formula}: one may rotate $U_{\theta,b}$, another may shift the distribution of $U_{\theta,b}^{\top}\bar\phi_\theta(x)$, and a third may alter the readout from chart coordinates to output behavior.

%% file: sections__supplement__b_data_and_features.tex
\section{Repetition Detection and Feature Extraction}
\label{app:data_features}
\subsection{Periodic \(n\)-gram Repetition Detection}
Let the generated token sequence be $x_{1:T}$.  For each position $i$, the detector forms an $n$-gram $g_i=(x_i,\ldots,x_{i+n-1})$ with $n=10$.  It checks whether this $n$-gram occurs at least $k=3$ times within the previous $r=100$ tokens.  Among the first three occurrences, the detector further requires equal spacing:
\begin{equation}
    p_2-p_1=p_3-p_2.
\end{equation}
A sample is retained only if the second occurrence begins after at least 50 non-repetitive prefix tokens.  This ensures that the detected repetition follows an initially non-repetitive prefix.

The stored JSONL record includes the prompt, generated sequence, repeated
$n$-gram, detected positions, and local windows around the repetition onset, which define the repetition-relevant regions used for feature extraction.

\subsection{Token Regions}
Feature extraction is region-aware.  The extraction script first constructs either a chat-template input or a fallback plain text input of the form \texttt{Question: ... Answer: ...}.  It then identifies input, output, answer-token, and non-answer-token regions.  For repetition and sycophancy, the behavior-relevant token span is used for mean pooling.  Samples with an unresolved required span are omitted from the corresponding regional analysis.

\subsection{Raw Activation Space}
For each selected token region, raw activations are the inputs to the MLP down-projection layers.  If $h_{\ell,t}\in\R^{d_{ff}}$ denotes the input to \texttt{down\_proj} at layer $\ell$ and token $t$, ACT features are mean-pooled over the designated token region, after which the selected coordinates are retained.

\subsection{Normalized Output-Contribution Magnitude}
For the experiments reported in this work, the directional NOC concept
introduced in Eq.~\eqref{eq:noc-token} is operationalized using the following
non-negative magnitude proxy:
\begin{equation}
    c_{\ell,t,j}
    =\frac{|h_{\ell,t,j}|\,\|W^{\mathrm{down}}_{\ell,:,j}\|_2}{\|o_{\ell,t}\|_2+10^{-8}},
\end{equation}
where $W^{\mathrm{down}}_{\ell,:,j}$ is the outgoing column of the down-projection matrix and $o_{\ell,t}$ is the output of the down-projection layer.  This normalization reduces sensitivity to raw activation and output scale. The implemented proxy is non-negative and does not preserve the sign or direction of an individual coordinate's contribution.

\subsection{Implementation Notes}
The hook registration matches modules whose names contain \texttt{down\_proj}, excluding vision modules when present.

%% file: sections__supplement__c_subspace_estimation.tex
\section{Neuron Selection, Subspace Estimation, and Cross-Model Alignment}
\label{app:subspace_estimation}

\subsection{Sparse Classifier}
For each behavior and model, we train a binary logistic regression classifier on flattened selected-region features.  The archived classifier runs predominantly use the \texttt{saga} solver,
with a small number of legacy runs using \texttt{liblinear}. The penalty is
$\ell_1$, with $C=1.0$, a maximum of $1000$ iterations, tolerance $10^{-3}$,
and random seed $42$.  The positive class is the behavior class: repetitive samples for repetition and sycophantic samples for sycophancy.  Coordinates whose classifier weights are positive are selected:
\begin{equation}
    R=\{j:w_j>0\}.
\end{equation}
The code also supports selecting all nonzero weights, but the reported main experiments use the positive-weight rule.

\subsection{Physical Index Recovery}
The classifier operates on flattened arrays.  For dense MLP models with reference shape $(L,d_{ff})$, a selected flat index $q$ is mapped back to layer and neuron by
\begin{equation}
    \ell=\left\lfloor \frac{q}{d_{ff}}\right\rfloor,
    \qquad
    j=q\bmod d_{ff}.
\end{equation}
The recovered mapping is stored as a JSON mapping from each layer index to its selected neuron indices.  This mapping is used by neuron steering, subspace steering, and qualitative generation experiments.

\subsection{PCA Behavioral Chart}
Let $X\in\R^{N\times |R|}$ be the feature matrix restricted to selected coordinates.  PCA is fit with up to $\min(N-1,|R|)$ components.  The implementation saves the centered projections
\(Z=(X-\mu)U\), the components \(U\), the PCA mean \(\mu\), the explained-variance ratios, the selected-neuron mask, the labels, and the logistic-regression direction.  The effective rank $k$ is
\begin{equation}
    k=\min\left\{m:\sum_{i=1}^{m}\mathrm{EVR}_i\ge0.9\right\}.
\end{equation}
All later geometry metrics use this $90\%$-variance cutoff unless otherwise stated.

\subsection{CCA Alignment}
For two models $a,b$, let $Z_a$ and $Z_b$ be their PCA projections truncated to $k_a$ and $k_b$ components.  CCA is fit with $q=\min(k_a,k_b)$ canonical components.  The first canonical correlation is reported as primary CCA, while the average over components is reported as mean CCA.  Because CCA directions have arbitrary signs, the code orients each component so that the positive class has the larger mean in both CCA views:
\begin{equation}
    \E[U_i\mid y=1] > \E[U_i\mid y=0],
    \quad
    \E[V_i\mid y=1] > \E[V_i\mid y=0].
\end{equation}
This orientation does not affect the canonical-correlation magnitudes, but it provides a consistent behavioral sign convention for visualization and cross-model aggregation of canonical coordinates.

\subsection{Classifier-Direction Alignment}
In addition to CCA, we align the sparse logistic-regression direction.  The classifier weight vector is projected into the PCA basis, and each sample is scored by its projection onto that direction.  The direction-correlation score between two models is the absolute Pearson correlation between their sample-level classifier-direction scores.

\subsection{Directed Linear Probing}
The linear-probing experiment asks whether model $A$'s behavioral chart can predict model $B$'s chart on matched samples.  In $k$-dimensional mode, the source and target use enough PCs to explain $90\%$ variance, and a linear regression fits $Z_A\mapsto Z_B$.  The score is the mean Pearson correlation between each true target dimension and its prediction.  The score is directed: $A\to B$ need not equal $B\to A$.

For sycophancy, where PC1 and PC2 can have similar variance, we also use a top-2-PC mode.  This searches over the four one-dimensional mappings PC1/PC2 of the source to PC1/PC2 of the target and reports the maximum signed Pearson correlation.  This avoids hard-coding which of the first two PCs carries the behavior-aligned direction.

%% file: sections__supplement__d_steering.tex
\section{Steering and Intervention Details}
\label{app:steering}

\subsection{Multiplicative Neuron Scaling}
The simplest intervention scales selected behavior-associated neurons at the input to each MLP down-projection layer.  If $R_\ell$ is the selected neuron set at layer $\ell$ and $h_{\ell,t}$ is the down-projection input, the intervention is
\begin{equation}
    h'_{\ell,t,j}=\alpha h_{\ell,t,j},\qquad j\in R_\ell.
\end{equation}
During likelihood-style evaluation on labeled spans, scaling is applied only to the behavior-relevant token region.  During generation, scaling is global over the current token because autoregressive decoding feeds one token at a time after prefill.

\subsection{Additive Subspace Steering}
For activation-space steering, we use the PCA chart itself.  Let $U_k$ be the first $k$ PCA components and $\mu$ the PCA mean over selected coordinates.  Given the current selected-neuron vector $x$, the projection onto the behavior chart is
\begin{equation}
    x_S = U_kU_k^\top(x-\mu).
\end{equation}
The steered vector is
\begin{equation}
    x' = x + \alpha x_S.
\end{equation}
Positive $\alpha$ amplifies the current component along the behavior chart; negative $\alpha$ subtracts it.

\subsection{Two-Pass Autoregressive Subspace Steering}
Generation-time subspace steering uses a two-pass decoding loop to avoid stale projections.  At each generated token:
\begin{enumerate}[leftmargin=1.1em,itemsep=1pt,topsep=2pt]
    \item A dry-run forward pass uses the current token and historical KV cache, registers hooks, and collects the selected-neuron vector $x$ for that token.
    \item The dry-run output is discarded and the KV cache is not updated.
    \item The projection $x_S=U_kU_k^\top(x-\mu)$ is computed once the full selected-neuron vector has been assembled across layers.
    \item A second forward pass is run with the same input token and same historical KV cache; hooks inject $\alpha x_S$ into the selected coordinates.
    \item The steered logits select the next token greedily, and only the steered pass updates the KV cache.
\end{enumerate}
This two-pass protocol is slower than ordinary generation but avoids using an intervention vector computed from an earlier hidden state.

\subsection{Greedy Challenge Prompts}
For qualitative repetition control, a manually designed challenge set contains prompts that strongly invite repetition but allow a semantically appropriate escape.  Baseline and steered outputs are generated greedily for 50 tokens.  Both neuron scaling and subspace steering are evaluated with model-specific \(\alpha\) values chosen from development sweeps.  Quantitative metrics are computed on generated tokens only, while prompt tokens are excluded from PPL.

\subsection{Generated-Text Metrics}
For a generated token sequence $x_{1:T}$, Rep-$n$ is
\begin{equation}
    \mathrm{Rep}\text{-}n=1-\frac{|\{(x_i,\ldots,x_{i+n-1})\}_{i=1}^{T-n+1}|}{T-n+1}.
\end{equation}
Normalized entropy is
\begin{equation}
    H_{norm}=\frac{-\sum_v p_v\log_2p_v}{\log_2 |\{v:p_v>0\}|}.
\end{equation}
Self-BLEU-1 is averaged over generations within each model--intervention setting, treating each generation as the hypothesis and the remaining
generations as references.  PPL is computed on the concatenated prompt and generation using the
corresponding model checkpoint without intervention hooks. Prompt labels are masked so that only generated tokens contribute to the loss.

%% file: sections__supplement__e_posttraining.tex
\section{Post-Training Geometry Metrics and Checkpoint Analysis}
\label{app:posttraining}

\subsection{Projector Metrics Without Forming Full Projectors}
Given two orthonormal bases $U_a\in\R^{d\times k_a}$ and $U_b\in\R^{d\times k_b}$, all geometry computations use the matched rank $r^\star=\min(k_a,k_b)$ and truncated bases.  After truncation to the first \(r^\star\) columns, we reuse \(U_a\) and \(U_b\) to denote the matched-rank bases. The implementation avoids constructing $d\times d$ projectors.  Let
\begin{equation}
    M=U_a^\top U_b\in\R^{r^\star\times r^\star}.
\end{equation}
Then
\begin{equation}
    S(a,b)=\frac{\|M\|_F^2}{r^\star},
\end{equation}
principal angles are obtained from singular values $\sigma_i(M)$, and projection distance is computed as
\begin{equation}
    d_{proj}(a,b)=\sqrt{r^\star-\|M\|_F^2}.
\end{equation}
This is equivalent to $2^{-1/2}\|P_a-P_b\|_F$ at matched rank.

\subsection{Anchored Fisher Retention}
For a base checkpoint $a$ and post-training checkpoint $b$, the native checkpoint-$b$ Fisher strength is computed in $b$'s own PCA coordinates.  The anchored Fisher strength uses the base chart on checkpoint-$b$ features.  The implementation loads post-checkpoint raw activations, applies the base selected-neuron mask, centers by the base PCA mean, and projects into the base PCA components:
\begin{equation}
    Z^{a\to b}=(X_b^{R_a}-\mu_a)U_a.
\end{equation}
Then
\begin{equation}
    R^G_{a\to b}=\frac{G(Z^{a\to b},y_b)}{G(Z^b,y_b)}.
\end{equation}
If $R^G$ is near one, the base chart remains almost as discriminative as the native chart; if it is small, the checkpoint has moved the behavior into new coordinates.

\subsection{Pairwise Universal Strength Aggregation}
For cross-model $G$ metrics, the code first extracts pairwise shared CCA coordinates.  For each pair file, it computes $G^{\mathrm{uniF}}$ and $G^{\mathrm{occ}}$ for both models using the same CCA reliability vector $\rho$.  Model-level metrics are then averaged over all pairwise appearances of that model.  This produces one row per model, space, and behavior.  Metadata such as parameter count, family, and instruction flag are appended for regression and visualization.

\subsection{Causal Gain From Steering Curves}
$G^{\mathrm{causal}}$ is estimated by centered finite differences on logit-transformed behavior probabilities.  For multiplicative neuron scaling, the no-intervention point is \(\alpha=1\). For additive subspace steering, the no-intervention point is \(\alpha=0\).  The sycophancy and repetition metric scripts default to an \(\alpha\)-step of $2.0$ when reading stored steering curves, while repetition strength summaries may use smaller local steps when available.  All causal-gain estimates should therefore be interpreted as local finite-difference sensitivities rather than exact derivatives.

\subsection{Controlled Training Pipelines}
The controlled training experiments use Llama-3.1-8B.  For each behavior, we train six trajectory types:
\begin{enumerate}[leftmargin=1.1em,itemsep=1pt,topsep=2pt]
    \item SFT-up: imitate behavior-positive outputs.
    \item SFT-down: imitate behavior-negative or repaired outputs.
    \item Reward-up: prefer behavior-positive outputs.
    \item Reward-down: prefer behavior-negative outputs.
    \item SFT+Reward-up: run reward-up after SFT-up.
    \item SFT+Reward-down: run reward-down after SFT-down.
\end{enumerate}
For sycophancy, the current reward runs are DPO-style preference optimization.  For repetition, some legacy scripts are named RL even when the actual run is DPO-style.  We therefore use \emph{reward optimization} as an umbrella term in the reported analyses, while noting that the audited optimization objective is DPO.

\subsection{End-to-End Fine-Tuning Analysis Pipeline}
Each checkpoint analysis executes the following sequence: extract activations for the base and checkpoint, train or reuse sparse classifiers, extract selected neurons, compute global padded indices, run PCA, zero-pad bases to a common ambient coordinate system, compute geometry metrics, evaluate behavior probabilities/PPL, and append behavior metrics to the geometry table.  This full sequence is orchestrated by the fine-tuning run scripts for repetition and sycophancy.

\subsection{Endpoint-Summary Reporting Conventions}
\label{app:endpoint-summary-conventions}

Table~\ref{tab:controlled_posttrain_summary} uses the following reporting
conventions. \emph{Dir.} indicates whether training enhances
(\emph{up}) or suppresses (\emph{down}) the target behavior.

The \emph{Geom.} entry reports
\(S(0,T)/\bar{\theta}(0,T)\), where \(S(0,T)\in[0,1]\) is the normalized
projector overlap between the Base and endpoint behavioral charts and
\(\bar{\theta}(0,T)\) is their mean principal angle in degrees.
The slash separates the two reported values and does not denote division.
Larger \(S(0,T)\) and smaller \(\bar{\theta}(0,T)\) indicate stronger
preservation of the Base chart. Anchored Fisher retention \(R^G\) is defined in the
\emph{Anchored Fisher Retention} subsection above.

The \emph{Behavior} entry reports \(p_b/m_b\), where \(p_b\) is the
evaluation probability assigned to the behavior-positive target and
\(m_b\) is the realized behavioral metric. For repetition, \(m_b\) is the
character 3-gram repetition rate; for sycophancy, it is the sycophantic
preference score.

The quantity \(\mathrm{PPL}_{-}\) denotes perplexity on the
behavior-negative target: non-repetitive responses for repetition and
truthful responses for sycophancy.

Within each behavior block, \(d_{\mathrm{union}}\) is the number of
distinct BAC/R-neuron coordinates selected across all audited checkpoints
and intervention directions. Bold geometry entries in
Table~\ref{tab:controlled_posttrain_summary} identify the DPO endpoints
with the strongest Base-chart preservation within the corresponding
behavior.

%% file: sections__supplement__i_Occupancy_Gain.tex
\section{Geometry-Occupancy-Gain Profiles of Controlled Post-Training}
\label{app:geometry-occupancy-gain-profiles}

\subsection{Derivation of Occupancy and Gain Estimators}
\label{app:occupancy_gain_derivation}

We derive the occupancy and gain estimators from the local behavioral model
defined in Eq.~(\ref{eq:local_behavior_model}):

\begin{equation}
\label{eq:appendix_local_model}
\mathrm{logit}\,p_\theta(y_b=1|x)
=
\beta_{\theta,b}^{\top}
U_{\theta,b}^{\top}
\bar\phi_\theta(x)
+
\tau_{\theta,b}.
\end{equation}

For simplicity, we denote

\begin{equation}
    z_\theta(x)
    =
    U_{\theta,b}^{\top}
    \bar\phi_\theta(x).
\end{equation}

The behavioral logit can therefore be written as

\begin{equation}
\label{eq:appendix_logit}
    \ell_\theta(x)
    =
    \beta_{\theta,b}^{\top}z_\theta(x)
    +
    \tau_{\theta,b}.
\end{equation}

This decomposition separates three independent factors:
(i) the behavioral chart $U_{\theta,b}$,
(ii) the distribution of states within the chart $z_\theta(x)$,
and
(iii) the readout function $(\beta_{\theta,b},\tau_{\theta,b})$.

\subsubsection{Occupancy as a Second-Moment Estimator}

The occupancy component corresponds to the distribution of projected states:

\begin{equation}
    z_\theta(x)
    =
    U_{\theta,b}^{\top}\bar\phi_\theta(x).
\end{equation}

Conditioned on the behavioral label, assume

\begin{equation}
    z_\theta(x)|y=c
    \sim
    (\mu_c,\Sigma_c),
    \qquad c\in\{0,1\}.
\end{equation}

The expected squared distance from the origin in the behavioral chart is

\begin{align}
\mathbb{E}
[
\|z_\theta(x)\|_2^2
|y=c
]
&=
\mathbb{E}
[
z_\theta(x)^\top z_\theta(x)
|y=c
]
\\
&=
\mathrm{tr}
(
\mathbb{E}
[
z_\theta(x)z_\theta(x)^\top
|y=c
]
)
\\
&=
\mathrm{tr}
(
\Sigma_c+\mu_c\mu_c^\top
)
\\
&=
\mathrm{tr}(\Sigma_c)
+
\|\mu_c\|_2^2 .
\end{align}

Therefore, the difference between behavior-positive and behavior-negative
states is

\begin{align}
&
\mathbb{E}
[
\|z_\theta(x)\|_2^2|y=1
]
-
\mathbb{E}
[
\|z_\theta(x)\|_2^2|y=0
]
\nonumber
\\
&=
\|\mu_1\|_2^2-\|\mu_0\|_2^2
+
\mathrm{tr}(\Sigma_1-\Sigma_0).
\end{align}

Thus,

\begin{equation}
\label{eq:occ_derivation}
G^{\mathrm{occ}}
=
\Delta_{\mathrm{mean}}
+
\Delta_{\mathrm{variance}},
\end{equation}

where

\[
\Delta_{\mathrm{mean}}
=
\|\mu_1\|_2^2-\|\mu_0\|_2^2
\]

captures displacement of behavior-positive states in the chart, and

\[
\Delta_{\mathrm{variance}}
=
\mathrm{tr}(\Sigma_1-\Sigma_0)
\]

captures distributional expansion or contraction.

Therefore, $G^{\mathrm{occ}}$ is a rotation-invariant estimator of
behavioral manifold occupancy.  It does not estimate the location of the
manifold itself; instead, it estimates how strongly model states occupy the
manifold after the geometry has been fixed.

\subsubsection{Gain as Effective Readout Sensitivity}

Now consider an intervention with strength $\alpha$.  The intervention
changes the projected chart coordinate:

\begin{equation}
    z_\theta(\alpha).
\end{equation}

The behavioral logit becomes

\begin{equation}
    \ell_\theta(\alpha)
    =
    \beta_{\theta,b}^{\top}
    z_\theta(\alpha)
    +
    \tau_{\theta,b}.
\end{equation}

Taking the derivative with respect to intervention strength gives

\begin{align}
\frac{\partial \ell_\theta(\alpha)}
{\partial\alpha}
&=
\frac{\partial}
{\partial\alpha}
\left(
\beta_{\theta,b}^{\top}
z_\theta(\alpha)
+
\tau_{\theta,b}
\right)
\\
&=
\beta_{\theta,b}^{\top}
\frac{\partial z_\theta(\alpha)}
{\partial\alpha}.
\end{align}

The threshold term disappears:

\begin{equation}
\frac{\partial\tau_{\theta,b}}
{\partial\alpha}
=
0.
\end{equation}

Therefore, causal gain does not estimate $\tau_{\theta,b}$; instead, it estimates
the effective projection of the behavioral readout vector onto the induced
movement direction.

For a direct chart intervention

\begin{equation}
z_\theta(\alpha)
=
z_0+\alpha u,
\end{equation}

we obtain

\begin{equation}
\frac{\partial\ell_\theta}
{\partial\alpha}
=
\beta_{\theta,b}^{\top}u.
\end{equation}

Hence, if $u$ is chosen as the normalized behavioral direction,

\begin{equation}
u=
\frac{\beta_{\theta,b}}
{\|\beta_{\theta,b}\|_2},
\end{equation}

then

\begin{equation}
G^{\mathrm{causal}}
=
\|\beta_{\theta,b}\|_2.
\end{equation}

More generally, when interventions are performed in activation space,
\[
h_\alpha=h+\alpha v,
\]
the induced chart movement is

\begin{equation}
\frac{\partial z_\theta}
{\partial\alpha}
=
U_{\theta,b}^{\top}
J_{\phi_\theta}(h)v,
\end{equation}

where $J_{\phi_\theta}$ is the local Jacobian of the representation map.
Therefore,

\begin{equation}
\label{eq:general_gain}
G^{\mathrm{causal}}
=
\beta_{\theta,b}^{\top}
U_{\theta,b}^{\top}
J_{\phi_\theta}(h)v.
\end{equation}

\subsubsection{Estimating the Threshold Term}

Although causal gain estimates the effective slope $\beta_{\theta,b}$, the
intercept $\tau_{\theta,b}$ cancels under differentiation.  The threshold term
can instead be estimated from a fitted logistic model:

\begin{equation}
    \hat\tau_{\theta,b}
\end{equation}

or equivalently,

\begin{equation}
\hat\tau_{\theta,b}
=
\mathbb{E}[\ell_\theta(x)]
-
\hat\beta_{\theta,b}^{\top}
\mathbb{E}[z_\theta(x)].
\end{equation}

Thus, the complete decomposition is:

\begin{center}
\begin{tabular}{c|c}
\textbf{Component} & \textbf{Estimator}\\
\hline
Behavioral geometry $U_{\theta,b}$
&
Grassmannian metrics
\\
Occupancy $z_\theta(x)$
&
$G^{\mathrm{occ}}$
\\
Readout gain $\beta_{\theta,b}$
&
$G^{\mathrm{causal}}$
\\
Threshold $\tau_{\theta,b}$
&
Logistic intercept
\end{tabular}
\end{center}

This decomposition provides a mechanistic interpretation of post-training:
SFT can alter the behavioral geometry itself, while reward optimization can
modify the occupancy distribution or readout strength within an inherited
behavioral chart.

%% file: sections__supplement__f_full_tables.tex
\section{Numerical Tables}
\label{app:full_tables}

\subsection{Repetition Strength Across Models}
Table~\ref{tab:rep_strength_act_app} gives representative ACT-space repetition strength measurements.  \(T_{\mathrm{base}}\) is the baseline repetition tendency at the
no-intervention point, and \(N_R^{\mathrm{eff}}\) is the PCA dimension at
the \(90\%\)-variance threshold.

\begin{table*}[t]
\centering
\scriptsize
\begin{tabular}{lrrrrr}
\toprule
Model & $G_{\mathrm{ACT}}^{\mathrm{uniF}}$ & $G_{\mathrm{ACT}}^{\mathrm{occ}}$ & $G^{\mathrm{causal}}$ & $T_{\mathrm{base}}$ & $N_R^{\mathrm{eff}}$ \\
\midrule
Gemma-2-2B & 6.8465 & 0.9809 & 0.2942 & 0.8632 & 13 \\
Gemma-2-2B-it & 8.0945 & 0.7189 & 0.2750 & 0.6800 & 15 \\
Gemma-2-9B & 6.8896 & 0.8550 & 0.1980 & 0.8704 & 16 \\
Gemma-2-9B-it & 7.7726 & 0.7816 & 0.0771 & 0.8136 & 14 \\
Qwen2.5-0.5B & 7.5528 & 0.7049 & 0.2108 & 0.8562 & 26 \\
Qwen2.5-0.5B-Inst. & 7.1556 & 0.7323 & 0.5994 & 0.8219 & 25 \\
Qwen2.5-7B & 6.9355 & 0.8142 & 0.2648 & 0.9054 & 16 \\
Qwen2.5-7B-Inst. & 8.0421 & 0.7786 & 0.2072 & 0.8801 & 21 \\
Mistral-7B-v0.3 & 10.3875 & 0.6061 & 0.1661 & 0.8833 & 25 \\
Mistral-7B-Inst. & 8.3211 & 0.6757 & 0.2480 & 0.8713 & 18 \\
Llama-3.1-8B & 3.6277 & 1.1152 & 0.0891 & 0.9082 & 9 \\
Llama-3.1-8B-Inst. & 4.3802 & 1.1844 & 0.1823 & 0.7599 & 8 \\
\bottomrule
\end{tabular}
\caption{Representative repetition strength measurements in ACT space.}
\label{tab:rep_strength_act_app}
\end{table*}

\subsection{Sycophancy Low-Rank Statistics}
Table~\ref{tab:syco_lowrank_app} summarizes selected sycophancy PCA results.  ACT-space PC1 explained variance is usually higher than NOC-space PC1 explained variance, consistent with the main-text claim that sycophancy is more concentrated as a runtime state than as a contribution-space basis.

\begin{table*}[t]
\centering
\scriptsize
\begin{tabular}{lrrrr}
\toprule
Model & NOC PC1 & NOC Dim90 & ACT PC1 & ACT Dim90 \\
\midrule
Qwen2.5-0.5B & 0.2025 & 38 & 0.6011 & 4 \\
Qwen2.5-7B & 0.2924 & 28 & 0.7673 & 3 \\
Qwen2.5-32B & 0.2603 & 32 & 0.7952 & 3 \\
Gemma-2-27B-it & 0.2919 & 23 & 0.6587 & 2 \\
Llama-3.1-8B & 0.3093 & 29 & 0.4262 & 8 \\
Mistral-Nemo-Inst. & 0.2799 & 25 & 0.6764 & 6 \\
Mistral-Small-24B-Base & 0.5269 & 12 & 0.3486 & 8 \\
Qwen3.5-27B & 0.3149 & 34 & 0.3035 & 22 \\
\bottomrule
\end{tabular}
\caption{Selected sycophancy low-rank statistics in NOC and ACT spaces.}
\label{tab:syco_lowrank_app}
\end{table*}

\subsection{Full Controlled Repetition Summary}
Table~\ref{tab:rep_control_app} repeats the final controlled repetition summary from the main text for convenience.

\begin{table*}[t]
\centering
\scriptsize
\begin{tabular}{llrrrrrrrr}
\toprule
Objective & Dir. & $S$ & Angle & $d_{\mathrm{proj}}$ & $R^G$ & $G$ & Repeat & PPL & n-gram \\
\midrule
SFT & up & 0.6934 & 27.78 & 2.7128 & 0.9380 & 76.5460 & 0.96803 & 3.70416 & 0.72299 \\
SFT & down & 0.7106 & 25.77 & 2.5800 & 0.9419 & 63.8125 & 0.92028 & 3.52084 & 0.37174 \\
Reward & up & 0.9600 & 3.59 & 0.9999 & 1.0955 & 67.1523 & 0.91801 & 2.94752 & 0.40325 \\
Reward & down & 0.9200 & 7.20 & 1.4142 & 1.0493 & 66.4614 & 0.88494 & 2.91139 & 0.43393 \\
SFT$\rightarrow$Reward & up & 0.5975 & 36.05 & 3.0427 & 0.8995 & 81.4551 & 0.96703 & 5.00357 & 0.94186 \\
SFT$\rightarrow$Reward & down & 0.6040 & 35.17 & 2.9515 & 0.7045 & 57.8396 & 0.59574 & 3.86605 & 0.56458 \\
\bottomrule
\end{tabular}
\caption{Final checkpoints of the controlled repetition runs.}
\label{tab:rep_control_app}
\end{table*}

\subsection{Full Controlled Sycophancy Summary}
Table~\ref{tab:syco_control_app} summarizes the final controlled sycophancy
results. Reward-only runs preserve the Base geometry most strongly, whereas
SFT-containing runs produce larger geometric changes and stronger directional
shifts in sycophancy behavior.

\begin{table*}[t]
\centering
\scriptsize
\begin{tabular}{llrrrrrrrr}
\toprule
Objective & Dir. & $S$ & Angle & $d_{\mathrm{proj}}$ & $R^G$ & $G$ & Syco Prob & Truth PPL & Syco Pref \\
\midrule
SFT & up & 0.7734 & 24.69 & 2.5636 & 0.9657 & 10.9596 & 0.88584 & 11.71793 & 0.64595 \\
SFT & down & 0.8035 & 22.04 & 2.3871 & 0.8144 & 17.7359 & 0.44221 & 1.28075 & 0.33044 \\
Reward & up & 0.9382 & 10.10 & 1.3384 & 0.9983 & 13.8834 & 0.37020 & 6.48420 & 0.47387 \\
Reward & down & 0.9389 & 9.68 & 1.3309 & 0.9949 & 13.8919 & 0.34155 & 5.68717 & 0.43599 \\
SFT$\rightarrow$Reward & up & 0.7211 & 29.03 & 2.7944 & 0.8845 & 10.3968 & 0.85073 & 1485.09937 & 0.66585 \\
SFT$\rightarrow$Reward & down & 0.6950 & 30.88 & 2.9224 & 0.8187 & 15.8135 & 0.38855 & 1.37858 & 0.29712 \\
\bottomrule
\end{tabular}
\caption{Final checkpoints of the controlled sycophancy runs.}
\label{tab:syco_control_app}
\end{table*}

\subsection{Checkpoint-Level Sycophancy Trajectory}
Table~\ref{tab:syco_rl_app} shows that reward-only sycophancy runs preserve geometry throughout the trajectory.  Behavior moves gradually, but overlap remains around $0.94$--$0.95$.

\begin{table}[t]
\centering
\scriptsize
\begin{tabular}{lrrrr}
\toprule
Run & Step & $S$ & Angle & Syco Prob \\
\midrule
Reward-up & 6 & 0.9498 & 9.06 & 0.3550 \\
Reward-up & 30 & 0.9451 & 9.47 & 0.3623 \\
Reward-up & 60 & 0.9382 & 10.10 & 0.3702 \\
Reward-down & 6 & 0.9498 & 9.06 & 0.3550 \\
Reward-down & 30 & 0.9518 & 8.84 & 0.3472 \\
Reward-down & 60 & 0.9389 & 9.68 & 0.3416 \\
\bottomrule
\end{tabular}
\caption{Reward-only sycophancy trajectories preserve geometry while moving behavior modestly.}
\label{tab:syco_rl_app}
\end{table}

%% file: sections__supplement__j_fisher_derivation.tex
\section{Derivation of Universal Fisher Strength}
\label{app:fisher_derivation}

This section provides the theoretical motivation for the Universal Fisher
Strength metric used in the main text.  The goal is to formalize when an
estimated behavioral chart constitutes a reliable coordinate system for a
behavior, rather than merely a low-dimensional projection with accidental
separability.

\subsection{Behavior Separation in the Local Chart}

Recall the local behavioral model introduced in
Eq.~(\ref{eq:local_behavior_model}):

\begin{equation}
\label{eq:appendix_behavior_model}
\mathrm{logit}\,p_\theta(y_b=1|x)
=
\beta_{\theta,b}^{\top}
z_{\theta,b}(x)
+
\tau_{\theta,b},
\end{equation}

where

\begin{equation}
    z_{\theta,b}(x)
    =
    U_{\theta,b}^{\top}
    \bar\phi_\theta(x)
\end{equation}

is the coordinate of the sample inside the estimated behavioral chart.

Assume that the projected behavioral coordinates follow two
class-conditional distributions:

\begin{equation}
    z|y=c
    \sim
    (\mu_c,\Sigma_c),
    \qquad c\in\{0,1\}.
\end{equation}

The difference between behavior-positive and behavior-negative states is
captured by the mean displacement:

\begin{equation}
    \Delta\mu
    =
    \mu_1-\mu_0 .
\end{equation}

However, the magnitude of $\Delta\mu$ alone is insufficient.  A large
displacement may simply reflect a noisy or highly variable representation
space.  Therefore, the quality of a behavioral chart depends on the ratio
between between-class displacement and within-class variability.

\subsection{Connection to Fisher Discriminant Analysis}

Classical Fisher discriminant analysis seeks a projection direction $v$
that maximizes the ratio

\begin{equation}
\label{eq:fisher_objective}
    J(v)
    =
    \frac{
    (v^\top(\mu_1-\mu_0))^2
    }
    {
    v^\top\Sigma_w v
    },
\end{equation}

where

\begin{equation}
    \Sigma_w
    =
    \frac{1}{2}\Sigma_1
    +
    \frac{1}{2}\Sigma_0
\end{equation}

is the pooled within-class covariance.

The optimal Fisher direction is

\begin{equation}
\label{eq:fisher_direction}
    v^\star
    \propto
    \Sigma_w^{-1}
    (\mu_1-\mu_0).
\end{equation}

Substituting Eq.~(\ref{eq:fisher_direction}) into
Eq.~(\ref{eq:fisher_objective}) gives the maximum Fisher criterion:

\begin{equation}
\label{eq:fisher_max}
    J(v^\star)
    =
    (\mu_1-\mu_0)^\top
    \Sigma_w^{-1}
    (\mu_1-\mu_0).
\end{equation}

Therefore,

\begin{equation}
\label{eq:basic_fisher}
    G^{\mathrm{Fisher}}
    =
    \Delta\mu^\top
    \Sigma_w^{-1}
    \Delta\mu
\end{equation}

is the optimal signal-to-noise ratio of the behavioral chart.

Intuitively, this quantity asks:

\begin{quote}
How far apart are behavior-positive and behavior-negative states after
normalizing by the intrinsic variability of the chart?
\end{quote}

A large value indicates that the chart contains a stable behavioral
coordinate.  A small value indicates that apparent separation is dominated
by within-class variation.

\subsection{Shrinkage Stabilization}

In practical language-model representations, the dimensionality of the chart
can still be large relative to the number of evaluation samples.  Directly
inverting $\Sigma_w$ may therefore be unstable.

We use a standard shrinkage estimator:

\begin{equation}
\label{eq:shrinkage}
    \Sigma_w^{(\lambda)}
    =
    (1-\lambda)\Sigma_w
    +
    \lambda
    \frac{\mathrm{tr}(\Sigma_w)}{q}I,
\end{equation}

where $q$ is the chart dimensionality.

The second term corresponds to an isotropic covariance model.  Therefore,
$\lambda$ controls the trade-off between the empirical covariance structure
and a well-conditioned isotropic approximation.

The resulting stabilized Fisher score is

\begin{equation}
\label{eq:shrinkage_fisher}
    G^{\mathrm{SF}}
    =
    \Delta\mu^\top
    (\Sigma_w^{(\lambda)})^{-1}
    \Delta\mu .
\end{equation}

This prevents artificially large scores caused by nearly singular covariance
directions.

\subsection{Incorporating Cross-Model Reliability}

A central goal of this work is not only to identify a behavioral chart within
one model, but to determine whether the chart represents a transferable
behavioral structure across models.

Suppose the chart is aligned across models using CCA.  Let

\begin{equation}
    \rho_i
\end{equation}

denote the canonical correlation of the $i$-th aligned coordinate.

A coordinate with high $\rho_i$ is consistently represented across models,
whereas a coordinate with low $\rho_i$ may correspond to model-specific noise
or idiosyncratic variation.

Therefore, before computing Fisher separation, we weight each shared
coordinate according to its cross-model reliability:

\begin{equation}
\label{eq:cca_weight}
    D_\rho
    =
    \mathrm{diag}
    (\rho_1,\rho_2,\ldots,\rho_q).
\end{equation}

The reliability-weighted behavioral displacement becomes

\begin{equation}
    \Delta\mu_\rho
    =
    D_\rho\Delta\mu .
\end{equation}

Substituting into Eq.~(\ref{eq:shrinkage_fisher}) gives

\begin{equation}
\label{eq:universal_fisher_derivation}
    G^{\mathrm{uniF}}
    =
    \Delta\mu^\top
    D_\rho
    (\Sigma_w^{(\lambda)})^{-1}
    D_\rho
    \Delta\mu .
\end{equation}

This is the Universal Fisher Strength used in the main text.

\subsection{Interpretation as Behavioral Chart Reliability}

The proposed metric combines three properties required for a meaningful
behavioral manifold.

\paragraph{1. Behavioral separation.}

The displacement term

\begin{equation}
    \Delta\mu
    =
    \mu_1-\mu_0
\end{equation}

requires that behavior-positive and behavior-negative states occupy different
regions of the chart.

A chart with no behavioral displacement receives a low score.

\paragraph{2. Representation stability.}

The covariance normalization

\begin{equation}
    (\Sigma_w^{(\lambda)})^{-1}
\end{equation}

penalizes directions where separation is caused by noisy or unstable
variation.

A chart where positive and negative states overlap substantially receives a
low score even if their raw means differ.

\paragraph{3. Cross-model universality.}

The CCA reliability weighting

\begin{equation}
    D_\rho
\end{equation}

suppresses directions that only appear in a single model.

Therefore, a high Universal Fisher Strength indicates that a behavioral chart
is simultaneously:

\begin{enumerate}
    \item behavior-discriminative;
    \item low-noise;
    \item transferable across models.
\end{enumerate}

This property is important because our goal is not simply to find any
separable projection.  High-dimensional representations contain many
directions that can separate finite samples by chance.  Instead, we seek
behavioral coordinate systems that correspond to stable and reusable
internal structures.

\subsection{One-Dimensional Case}

When the aligned behavioral chart is dominated by a single canonical
coordinate, $q=1$, we have

\begin{equation}
    D_\rho=\rho_1,
\end{equation}

and the Universal Fisher Strength reduces to

\begin{equation}
\label{eq:one_dimension_fisher}
    G^{\mathrm{uniF}\text{-}1D}
    =
    \rho_1^2
    \frac{
    (\mu_1-\mu_0)^2
    }
    {
    \sigma_w^2+\epsilon
    }.
\end{equation}

Thus the metric measures whether the first behavioral direction is not only
predictive within a single model, but also represents a shared behavioral
coordinate across different models.

%% file: sections__supplement__k_Leakage_Control.tex
\section{Leakage Control and Robustness Analysis}
\label{app:robustness}

\subsection{Coordinate Selection Robustness}

The first stage of our framework identifies a sparse coordinate scaffold before
estimating behavioral geometry.  Since this stage may potentially introduce
selection bias, we perform several controls.

\paragraph{Random-coordinate baseline.}

For each behavior-associated coordinate set
\[
    \mathcal{N}_{\theta,b},
\]
we construct matched-size random coordinate sets sampled from the same model
and layer distribution. 

A genuine behavioral chart should exhibit stronger separability, cross-model
alignment, and causal steering effects than a random coordinate population of
the same size.

In particular, the random baseline is used for:

\begin{itemize}
    \item PCA variance comparison;
    \item cross-model CCA alignment;
    \item linear probing transfer accuracy;
    \item steering effectiveness.
\end{itemize}

\paragraph{Label permutation control.}

To test whether the discovered geometry depends on genuine behavioral structure,
we repeat coordinate selection and subspace estimation after randomly
permuting behavioral labels.  Under shuffled labels, the sparse classifier
should fail to identify stable behavior-associated coordinates, and subsequent
CCA alignment, Fisher separation, and steering effects should approach random
levels.

\subsection{Subspace Estimation Robustness}

Our main behavioral chart is estimated by PCA on selected coordinates.  We
evaluate whether conclusions depend on this specific dimensionality choice.

\paragraph{Rank sensitivity.}

The default chart dimension is selected by the cumulative explained variance
criterion:
\[
    \frac{\sum_{i=1}^{k}\lambda_i}
    {\sum_i\lambda_i}
    \geq0.90.
\]

We additionally evaluate fixed-rank charts:
\[
    k=1,\qquad k=3,
\]
as well as alternative variance thresholds.  The goal is to determine whether
the observed geometric movement between SFT and reward optimization is a stable
property of the behavioral geometry or an artifact of a particular PCA cutoff.

\paragraph{Alternative directions.}

The PCA chart is an unsupervised estimate of dominant behavioral variation.
To verify that it captures behavior rather than generic variance, we compare it
with supervised alternatives:

\begin{enumerate}
    \item logistic-regression direction:
    \[
    v_{\mathrm{logit}}
    =
    \frac{w}{\|w\|_2};
    \]

    \item Fisher discriminant direction:
    \[
    v_{\mathrm{Fisher}}
    \propto
    \Sigma_w^{-1}(\mu_1-\mu_0);
    \]

    \item CCA-aligned shared directions;

    \item SAE decoder directions when sparse feature representations are used.
\end{enumerate}

Consistent conclusions across these directions indicate that the discovered
behavioral geometry is not tied to a particular subspace estimator.

\subsection{Behavior Specificity Controls}

A central assumption of our framework is that different behaviors correspond to
distinct behavioral geometries rather than a single generic ``undesirable
behavior'' direction.  We therefore perform behavior-mismatch experiments.

For two behaviors $b_1$ and $b_2$, we evaluate:

\begin{equation}
    U_{\theta,b_1}
    \rightarrow
    y_{b_2}.
\end{equation}

For example, repetition charts are evaluated on sycophancy labels, and
sycophancy charts are evaluated on repetition labels.

A genuine behavior-specific manifold should satisfy:

\begin{itemize}
    \item strong within-behavior separability;
    \item weak cross-behavior transfer;
    \item strong within-behavior causal steering;
    \item weak cross-behavior causal effect.
\end{itemize}

This control prevents the framework from merely identifying generic
degeneration, refusal, or alignment-related directions.

\subsection{Token Extraction Robustness}

Behavioral expression may depend on the choice of token window.  We therefore
evaluate multiple extraction strategies.

For a behavior-relevant instance, the feature representation can be extracted
from:

\begin{enumerate}
    \item the complete response span;
    \item the behavior onset region;
    \item an event-centered local window;
    \item a fixed-length pre-event window.
\end{enumerate}

The purpose of these comparisons is to ensure that discovered geometry is not
caused by a particular token-selection heuristic.

For dynamic analyses, such as generation trajectories, we additionally compare
representations before and after the visible behavioral event.  This tests
whether the internal behavioral state emerges before the surface-level output
phenomenon.

\subsection{Cross-Model Alignment Robustness}

Because neuron indices are not comparable across architectures, our framework
uses CCA-based alignment to construct shared behavioral charts.  We verify that
the observed alignment is not caused by arbitrary high-dimensional matching.

The following controls are performed:

\begin{enumerate}
    \item matched-size random coordinate alignment;
    \item shuffled-label alignment;
    \item different CCA ranks;
    \item different reference models;
    \item pairwise family-stratified comparisons.
\end{enumerate}

For each comparison we report:

\begin{itemize}
    \item canonical correlation;
    \item linear probing transfer performance;
    \item projector overlap;
    \item principal angles.
\end{itemize}

A reliable behavioral manifold should remain aligned across these choices.

\subsection{Causal Intervention Controls}

The causal experiments use three intervention levels:

\begin{enumerate}
    \item sparse coordinate scaling;
    \item subspace projection editing;
    \item direction-level steering.
\end{enumerate}

For each intervention, we compare:

\begin{itemize}
    \item behavior-associated directions;
    \item random directions;
    \item opposite-direction controls;
    \item matched-magnitude non-behavior directions.
\end{itemize}

The intervention strength is varied to obtain dose-response curves rather than
single-point comparisons.  A genuine behavioral chart should produce a
monotonic or directionally consistent response when amplified or suppressed.

\subsection{Post-Training Geometry Robustness}

To ensure that the observed difference between SFT and reward optimization is
not caused by a particular checkpoint or metric, we analyze post-training
trajectories using multiple geometric quantities:

\begin{itemize}
    \item projector overlap:
    \[
    S(a,b);
    \]

    \item principal angles:
    \[
    \vartheta_i(a,b);
    \]

    \item projection distance:
    \[
    d_{\mathrm{proj}}(a,b);
    \]

    \item orthogonal novelty:
    \[
    N_{\perp}(a\rightarrow b);
    \]

    \item anchored Fisher retention:
    \[
    R^G_{0\rightarrow t}.
    \]
\end{itemize}

Agreement across these measures supports the interpretation that SFT and reward
optimization induce different geometric trajectories rather than merely
different behavioral magnitudes.

%% file: sections__supplement__l_Vis_Behavioral_Charts.tex
\section{Visualization of Behavioral Charts}
\label{app:behavioral-chart-visualization}

The behavioral manifold framework assumes that behavior-associated internal
states are not arbitrarily distributed in the ambient representation space, but
rather concentrate around low-dimensional local charts.  While quantitative
metrics such as projector overlap, Fisher strength, and cross-model alignment
measure geometric properties numerically, directly visualizing the estimated
charts provides an intuitive validation of this assumption.

This appendix visualizes the learned behavioral charts for two qualitatively
different behaviors: repetition and sycophancy.  These behaviors are selected
because they represent distinct classes of model failures: repetition reflects a
generation-dynamics failure, whereas sycophancy reflects an alignment-related
preference failure.  The fact that both exhibit similar low-dimensional
geometric organization provides evidence that the proposed framework captures a
general property of behavior representation rather than a behavior-specific
artifact.

\subsection{Visualization Protocol}

For a model $f_\theta$, behavior $b$, and representation space $\Omega$, the
behavioral chart is estimated as described in
Section~\ref{sec:framework}:

\[
    U_{\theta,b,\Omega}
    =
    [v_1,\ldots,v_k].
\]

Given a standardized representation vector
\[
    \bar\phi_{\theta,b}^{(\Omega)}(x),
\]
we project each instance into the learned chart coordinate system:

\begin{equation}
    z_{\theta,b,\Omega}(x)
    =
    U_{\theta,b,\Omega}^{\top}
    \bar\phi_{\theta,b}^{(\Omega)}(x).
\end{equation}

For visualization, we display the first two chart coordinates:

\[
    (z_1,z_2).
\]

Positive and negative behavior examples are shown using different colors.  We
also project two independently estimated supervised directions into the same
coordinate system:

\begin{itemize}
    \item the Fisher discriminant direction
    \[
    v_{\mathrm{Fisher}}
    \propto
    \Sigma_w^{-1}(\mu_1-\mu_0),
    \]
    which maximizes class separation under the shared covariance assumption;

    \item the behavior-positive logistic direction obtained from the sparse
    coordinate classifier.
\end{itemize}

The alignment between these supervised directions and the dominant chart axis
provides an additional sanity check that the visualization reflects
behavior-relevant geometry rather than arbitrary variance captured by PCA.

We visualize two complementary representation spaces:

\begin{enumerate}
    \item \textbf{Activation space (ACT):} 
    the selected neuron activations directly measure whether
    behavior-associated coordinates are active.

    \item \textbf{Normalized output-contribution space (NOC):}
    the selected coordinates are weighted according to their signed
    contribution to the realized MLP update, measuring whether the coordinates
    functionally participate in the layer computation.
\end{enumerate}

Consistency between ACT and NOC visualizations indicates that the discovered
behavioral geometry is not solely caused by activation magnitude, but is also
reflected in functionally relevant computation.

\subsection{Repetition Behavioral Charts}

\begin{figure*}[t]
    \centering
    \begin{subfigure}{0.48\linewidth}
        \centering
        \includegraphics[width=\linewidth]{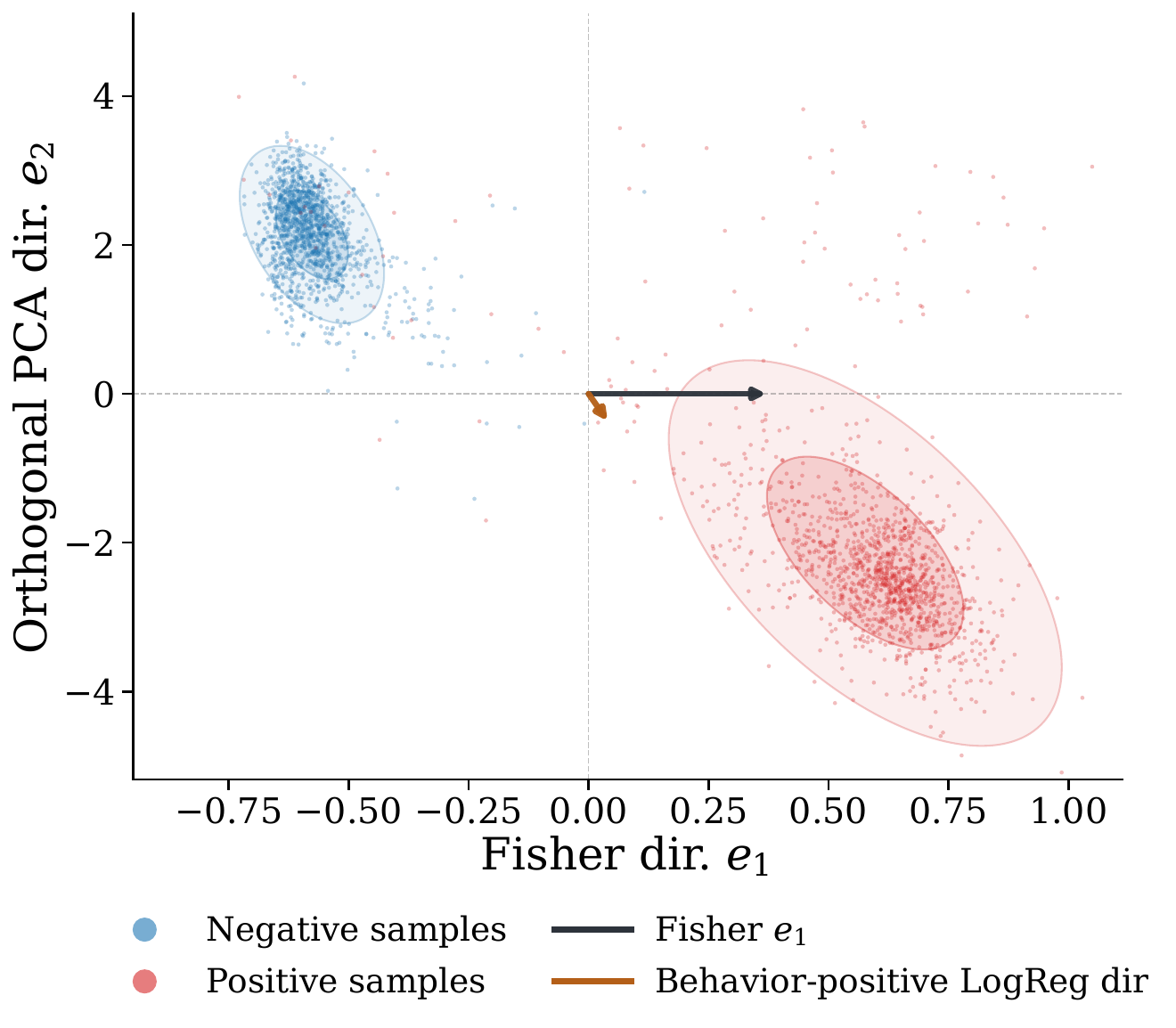}
        \caption{Activation space (ACT).}
    \end{subfigure}
    \hfill
    \begin{subfigure}{0.48\linewidth}
        \centering
        \includegraphics[width=\linewidth]{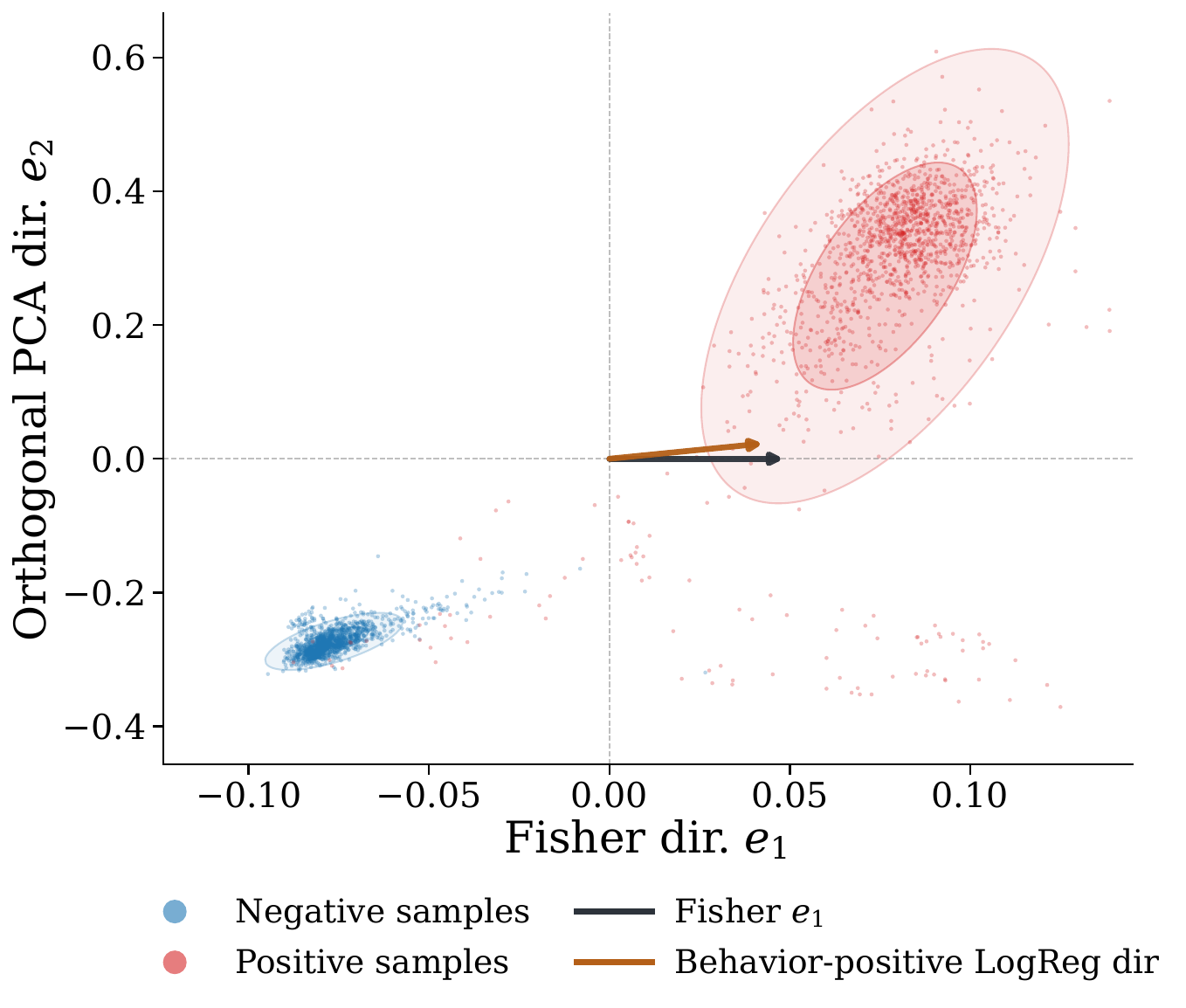}
        \caption{Normalized output-contribution space (NOC).}
    \end{subfigure}

    \caption{
    Visualization of the repetition behavioral chart.
    Behavior-positive and behavior-negative examples form distinguishable
    regions in both ACT and NOC spaces.  The Fisher and logistic directions
    align with the dominant separation axis, suggesting that repetition is
    represented by a structured low-dimensional geometry rather than isolated
    activation events.
    }
    \label{fig:repetition-behavioral-chart}
\end{figure*}

Repetition provides a test case for a dynamically generated failure mode.
Figure~\ref{fig:repetition-behavioral-chart} compares the learned repetition
charts in ACT and NOC spaces, showing that repetitive and non-repetitive states
occupy different regions of the chart.

In both representation spaces, repetition-positive states occupy a distinct
region from repetition-negative states.  The agreement between activation
space and normalized output-contribution space is important: if the separation
were caused only by large activation magnitudes, the structure would disappear
after accounting for functional contribution.  Instead, the same qualitative
geometry persists in both spaces, suggesting that repetition corresponds to a
stable computational state supported by a coordinated population of
coordinates.

The alignment of Fisher and logistic directions further indicates that the
dominant variance direction of the chart is also behavior-discriminative.  In
other words, the estimated chart is not merely a low-dimensional compression of
activation variation; it captures directions along which the model state
systematically changes its probability of entering the repetition regime.

\subsection{Sycophancy Behavioral Charts}

\begin{figure*}[t]
    \centering
    \begin{subfigure}{0.48\linewidth}
        \centering
        \includegraphics[width=\linewidth]{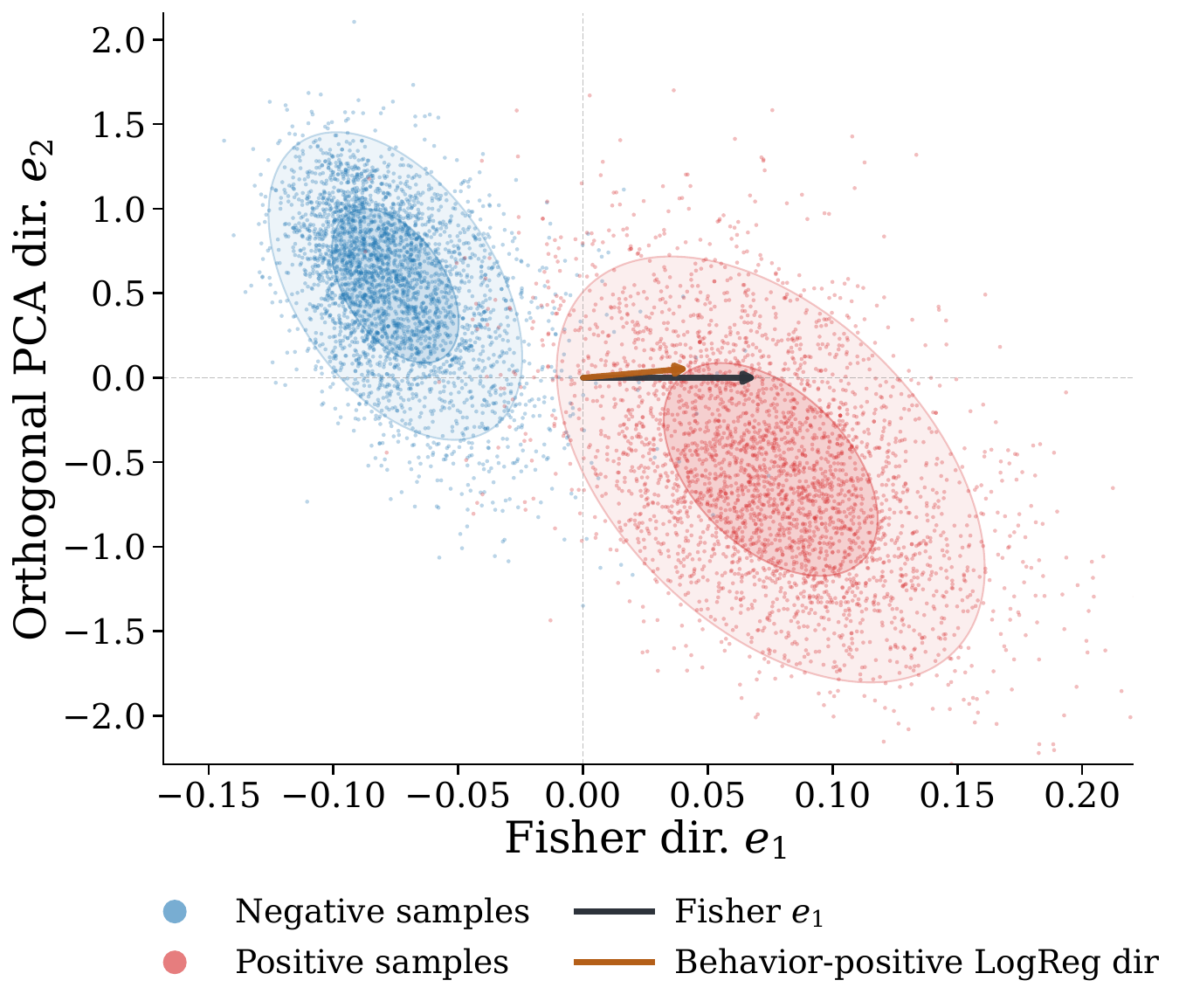}
        \caption{Activation space (ACT).}
    \end{subfigure}
    \hfill
    \begin{subfigure}{0.48\linewidth}
        \centering
        \includegraphics[width=\linewidth]{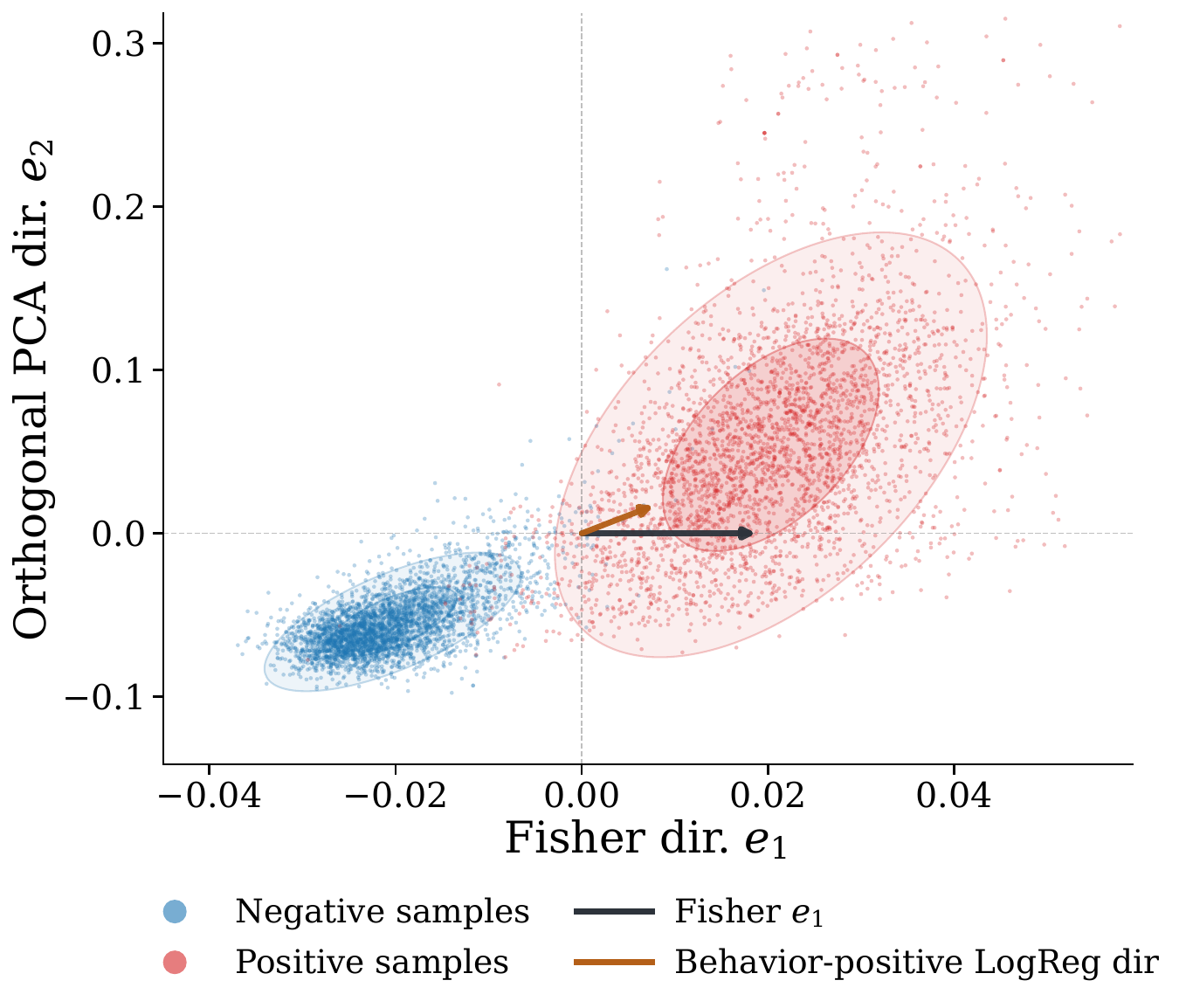}
        \caption{Normalized output-contribution space (NOC).}
    \end{subfigure}

    \caption{
    Visualization of the sycophancy behavioral chart.
    Sycophantic and non-sycophantic states occupy different regions in the
    learned chart.  The agreement between ACT and NOC representations suggests
    that sycophancy is supported by structured internal computation rather than
    merely superficial output statistics.
    }
    \label{fig:sycophancy-behavioral-chart}
\end{figure*}

Sycophancy provides an independent validation on a qualitatively different
behavior.  Unlike repetition, which emerges from generation dynamics,
sycophancy reflects preference and alignment behavior.  Nevertheless,
Figure~\ref{fig:sycophancy-behavioral-chart} shows that the same framework
identifies a structured low-dimensional chart in both ACT and NOC spaces.

The sycophancy charts demonstrate that the behavioral manifold framework
extends beyond generation degeneration.  Behavior-positive and
behavior-negative states form distinguishable regions even though sycophancy
involves a different causal mechanism from repetition.  The consistency between
the two feature spaces suggests that the discovered structure reflects a
distributed behavioral representation rather than a single activation
correlate.

Moreover, the supervised directions consistently align with the chart
separation axis.  This supports the interpretation that the estimated
behavioral chart $U_{\theta,b}$ captures a meaningful local coordinate system
for the behavior.

%% file: sections__supplement__m_Trajectory.tex
\section{Token-Level Dynamics in Behavioral Landscapes}
\label{app:trajectory-visualization}

The main behavioral-chart analysis treats behavior-positive and
behavior-negative representations as class-conditional regions in a learned
low-dimensional coordinate system.  This static view establishes that a
behavior has an identifiable representation geometry, but it does not reveal
how the model reaches that region while processing or generating a sequence.
We therefore complement the static analysis with token-level trajectories
through the learned behavioral landscapes.

The trajectory analysis addresses a dynamic question:
\emph{how does the internal state evolve relative to a behavioral region before,
during, and after the behavior becomes visible in the output?}  It also provides
a qualitative test of whether the estimated chart captures temporally coherent
state evolution rather than only aggregate class separation.

\subsection{Behavioral Energy Landscape}

For a fixed model, behavior $b$, and feature space
$\Omega\in\{\mathrm{ACT},\mathrm{NOC}\}$, let
\[
    z_{\theta,b,\Omega}(x)
    =
    U_{\theta,b,\Omega}^{\top}
    \bar\phi_{\theta,b}^{(\Omega)}(x)
\]
be the representation of an instance in the learned behavioral chart.  To
visualize the first two chart coordinates, we estimate class-conditional
densities
\[
    \hat p^+_{b,\Omega}(z_1,z_2),
    \qquad
    \hat p^-_{b,\Omega}(z_1,z_2)
\]
from behavior-positive and behavior-negative chart samples.  The displayed
background is the log-density ratio
\begin{equation}
\label{eq:trajectory_energy_landscape}
    \mathcal{E}_{b,\Omega}(z)
    =
    \log
    \frac{
    \hat p^+_{b,\Omega}(z)+\epsilon
    }{
    \hat p^-_{b,\Omega}(z)+\epsilon
    }.
\end{equation}
Positive values indicate regions relatively more characteristic of
behavior-positive states, while negative values indicate regions relatively
more characteristic of behavior-negative states.  The red and blue background
should therefore be interpreted as a relative behavioral-density landscape,
not as a calibrated behavioral probability.

To prevent the visualized trajectory from determining its own background, the
trajectory instance is excluded from the density estimate whenever it belongs
to the chart-estimation set.  The behavior-positive steering direction is also
projected into the two-dimensional chart and shown as a black arrow.

\subsection{Token-Level Trajectory Construction}

Given a teacher-forced sequence
\[
    x_1,x_2,\ldots,x_T,
\]
causal masking ensures that the hidden state at position $t$ depends only on
$x_{\leq t}$.  We extract the behavior-associated representation at each token:
\begin{equation}
    \phi_t^{(\Omega)}
    =
    \phi_{\theta,b}^{(\Omega)}(x_{\leq t}),
\end{equation}
standardize it using the chart-estimation statistics, and project it into the
learned chart:
\begin{equation}
\label{eq:token_trajectory_projection}
    z_t
    =
    U_{\theta,b,\Omega}^{\top}
    \bar\phi_t^{(\Omega)}.
\end{equation}
The token-level behavioral trajectory is the ordered sequence
\begin{equation}
    \mathcal{T}_{b,\Omega}(x)
    =
    (z_1,z_2,\ldots,z_T).
\end{equation}
Only the first two chart coordinates are displayed.  Consecutive token
projections are joined in temporal order.  A shape-preserving piecewise cubic
interpolant is used only to render the path smoothly; all markers and all
quantitative analyses are based on the original token projections.

For an annotated behavioral span
$\mathcal{I}_b\subseteq\{1,\ldots,T\}$, we additionally show its mean chart
coordinate:
\begin{equation}
    \bar z_{\mathcal{I}_b}
    =
    \frac{1}{|\mathcal{I}_b|}
    \sum_{t\in\mathcal{I}_b}z_t.
\end{equation}
This mean is displayed as a white diamond and summarizes where the established
behavioral span lies relative to the positive and negative chart regions.

\subsection{Visual Encoding}

The trajectories use a shared visual grammar across behaviors and feature
spaces.  The black circle marks the first displayed token, and the black square
marks the final displayed token.  A red star marks the annotated behavioral
onset.  The white diamond denotes the mean representation of the annotated
behavioral span.

For repetition, gray denotes the input prefix, blue denotes ordinary
non-repetitive generation, yellow denotes the first occurrence of the repeated
unit that leads into the loop, and red denotes established repetition beginning
with the second consecutive occurrence of that unit.

For sycophancy, gray denotes the input prefix, yellow denotes response tokens
preceding the annotated sycophantic span, red denotes the sycophantic span, and
blue denotes any displayed post-span tokens.  The colors describe temporal
segments of one trajectory and are distinct from the red/blue class-density
background.

\subsection{Repetition Trajectories in Activation Space}

\begin{figure*}[t]
    \centering

    \begin{minipage}[t]{0.315\textwidth}
        \centering
        \includegraphics[width=\linewidth]
        {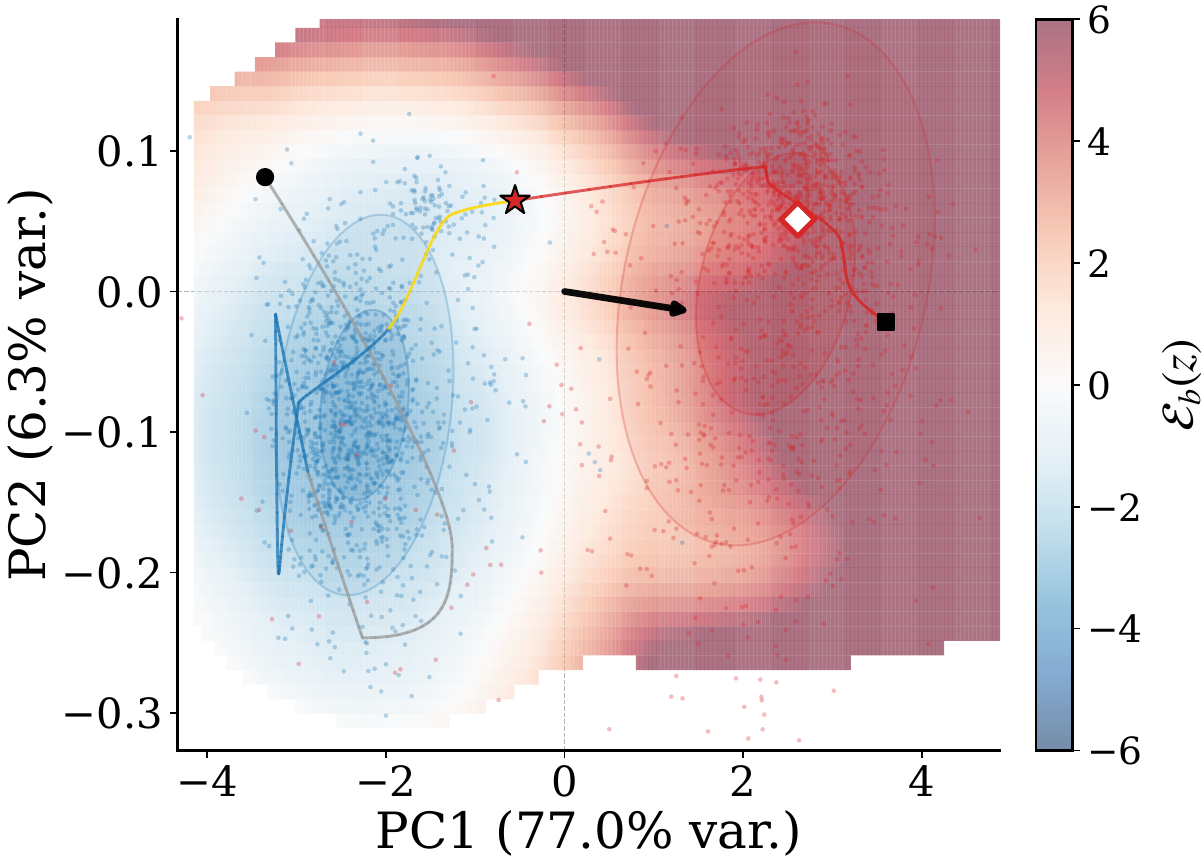}
        \vspace{-1mm}

        {\scriptsize (a) Example 1}
    \end{minipage}
    \hfill
    \begin{minipage}[t]{0.315\textwidth}
        \centering
        \includegraphics[width=\linewidth]
        {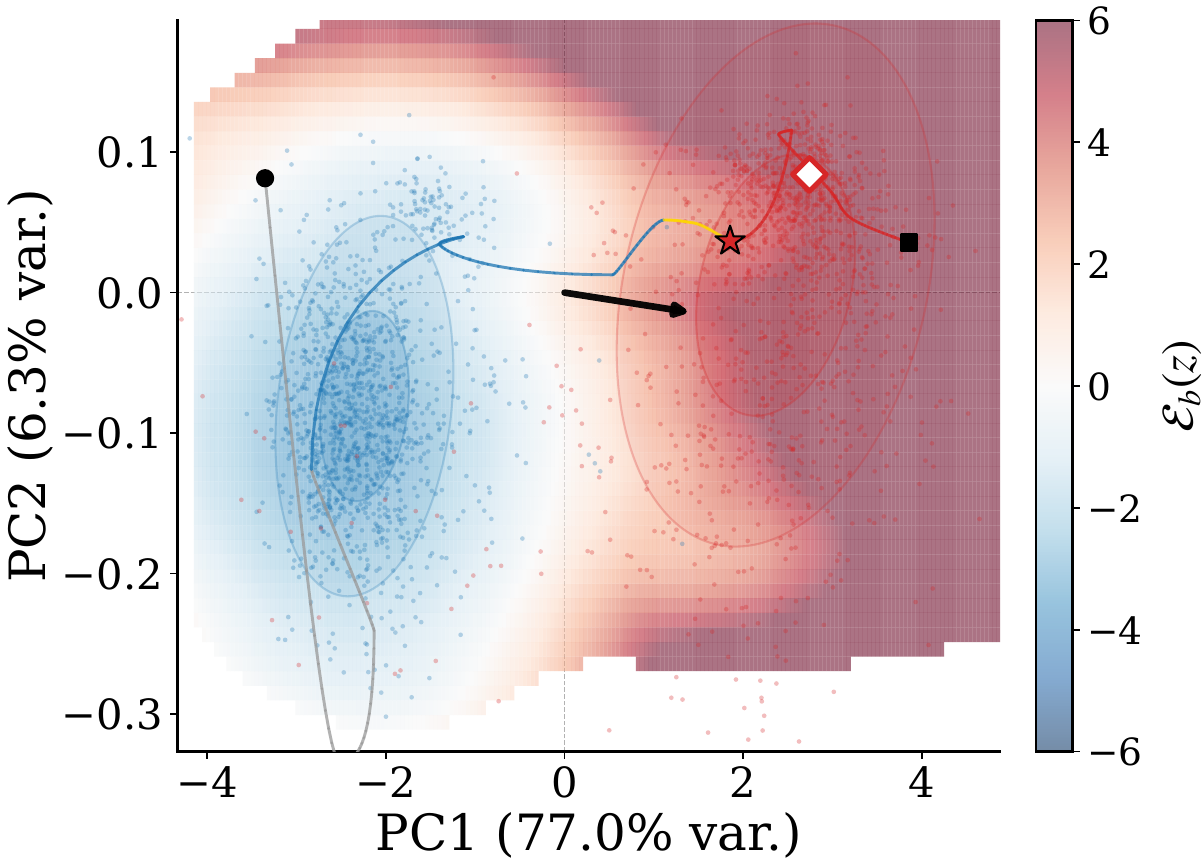}
        \vspace{-1mm}

        {\scriptsize (b) Example 2}
    \end{minipage}
    \hfill
    \begin{minipage}[t]{0.315\textwidth}
        \centering
        \includegraphics[width=\linewidth]
        {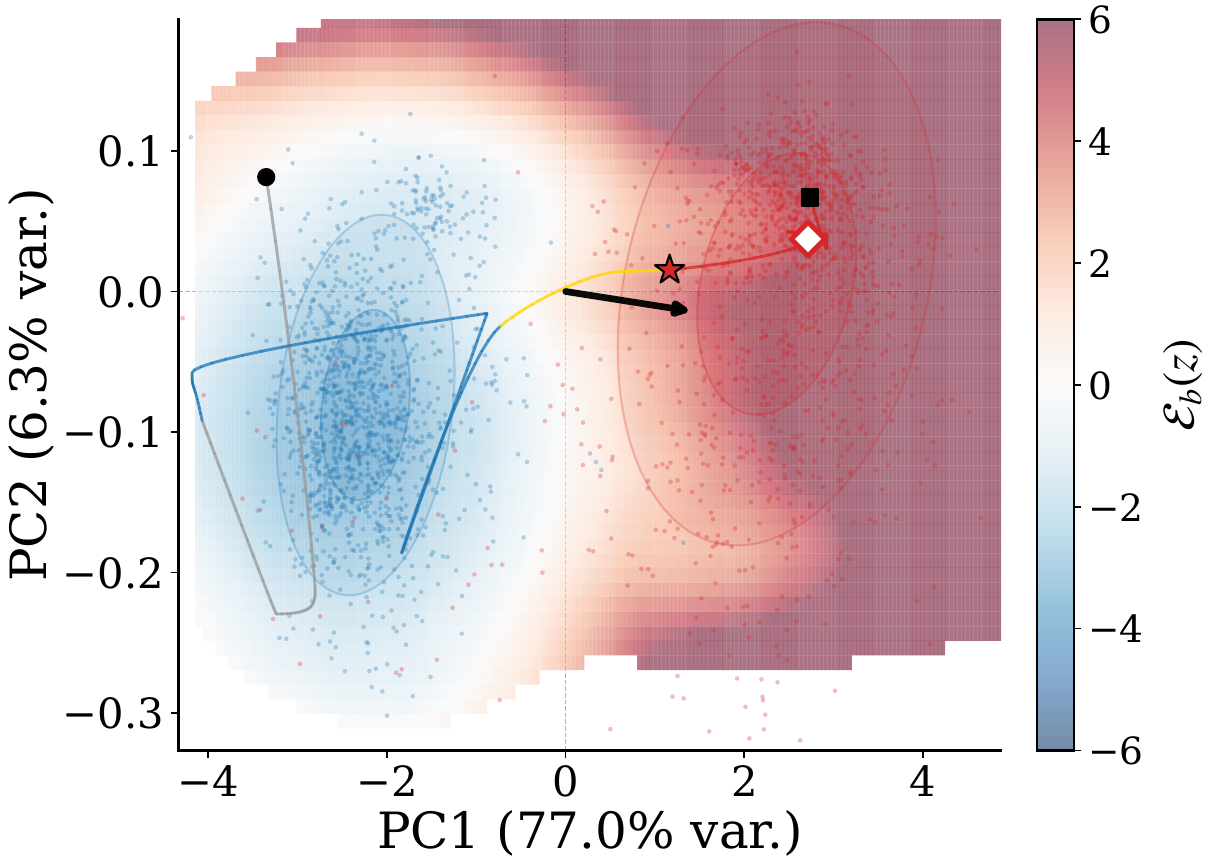}
        \vspace{-1mm}

        {\scriptsize (c) Example 3}
    \end{minipage}

    \vspace{1.5mm}

    \begin{minipage}[t]{0.315\textwidth}
        \centering
        \includegraphics[width=\linewidth]
        {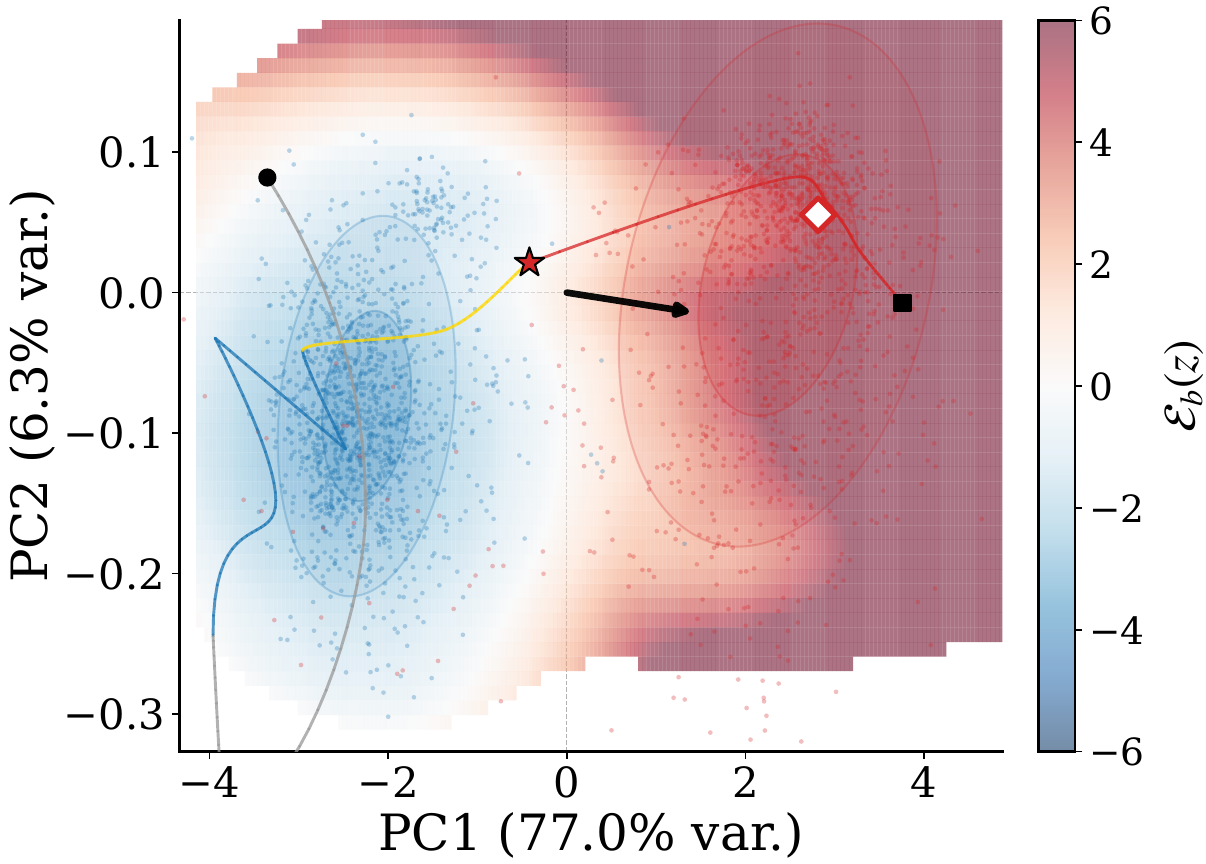}
        \vspace{-1mm}

        {\scriptsize (d) Example 4}
    \end{minipage}
    \hfill
    \begin{minipage}[t]{0.315\textwidth}
        \centering
        \includegraphics[width=\linewidth]
        {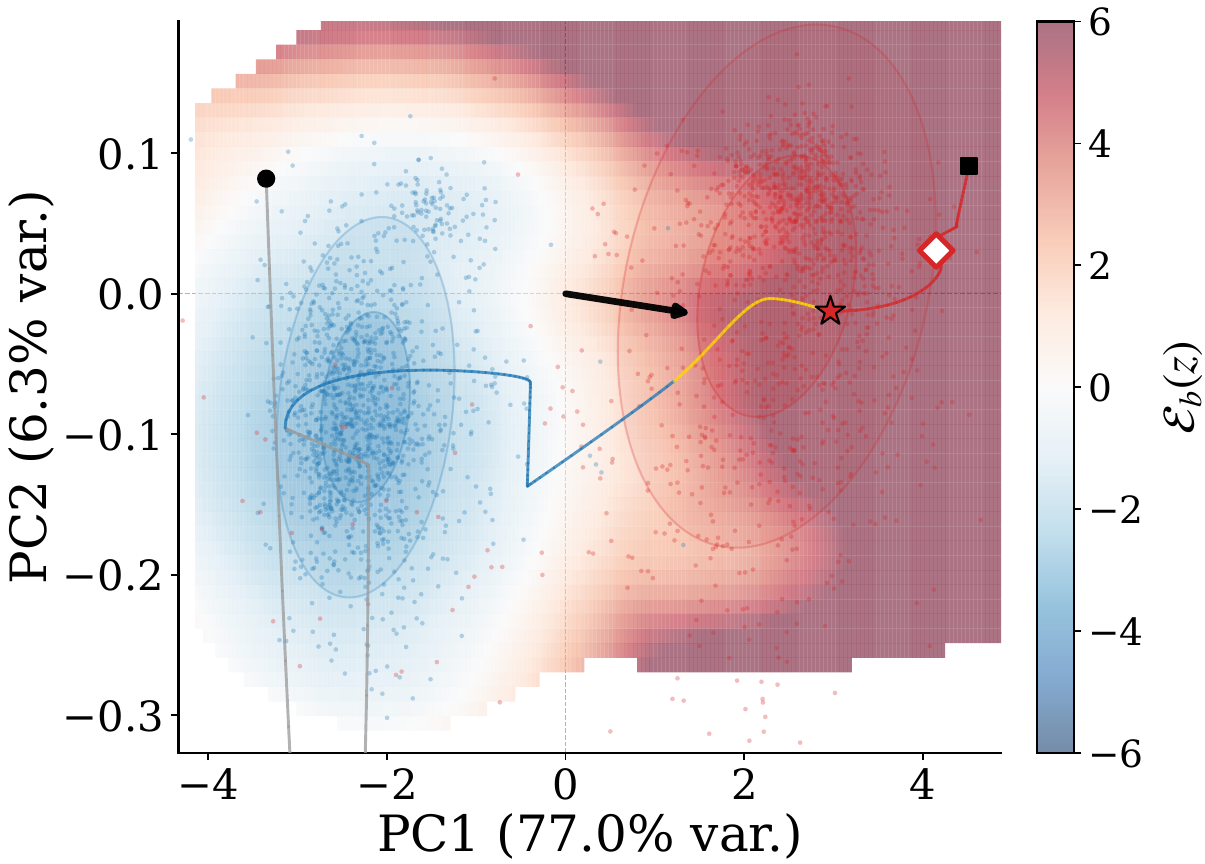}
        \vspace{-1mm}

        {\scriptsize (e) Example 5}
    \end{minipage}
    \hfill
    \begin{minipage}[t]{0.315\textwidth}
        \centering
        \includegraphics[width=\linewidth]
        {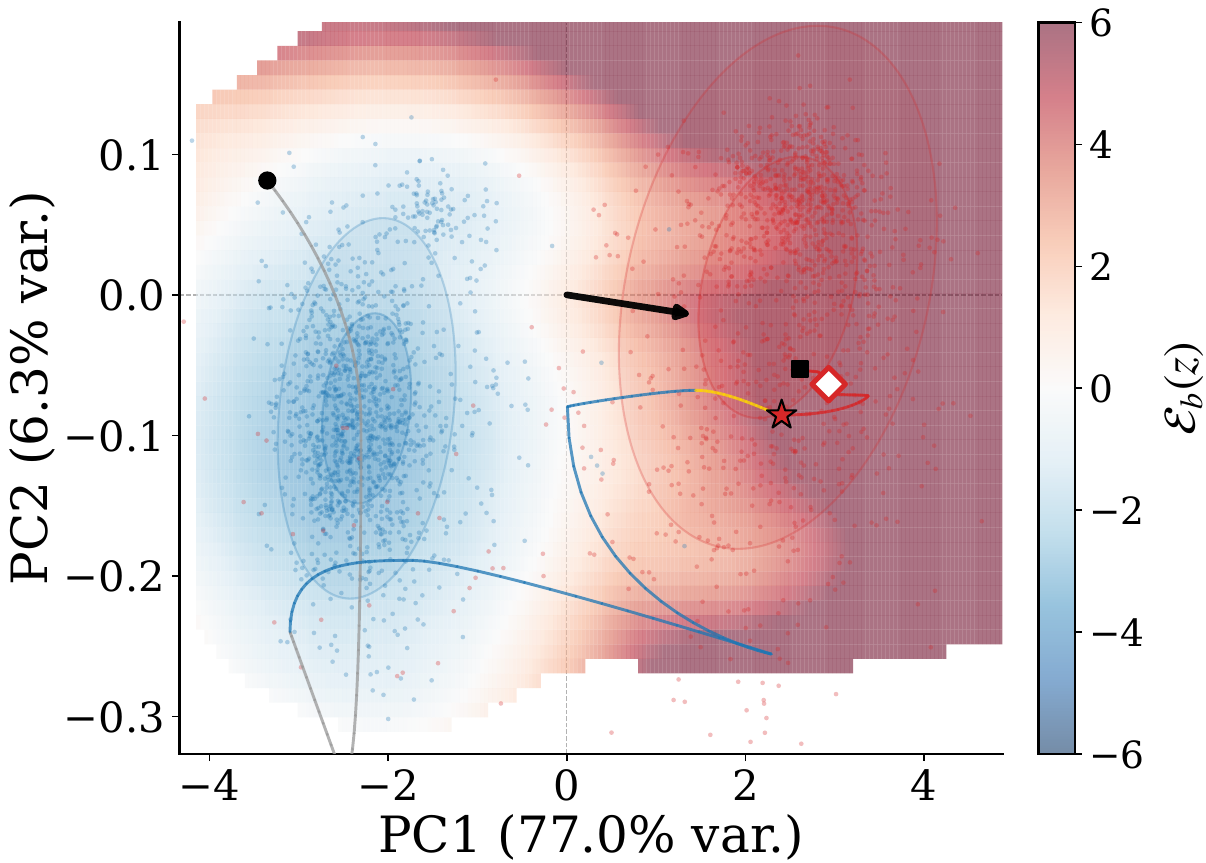}
        \vspace{-1mm}

        {\scriptsize (f) Example 6}
    \end{minipage}

    \caption{
    Repetition trajectories in ACT space for examples with matched NOC
    visualizations.  Across examples, prefix and ordinary-generation states
    predominantly occupy the negative or transitional region, whereas repeated
    spans move toward and remain in the behavior-positive region.  Individual
    trajectories differ in curvature, transition speed, and the extent to which
    the first repeated unit already enters the positive basin.
    }
    \label{fig:rep_act_trajectory_gallery_a}
\end{figure*}

\begin{figure*}[t]
    \centering

    \begin{minipage}[t]{0.315\textwidth}
        \centering
        \includegraphics[width=\linewidth]
        {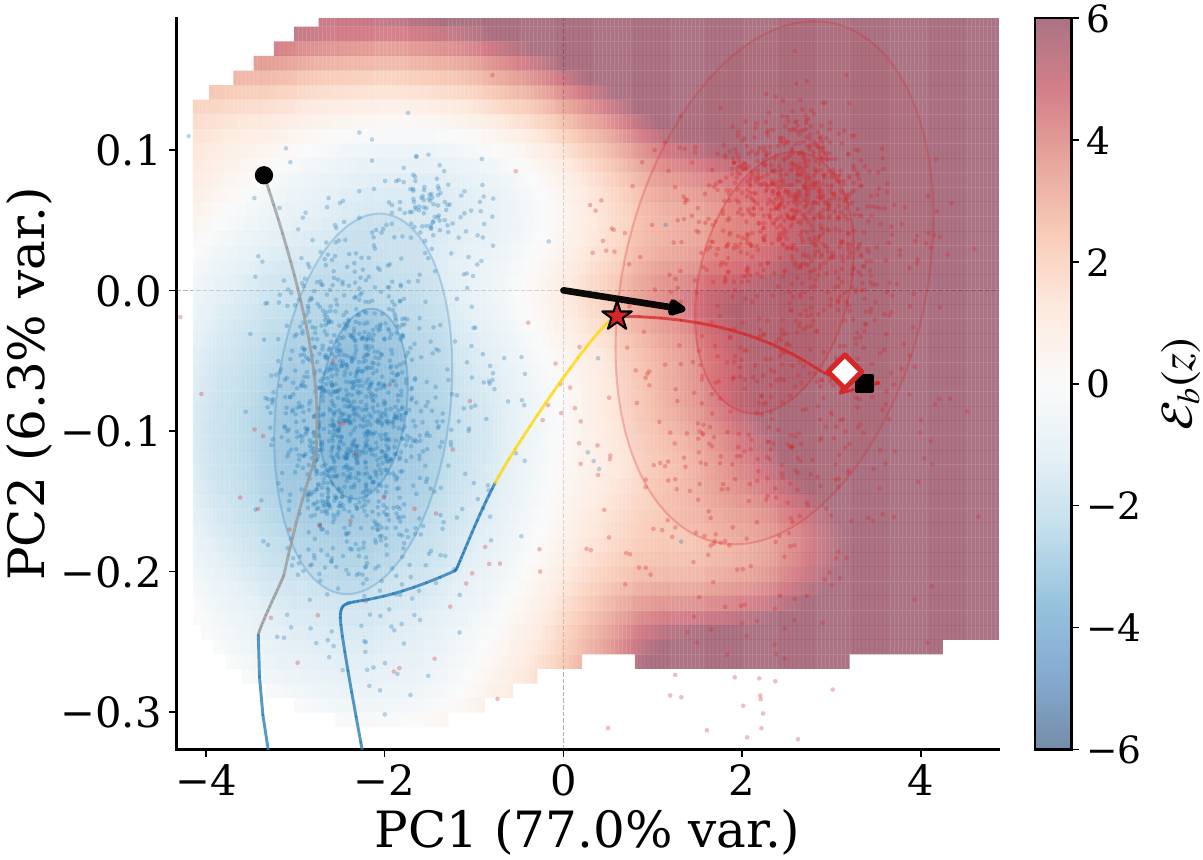}
        \vspace{-1mm}

        {\scriptsize (a) Example 1}
    \end{minipage}
    \hfill
    \begin{minipage}[t]{0.315\textwidth}
        \centering
        \includegraphics[width=\linewidth]
        {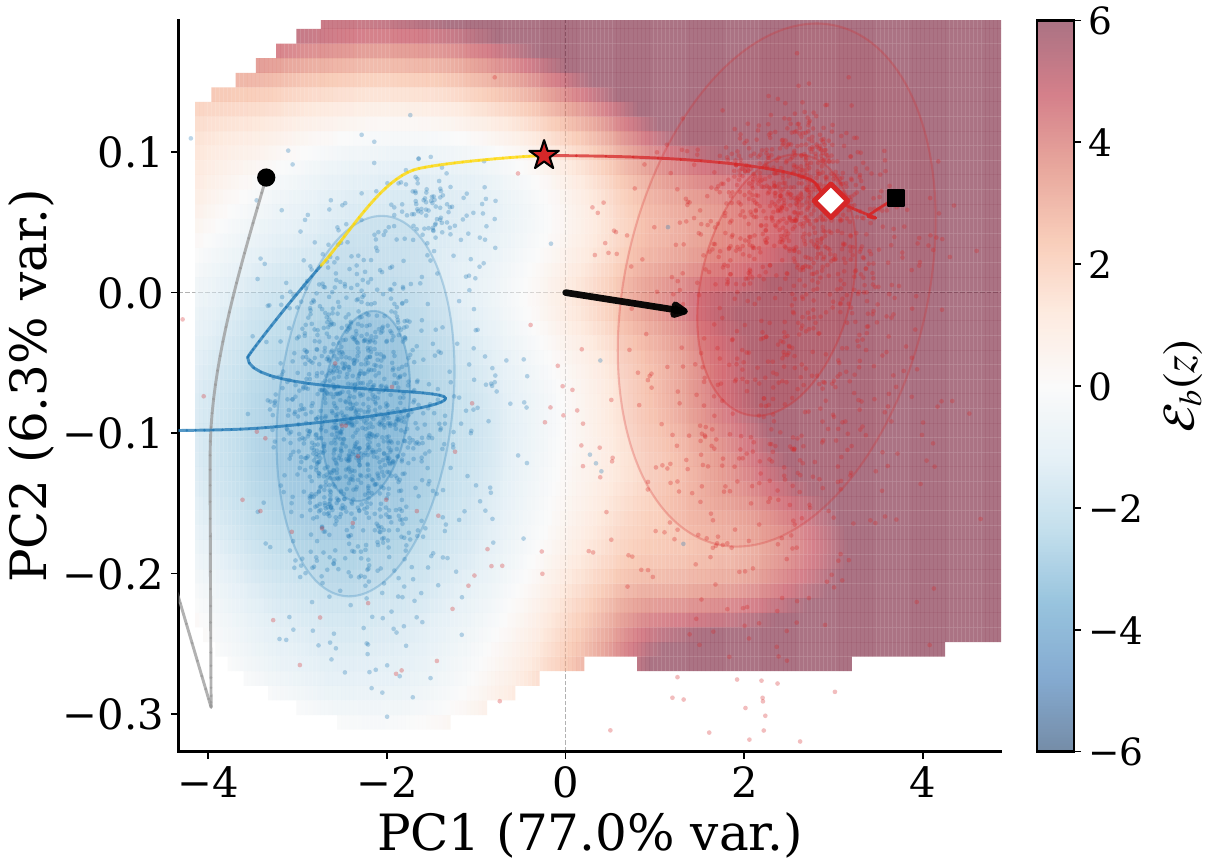}
        \vspace{-1mm}

        {\scriptsize (b) Example 2}
    \end{minipage}
    \hfill
    \begin{minipage}[t]{0.315\textwidth}
        \centering
        \includegraphics[width=\linewidth]
        {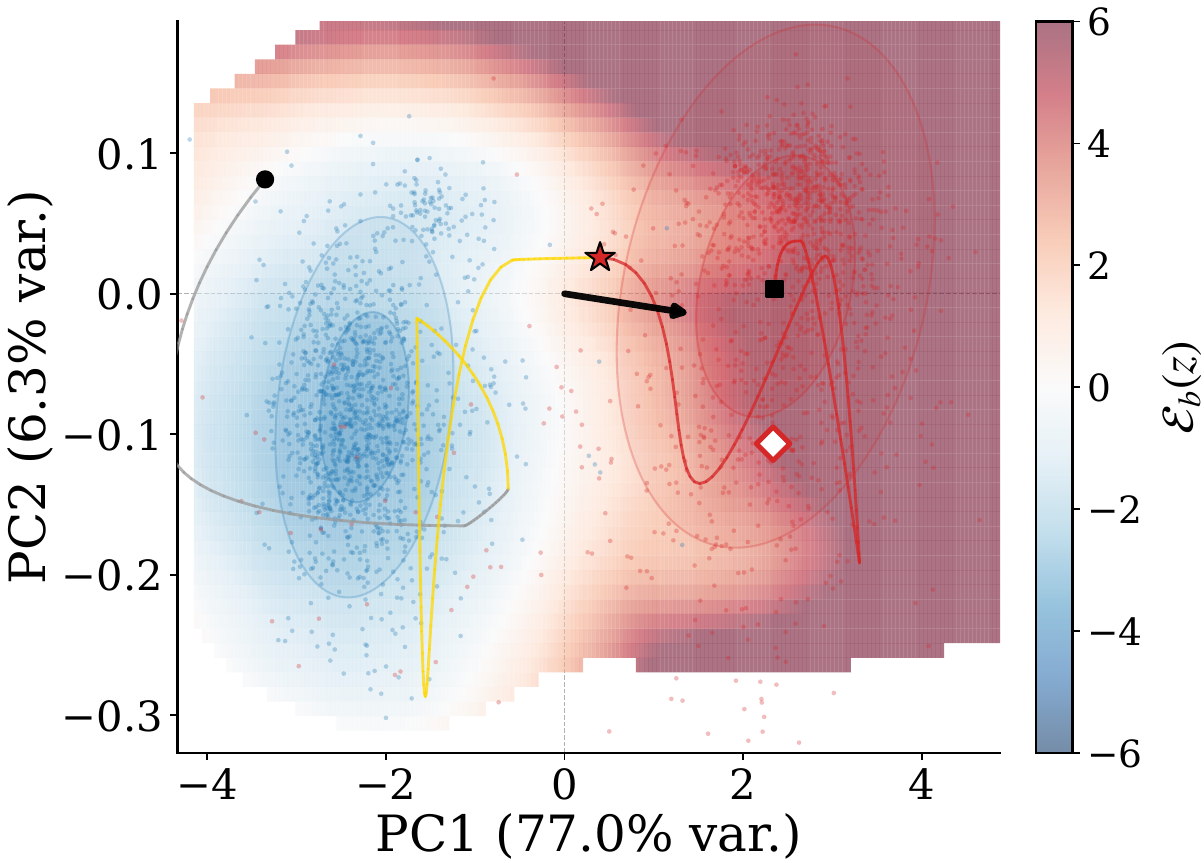}
        \vspace{-1mm}

        {\scriptsize (c) Paragraph-level example 3}
    \end{minipage}

    \caption{
    Additional repetition trajectories in ACT space.  The gallery demonstrates
    that entry into the repetition-positive region need not follow a single
    path: some trajectories cross the chart boundary directly, while others
    approach it through a curved or partially recurrent path.  The common
    property is the relocation of established repeated tokens toward the
    positive-density region.
    }
    \label{fig:rep_act_trajectory_gallery_b}
\end{figure*}

Figure~\ref{fig:rep_act_trajectory_gallery_a} shows six repetition trajectories
in ACT space for examples that also have corresponding NOC visualizations.
Figure~\ref{fig:rep_act_trajectory_gallery_b} shows three additional ACT
trajectories, including a paragraph-level example.  Together, these examples
illustrate both the recurring geometric pattern and the variability of
individual decoding paths.

The ACT trajectories support an attractor-like interpretation of repetition,
but they do not imply a single deterministic route into repetition.  The
ordinary-generation segment often occupies the negative basin or its boundary.
The lead-in segment then moves toward the positive region, and the established
repetition segment is concentrated more strongly on the positive side.  The
white repeated-span means are consistently displaced toward the positive
region, even when the token-level path contains local reversals or loops.

The trajectories also show why a static endpoint analysis is incomplete.  In
several examples, movement toward the positive region begins during the first
occurrence of the repeated unit, before the detector marks established
repetition at the second occurrence.  The internal transition may therefore
precede the output-level criterion used to declare repetition onset.

\subsection{Repetition Trajectories in NOC Space}

\begin{figure*}[t]
    \centering

    \begin{minipage}[t]{0.315\textwidth}
        \centering
        \includegraphics[width=\linewidth]
        {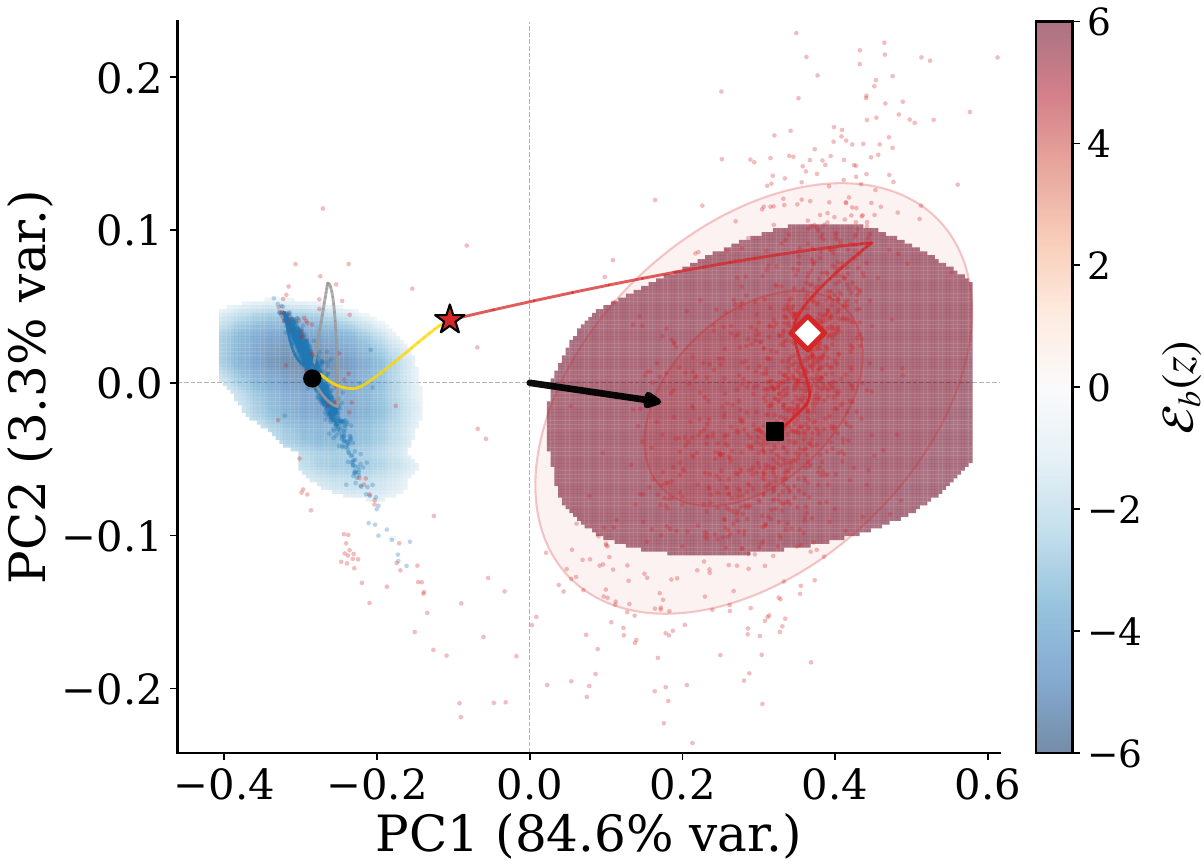}
        \vspace{-1mm}

        {\scriptsize (a) Example 1}
    \end{minipage}
    \hfill
    \begin{minipage}[t]{0.315\textwidth}
        \centering
        \includegraphics[width=\linewidth]
        {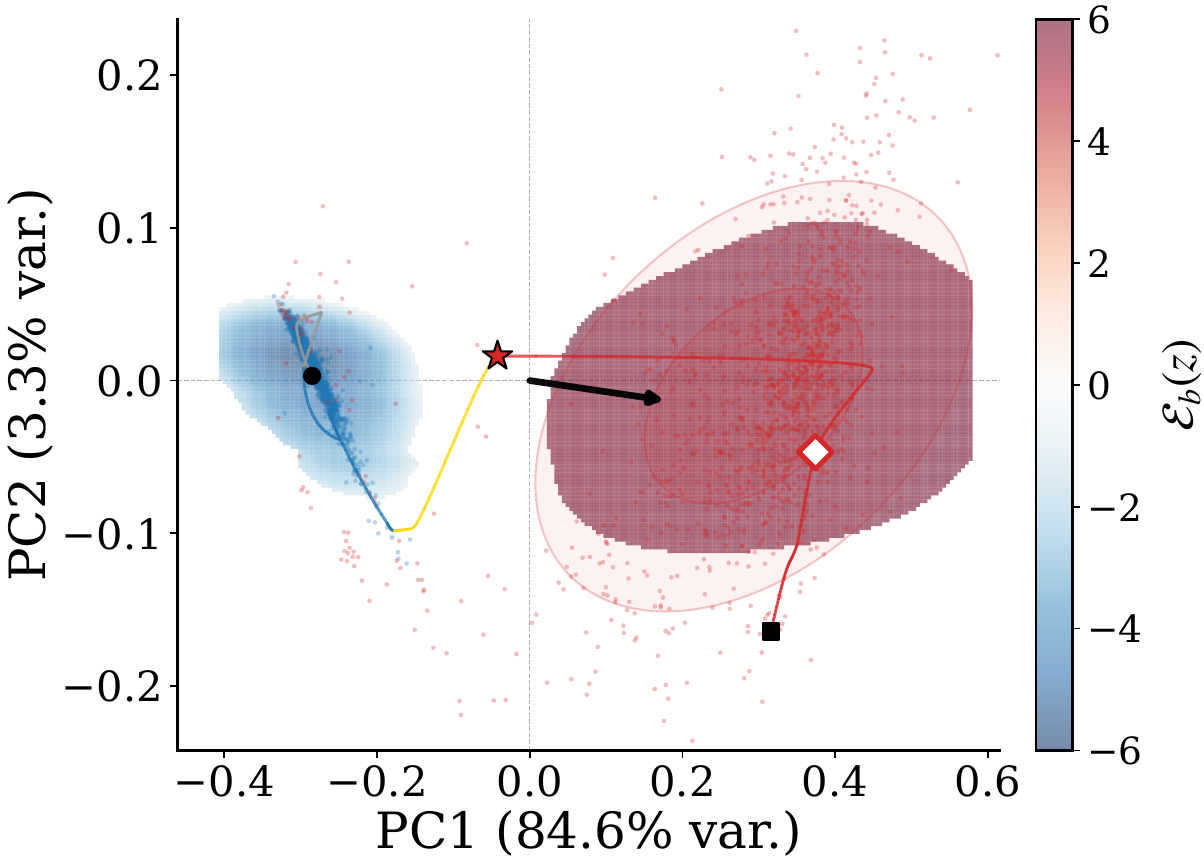}
        \vspace{-1mm}

        {\scriptsize (b) Example 2}
    \end{minipage}
    \hfill
    \begin{minipage}[t]{0.315\textwidth}
        \centering
        \includegraphics[width=\linewidth]
        {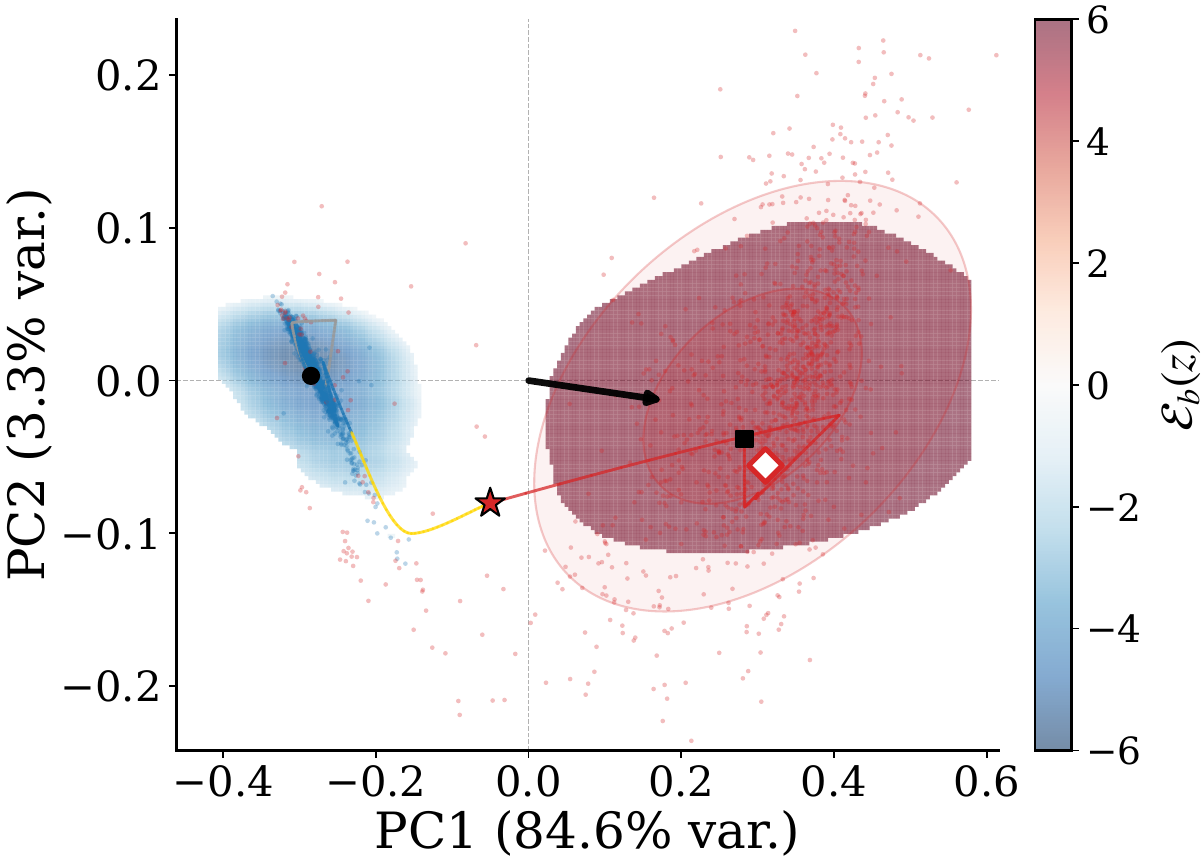}
        \vspace{-1mm}

        {\scriptsize (c) Example 3}
    \end{minipage}

    \vspace{1.5mm}

    \begin{minipage}[t]{0.315\textwidth}
        \centering
        \includegraphics[width=\linewidth]
        {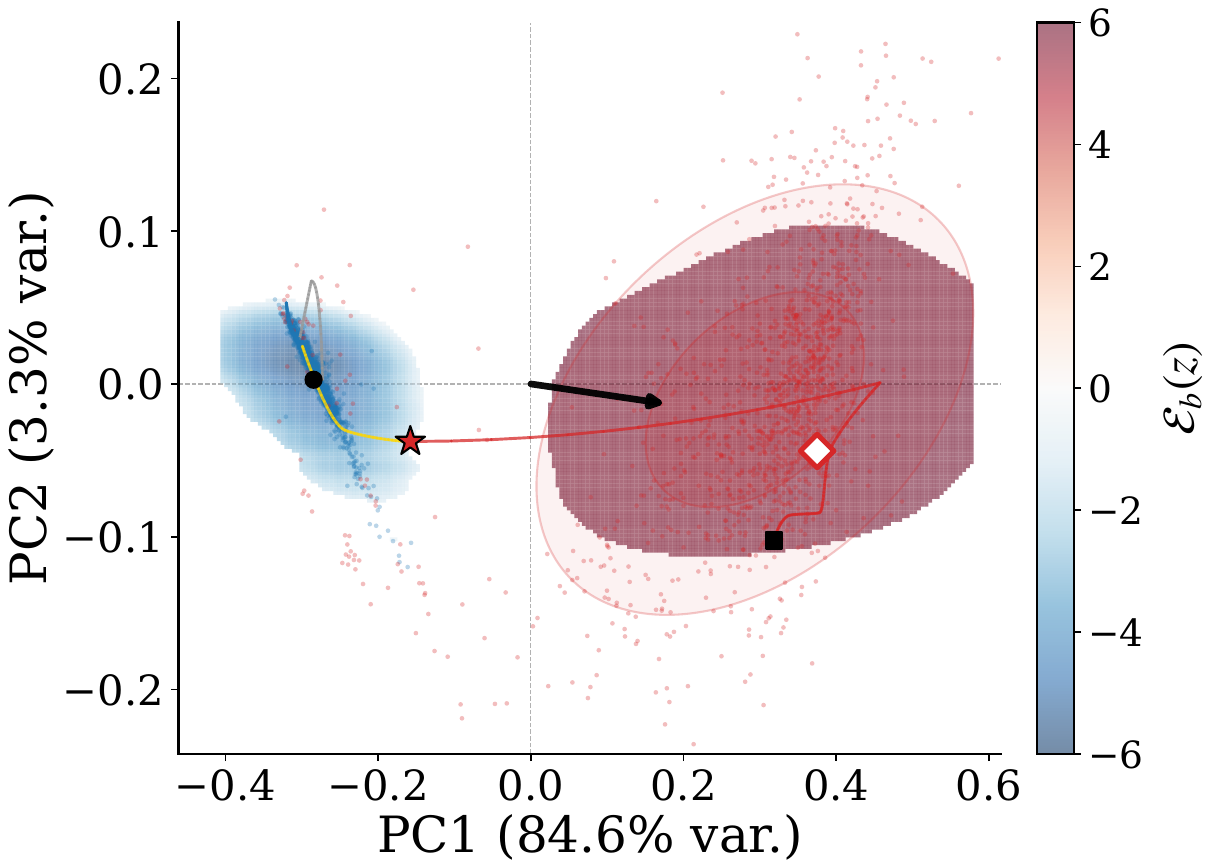}
        \vspace{-1mm}

        {\scriptsize (d) Example 4}
    \end{minipage}
    \hfill
    \begin{minipage}[t]{0.315\textwidth}
        \centering
        \includegraphics[width=\linewidth]
        {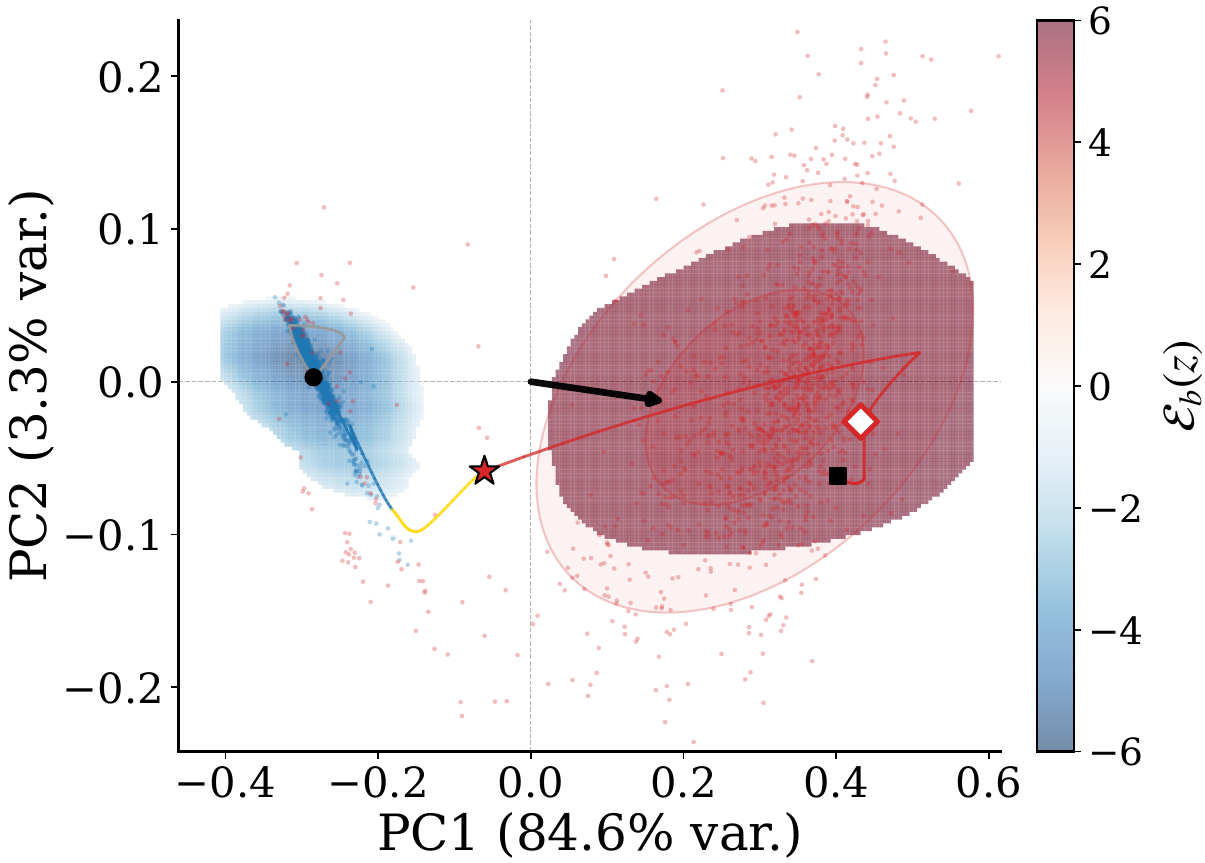}
        \vspace{-1mm}

        {\scriptsize (e) Example 5}
    \end{minipage}
    \hfill
    \begin{minipage}[t]{0.315\textwidth}
        \centering
        \includegraphics[width=\linewidth]
        {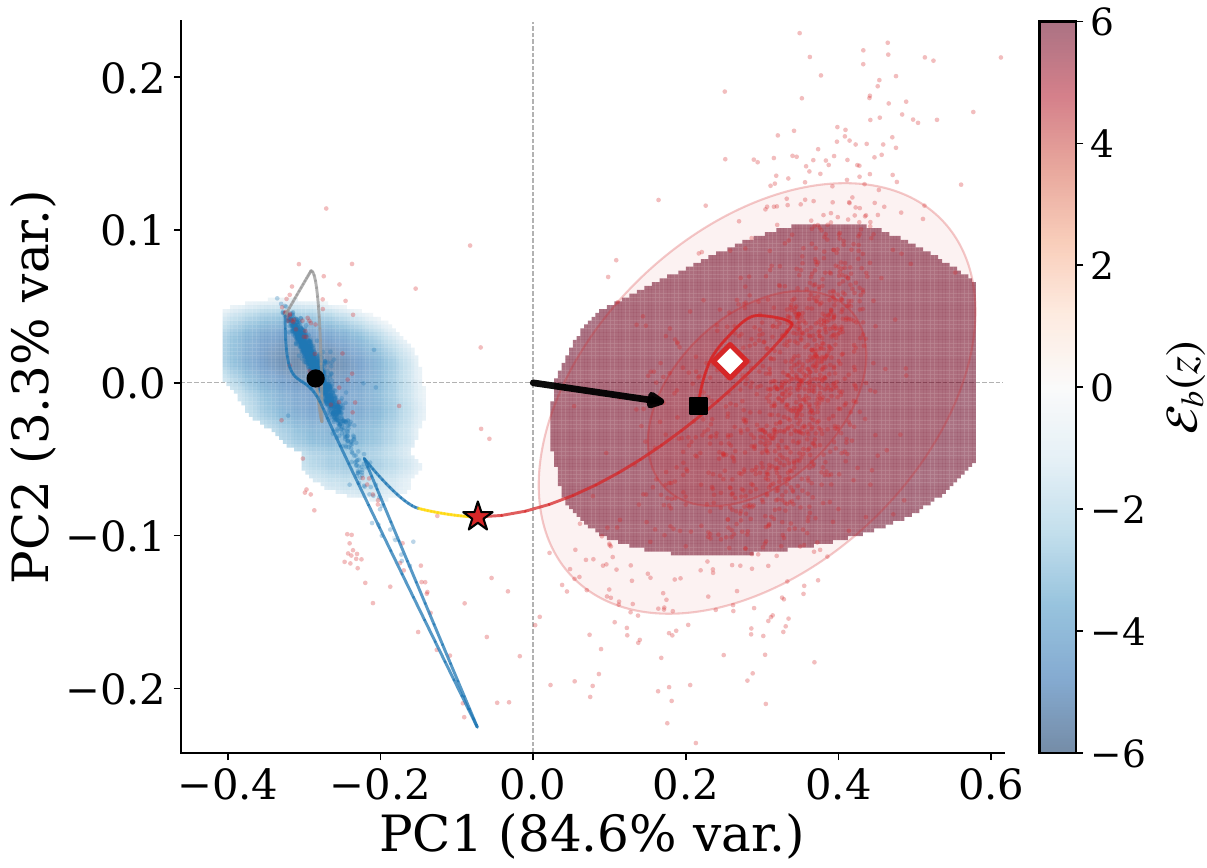}
        \vspace{-1mm}

        {\scriptsize (f) Example 6}
    \end{minipage}

    \caption{
    Matched repetition trajectories in NOC space.  The NOC representation
    emphasizes the transition from a compact behavior-negative contribution
    state to a behavior-positive contribution state.  Although the exact path
    differs across instances, the established repetition span and its mean are
    consistently located on the positive side of the landscape.
    }
    \label{fig:rep_noc_trajectory_gallery}
\end{figure*}

Figure~\ref{fig:rep_noc_trajectory_gallery} shows the corresponding repetition
trajectories in NOC space.  Compared with ACT, the NOC chart is more strongly
dominated by its first coordinate in these examples, producing a compact
negative region and a clearly separated positive contribution region.

The agreement between ACT and NOC is a representation-space robustness result,
not a claim that their coordinates or trajectory shapes should be identical.
ACT describes which behavior-associated coordinates are active, whereas NOC
describes how strongly those coordinates contribute to the realized MLP update.
A trajectory can therefore curve differently in the two spaces while preserving
the same qualitative transition from a negative or transitional region to a
positive region.  The stronger first-coordinate organization in NOC suggests
that the functional expression of repetition may be more nearly one-dimensional
than its raw activation dynamics.

\subsection{Sycophancy Trajectories in Activation Space}

Sycophancy is not a periodic decoding attractor.  Its trajectory should
therefore not be expected to exhibit the same narrow basin-crossing dynamics as
repetition.  Instead, the trajectory records how response representations move
relative to a broader preference-related region as the answer develops.

\begin{figure*}[t]
    \centering

    \begin{minipage}[t]{0.315\textwidth}
        \centering
        \includegraphics[width=\linewidth]
        {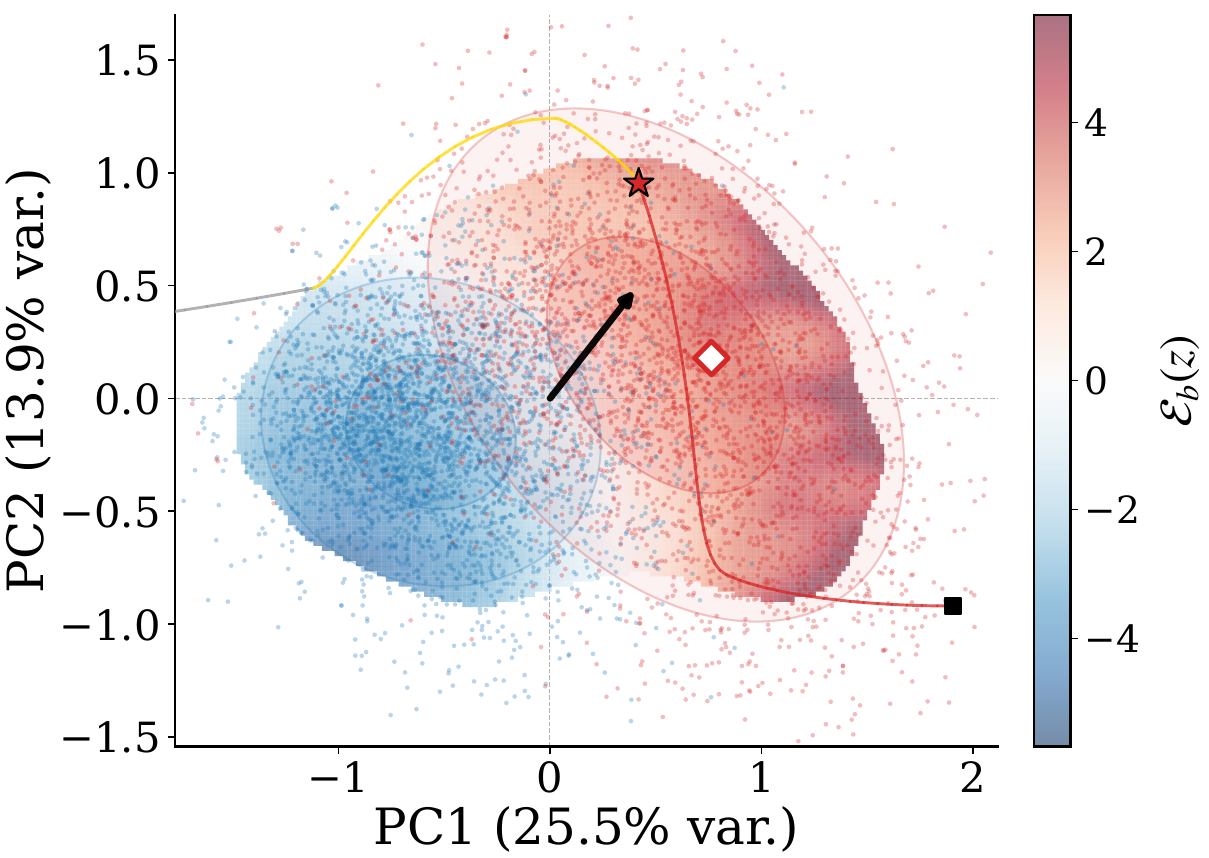}
        \vspace{-1mm}

        {\scriptsize (a) Example 1}
    \end{minipage}
    \hfill
    \begin{minipage}[t]{0.315\textwidth}
        \centering
        \includegraphics[width=\linewidth]
        {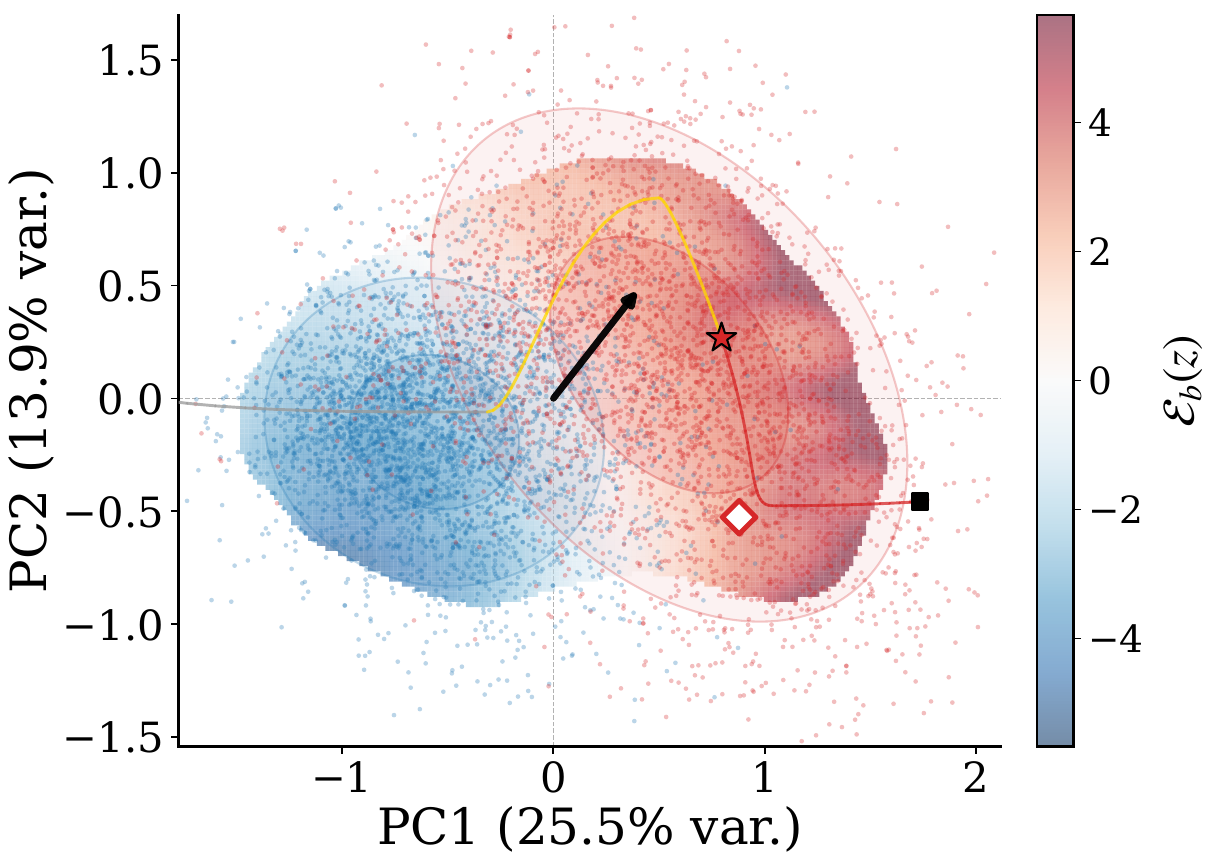}
        \vspace{-1mm}

        {\scriptsize (b) Example 2}
    \end{minipage}
    \hfill
    \begin{minipage}[t]{0.315\textwidth}
        \centering
        \includegraphics[width=\linewidth]
        {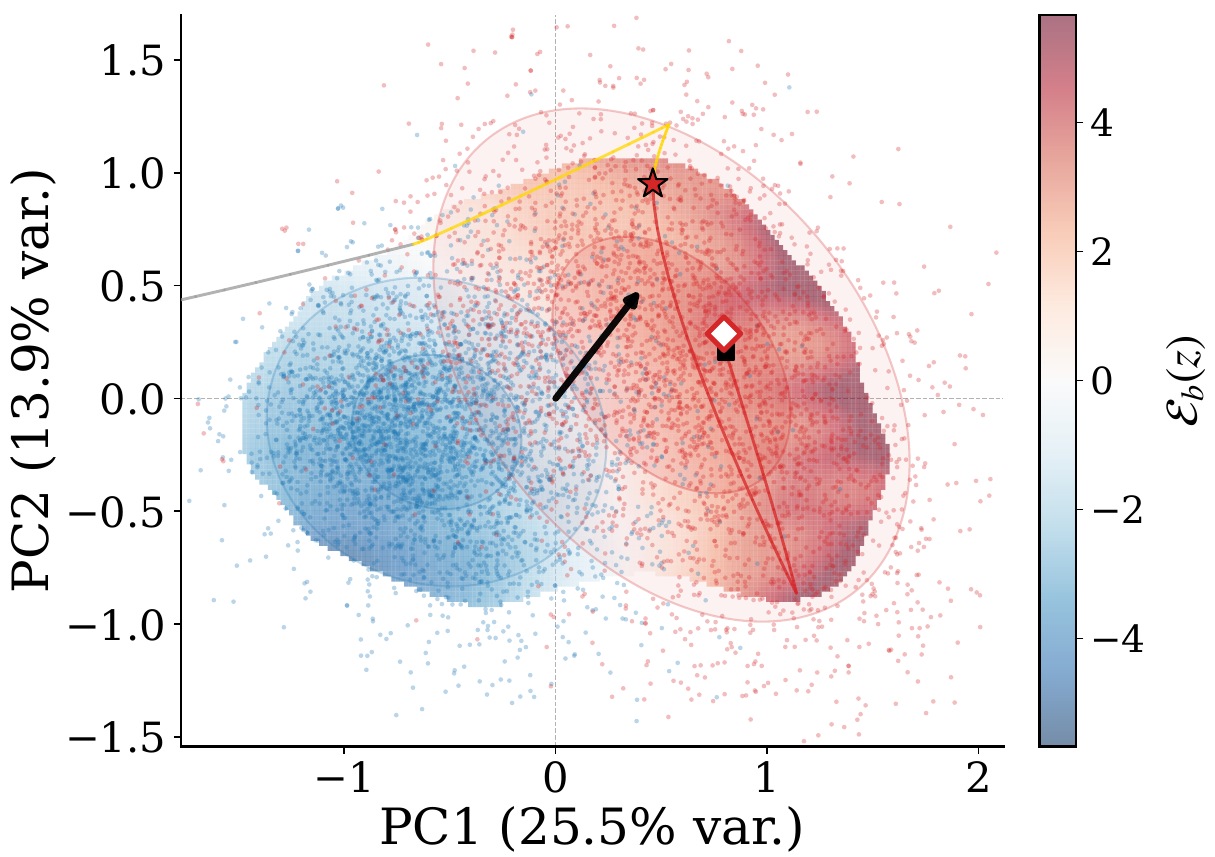}
        \vspace{-1mm}

        {\scriptsize (c) Example 3}
    \end{minipage}

    \caption{
    Sycophancy trajectories in ACT space.  Prefix and pre-onset states traverse
    the overlapping or transitional part of the chart, while the annotated
    sycophantic spans evolve within the positive-density region.  Compared with
    repetition, the trajectories are more heterogeneous and two-dimensional,
    consistent with sycophancy being a broader response-level alignment mode
    rather than a periodic decoding attractor.
    }
    \label{fig:syco_act_trajectory_gallery}
\end{figure*}

Figure~\ref{fig:syco_act_trajectory_gallery} shows three ACT examples.  The
remaining ACT example is presented together with its matched NOC trajectory in
Figure~\ref{fig:syco_cross_space_trajectory}.

The ACT trajectories show that the onset marker need not coincide with the first
entry into a positive-density region.  The pre-onset response can already move
toward or through that region, while the annotation marks the point at which
sycophancy becomes linguistically identifiable.  This distinction is expected:
the internal state may begin organizing an agreement-seeking response before a
specific sycophantic phrase appears.

Unlike repetition, sycophancy trajectories are not monotone along PC1 and do
not remain confined to a narrow positive basin.  They often curve through the
positive region, reflecting the fact that a response must simultaneously encode
semantic content, linguistic realization, and user-directed preference.  The
appropriate conclusion is therefore not that sycophancy follows one universal
path, but that its annotated span is organized relative to a reproducible
positive behavioral region.

\subsection{Matched Sycophancy Trajectory Across ACT and NOC}

\begin{figure*}[t]
    \centering

    \begin{minipage}[t]{0.485\textwidth}
        \centering
        \includegraphics[width=\linewidth]
        {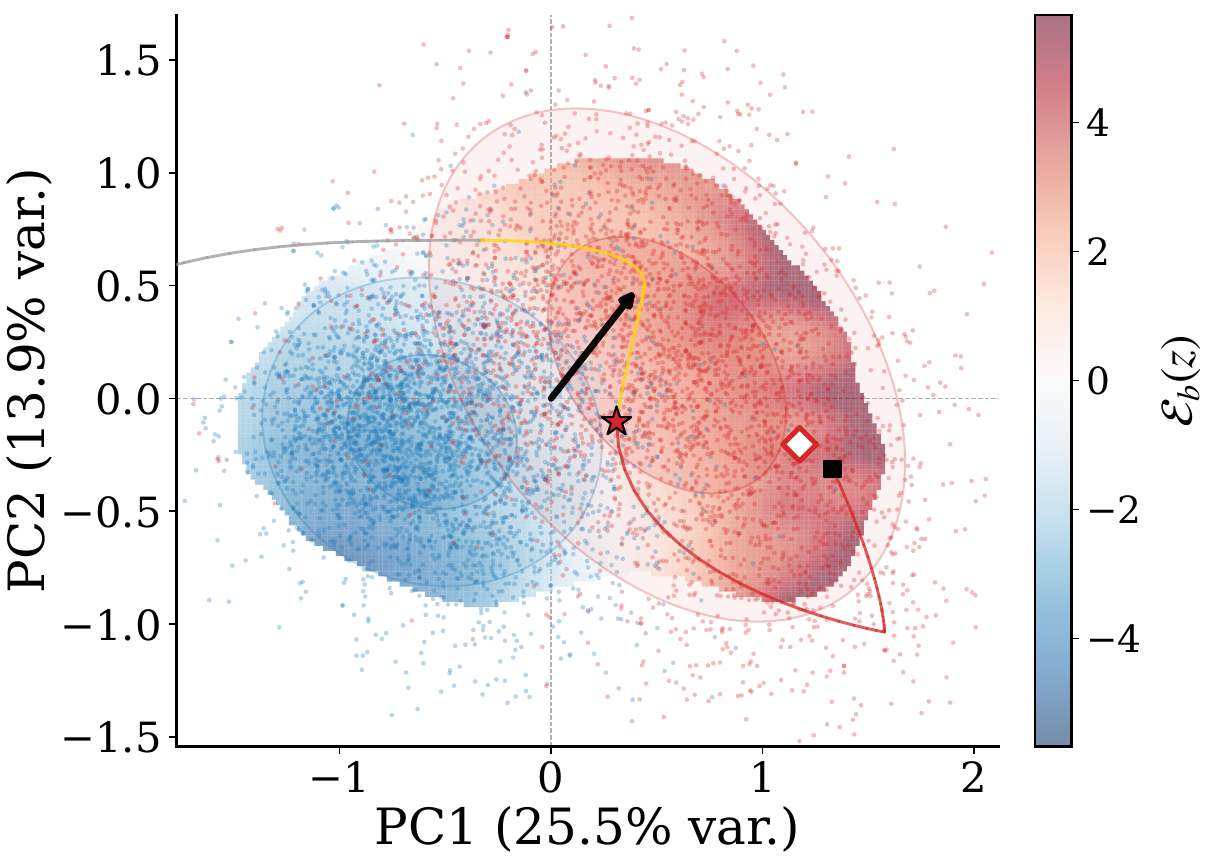}
        \vspace{-1.5mm}

        {\small (a) ACT space}
    \end{minipage}
    \hfill
    \begin{minipage}[t]{0.485\textwidth}
        \centering
        \includegraphics[width=\linewidth]
        {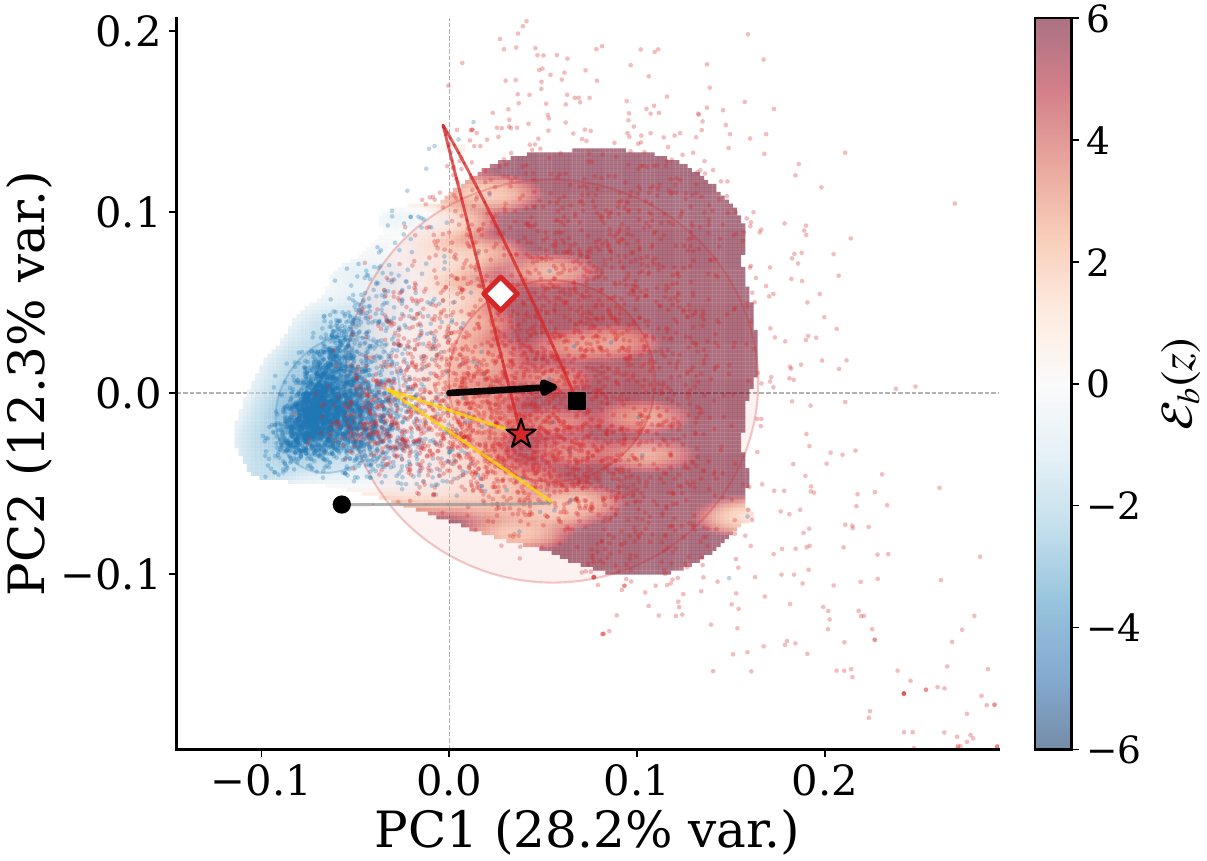}
        \vspace{-1.5mm}

        {\small (b) NOC space}
    \end{minipage}

    \caption{
    Matched ACT and NOC trajectories for the same sycophancy instance.
    The two spaces produce different local path shapes because they encode
    different aspects of the computation.  Nevertheless, in both spaces the
    sycophantic span and its mean are located in the behavior-positive region,
    supporting the representation robustness of the learned sycophancy chart.
    }
    \label{fig:syco_cross_space_trajectory}
\end{figure*}

Figure~\ref{fig:syco_cross_space_trajectory} compares ACT and NOC trajectories
for the same sycophancy instance.  This matched comparison isolates the effect
of representation choice from differences between examples.

The ACT trajectory emphasizes the evolution of the model's internal state,
whereas the NOC trajectory emphasizes the functional contribution of the
selected coordinates.  Their numerical coordinates and curvature are not
directly comparable.  The relevant invariant is qualitative: in both
representations, the annotated sycophantic span is associated with the
positive-density side of the learned chart.

The NOC landscape is more compressed and exhibits a different dominant
orientation.  This indicates that a broad activation-space trajectory can map
to a more compact pattern of functional contribution.  Such differences
motivate treating ACT and NOC as complementary views rather than expecting one
to reproduce the other point by point.

%% file: sections__supplement__n_Sycophancy_Path.tex
\section{Base-Anchored Sycophancy Paths Across Post-Training Regimes}
\label{app:base-anchored-sycophancy-paths}

The main post-training analysis compares behavioral charts at the level of
subspace geometry, occupancy, and behavioral readout.  We complement these
aggregate measurements with a controlled token-level analysis of how the same
candidate answers are represented across checkpoints.

For each example, we fix the prompt and a paired truthful and sycophantic
candidate response.  The two candidates are teacher-forced through the Base,
RL-up, SFT-up, and SFT$\rightarrow$RL-up checkpoints.  Thus, within each row of
Figure~\ref{fig:base-anchored-syco-path-grid}, the textual input and candidate
answers are identical; only the post-training state of the model changes.

This design isolates training-induced representational change from lexical,
semantic, and sampling variation.  In particular, a displacement of the shared
prompt state or of either candidate path cannot be attributed to a different
prompt, a different response, or stochastic generation.

\subsection{Base-Anchored Coordinate System}

All trajectories are displayed in a single coordinate system defined by the
Base model.  Let
\[
    U_{0,b,\mathrm{ACT}}
\]
denote the Base-model ACT chart for sycophancy, and let
\[
    \mu_{0,b,\mathrm{ACT}},
    \qquad
    \sigma_{0,b,\mathrm{ACT}}
\]
denote the corresponding Base-model standardization statistics.  For checkpoint
$m$, candidate type
\[
    c\in\{\mathrm{syco},\mathrm{truth}\},
\]
and candidate-prefix position $t$, we compute the anchored coordinate
\begin{equation}
\label{eq:base-anchored-syco-path}
    z_{m,c,t}^{(0)}
    =
    U_{0,b,\mathrm{ACT}}^{\top}
    \left(
    \frac{
        \phi_{m,b}^{(\mathrm{ACT})}
        \left(x,y_{\leq t}^{c}\right)
        -
        \mu_{0,b,\mathrm{ACT}}
    }{
        \sigma_{0,b,\mathrm{ACT}}+\epsilon
    }
    \right).
\end{equation}
The superscript $(0)$ emphasizes that every checkpoint is
evaluated using the Base coordinate scaffold, standardization, and PCA basis.

The background distribution is also fixed across panels.  The pale red points
are Base-model sycophancy-positive samples, and the pale blue points are
Base-model sycophancy-negative samples.  Consequently, the four columns are
directly comparable: movement across panels reflects how a checkpoint represents
the same prompt and candidates relative to the inherited Base chart.

This is an \emph{anchored} visualization rather than a visualization in each
checkpoint's native chart.  It therefore answers whether the Base chart
continues to organize the internal states of a post-trained model.  A trajectory
that leaves the high-density Base support does not necessarily indicate weaker
sycophancy.  Instead, it may indicate that the post-trained model represents the
behavior in a chart that is poorly aligned with the Base chart.

\subsection{Candidate-Path and Preference Encoding}

For each candidate, the line connects the projected representations of
successive answer prefixes.  The red path represents the sycophantic candidate,
and the blue path represents the truthful candidate.  Both begin from the same
prompt representation, shown by the gray star.

For a candidate $c$ at expansion length $t$, define its length-normalized
teacher-forced log-likelihood as
\begin{equation}
\label{eq:length-normalized-candidate-score}
    \ell_{m,c,t}
    =
    \frac{1}{t}
    \sum_{i=1}^{t}
    \log
    p_m
    \left(
        y_i^c
        \mid
        x,y_{<i}^c
    \right).
\end{equation}
The relative preference between the two candidates can be summarized by
\begin{equation}
\label{eq:prefix-preference-margin}
    \Delta_{m,t}
    =
    \ell_{m,\mathrm{syco},t}
    -
    \ell_{m,\mathrm{truth},t}.
\end{equation}
Positive values indicate a relative preference for the sycophantic candidate,
whereas negative values indicate a relative preference for the truthful
candidate.

Node area is a monotone visualization of the corresponding
length-normalized likelihood score during candidate expansion.  Larger red
nodes therefore indicate stronger support for the sycophantic candidate at
that stage, while larger blue nodes indicate stronger support for the truthful
candidate.  Node size encodes preference and should be interpreted separately
from spatial position: position describes internal representation in the
Base chart, whereas node size describes the model's evolving candidate
likelihood.

\subsection{Matched Candidate-Path Gallery}

\begin{figure*}[p]
    \centering

    \begingroup
    \setlength{\tabcolsep}{1pt}
    \renewcommand{\arraystretch}{0.92}

    \begin{tabular}{@{}c@{\hspace{1mm}}c@{\hspace{1mm}}c@{\hspace{1mm}}c@{\hspace{1mm}}c@{}}

        &
        {\scriptsize\bfseries Base}
        &
        {\scriptsize\bfseries RL-up}
        &
        {\scriptsize\bfseries SFT-up}
        &
        {\scriptsize\bfseries SFT$\rightarrow$RL-up}
        \\[-0.5mm]

        \rotatebox[origin=c]{90}{\scriptsize Pair 1}
        &
        \sycopathpanel{
            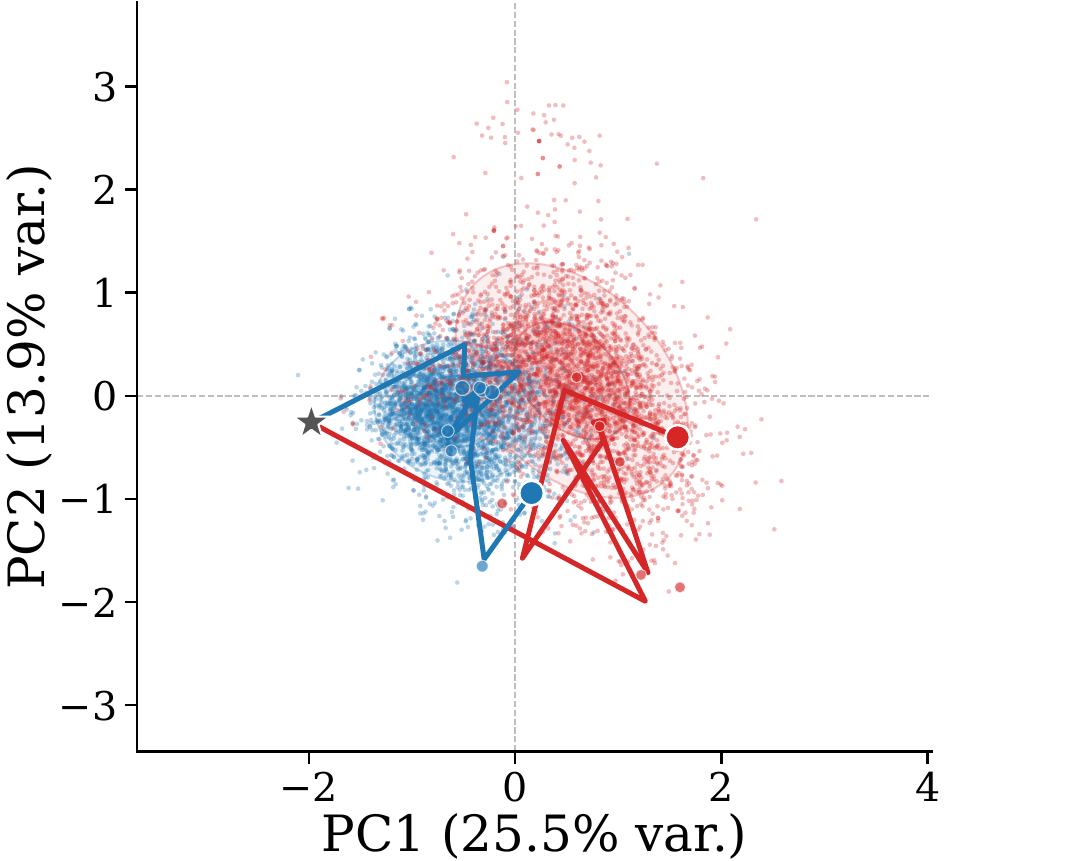
        }
        &
        \sycopathpanel{
            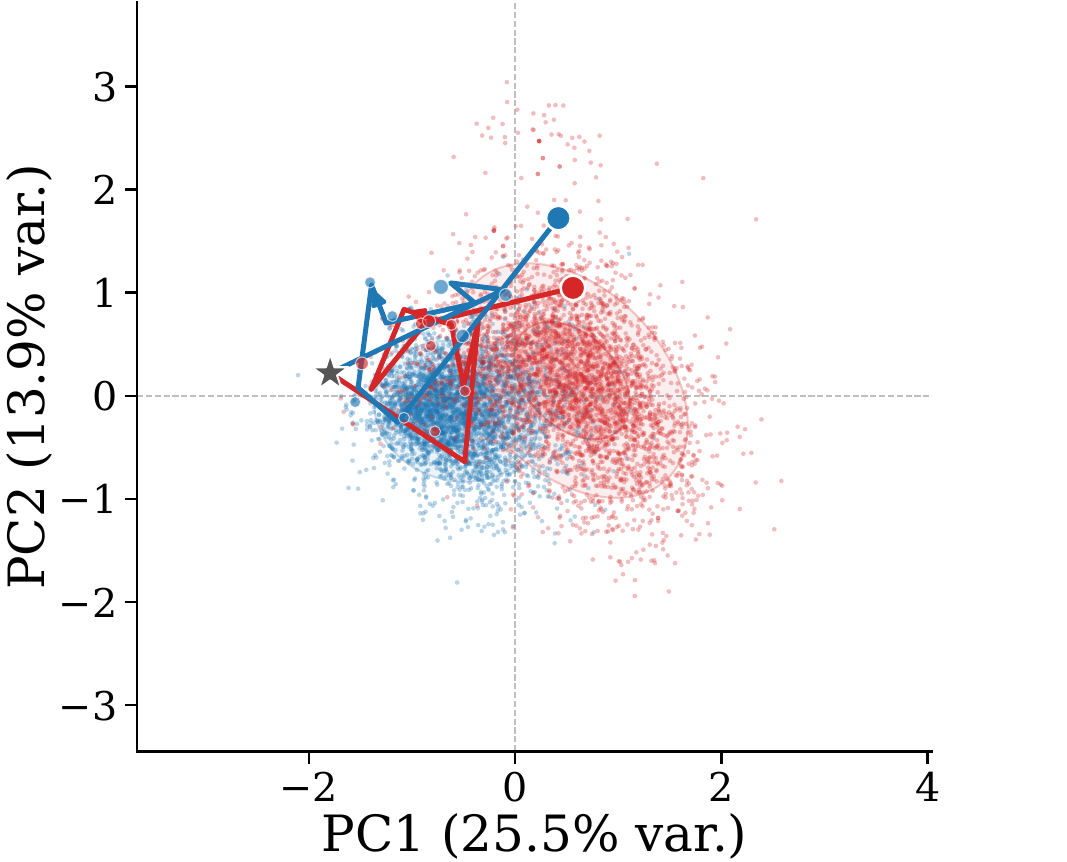
        }
        &
        \sycopathpanel{
            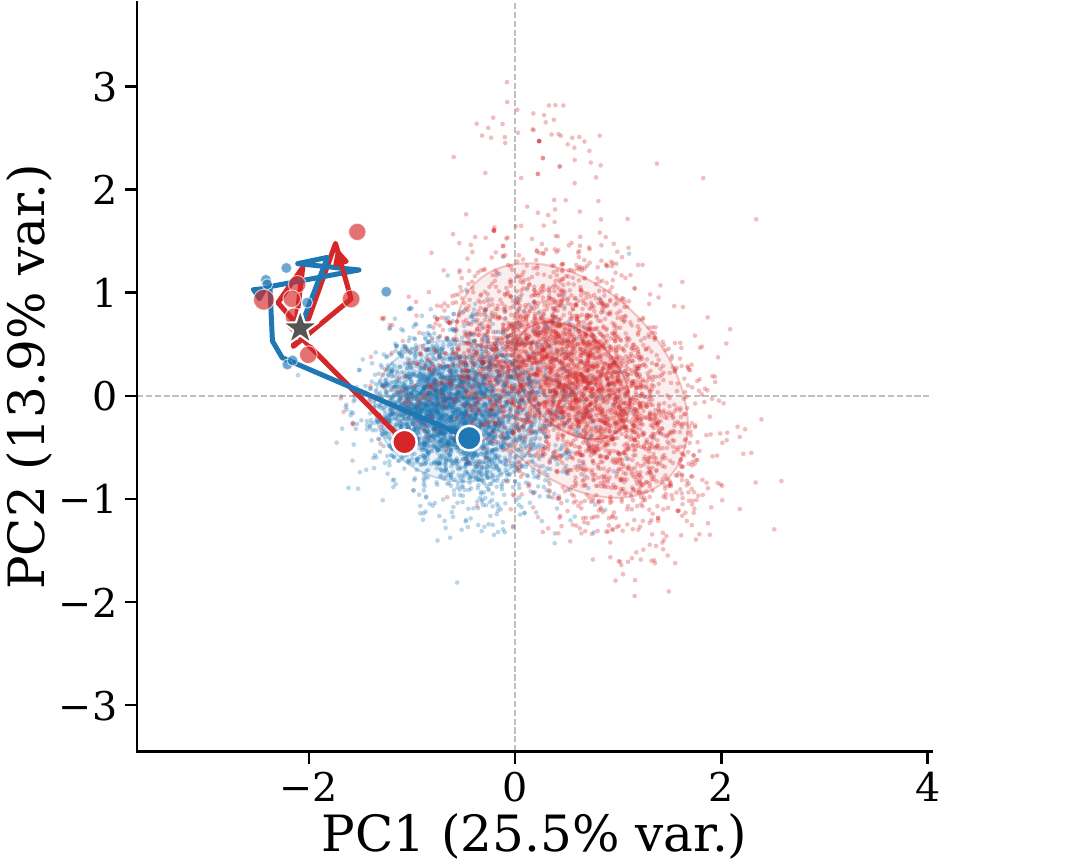
        }
        &
        \sycopathpanel{
            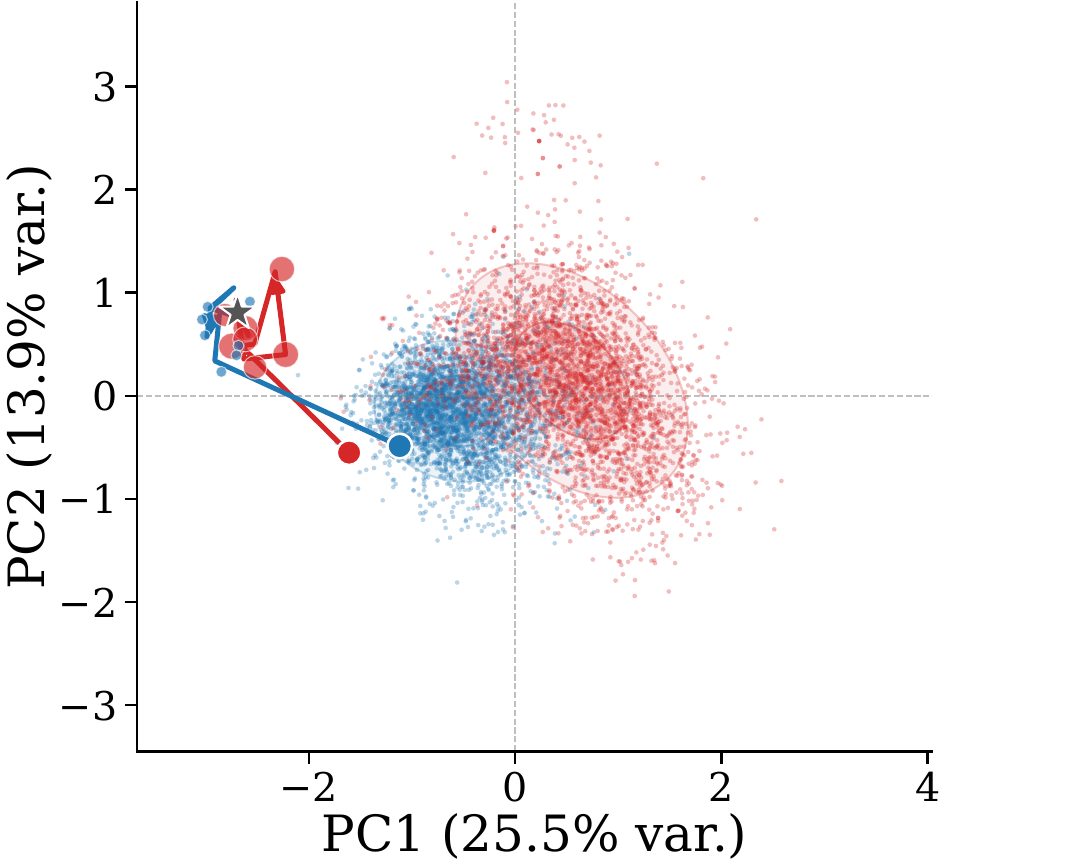
        }
        \\[-1mm]

        \rotatebox[origin=c]{90}{\scriptsize Pair 2}
        &
        \sycopathpanel{
            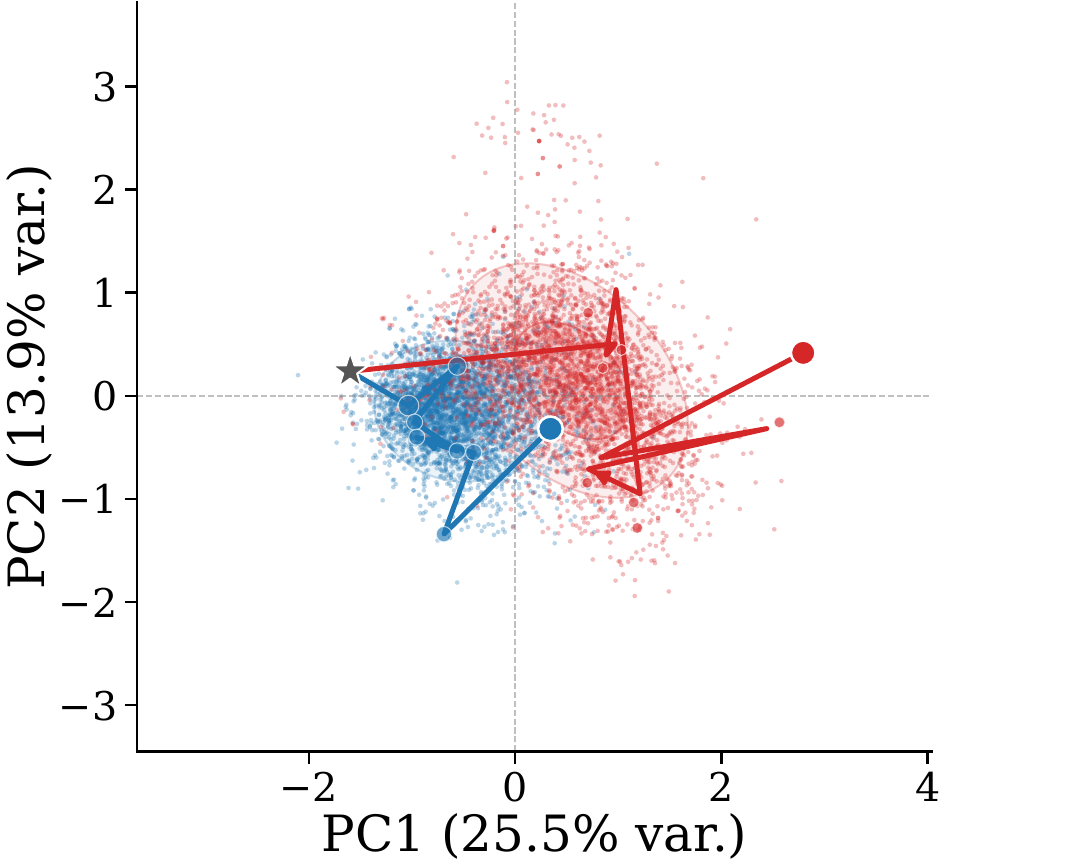
        }
        &
        \sycopathpanel{
            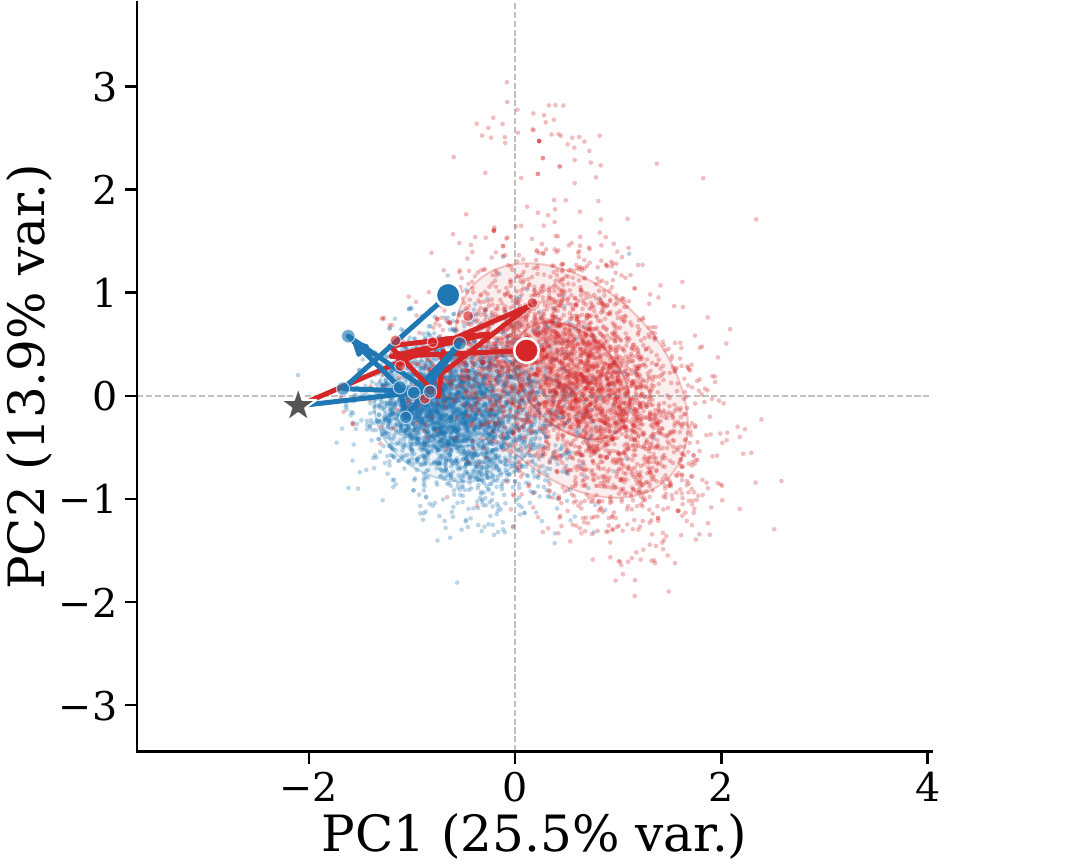
        }
        &
        \sycopathpanel{
            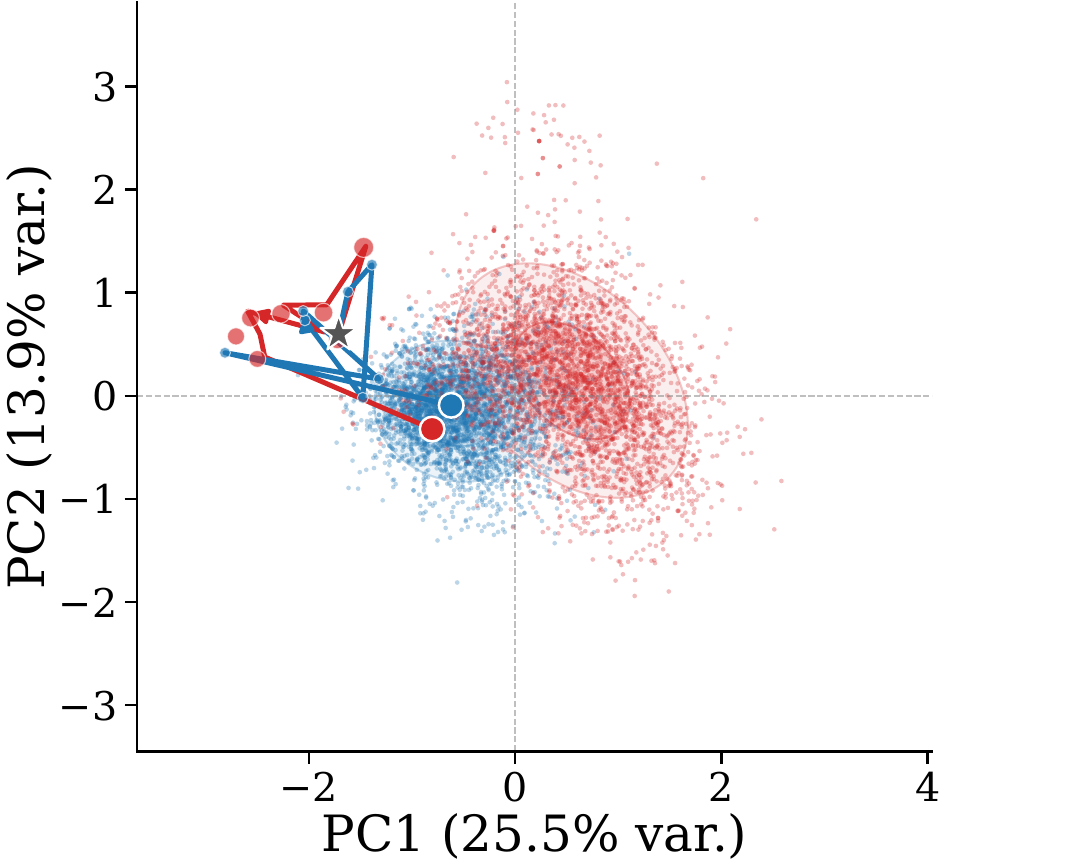
        }
        &
        \sycopathpanel{
            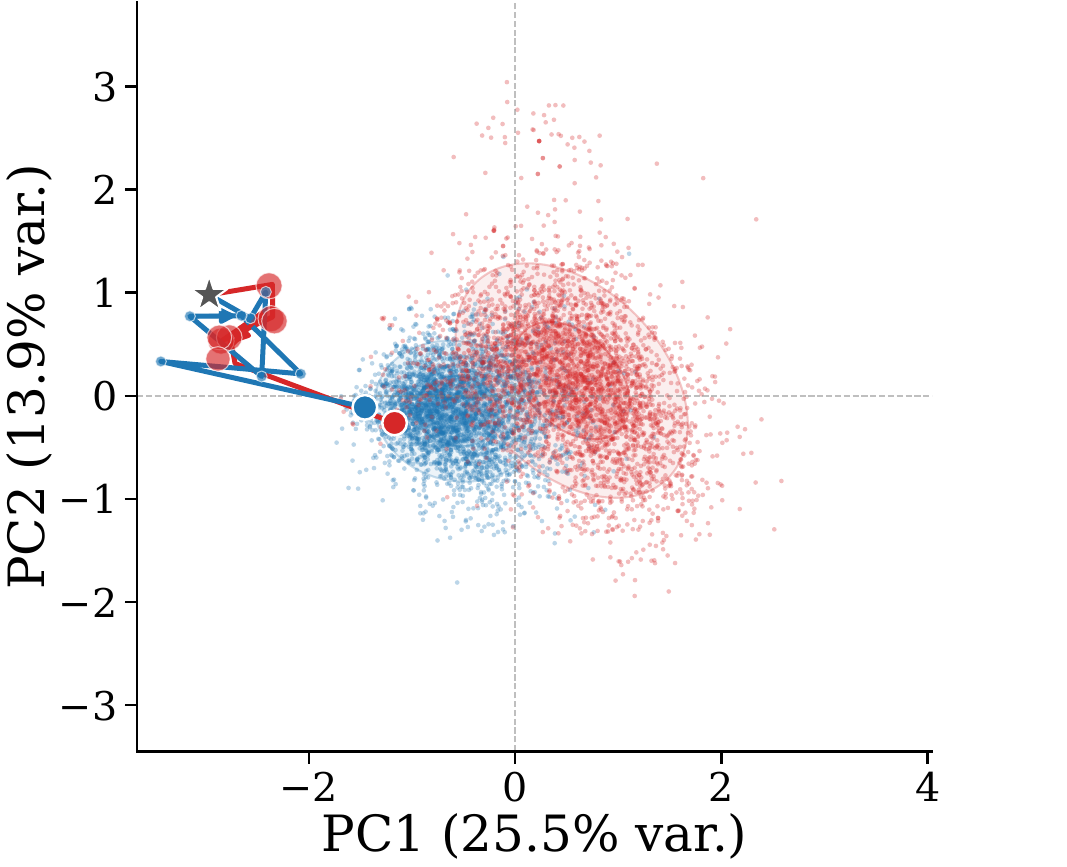
        }
        \\[-1mm]

        \rotatebox[origin=c]{90}{\scriptsize Pair 3}
        &
        \sycopathpanel{
            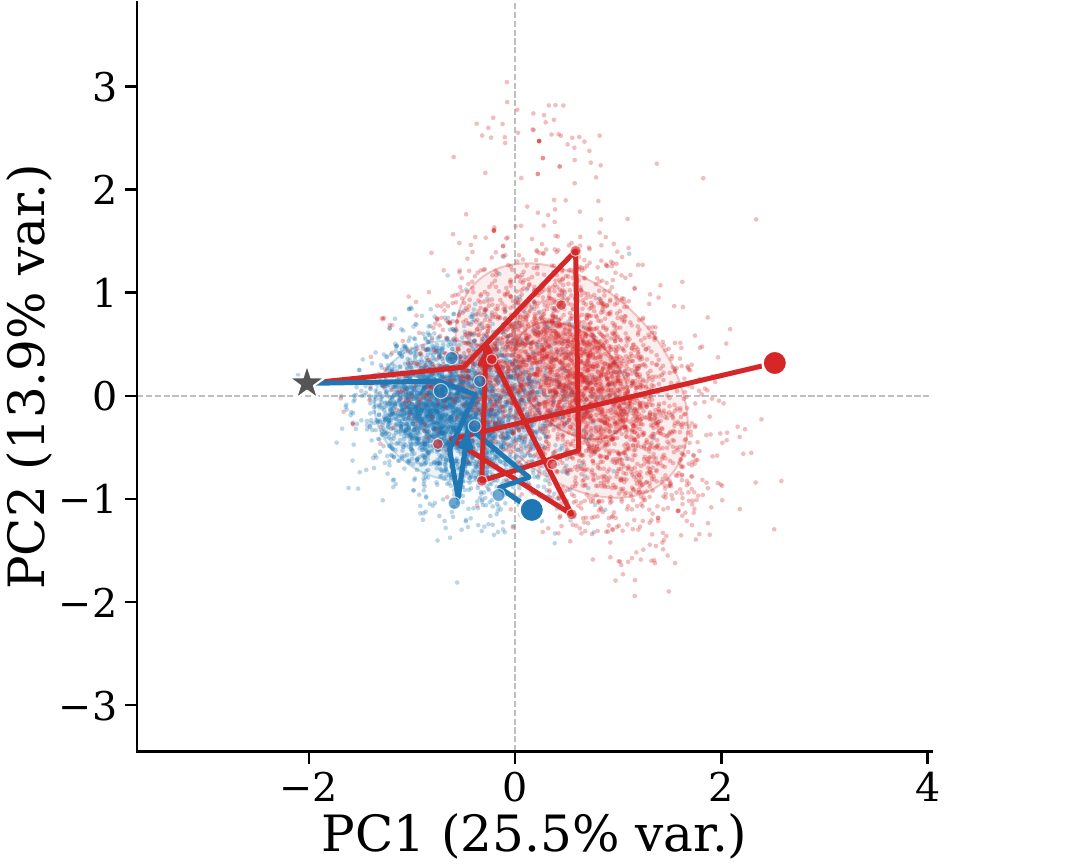
        }
        &
        \sycopathpanel{
            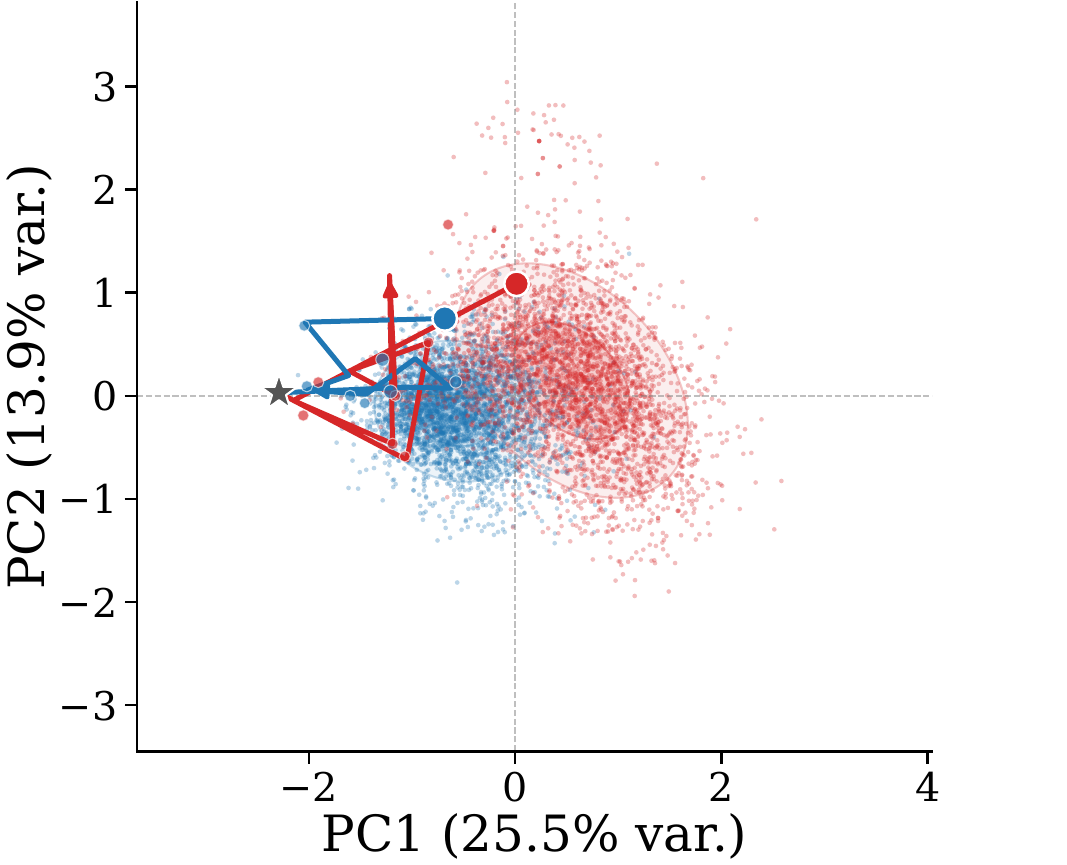
        }
        &
        \sycopathpanel{
            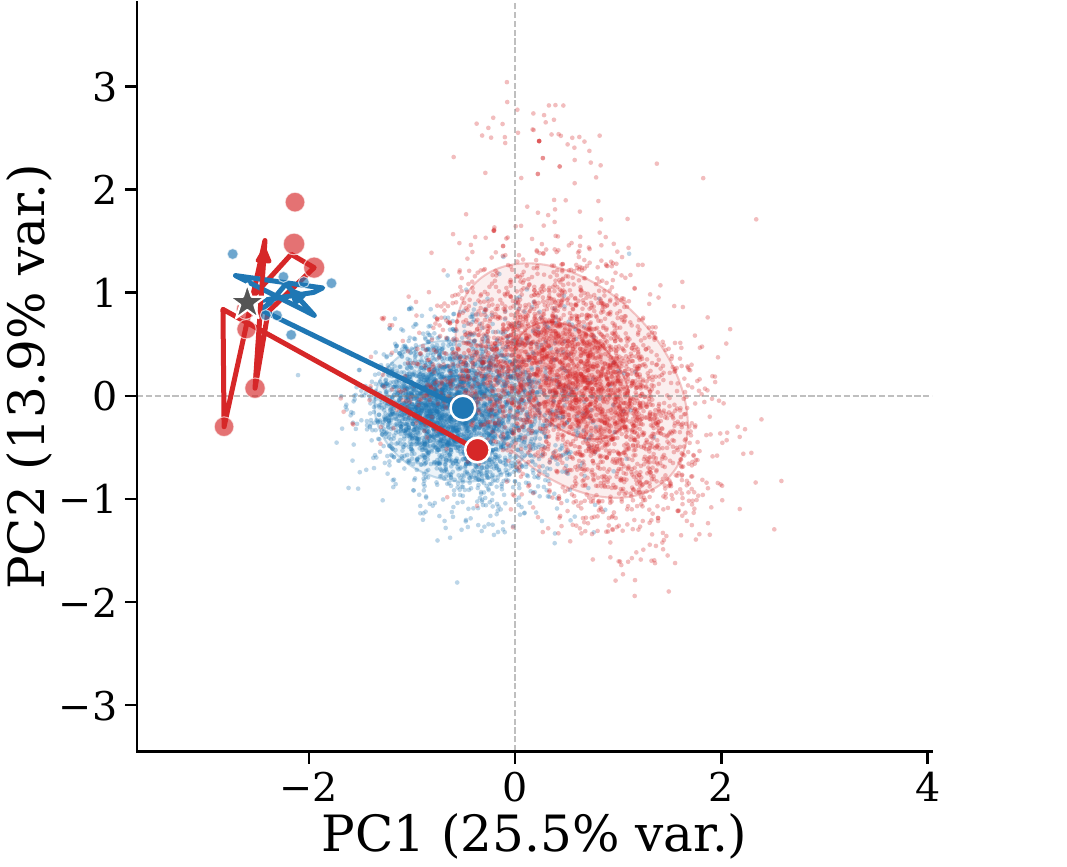
        }
        &
        \sycopathpanel{
            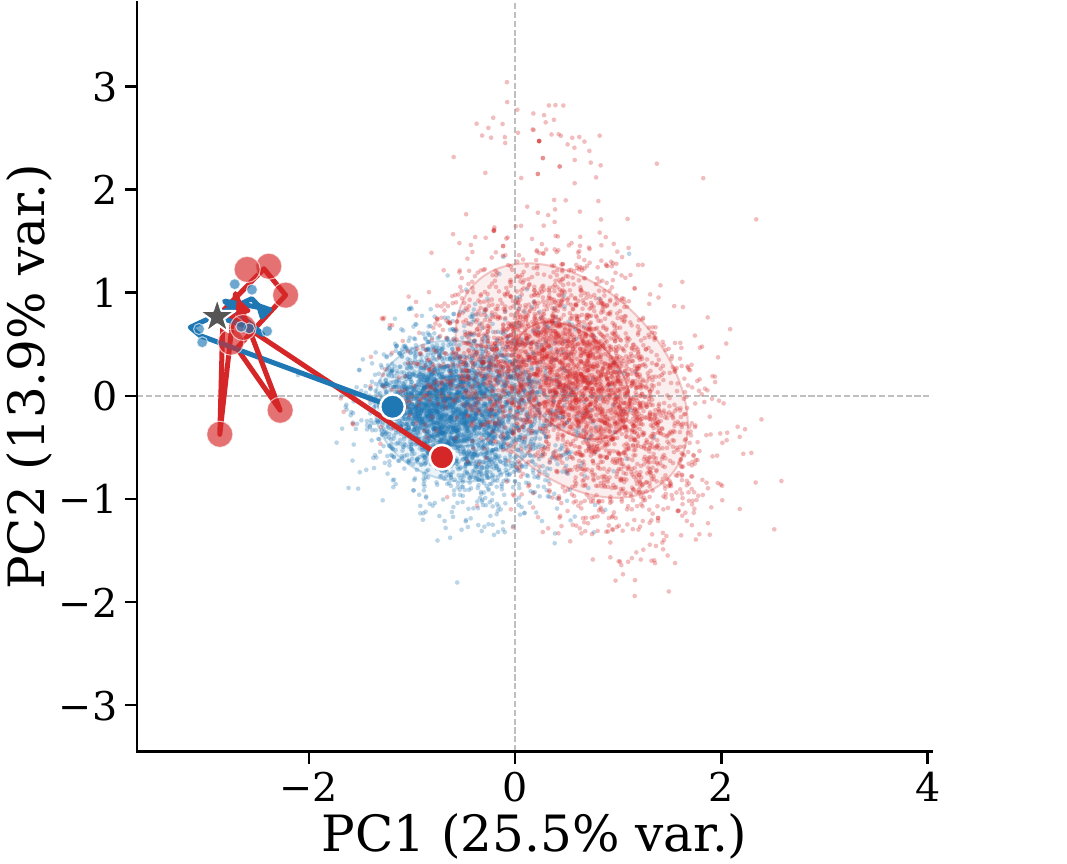
        }
        \\[-1mm]

        \rotatebox[origin=c]{90}{\scriptsize Pair 4}
        &
        \sycopathpanel{
            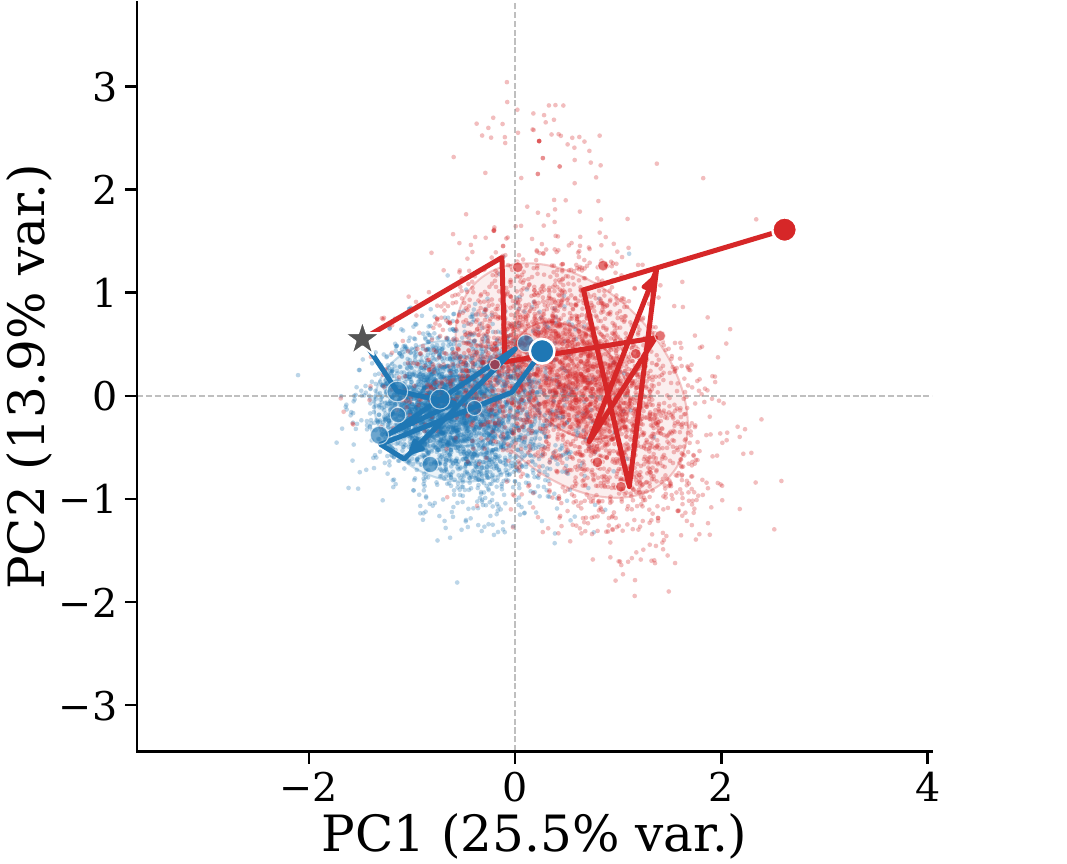
        }
        &
        \sycopathpanel{
            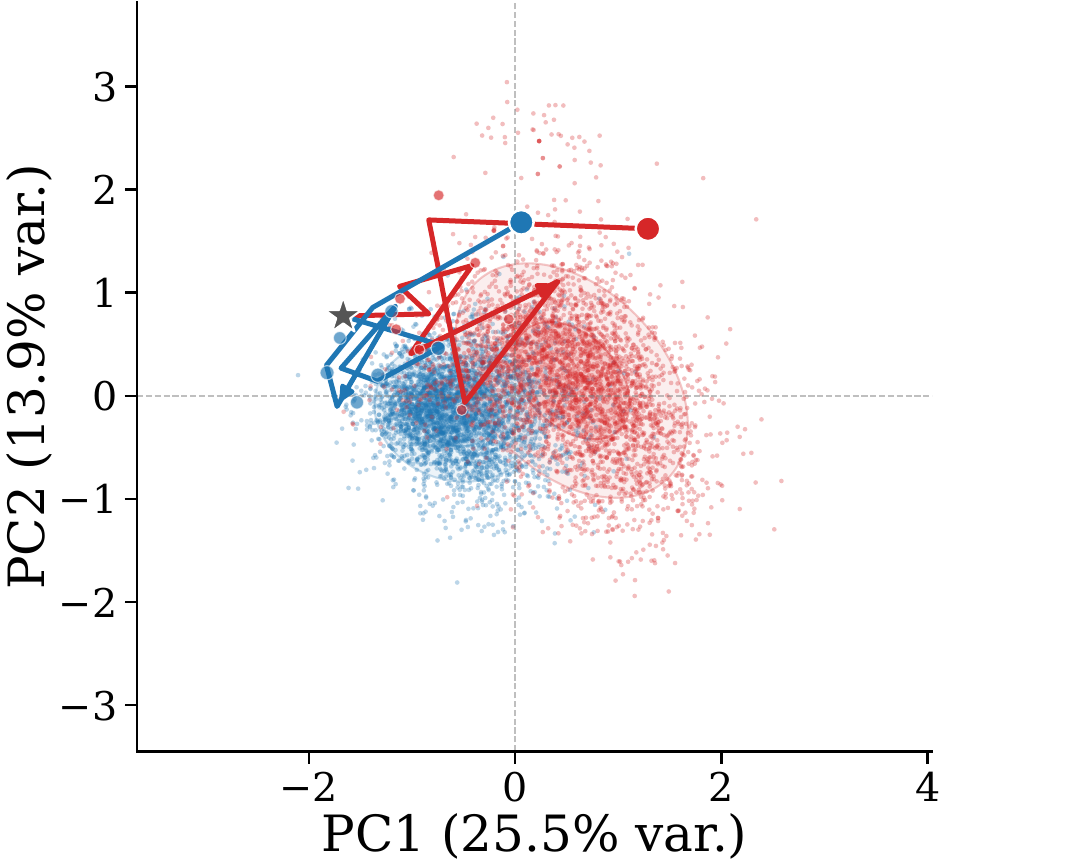
        }
        &
        \sycopathpanel{
            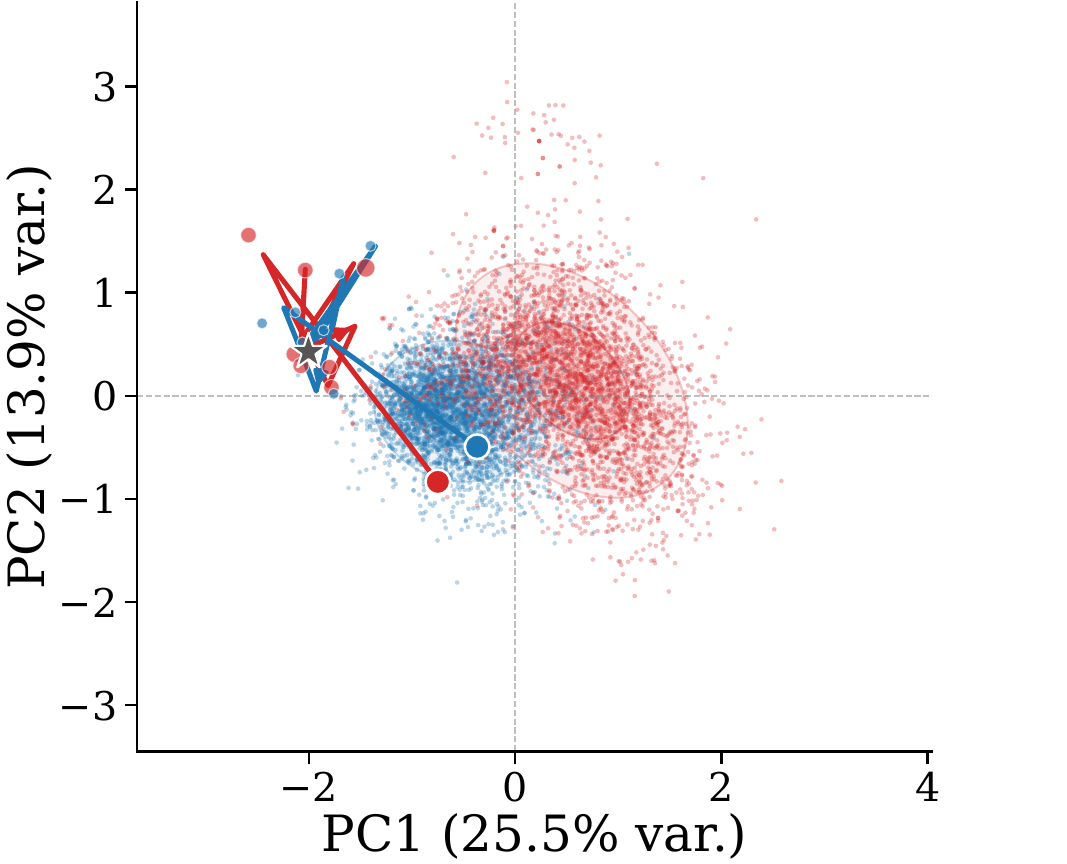
        }
        &
        \sycopathpanel{
            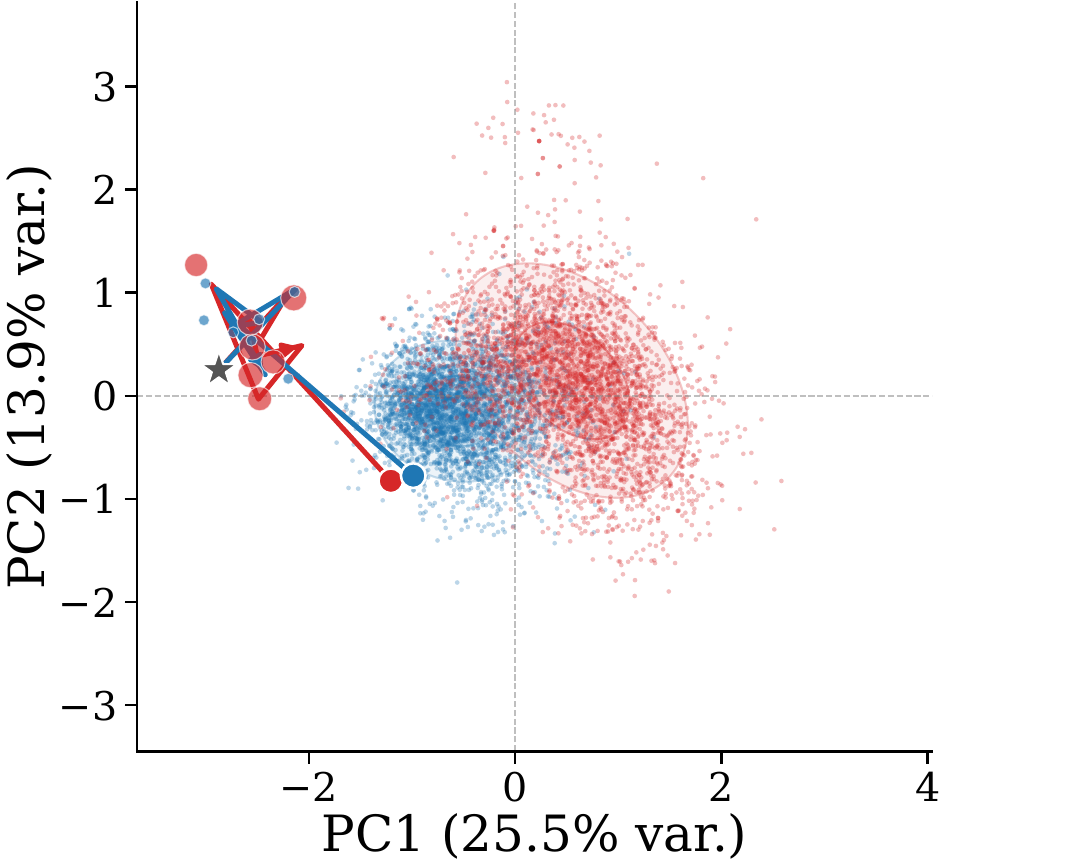
        }

    \end{tabular}
    \endgroup

    \caption{
    Base-anchored ACT candidate-answer paths for four fixed
    truthful--sycophantic pairs across post-training regimes.  Each row uses
    exactly the same prompt, truthful candidate, and sycophantic candidate;
    columns show the Base, RL-up, SFT-up, and
    SFT$\rightarrow$RL-up checkpoints.  All panels use the Base-model selected
    coordinates, standardization statistics, PCA basis, axis orientation, and
    sample background.  Pale red and pale blue points denote Base-model
    sycophancy-positive and sycophancy-negative samples, respectively.  The red
    polyline is the teacher-forced sycophantic candidate, the blue polyline is
    the teacher-forced truthful candidate, and the gray star is their shared
    prompt representation.  Successive nodes correspond to answer-prefix
    representations.  Larger red or blue nodes indicate stronger
    length-normalized likelihood assigned to the corresponding candidate during
    answer expansion.  The fixed background makes representation drift directly
    visible: RL-only trajectories remain largely inside the inherited Base
    support, whereas SFT moves the prompt-to-answer trajectories toward a
    different region of the anchored chart.  The
    SFT$\rightarrow$RL-up paths remain closer to the SFT regime than to the
    original Base regime.
    }
    \label{fig:base-anchored-syco-path-grid}
\end{figure*}

Figure~\ref{fig:base-anchored-syco-path-grid} presents four matched candidate
pairs. Columns correspond to the Base, RL-up, SFT-up, and
SFT$\rightarrow$RL-up checkpoints.  The intended reading order is horizontal:
each row traces the evolution of one fixed prompt--candidate pair through the
post-training sequence.

\subsection{Cross-Pair Interpretation}

Despite substantial instance-level variation, the four rows exhibit a common
training-dependent pattern.

\paragraph{The Base chart organizes Base-model candidate divergence.}
For the Base model, the sycophantic candidates generally move farther toward
the positive side of the chart, whereas truthful candidates remain closer to
the negative or overlapping region.  The exact paths are non-monotonic and can
contain local loops, but their broad organization is consistent with the
positive and negative sample clouds used to estimate the chart.

\paragraph{RL primarily relocates paths within inherited support.}
Across the RL-up column, the candidate paths change their curvature, endpoints,
and preference markers, but remain largely inside the high-density support of
the Base chart.  The Base coordinates therefore continue to provide an
interpretable description of the RL-up model.

This visual pattern is consistent with the quantitative observation that
RL-only checkpoints retain high projector overlap and small principal angles
relative to the Base behavioral chart.  It supports an interpretation in which
RL modifies the distribution of visited chart coordinates and the behavioral
readout applied to those coordinates.

\paragraph{SFT produces a systematic mismatch with the Base chart.}
Across all four pairs, SFT-up changes the shared contextual state and/or the
early answer transition and then sends both candidate paths toward a region
that differs from their Base-model routes.  The repeated upper-left displacement
is unlikely to result from the semantic content of a particular pair because
the same pattern occurs across four distinct prompts and candidate sentences.

The fact that sycophantic preference can increase while the corresponding path
moves away from the Base positive direction is particularly informative.  It
suggests that the Base chart is no longer the native behavioral coordinate
system after SFT.  In terms of the local behavioral model,
\begin{equation}
    p_\theta(y_b=1\mid x)
    \approx
    \sigma
    \left(
        \beta_{\theta,b}^{\top}
        U_{\theta,b}^{\top}
        \bar\phi_\theta(x)
        +
        \tau_{\theta,b}
    \right),
\end{equation}
the SFT result is compatible with a change in
$U_{\theta,b}$ and not merely a translation along the fixed Base readout
direction.

\paragraph{Sequential RL inherits the SFT-induced regime.}
The SFT$\rightarrow$RL-up column is consistently more similar to the SFT-up
column than to the Base or RL-only columns.  The sequential RL stage changes
local path details and evolving candidate preference, but does not return the
representations to the original Base path geometry.

This provides a token-level counterpart to the checkpoint-level Grassmannian
results: SFT establishes a shifted representation regime, and subsequent RL
largely optimizes within the regime inherited from SFT.

\paragraph{Representation and preference are distinct axes of analysis.}
Spatial coordinates and node size encode different quantities.  A path can
remain within the Base support while its likelihood preference changes, as is
often observed under RL-up.  Conversely, a strongly preferred sycophantic
candidate can occupy a region that appears non-positive under the Base chart
after SFT has changed the representation geometry.

The figures therefore provide an intuitive visualization of the
geometry--occupancy--readout decomposition.  Path location reflects anchored
state occupancy; cross-checkpoint path displacement reflects representation
change relative to the Base chart; and node size reflects the evolving
candidate preference induced by the model's readout.

%% file: sections__supplement__o_Coordinate_Statistics.tex
\providecommand{\BAC}{\textsc{bac}}
\providecommand{\BACs}{\textsc{bac}s}


\definecolor{GemmaAccent}{HTML}{4285F4}
\definecolor{QwenAccent}{HTML}{7B61FF}
\definecolor{MistralAccent}{HTML}{F28E2B}
\definecolor{LlamaAccent}{HTML}{59A14F}

\definecolor{GemmaBand}{HTML}{EAF2FF}
\definecolor{QwenBand}{HTML}{F0ECFF}
\definecolor{MistralBand}{HTML}{FFF1E6}
\definecolor{LlamaBand}{HTML}{EAF7EC}

\section{Full Behavioral Anchor Coordinate Statistics}
\label{app:full_bac_statistics}

This appendix reports the complete per-model statistics underlying
Section~\ref{sec:behavioral_anchor_results}.  We use the term
\emph{Behavioral Anchor Coordinates} (\BACs) for the sparse coordinates
selected by the behavior-discriminative classifier.  $N_{\mathrm{BAC}}$
denotes the scaffold size, and $k_{0.9}$ denotes the minimum PCA rank required
to explain $90\%$ of variance.

For repetition, $G^{\mathrm{uniF}}$ measures reliable behavior separation in
the shared chart, $G^{\mathrm{occ}}$ measures class-conditional chart
occupancy, and $G^{\mathrm{causal}}$ measures the local behavioral response to
intervention.  $T_{\mathrm{base}}$ is the behavior probability at the
no-intervention point.  All reported repetition strength results use the
leading shared coordinate, $q=1$.  Pairwise CCA values $\rho_1$ and
$\bar\rho_q$ are matrix-valued cross-model results and are reported separately
as CCA heatmaps rather than repeated in the per-model tables.

For sycophancy, PC1 and Top-3 denote explained-variance ratios, $k_{0.9}$
denotes the $90\%$-variance dimension, and ``Cos.'' denotes the reported signed
cosine similarity between PC1 and the supervised classifier direction.  Each
sycophancy result uses 6,678 examples.  Since the number of extracted features
and fitted PCA components equals $N_{\mathrm{BAC}}$, those redundant columns
are omitted.

\subsection{Initial Repetition Probe}

Table~\ref{tab:app_bac_probe} reports the initial repetition-probe accuracy
and the number of selected Behavioral Anchor Coordinates across model
families.  The results test whether a sparse set of \BACs is sufficient to
identify on-policy repetition before the full geometric analysis.

\begin{table*}[t]
\centering
\small
\setlength{\tabcolsep}{7pt}
\renewcommand{\arraystretch}{1.08}
\begin{tabular}{@{}l c r r r@{}}
\toprule
Model & Type & Parameters & $\#\BAC$ & Accuracy \\
\midrule

\familyheader{GemmaBand}
  {\GemmaLogo\quad\textbf{Gemma family}}{5}
Gemma-3-4B-it
    & Inst. & 4B & 27 & 0.9740 \\

\familyheader{QwenBand}
  {\QwenLogo\quad\textbf{Qwen family}}{5}
Qwen3.5-27B
    & Base & 27B & 41 & 0.9646 \\

\familyheader{MistralBand}
  {\MistralLogo\quad\textbf{Mistral family}}{5}
Mistral-7B-v0.3
    & Base & 7B & 53 & 0.9744 \\
Mistral-Small-3.1-24B-Base-2503 
    & Base & 24B & 29 & 0.9658 \\

\familyheader{LlamaBand}
  {\LlamaLogo\quad\textbf{Llama family}}{5}
Llama-3.1-8B
    & Base & 8B & 26 & 0.9811 \\
Llama-3.1-8B-Instruct
    & Inst. & 8B & 24 & 0.9879 \\
Llama-3.3-70B-Instruct
    & Inst. & 70B & 34 & 0.9853 \\
\midrule
\textbf{Mean}
    & -- & -- & 33.4 & 0.9767 \\
\bottomrule
\end{tabular}
\caption{
Initial repetition probe.  A small set of repetition
Behavioral Anchor Coordinates predicts on-policy repetition with high accuracy
across model families.
}
\label{tab:app_bac_probe}
\end{table*}

\subsection{Repetition Geometry in ACT Space}

In ACT space, causal gain is estimated from additive subspace steering.
Consequently, $G^{\mathrm{causal}}$ should be interpreted as sensitivity to
movement along the estimated activation chart. Table~\ref{tab:app_rep_act} reports the complete per-model repetition-chart
statistics in ACT space.

\begin{table*}[!t]
\centering
\scriptsize
\setlength{\tabcolsep}{3.0pt}
\renewcommand{\arraystretch}{1.05}
\begin{adjustbox}{max width=\textwidth}
\begin{tabular}{@{}l c r r r r r r r@{}}
\toprule
Model & Type &
$G^{\mathrm{uniF}}$ &
$G^{\mathrm{occ}}$ &
$G^{\mathrm{causal}}$ &
$T_{\mathrm{base}}$ &
$\#\BAC$ &
$k_{0.9}$ &
$\log P$ \\
\midrule

\familyheader{GemmaBand}
  {\GemmaLogo\ \textbf{Gemma family}}{9}
Gemma-2-2B
    & Base & 6.8465 & 0.9809 & 0.2942 & 0.8632 & 32 & 13 & 21.4164 \\
Gemma-2-2B-it
    & Inst. & 8.0945 & 0.7189 & 0.2750 & 0.6800 & 28 & 15 & 21.4164 \\
Gemma-3-4B-it
    & Inst. & 7.8447 & 0.7501 & 0.2230 & 0.8696 & 26 & 13 & 22.1096 \\
Gemma-2-9B
    & Base & 6.8896 & 0.8550 & 0.1980 & 0.8704 & 31 & 16 & 22.9205 \\
Gemma-2-9B-it
    & Inst. & 7.7726 & 0.7816 & 0.0771 & 0.8136 & 26 & 14 & 22.9205 \\
Gemma-2-27B
    & Base & 8.8439 & 0.6566 & 0.1306 & 0.8886 & 38 & 14 & 24.0191 \\
Gemma-2-27B-it
    & Inst. & 6.9798 & 0.8328 & 0.1554 & 0.7594 & 34 & 13 & 24.0191 \\

\familyheader{QwenBand}
  {\QwenLogo\ \textbf{Qwen family}}{9}
Qwen2.5-0.5B
    & Base & 7.5528 & 0.7049 & 0.2108 & 0.8562 & 48 & 26 & 20.0301 \\
Qwen2.5-0.5B-Instruct
    & Inst. & 7.1556 & 0.7323 & 0.5994 & 0.8219 & 48 & 25 & 20.0301 \\
Qwen2.5-7B
    & Base & 6.9355 & 0.8142 & 0.2648 & 0.9054 & 37 & 16 & 22.6692 \\
Qwen2.5-7B-Instruct
    & Inst. & 8.0421 & 0.7786 & 0.2072 & 0.8801 & 42 & 21 & 22.6692 \\
Qwen3.5-27B
    & Base & 10.3330 & 0.5082 & 0.0831 & 0.8978 & 45 & 22 & 24.0191 \\
Qwen2.5-32B
    & Base & 6.8184 & 0.8959 & 0.0881 & 0.9163 & 61 & 27 & 24.1890 \\
Qwen2.5-32B-Instruct
    & Inst. & 6.3355 & 1.0557 & 0.2112 & 0.8817 & 44 & 20 & 24.1890 \\

\familyheader{MistralBand}
  {\MistralLogo\ \textbf{Mistral family}}{9}
Mistral-7B-v0.3
    & Base & 10.3875 & 0.6061 & 0.1661 & 0.8833 & 56 & 25 & 22.6692 \\
Mistral-7B-Instruct-v0.3
    & Inst. & 8.3211 & 0.6757 & 0.2480 & 0.8713 & 32 & 18 & 22.6692 \\
Mistral-Nemo-Base-2407
    & Base & 5.6082 & 1.0666 & 0.0776 & 0.8940 & 43 & 21 & 23.2082 \\
Mistral-Nemo-Instruct-2407
    & Inst. & 4.6797 & 1.2299 & 0.1026 & 0.8975 & 42 & 22 & 23.2082 \\
Mistral-Small-3.1-24B-Base-2503 
    & Base & 8.9275 & 0.5998 & 0.1051 & 0.8963 & 28 & 12 & 23.9013 \\
Mistral-Small-3.1-24B-Instruct-2503
    & Inst. & 7.8805 & 0.6776 & 0.3037 & 0.8290 & 34 & 13 & 23.9013 \\

\familyheader{LlamaBand}
  {\LlamaLogo\ \textbf{Llama family}}{9}
Llama-3.1-8B
    & Base & 3.6277 & 1.1152 & 0.0891 & 0.9082 & 26 & 9 & 22.8027 \\
Llama-3.1-8B-Instruct
    & Inst. & 4.3802 & 1.1844 & 0.1823 & 0.7599 & 20 & 8 & 22.8027 \\
Llama-3.3-70B-Instruct
    & Inst. & 4.5069 & 1.1109 & 0.1106 & 0.8477 & 31 & 9 & 24.9718 \\
\midrule
\textbf{Mean}
    & -- & 7.1636 & 0.8405 & 0.1914 & 0.8561 & 37.04 & 17.04 & -- \\
\textbf{Median}
    & -- & 7.1556 & 0.7816 & 0.1823 & 0.8713 & 34 & 16 & -- \\
\bottomrule
\end{tabular}
\end{adjustbox}
\caption{
Complete repetition-chart statistics in ACT space across 23 models.
All rows use $q=1$.  $\log P$ is the natural logarithm of parameter count.
ACT causal gain is estimated from subspace steering.
}
\label{tab:app_rep_act}
\end{table*}

\subsection{Repetition Geometry in NOC Space}

NOC denotes the normalized outgoing-contribution space; the same quantity is
named NOC in the original experimental exports.  In NOC space, causal gain is
estimated using coordinate-level scaling of the selected scaffold.
Table~\ref{tab:app_rep_noc} reports the corresponding repetition-chart
statistics in NOC space.

\begin{table*}[!t]
\centering
\scriptsize
\setlength{\tabcolsep}{3.0pt}
\renewcommand{\arraystretch}{1.05}
\begin{adjustbox}{max width=\textwidth}
\begin{tabular}{@{}l c r r r r r r r@{}}
\toprule
Model & Type &
$G^{\mathrm{uniF}}$ &
$G^{\mathrm{occ}}$ &
$G^{\mathrm{causal}}$ &
$T_{\mathrm{base}}$ &
$\#\BAC$ &
$k_{0.9}$ &
$\log P$ \\
\midrule

\familyheader{GemmaBand}
  {\GemmaLogo\ \textbf{Gemma family}}{9}
Gemma-2-2B
    & Base & 6.2492 & 1.0786 & 0.1204 & 0.8837 & 32 & 13 & 21.4164 \\
Gemma-2-2B-it
    & Inst. & 5.6593 & 0.8830 & 0.2333 & 0.7581 & 28 & 15 & 21.4164 \\
Gemma-3-4B-it
    & Inst. & 8.4143 & 0.6596 & 0.1513 & 0.8759 & 26 & 13 & 22.1096 \\
Gemma-2-9B
    & Base & 4.2760 & 0.9555 & 0.1603 & 0.8875 & 31 & 15 & 22.9205 \\
Gemma-2-9B-it
    & Inst. & 4.2142 & 0.9589 & 0.0477 & 0.8157 & 26 & 13 & 22.9205 \\
Gemma-2-27B
    & Base & 8.8803 & 0.6375 & 0.1021 & 0.8952 & 38 & 15 & 24.0191 \\
Gemma-2-27B-it
    & Inst. & 5.1277 & 1.0103 & 0.1138 & 0.7796 & 34 & 13 & 24.0191 \\

\familyheader{QwenBand}
  {\QwenLogo\ \textbf{Qwen family}}{9}
Qwen2.5-0.5B
    & Base & 9.5037 & 0.6870 & 0.0216 & 0.8695 & 48 & 22 & 20.0301 \\
Qwen2.5-0.5B-Instruct
    & Inst. & 10.1911 & 0.6394 & 0.1244 & 0.8484 & 48 & 22 & 20.0301 \\
Qwen2.5-7B
    & Base & 8.1407 & 0.6886 & 0.1100 & 0.9145 & 37 & 15 & 22.6692 \\
Qwen2.5-7B-Instruct
    & Inst. & 8.2838 & 0.7411 & 0.1344 & 0.8962 & 42 & 20 & 22.6692 \\
Qwen3.5-27B
    & Base & 11.2543 & 0.4830 & 0.0322 & 0.9023 & 45 & 19 & 24.0191 \\
Qwen2.5-32B
    & Base & 7.3041 & 0.8637 & 0.0487 & 0.9187 & 61 & 24 & 24.1890 \\
Qwen2.5-32B-Instruct
    & Inst. & 4.6760 & 1.1594 & 0.0365 & 0.8874 & 44 & 18 & 24.1890 \\

\familyheader{MistralBand}
  {\MistralLogo\ \textbf{Mistral family}}{9}
Mistral-7B-v0.3
    & Base & 12.4879 & 0.4643 & 0.0901 & 0.8935 & 56 & 21 & 22.6692 \\
Mistral-7B-Instruct-v0.3
    & Inst. & 11.2649 & 0.5000 & 0.1500 & 0.8914 & 32 & 17 & 22.6692 \\
Mistral-Nemo-Base-2407
    & Base & 4.5099 & 1.0969 & 0.0247 & 0.8955 & 43 & 20 & 23.2082 \\
Mistral-Nemo-Instruct-2407
    & Inst. & 6.0374 & 0.9124 & 0.0222 & 0.9009 & 42 & 20 & 23.2082 \\
Mistral-Small-3.1-24B-Base-2503 
    & Base & 5.1098 & 0.9186 & 0.0698 & 0.9058 & 28 & 12 & 23.9013 \\
Mistral-Small-3.1-24B-Instruct-2503
    & Inst. & 6.3496 & 0.7855 & 0.0769 & 0.8346 & 34 & 13 & 23.9013 \\

\familyheader{LlamaBand}
  {\LlamaLogo\ \textbf{Llama family}}{9}
Llama-3.1-8B
    & Base & 4.1244 & 1.0646 & 0.0463 & 0.9115 & 26 & 9 & 22.8027 \\
Llama-3.1-8B-Instruct
    & Inst. & 4.6608 & 1.1588 & 0.1229 & 0.7819 & 20 & 7 & 22.8027 \\
Llama-3.3-70B-Instruct
    & Inst. & 4.6690 & 1.1443 & 0.1029 & 0.8598 & 31 & 8 & 24.9718 \\
\midrule
\textbf{Mean}
    & -- & 7.0169 & 0.8474 & 0.0932 & 0.8699 & 37.04 & 15.83 & -- \\
\textbf{Median}
    & -- & 6.2492 & 0.8830 & 0.1021 & 0.8875 & 34 & 15 & -- \\
\bottomrule
\end{tabular}
\end{adjustbox}
\caption{
Complete repetition-chart statistics in normalized outgoing-contribution
(NOC) space across 23 models.  All rows use $q=1$.  NOC causal gain is
estimated by scaling selected Behavioral Anchor Coordinates.
}
\label{tab:app_rep_noc}
\end{table*}

\subsection{Sycophancy Geometry in ACT Space}

Table~\ref{tab:app_syco_act} reports the complete per-model sycophancy PCA
statistics in ACT space.  These measurements characterize the dimensional
compression of the activation chart and the alignment between its leading
axis and the supervised classifier direction.

\begin{table*}[!t]
\centering
\scriptsize
\setlength{\tabcolsep}{3.5pt}
\renewcommand{\arraystretch}{1.05}
\begin{adjustbox}{max width=\textwidth}
\begin{tabular}{@{}l c r r r r r r@{}}
\toprule
Model & Type &
$\#\BAC$ &
PC1 &
Top-3 &
$k_{0.9}$ &
$k_{0.9}/N_{\mathrm{BAC}}$ &
Cos. \\
\midrule

\familyheader{QwenBand}
  {\QwenLogo\ \textbf{Qwen family}}{8}
Qwen2.5-0.5B-Instruct
    & Inst. & 78 & 0.6663 & 0.8460 & 6 & 0.077 & 0.0579 \\
Qwen2.5-0.5B
    & Base & 83 & 0.6011 & 0.8782 & 4 & 0.048 & 0.0317 \\
Qwen2.5-32B-Instruct
    & Inst. & 62 & 0.8610 & 0.9620 & 2 & 0.032 & 0.2737 \\
Qwen2.5-32B
    & Base & 75 & 0.7952 & 0.9003 & 3 & 0.040 & 0.1419 \\
Qwen2.5-7B-Instruct
    & Inst. & 61 & 0.8599 & 0.9646 & 2 & 0.033 & 0.0336 \\
Qwen2.5-7B
    & Base & 69 & 0.7673 & 0.9606 & 3 & 0.043 & 0.0780 \\
Qwen3.5-27B
    & Base & 101 & 0.3035 & 0.5806 & 22 & 0.218 & 0.0000 \\

\familyheader{GemmaBand}
  {\GemmaLogo\ \textbf{Gemma family}}{8}
Gemma-2-27B-it
    & Inst. & 53 & 0.6587 & 0.9440 & 2 & 0.038 & 0.0917 \\
Gemma-2-27B
    & Base & 55 & 0.4101 & 0.7853 & 4 & 0.073 & 0.0300 \\
Gemma-2-2B-it
    & Inst. & 42 & 0.2469 & 0.5118 & 14 & 0.333 & 0.0972 \\
Gemma-2-2B
    & Base & 75 & 0.1822 & 0.4055 & 19 & 0.253 & 0.0866 \\
Gemma-2-9B-it
    & Inst. & 54 & 0.3213 & 0.6398 & 7 & 0.130 & 0.0000 \\
Gemma-2-9B
    & Base & 51 & 0.2023 & 0.5339 & 10 & 0.196 & 0.0728 \\
Gemma-3-4B-it
    & Inst. & 54 & 0.3093 & 0.6828 & 7 & 0.130 & 0.0000 \\

\familyheader{LlamaBand}
  {\LlamaLogo\ \textbf{Llama family}}{8}
Llama-3.1-8B-Instruct
    & Inst. & 78 & 0.2552 & 0.5008 & 20 & 0.256 & 0.0995 \\
Llama-3.1-8B
    & Base & 92 & 0.4262 & 0.7749 & 8 & 0.087 & 0.0029 \\
Llama-3.3-70B-Instruct
    & Inst. & 83 & 0.4069 & 0.6925 & 15 & 0.181 & 0.0326 \\

\familyheader{MistralBand}
  {\MistralLogo\ \textbf{Mistral family}}{8}
Mistral-7B-Instruct-v0.3
    & Inst. & 50 & 0.3278 & 0.5661 & 13 & 0.260 & 0.0445 \\
Mistral-7B-v0.3
    & Base & 46 & 0.3466 & 0.6108 & 15 & 0.326 & 0.0000 \\
Mistral-Nemo-Base-2407
    & Base & 78 & 0.5083 & 0.6841 & 12 & 0.154 & 0.0190 \\
Mistral-Nemo-Instruct-2407
    & Inst. & 67 & 0.6764 & 0.8163 & 6 & 0.090 & 0.0210 \\
Mistral-Small-3.1-24B-Base-2503 
    & Base & 61 & 0.3486 & 0.7351 & 8 & 0.131 & 0.1178 \\
Mistral-Small-3.1-24B-Instruct-2503
    & Inst. & 52 & 0.3563 & 0.7094 & 9 & 0.173 & 0.1002 \\
\midrule
\textbf{Mean}
    & -- & 66.09 & 0.4712 & 0.7255 & 9.17 & 0.144 & 0.0623 \\
\textbf{Median}
    & -- & 62 & 0.4069 & 0.7094 & 8 & 0.130 & 0.0445 \\
\bottomrule
\end{tabular}
\end{adjustbox}
\caption{
Complete sycophancy PCA statistics in ACT space.  Each model is evaluated on
6,678 examples.  ACT representations are strongly compressed, requiring only
$14.4\%$ of the selected scaffold dimensions on average to explain $90\%$ of
variance.
}
\label{tab:app_syco_act}
\end{table*}

\subsection{Sycophancy Geometry in NOC Space}

Table~\ref{tab:app_syco_noc} reports the corresponding per-model sycophancy
PCA statistics in NOC space.  Comparison with the ACT results shows how chart
compression and supervised-direction alignment change when coordinates are
represented by their normalized outgoing contributions.

\begin{table*}[!t]
\centering
\scriptsize
\setlength{\tabcolsep}{3.5pt}
\renewcommand{\arraystretch}{1.05}
\begin{adjustbox}{max width=\textwidth}
\begin{tabular}{@{}l c r r r r r r@{}}
\toprule
Model & Type &
$\#\BAC$ &
PC1 &
Top-3 &
$k_{0.9}$ &
$k_{0.9}/N_{\mathrm{BAC}}$ &
Cos. \\
\midrule

\familyheader{GemmaBand}
  {\GemmaLogo\ \textbf{Gemma family}}{8}
Gemma-2-2B
    & Base & 75 & 0.2606 & 0.4738 & 29 & 0.387 & 0.0511 \\
Gemma-2-2B-it
    & Inst. & 42 & 0.2884 & 0.5270 & 18 & 0.429 & 0.7440 \\
Gemma-2-9B
    & Base & 51 & 0.3869 & 0.5920 & 18 & 0.353 & 0.1225 \\
Gemma-2-9B-it
    & Inst. & 54 & 0.3111 & 0.5153 & 24 & 0.444 & 0.7365 \\
Gemma-3-4B-it
    & Inst. & 54 & 0.3615 & 0.5434 & 22 & 0.407 & 0.5338 \\
Gemma-2-27B
    & Base & 55 & 0.2501 & 0.4926 & 24 & 0.436 & 0.5971 \\
Gemma-2-27B-it
    & Inst. & 53 & 0.2919 & 0.5084 & 23 & 0.434 & 0.6870 \\

\familyheader{MistralBand}
  {\MistralLogo\ \textbf{Mistral family}}{8}
Mistral-7B-Instruct-v0.3
    & Inst. & 50 & 0.3990 & 0.5824 & 21 & 0.420 & 0.4227 \\
Mistral-7B-v0.3
    & Base & 46 & 0.4061 & 0.6286 & 15 & 0.326 & 0.4161 \\
Mistral-Nemo-Base-2407
    & Base & 78 & 0.2282 & 0.5039 & 26 & 0.333 & -0.1646 \\
Mistral-Nemo-Instruct-2407
    & Inst. & 67 & 0.2799 & 0.5080 & 25 & 0.373 & 0.5667 \\
Mistral-Small-3.1-24B-Instruct-2503
    & Inst. & 52 & 0.3440 & 0.6273 & 15 & 0.288 & 0.2482 \\
Mistral-Small-3.1-24B-Base-2503 
    & Base & 61 & 0.5269 & 0.7282 & 12 & 0.197 & 0.2025 \\

\familyheader{LlamaBand}
  {\LlamaLogo\ \textbf{Llama family}}{8}
Llama-3.1-8B
    & Base & 92 & 0.3093 & 0.5056 & 29 & 0.315 & 0.1757 \\
Llama-3.1-8B-Instruct
    & Inst. & 78 & 0.2816 & 0.4952 & 30 & 0.385 & 0.6781 \\
Llama-3.3-70B-Instruct
    & Inst. & 83 & 0.2859 & 0.4795 & 30 & 0.361 & 0.6731 \\

\familyheader{QwenBand}
  {\QwenLogo\ \textbf{Qwen family}}{8}
Qwen2.5-0.5B
    & Base & 83 & 0.2025 & 0.3670 & 38 & 0.458 & 0.6242 \\
Qwen2.5-0.5B-Instruct
    & Inst. & 78 & 0.1917 & 0.3517 & 36 & 0.462 & 0.6241 \\
Qwen2.5-7B
    & Base & 69 & 0.2924 & 0.4997 & 28 & 0.406 & 0.5060 \\
Qwen2.5-7B-Instruct
    & Inst. & 61 & 0.2759 & 0.4415 & 30 & 0.492 & 0.5745 \\
Qwen2.5-32B
    & Base & 75 & 0.2603 & 0.4714 & 32 & 0.427 & 0.6337 \\
Qwen2.5-32B-Instruct
    & Inst. & 62 & 0.2865 & 0.6132 & 24 & 0.387 & 0.1861 \\
Qwen3.5-27B
    & Base & 101 & 0.3149 & 0.5026 & 34 & 0.337 & 0.6593 \\
\midrule
\textbf{Mean}
    & -- & 66.09 & 0.3059 & 0.5199 & 25.35 & 0.385 & 0.4565 \\
\textbf{Median}
    & -- & 62 & 0.2884 & 0.5056 & 25 & 0.387 & 0.5667 \\
\bottomrule
\end{tabular}
\end{adjustbox}
\caption{
Complete sycophancy PCA statistics in normalized outgoing-contribution
(NOC) space.  NOC charts are less compressed than ACT charts but often align
more strongly with the supervised classifier direction.
}
\label{tab:app_syco_noc}
\end{table*}

\subsection{Unified NOC Statistics}

Table~\ref{tab:app_all_bac_statistics_noc} consolidates the repetition and
sycophancy statistics in NOC space using a shared per-model schema.  This view
supports direct comparison of scaffold size, chart strength, causal gain, and
low-rank structure across the evaluated models.  For both behaviors, the three
strength metrics follow the definitions given above and are evaluated in NOC
space.

\begin{table*}[!t]
\centering
\small
\setlength{\tabcolsep}{2.45pt}
\renewcommand{\arraystretch}{1.04}

\begin{adjustbox}{
  max width=\linewidth,
  max totalheight=0.80\textheight,
  keepaspectratio,
  center
}
\begin{tabular}{
@{}
>{\raggedright\arraybackslash}p{3.55cm}
r *{6}{r}
r *{6}{r}
@{}
}
\toprule

\multirow{3}{*}{Model}
&
\multicolumn{7}{c}{\textbf{Repetition geometry} ($q=1$)}
&
\multicolumn{7}{c}{\textbf{Sycophancy geometry} (6,678 examples)}
\\

\cmidrule(lr){2-8}
\cmidrule(lr){9-15}

&
\multirow{2}{*}{$\#\BAC$}
&
\multicolumn{6}{c}{NOC space}
&
\multirow{2}{*}{$\#\BAC$}
&
\multicolumn{6}{c}{NOC space}
\\

\cmidrule(lr){3-8}
\cmidrule(lr){10-15}

& &
$G^{\mathrm{uniF}}$
&
$G^{\mathrm{occ}}$
&
$G^{\mathrm{causal}}$
&
PC1
&
Top-3
&
$k_{0.9}$
&
&
$G^{\mathrm{uniF}}$
&
$G^{\mathrm{occ}}$
&
$G^{\mathrm{causal}}$
&
PC1
&
Top-3
&
$k_{0.9}$
\\

\midrule
\familyheader{GemmaBand}{\GemmaLogo\quad\textbf{Gemma family}}{15}
Gemma-2-2B & 32 & 6.2492 & 1.0786 & 0.1204 & 0.7199 & 0.8741 & 13 & 75 & 8.23263 & 0.493434 & 0.0452983 & 0.2606 & 0.4738 & 29 \\
Gemma-2-2B-it & 28 & 5.6593 & 0.8830 & 0.2333 & 0.7369 & 0.8398 & 15 & 42 & 7.54262 & 0.469359 & 0.0189962 & 0.2884 & 0.5270 & 18 \\
Gemma-3-4B-it & 26 & 8.4143 & 0.6596 & 0.1513 & 0.7059 & 0.8165 & 13 & 54 & 7.03847 & 0.5389 & 0.307394 & 0.3615 & 0.5434 & 22 \\
Gemma-2-9B & 31 & 4.2760 & 0.9555 & 0.1603 & 0.7752 & 0.8912 & 15 & 51 & 9.47754 & 0.492795 & 0.111633 & 0.3869 & 0.5920 & 18 \\
Gemma-2-9B-it & 26 & 4.2142 & 0.9589 & 0.0477 & 0.8345 & 0.8969 & 13 & 54 & 8.80361 & 0.474639 & 0.0128778 & 0.3111 & 0.5153 & 24 \\
Gemma-2-27B & 38 & 8.8803 & 0.6375 & 0.1021 & 0.7261 & 0.8487 & 15 & 55 & 8.33293 & 0.573001 & 0.0558019 & 0.2501 & 0.4926 & 24 \\
Gemma-2-27B-it & 34 & 5.1277 & 1.0103 & 0.1138 & 0.7731 & 0.8772 & 13 & 53 & 8.00736 & 0.632828 & 0.133148 & 0.2919 & 0.5084 & 23 \\

\familyheader{QwenBand}{\QwenLogo\quad\textbf{Qwen family}}{15}
Qwen2.5-0.5B & 48 & 9.5037 & 0.6870 & 0.0216 & 0.5832 & 0.7412 & 22 & 83 & 7.08938 & 0.491728 & 0.0295526 & 0.2025 & 0.3670 & 38 \\
Qwen2.5-0.5B-Instruct & 48 & 10.1911 & 0.6394 & 0.1244 & 0.6390 & 0.7670 & 22 & 78 & 7.4928 & 0.546409 & 0.0207409 & 0.1917 & 0.3517 & 36 \\
Qwen2.5-7B & 37 & 8.1407 & 0.6886 & 0.1100 & 0.7271 & 0.8594 & 15 & 69 & 8.68511 & 0.583384 & -0.000877529 & 0.2924 & 0.4997 & 28 \\
Qwen2.5-7B-Instruct & 42 & 8.2838 & 0.7411 & 0.1344 & 0.7333 & 0.8512 & 20 & 61 & 8.50326 & 0.624147 & 0.0223223 & 0.2759 & 0.4415 & 30 \\
Qwen3.5-27B & 45 & 11.2543 & 0.4830 & 0.0322 & 0.6902 & 0.8345 & 19 & 101 & 9.33057 & 0.553323 & 0.01328 & 0.3149 & 0.5026 & 34 \\
Qwen2.5-32B & 61 & 7.3041 & 0.8637 & 0.0487 & 0.6887 & 0.8035 & 24 & 75 & 9.52139 & 0.44551 & 0.00741185 & 0.2603 & 0.4714 & 32 \\
Qwen2.5-32B-Instruct & 44 & 4.6760 & 1.1594 & 0.0365 & 0.6992 & 0.8235 & 18 & 62 & 9.92692 & 0.510514 & -0.000686973 & 0.2865 & 0.6132 & 24 \\

\familyheader{MistralBand}{\MistralLogo\quad\textbf{Mistral family}}{15}
Mistral-7B-v0.3 & 56 & 12.4879 & 0.4643 & 0.0901 & 0.7102 & 0.7996 & 21 & 46 & 7.06802 & 0.648702 & 0.0483211 & 0.4061 & 0.6286 & 15 \\
Mistral-7B-Instruct-v0.3 & 32 & 11.2649 & 0.5000 & 0.1500 & 0.7132 & 0.8112 & 17 & 50 & 9.17742 & 0.571795 & 0.0266025 & 0.3990 & 0.5824 & 21 \\
Mistral-Nemo-Base-2407 & 43 & 4.5099 & 1.0969 & 0.0247 & 0.7907 & 0.8568 & 20 & 78 & 6.82295 & 0.675682 & 0.0237964 & 0.2282 & 0.5039 & 26 \\
Mistral-Nemo-Instruct-2407 & 42 & 6.0374 & 0.9124 & 0.0222 & 0.7391 & 0.8168 & 20 & 67 & 7.33621 & 0.682483 & 0.0729401 & 0.2799 & 0.5080 & 25 \\
Mistral-Small-3.1-24B-Base-2503  & 28 & 5.1098 & 0.9186 & 0.0698 & 0.7466 & 0.8556 & 12 & 61 & 7.2653 & 0.692815 & 0.200294 & 0.5269 & 0.7282 & 12 \\
Mistral-Small-3.1-24B-Instruct-2503 & 34 & 6.3496 & 0.7855 & 0.0769 & 0.7206 & 0.8402 & 13 & 52 & 6.2152 & 0.75018 & 0.172712 & 0.3440 & 0.6273 & 15 \\

\familyheader{LlamaBand}{\LlamaLogo\quad\textbf{Llama family}}{15}
Llama-3.1-8B & 26 & 4.1244 & 1.0646 & 0.0463 & 0.8063 & 0.8844 & 9 & 92 & 8.30751 & 0.513425 & 0.0395678 & 0.3093 & 0.5056 & 29 \\
Llama-3.1-8B-Instruct & 20 & 4.6608 & 1.1588 & 0.1229 & 0.8462 & 0.9031 & 7 & 78 & 7.62933 & 0.531903 & 0.0438382 & 0.2816 & 0.4952 & 30 \\
Llama-3.3-70B-Instruct & 31 & 4.6690 & 1.1443 & 0.1029 & 0.8594 & 0.9191 & 8 & 83 & 10.6307 & 0.502569 & 0.0203946 & 0.2859 & 0.4795 & 30 \\
\textbf{Mean} & 37.04 & 7.0169 & 0.8474 & 0.0932 & 0.7376 & 0.8440 & 15.83 & 66.09 & 8.19292 & 0.565197 & 0.0619721 & 0.3059 & 0.5199 & 25.35 \\
\textbf{Median} & 34 & 6.2492 & 0.8830 & 0.1021 & 0.7271 & 0.8487 & 15 & 62 & 8.23263 & 0.546409 & 0.0295526 & 0.2884 & 0.5056 & 25 \\
\bottomrule
\end{tabular}
\end{adjustbox}

\caption{
Unified per-model Behavioral Anchor Coordinate statistics in the
NOC representation space. Repetition is shown on the left and
sycophancy on the right using the same retained column schema.
}
\label{tab:app_all_bac_statistics_noc}
\end{table*}

%% file: sections__supplement__p_bac_layer_distri.tex
\section{Layer-Wise Distribution of Behavioral Anchor Coordinates}
\label{app:bac_layer_distribution}

\begin{figure}[t]
    \centering
    \includegraphics[
        width=\columnwidth
    ]{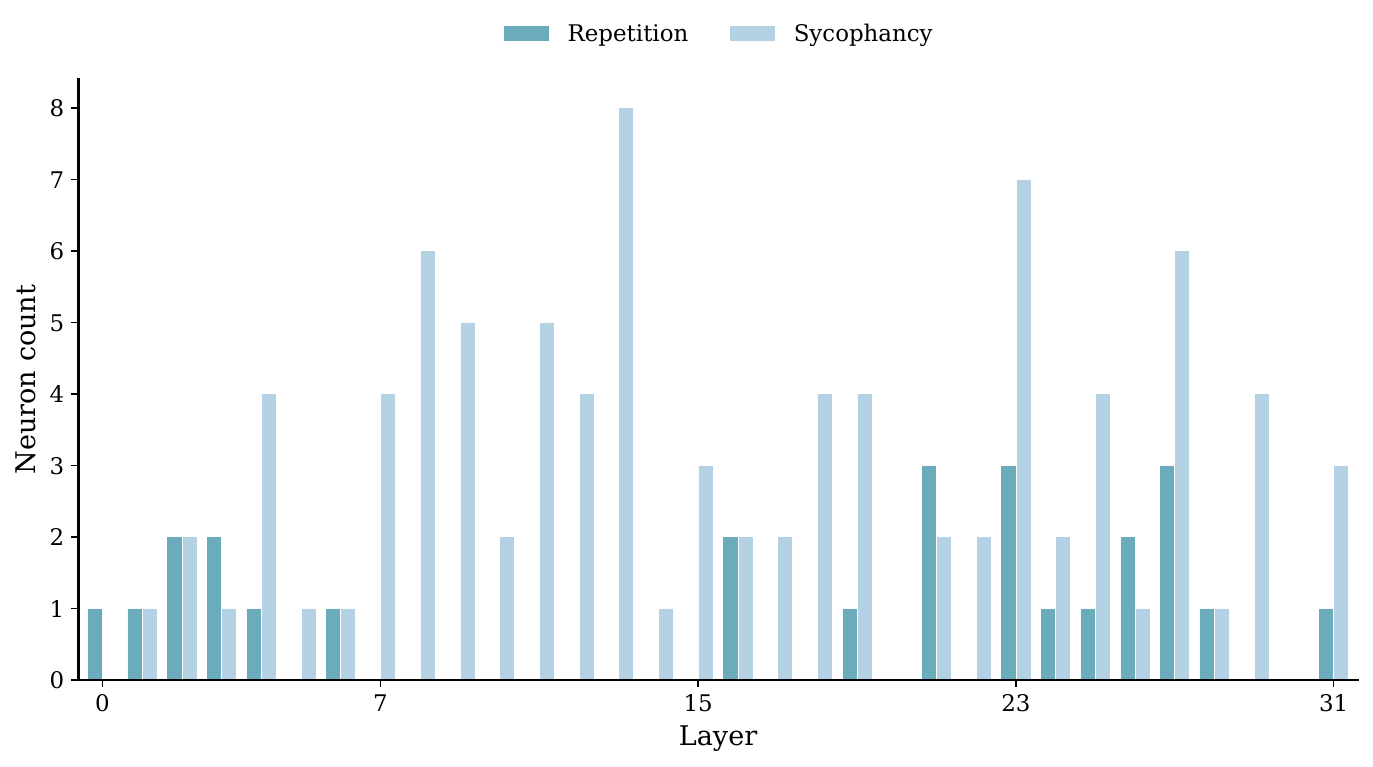}
    \caption{
    Layer-wise distribution of repetition and sycophancy Behavioral Anchor
    Coordinates in Llama-3.1-8B.  Each bar reports the number of positively
    behavior-associated coordinates selected in the corresponding transformer
    layer.  Repetition is represented by a comparatively sparse scaffold,
    whereas sycophancy is supported by a broader and more widely distributed
    coordinate population.  These counts visualize the coordinate scaffolds
    used to estimate the corresponding low-dimensional behavioral charts; they
    should not be interpreted as direct measurements of layer-wise causal
    importance.
    }
    \label{fig:llama_bac_layer_distribution}
\end{figure}

To provide a concrete view of the coordinate scaffolds used in the geometric
analysis, Figure~\ref{fig:llama_bac_layer_distribution} shows the layer-wise
locations of the repetition and sycophancy Behavioral Anchor Coordinates
(\BACs) identified in Llama-3.1-8B.  For each transformer layer, the figure
reports the number of coordinates assigned a positive coefficient by the
corresponding sparse behavioral classifier.

The selected coordinates are not uniformly distributed across model depth.
Repetition is supported by a comparatively sparse scaffold, with only a small
number of selected coordinates in most layers.  Sycophancy is associated with
a broader scaffold and exhibits several layer-localized concentrations,
particularly in the middle and later portions of the model.  The two
distributions also differ substantially from one another, consistent with the
interpretation that the coordinate-selection procedure identifies
behavior-specific scaffolds rather than a single generic behavioral axis.

This visualization is intended as a localization diagnostic rather than a
complete measure of layer importance.  A larger coordinate count does not by
itself imply greater causal contribution, since coordinates may differ in
activation magnitude, outgoing contribution, discriminative weight, and
steering sensitivity.  The subsequent PCA, Fisher, CCA, and intervention
analyses therefore operate on the extracted feature values rather than using
layer-wise counts as the behavioral geometry itself.

%% file: sections__supplement__q_Cross-ModelBehavioralAtlas.tex
\section{Full Cross-Model Behavioral Atlas}
\label{app:cross-model-atlas-results}

This appendix reports the complete cross-model alignment results underlying
Section~\ref{sec:cross-model-results}.  The purpose of the analysis is to
determine whether behavioral charts estimated independently in models with
different architectures, parameter scales, and post-training histories exhibit a shared cross-model geometric core.

We emphasize that cross-model alignment does not require neuron-index
correspondence.  Each model is first represented in its own behavior-enriched
coordinate scaffold and its own low-dimensional PCA chart.  Alignment is then
measured through matched sample-level coordinates.  The relevant question is
therefore whether two models organize the same behavioral examples in linearly
related ways, rather than whether they use identical hidden units.

\subsection{Model Roster and Common Ordering}

All heatmaps use the common A--W model order in
Table~\ref{tab:cross-model-roster}.  This ordering must also be used for the
random-coordinate panels so that corresponding cells represent the same model
pair.

\begin{table*}[t]
\centering
\scriptsize
\setlength{\tabcolsep}{3.5pt}
\begin{tabular}{clclcl}
\toprule
ID & Model & ID & Model & ID & Model \\
\midrule
A & Qwen2.5-0.5B-Instruct
&
I & Gemma-2-27B
&
Q & Llama-3.3-70B-Instruct
\\
B & Qwen2.5-0.5B
&
J & Gemma-2-2B-it
&
R & Mistral-7B-Instruct-v0.3
\\
C & Qwen2.5-32B-Instruct
&
K & Gemma-2-2B
&
S & Mistral-7B-v0.3
\\
D & Qwen2.5-32B
&
L & Gemma-2-9B-it
&
T & Mistral-Nemo-Base-2407
\\
E & Qwen2.5-7B-Instruct
&
M & Gemma-2-9B
&
U & Mistral-Nemo-Instruct-2407
\\
F & Qwen2.5-7B
&
N & Gemma-3-4B-it
&
V & Mistral-Small-3.1-24B-Base-2503
\\
G & Qwen3.5-27B
&
O & Llama-3.1-8B-Instruct
&
W & Mistral-Small-3.1-24B-Instruct-2503
\\
H & Gemma-2-27B-it
&
P & Llama-3.1-8B
&
 & 
\\
\bottomrule
\end{tabular}
\caption{
Model identifiers used in all cross-model heatmaps.  A common ordering is
necessary for direct visual comparison between behavior-associated and
random-coordinate matrices.
}
\label{tab:cross-model-roster}
\end{table*}

\subsection{Evaluation Protocol}

For every model $m$, let
\[
    Z_m
    =
    \begin{bmatrix}
    z_m(x_1)^\top \\
    \vdots \\
    z_m(x_n)^\top
    \end{bmatrix}
    \in\mathbb{R}^{n\times k_m}
\]
denote its chart coordinates on the common matched sample set.
The chart dimension $k_m$ is the minimum number of PCA components required
to explain $90\%$ of the variance in the corresponding chart.  All alignment
and linear-association summaries are computed from these matched sample-level
coordinates and are interpreted as descriptive cross-model associations.

\paragraph{Classifier-direction alignment.}

Let $w_m$ be the sparse logistic-classifier direction for model $m$, projected
into the corresponding PCA coordinate system.  The sample-level classifier
score is
\begin{equation}
    s_m(x)
    =
    w_m^\top z_m(x).
\end{equation}
For models $m$ and $m'$, we report
\begin{equation}
\label{eq:classifier-direction-alignment}
    A_{\mathrm{cls}}(m,m')
    =
    \left|
    \operatorname{corr}
    \left(
        s_m(x),
        s_{m'}(x)
    \right)
    \right|.
\end{equation}
This measures the strength of cross-model score correspondence up to a global
sign.

\paragraph{Canonical-correlation alignment.}

For a pair of models, CCA finds matrices $C_m$ and $C_{m'}$ such that
corresponding coordinates in
\[
    Z_mC_m
    \qquad\text{and}\qquad
    Z_{m'}C_{m'}
\]
have maximal correlation under the canonical orthogonality constraints.  Let
\[
    \rho_1\geq\rho_2\geq\cdots\geq\rho_q,
    \qquad
    q=\min(k_m,k_{m'}),
\]
be the resulting canonical correlations on the matched sample set.  We report the primary score
\begin{equation}
\label{eq:primary-cca-score}
    A_{\mathrm{CCA}}^{(1)}(m,m')
    =
    \rho_1,
\end{equation}
and the mean score
\begin{equation}
\label{eq:mean-cca-score}
    A_{\mathrm{CCA}}^{(\mathrm{mean})}(m,m')
    =
    \frac{1}{q}
    \sum_{i=1}^{q}\rho_i.
\end{equation}
The primary score asks whether at least one strongly shared direction exists.
The mean score asks whether alignment extends across the retained chart rather
than being concentrated in a single canonical axis.

\paragraph{Top-two-PC linear association.}

For source model $m$ and target model $m'$, we consider the four
one-dimensional mappings
\[
    z_{m,i}
    \longrightarrow
    z_{m',j},
    \qquad
    i,j\in\{1,2\}.
\]
For each mapping, we report the absolute correlation between the fitted target
coordinate and the observed target coordinate on the common matched sample
set. The top-two-PC score is
\begin{equation}
\label{eq:top2-probing-score}
    A_{\mathrm{probe}}^{(2)}(m,m')
    =
    \max_{i,j\in\{1,2\}}
    \left|
    \operatorname{corr}
    \left(
        \widehat z_{m',j},
        z_{m',j}
    \right)
    \right|.
\end{equation}
This score is useful when the behavior-aligned direction may be PC1 in one
model and PC2 in another.

\paragraph{Full-chart directed linear association.}

We additionally fit a linear map
\begin{equation}
    \widehat Z_{m'}
    =
    Z_m W_{m\rightarrow m'}+b_{m\rightarrow m'}
\end{equation}
from the source chart to the complete variance-thresholded target chart.  Its
score is
\begin{equation}
\label{eq:kdim-probing-score}
    A_{\mathrm{probe}}^{(k)}(m\rightarrow m')
    =
    \frac{1}{k_{m'}}
    \sum_{j=1}^{k_{m'}}
    \operatorname{corr}
    \left(
        \widehat Z_{m'}[:,j],
        Z_{m'}[:,j]
    \right).
\end{equation}
Unlike pairwise CCA and one-dimensional association, this quantity is directed:
\[
    A_{\mathrm{probe}}^{(k)}(m\rightarrow m')
    \neq
    A_{\mathrm{probe}}^{(k)}(m'\rightarrow m)
\]
in general.  The asymmetry reflects differences in chart rank, redundancy,
and how readily one model's representation spans another model's residual
coordinates.

\paragraph{Random-coordinate controls.}

For every model and feature space, we replace the behavior-associated
coordinate scaffold with a random coordinate set of the same cardinality and
repeat standardization, PCA, CCA, and linear-association analysis.  These
controls preserve the ambient model, sample identities,
dimensionality-selection procedure, and evaluation protocol while removing
behavior-informed coordinate selection.

The random controls are especially important for CCA.  In finite samples,
CCA can find a strongly correlated leading pair between unrelated
high-dimensional feature sets.  A meaningful behavioral result should therefore exceed the random baseline not
only for $\rho_1$, but also for mean CCA and matched-sample linear association.

\subsection{Aggregate Results}

Table~\ref{tab:cross-model-atlas-summary-app} reports the mean off-diagonal
score for each analysis.  For the directed $k$-dimensional association scores, the mean is
taken over all ordered source--target pairs.

\begin{table}[t]
\centering
\small
\setlength{\tabcolsep}{3.7pt}
\begin{tabular}{lrrrr}
\toprule
& \multicolumn{2}{c}{ACT} & \multicolumn{2}{c}{NOC} \\
\cmidrule(lr){2-3}
\cmidrule(lr){4-5}
Metric & BAC & Random & BAC & Random \\
\midrule
Primary CCA       & 0.938 & 0.667 & 0.974 & 0.817 \\
Mean CCA          & 0.515 & 0.222 & 0.648 & 0.233 \\
Top-2-PC probing  & 0.784 & 0.266 & 0.958 & 0.495 \\
$k$-dim. probing  & 0.473 & 0.215 & 0.581 & 0.234 \\
\bottomrule
\end{tabular}
\caption{
Mean off-diagonal cross-model alignment.  Behavior-associated charts exceed
matched-size random-coordinate baselines in both ACT and NOC.  The difference
is most diagnostic for mean CCA and full-chart linear association, because primary CCA is
also elevated for random coordinates.
}
\label{tab:cross-model-atlas-summary-app}
\end{table}

The leading canonical coordinate is nearly saturated in both spaces.  However,
the random primary-CCA values are also large, particularly in NOC.  Primary
CCA therefore establishes the existence of a highly alignable axis but does
not by itself show that this axis is behavior-specific.

Mean CCA provides a stronger test.  The ACT score increases from $0.222$ under
random selection to $0.515$ under behavior-associated selection, an absolute increase of $0.293$.  In NOC, the score increases from $0.233$ to $0.648$, an
increase of $0.415$.  Thus, behavior-informed selection improves alignment
across multiple canonical coordinates, not only the most favorable pair.

The linear-association summaries provide a complementary descriptive
comparison.  In ACT, the top-two-PC score increases from $0.266$ to $0.784$,
while the complete-chart score increases from $0.215$ to $0.473$.  In NOC, the
top-two-PC score increases from $0.495$ to $0.958$, and the complete-chart
score increases from $0.234$ to $0.581$. The particularly strong NOC result suggests that functional
contribution coordinates expose a shared behavior axis that is less dependent
on a model's raw activation basis.

\subsection{Classifier-Direction Alignment}

\begin{figure*}[t]
    \centering
    \includegraphics[
        width=0.72\textwidth,
        keepaspectratio
    ]{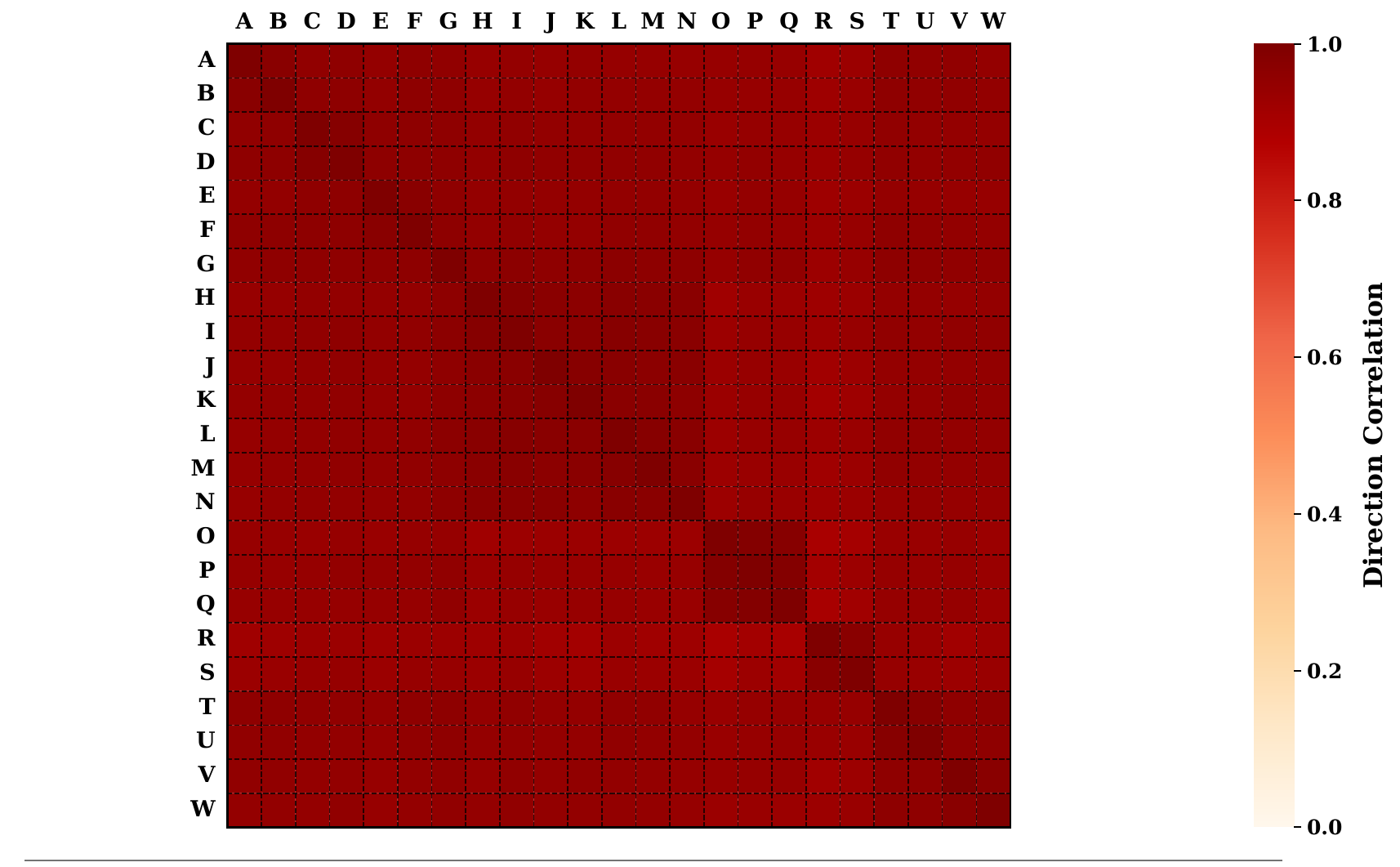}
    \caption{
    Cross-model alignment of sparse classifier-direction scores.  Each cell is
    the absolute Pearson correlation between the two models' matched-sample classifier scores.  The high absolute correlations show that independently fitted supervised axes
    exhibit strong score correspondence across architectures and scales, up to a
    global sign. Because the directions are
    label-supervised, this result is treated as a complementary sanity check
    rather than the primary evidence for unsupervised chart universality.
    }
    \label{fig:classifier-direction-atlas}
\end{figure*}

Figure~\ref{fig:classifier-direction-atlas} compares the sample-level scores of
the sparse supervised directions.  The mean off-diagonal absolute correlation
is $0.950$, and the mean for matched base/instruction pairs is $0.980$.

The near-universal absolute score correspondence does not require literal
neuron correspondence and does not imply that the classifier vectors occupy
equivalent ambient coordinates.  Rather, after each model is expressed in its
own chart, the supervised behavior-positive axes induce strongly corresponding
score patterns over the common examples, up to a global sign.

\subsection{Canonical-Correlation Atlas}

\begin{figure*}[t]
    \centering

    \begin{minipage}[t]{0.485\textwidth}
        \centering
        \atlaspanel{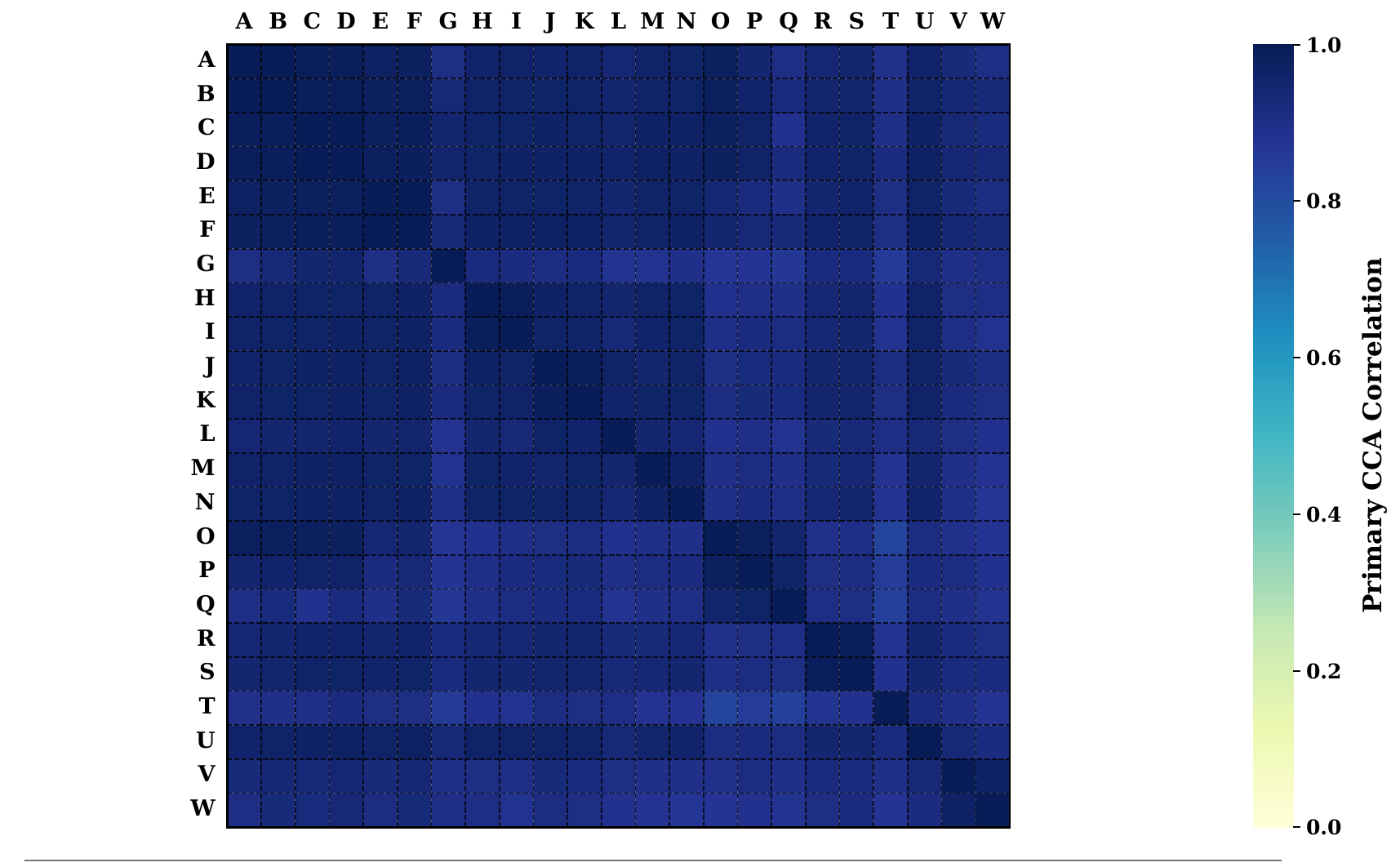}
        \vspace{-1mm}

        {\small (a) Primary CCA, ACT}
    \end{minipage}
    \hfill
    \begin{minipage}[t]{0.485\textwidth}
        \centering
        \atlaspanel{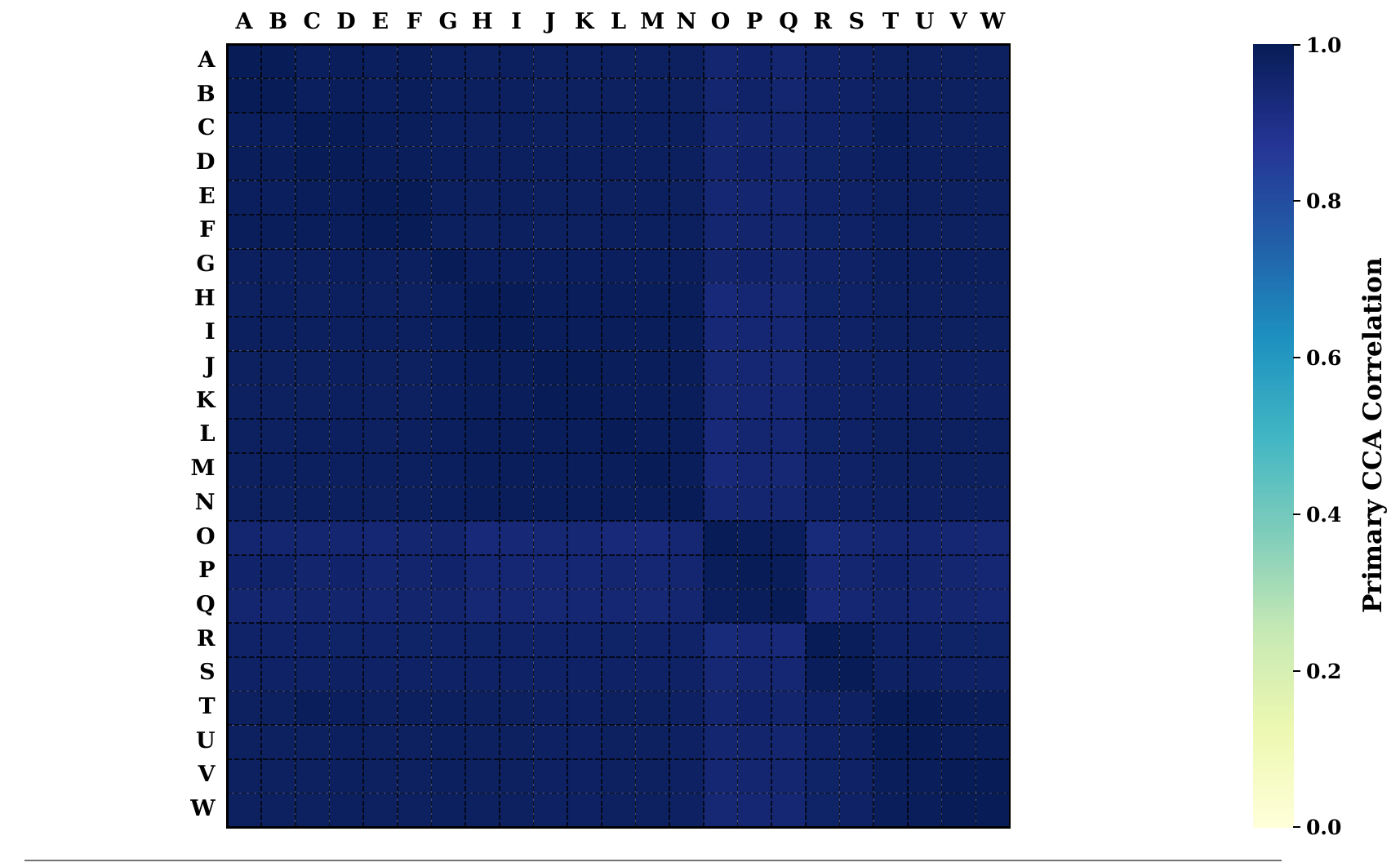}
        \vspace{-1mm}

        {\small (b) Primary CCA, NOC}
    \end{minipage}

    \vspace{2mm}

    \begin{minipage}[t]{0.485\textwidth}
        \centering
        \atlaspanel{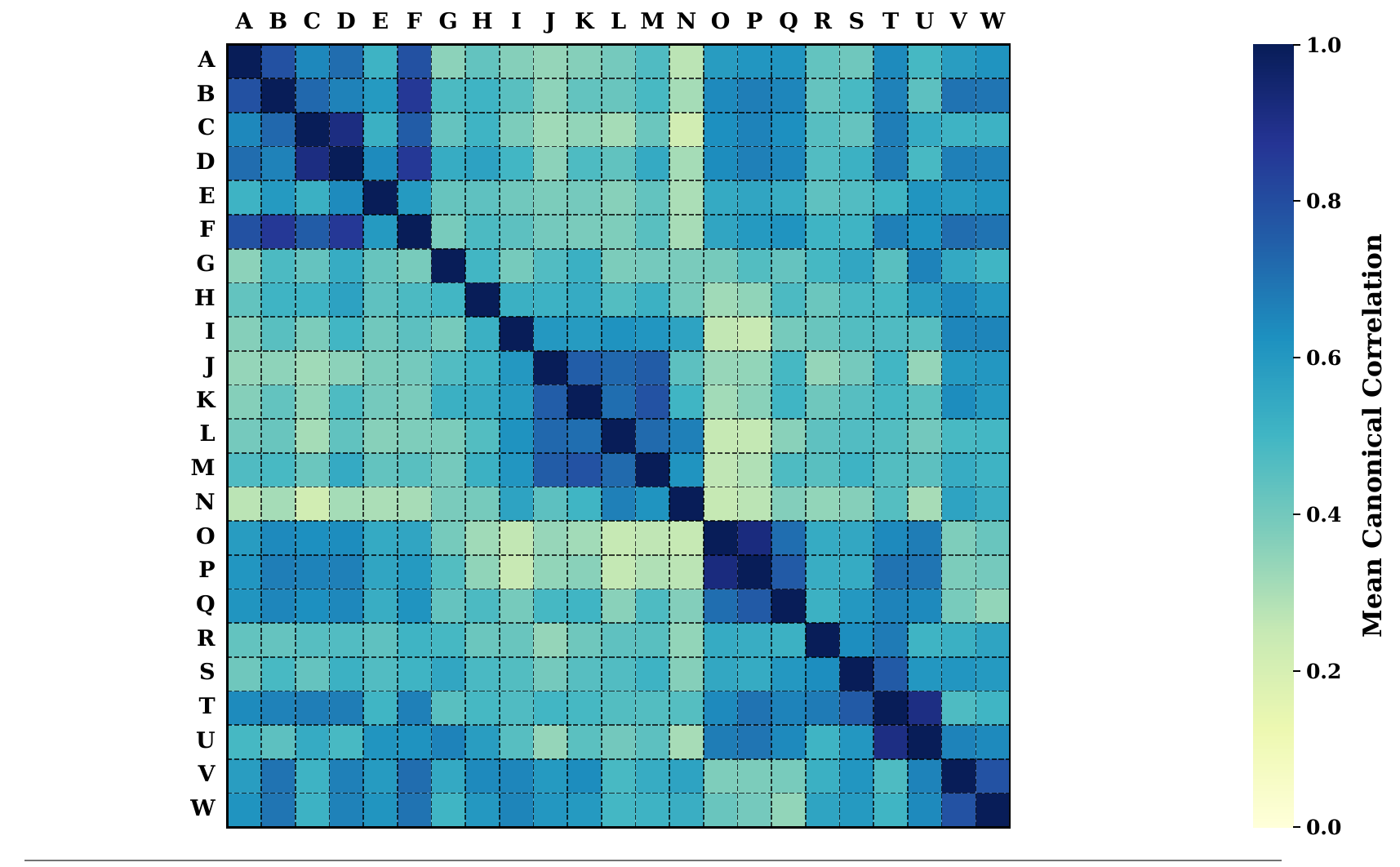}
        \vspace{-1mm}

        {\small (c) Mean CCA, ACT}
    \end{minipage}
    \hfill
    \begin{minipage}[t]{0.485\textwidth}
        \centering
        \atlaspanel{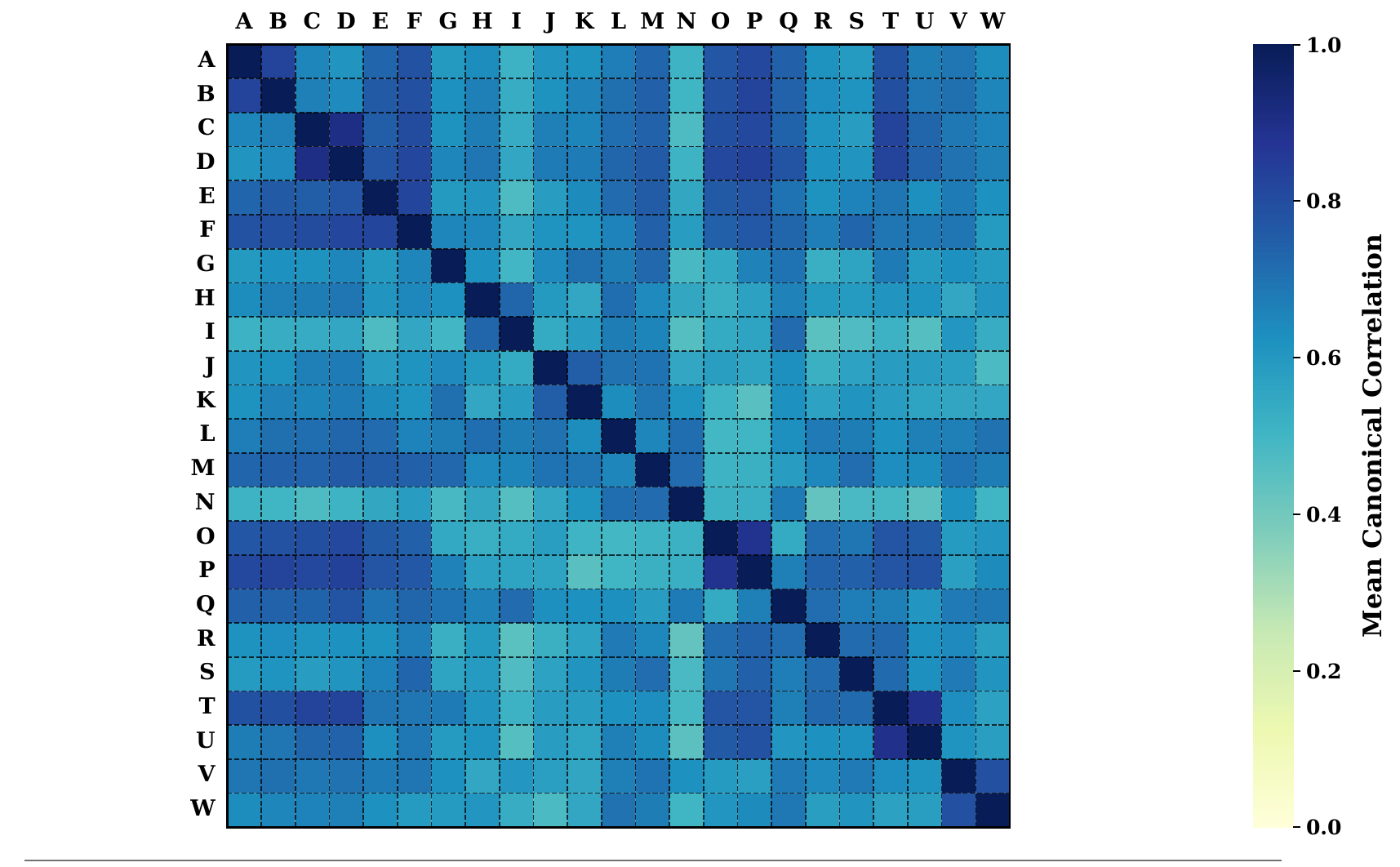}
        \vspace{-1mm}

        {\small (d) Mean CCA, NOC}
    \end{minipage}

    \caption{
    Pairwise CCA alignment of the behavior-associated sycophancy charts.
    Primary CCA measures the strongest canonical coordinate, whereas Mean CCA
    averages across the complete retained canonical chart.  Primary CCA is
    uniformly high in both spaces.  Mean CCA reveals more residual structure:
    ACT exhibits pronounced family-dependent blocks, while NOC remains more
    uniformly aligned across families.
    }
    \label{fig:cca-atlas-full}
\end{figure*}

Figure~\ref{fig:cca-atlas-full} reports primary and mean CCA in ACT and NOC.

The difference between primary and mean CCA is scientifically informative.
Most model pairs possess at least one strongly alignable direction, explaining
the nearly saturated primary matrix.  The lower mean correlations show that the
remaining chart coordinates are not identical.  The atlas is therefore better
described as a shared leading core with model-specific residual dimensions
than as one globally identical subspace.

The representation spaces also differ.  Mean CCA is $0.515$ in ACT and
$0.648$ in NOC.  ACT preserves more architecture-specific information about
the model's runtime state, while NOC removes some basis-specific variability by
expressing selected coordinates through their normalized functional
contribution.

\subsection{Random-Coordinate CCA Controls}

\begin{figure*}[t]
    \centering

    \begin{minipage}[t]{0.485\textwidth}
        \centering
        \atlaspanel{
            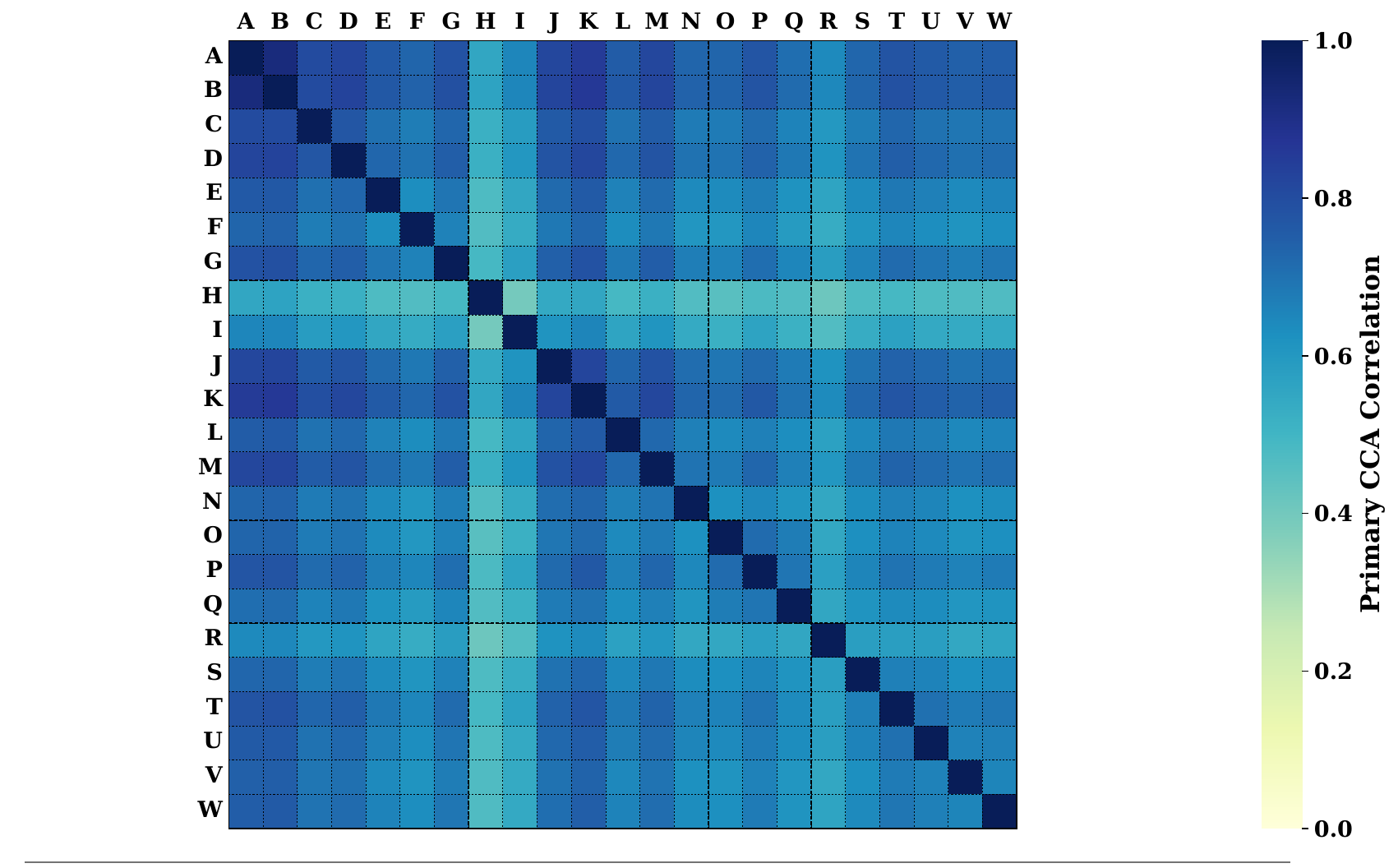
        }
        \vspace{-1mm}

        {\small (a) Random primary CCA, ACT}
    \end{minipage}
    \hfill
    \begin{minipage}[t]{0.485\textwidth}
        \centering
        \atlaspanel{
            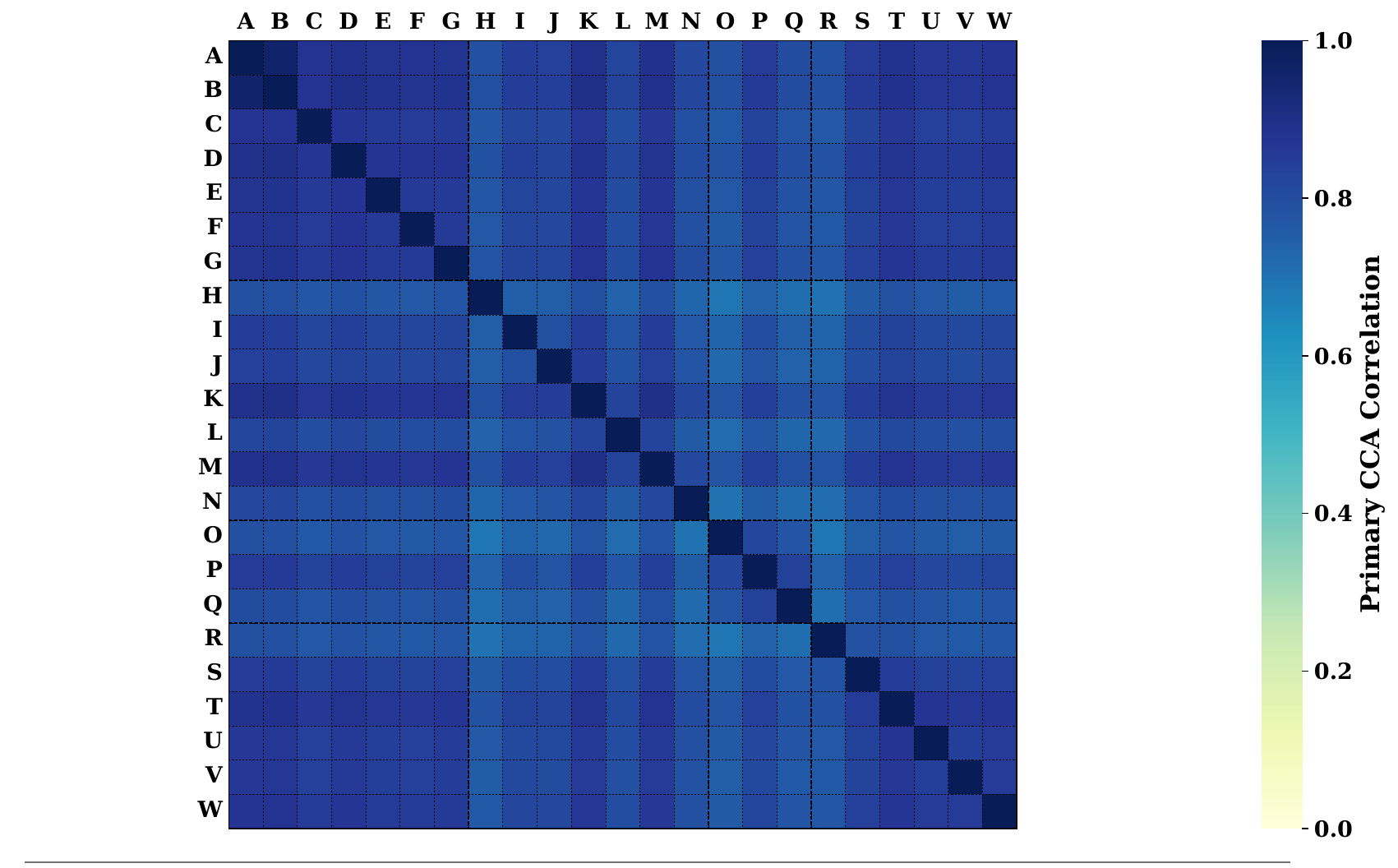
        }
        \vspace{-1mm}

        {\small (b) Random primary CCA, NOC}
    \end{minipage}

    \vspace{2mm}

    \begin{minipage}[t]{0.485\textwidth}
        \centering
        \atlaspanel{
            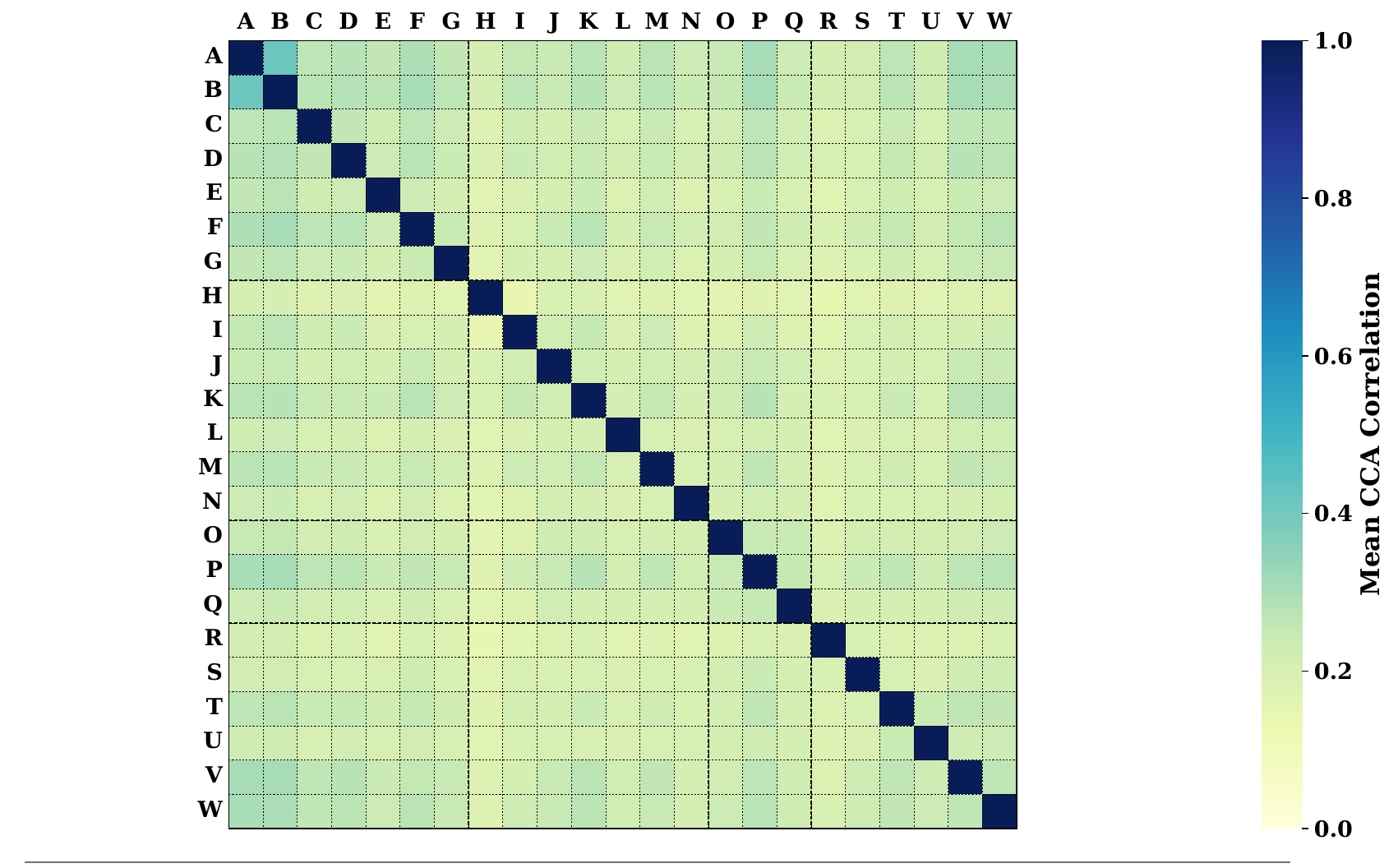
        }
        \vspace{-1mm}

        {\small (c) Random mean CCA, ACT}
    \end{minipage}
    \hfill
    \begin{minipage}[t]{0.485\textwidth}
        \centering
        \atlaspanel{
            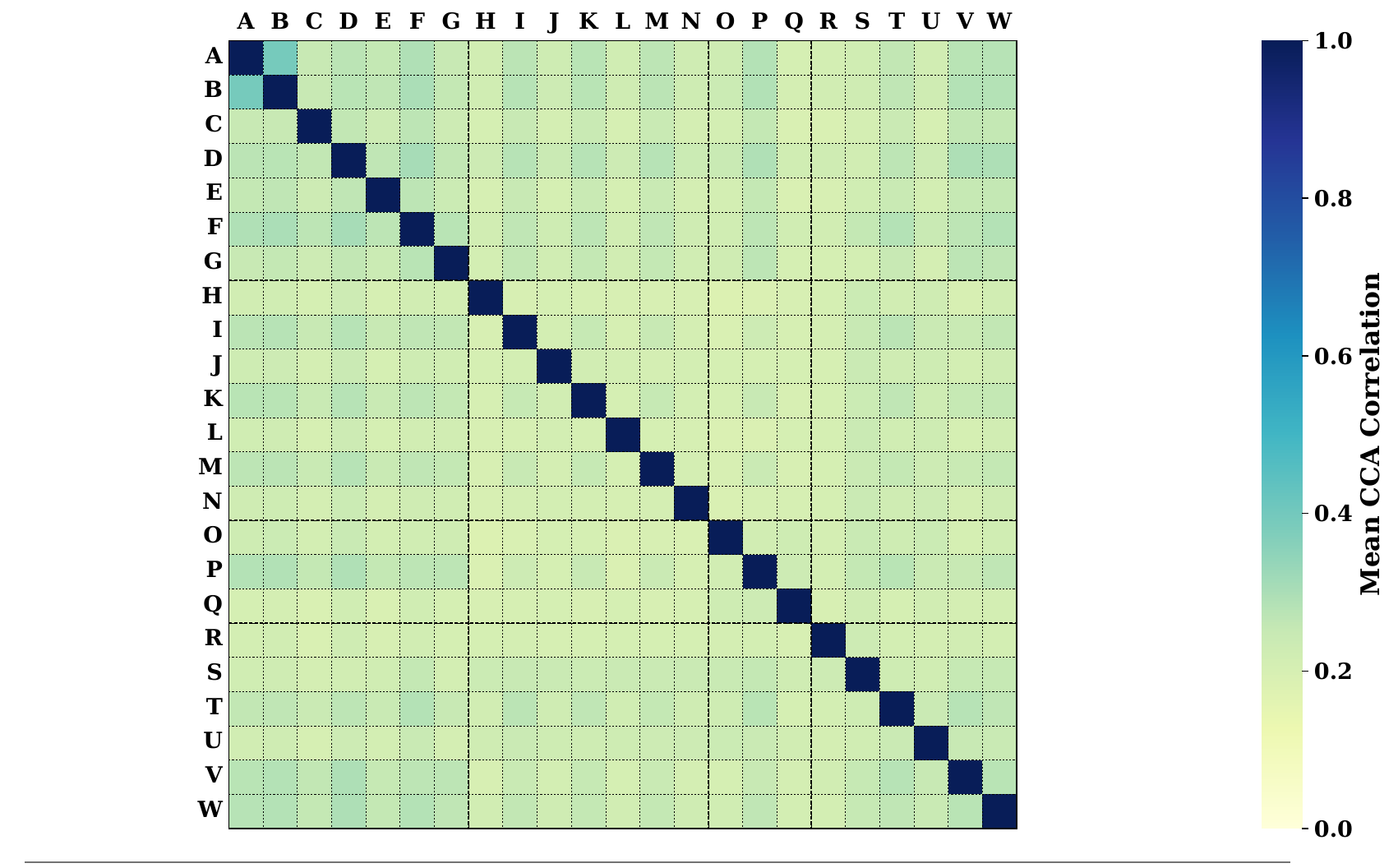
        }
        \vspace{-1mm}

        {\small (d) Random mean CCA, NOC}
    \end{minipage}

    \caption{
    CCA alignment after matched-size random coordinate selection.  Random
    coordinates can still produce a large leading canonical correlation,
    especially in NOC, demonstrating that primary CCA alone is not a sufficient
    universality test.  In contrast, random mean CCA is low in both spaces,
    showing that the multi-dimensional alignment in
    Figure~\ref{fig:cca-atlas-full} depends on behavior-enriched coordinate
    selection.
    }
    \label{fig:cca-random-atlas}
\end{figure*}

Figure~\ref{fig:cca-random-atlas} repeats the CCA analysis after replacing the
behavior-associated scaffolds with matched-size random coordinates.

The random controls change the interpretation of primary CCA.  The random
primary scores of $0.667$ in ACT and $0.817$ in NOC are too large to regard a
single canonical correlation as decisive evidence.  This effect can arise
because CCA selects the most favorable linear combination from two
high-dimensional feature populations evaluated on matched examples.

The random mean correlations, by contrast, are only $0.222$ and $0.233$.
The large behavior-associated margins for mean CCA therefore indicate that
alignment extends beyond a chance-optimized first pair.

\subsection{Cross-Model Linear-Association Atlas}

\begin{figure*}[t]
    \centering

    \begin{minipage}[t]{0.485\textwidth}
        \centering
        \atlaspanel{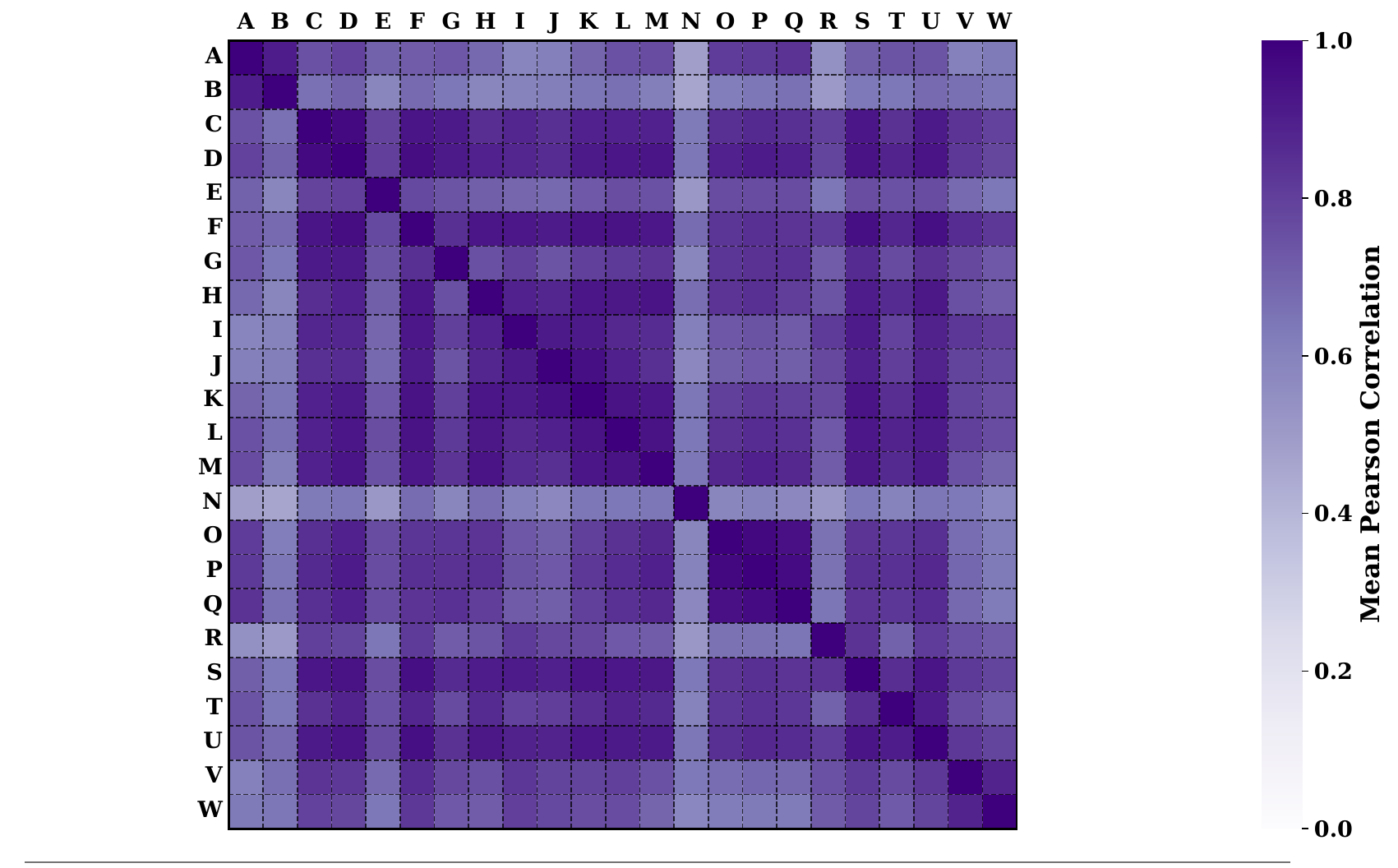}
        \vspace{-1mm}

        {\small (a) Top-2-PC probing, ACT}
    \end{minipage}
    \hfill
    \begin{minipage}[t]{0.485\textwidth}
        \centering
        \atlaspanel{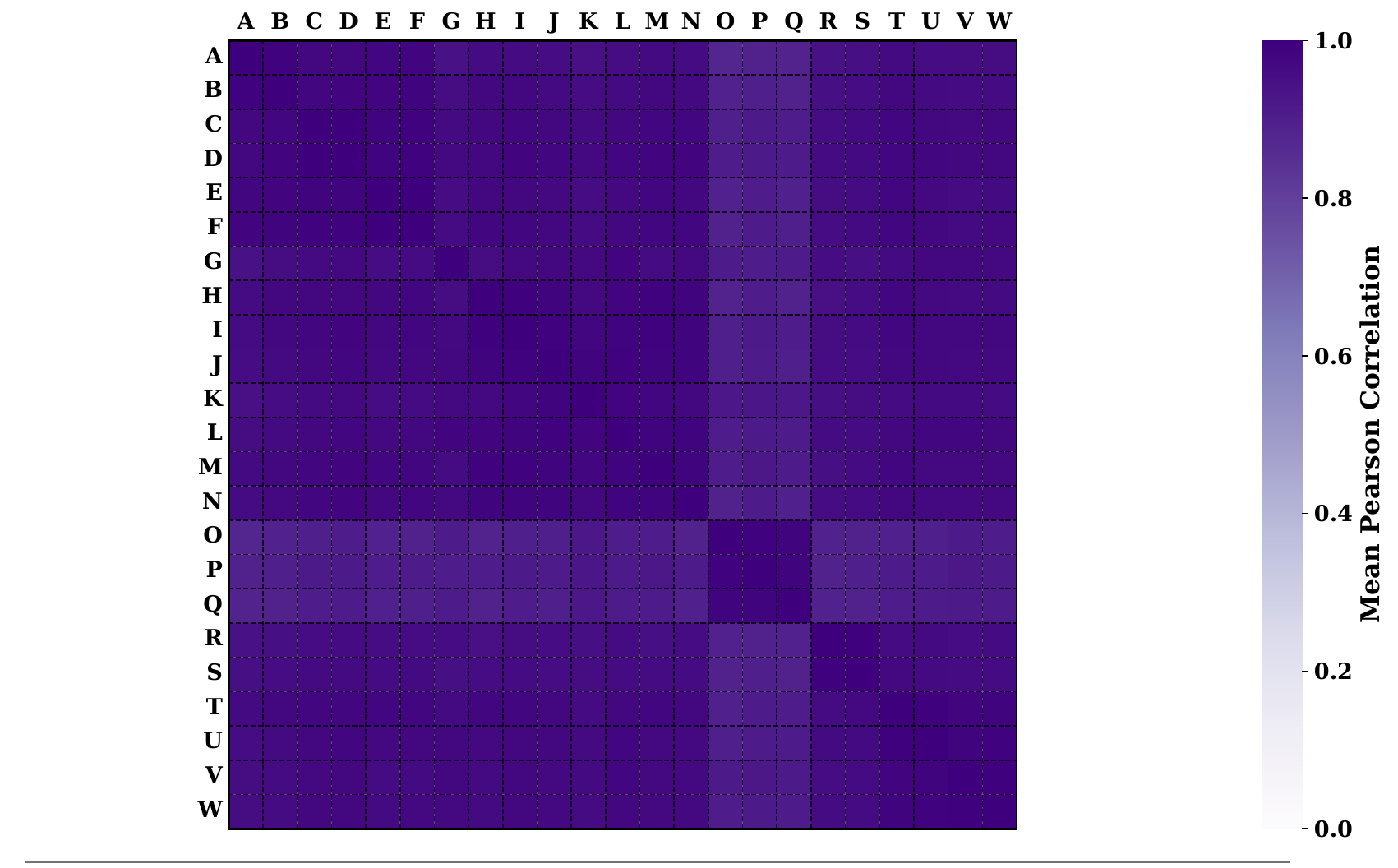}
        \vspace{-1mm}

        {\small (b) Top-2-PC probing, NOC}
    \end{minipage}

    \vspace{2mm}

    \begin{minipage}[t]{0.485\textwidth}
        \centering
        \atlaspanel{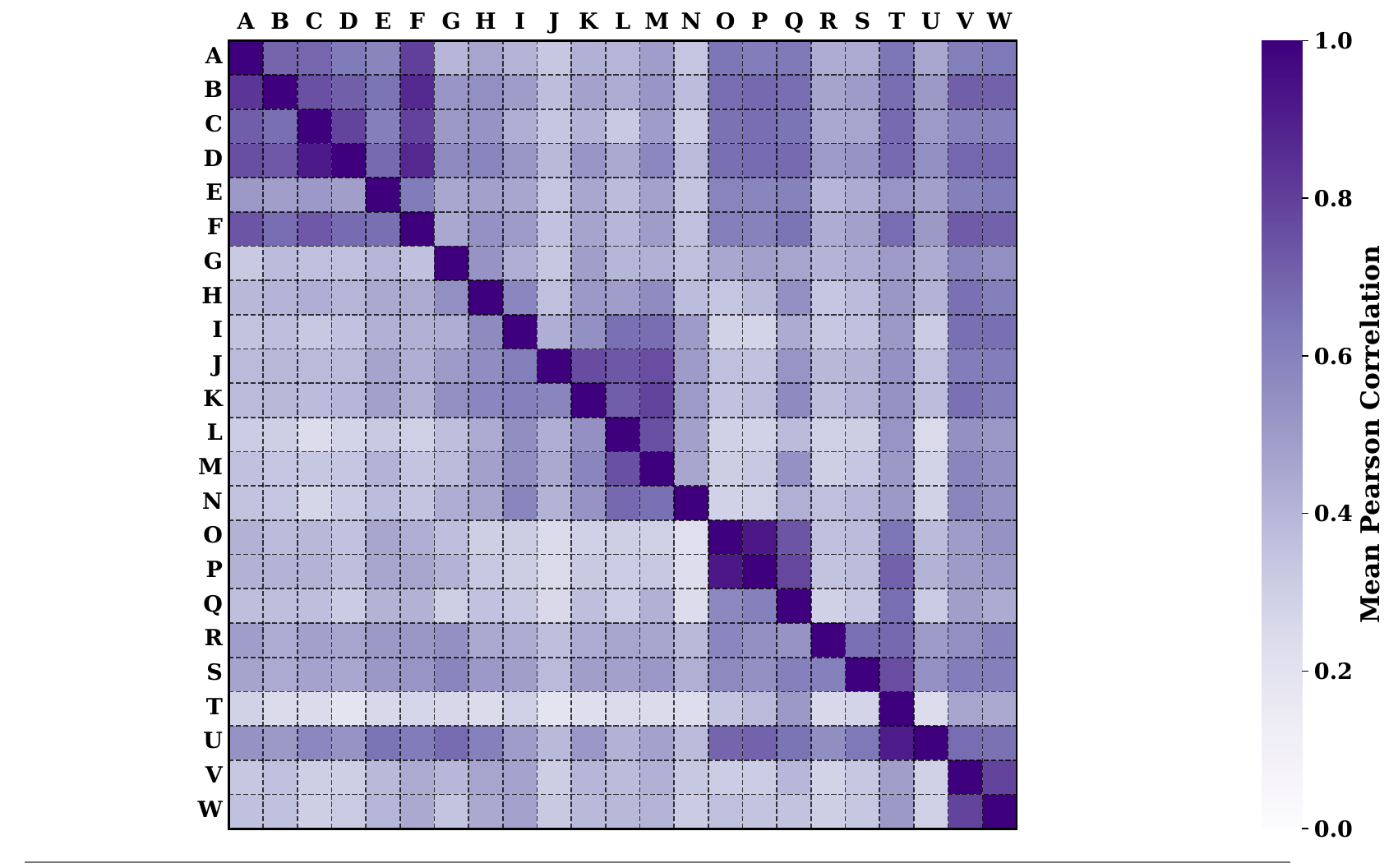}
        \vspace{-1mm}

        {\small (c) $k$-dimensional probing, ACT}
    \end{minipage}
    \hfill
    \begin{minipage}[t]{0.485\textwidth}
        \centering
        \atlaspanel{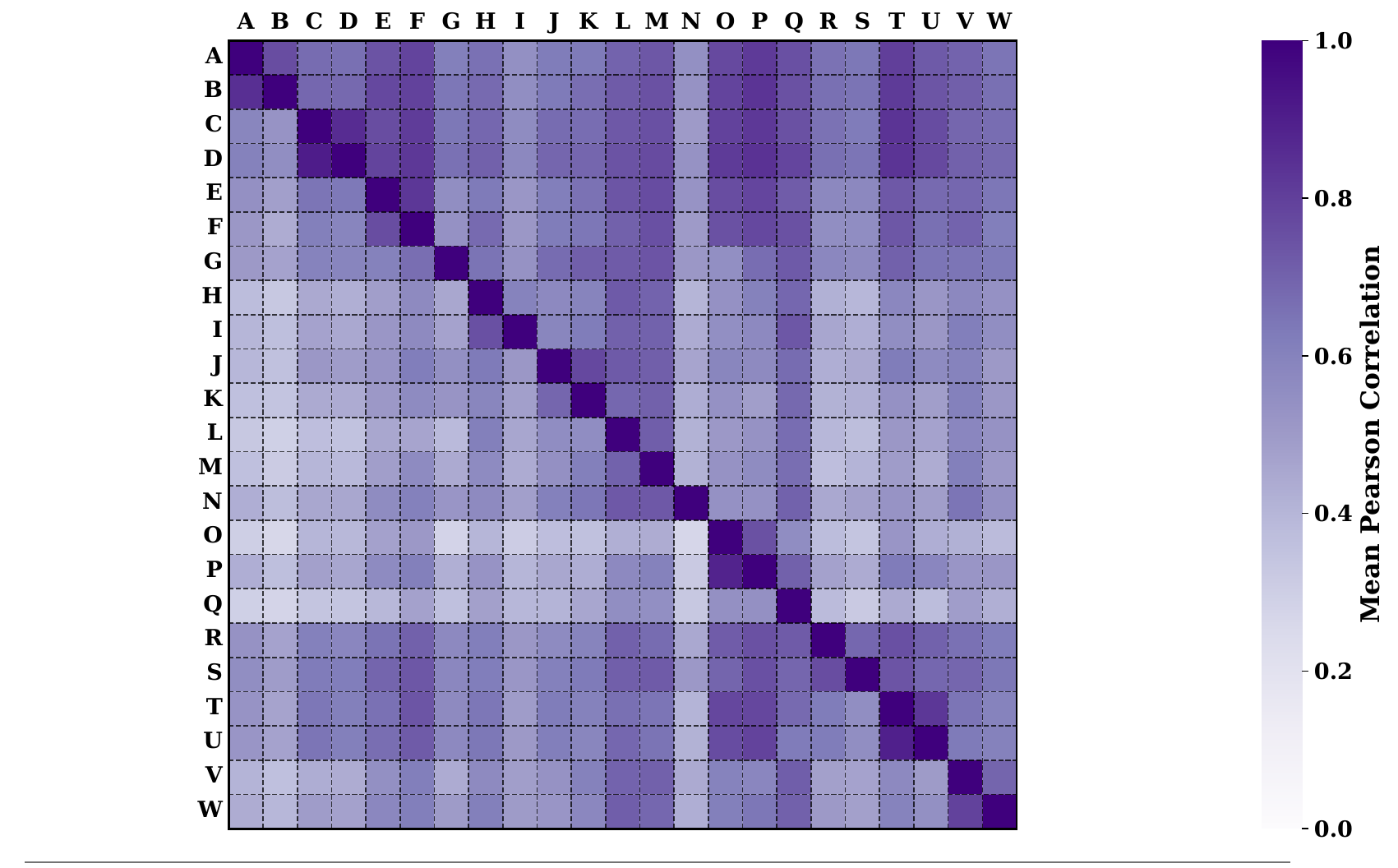}
        \vspace{-1mm}

        {\small (d) $k$-dimensional probing, NOC}
    \end{minipage}

    \caption{
    Cross-model linear association over matched examples.  Rows are source models and columns
    are target models.  Top-2-PC probing selects the most predictable mapping
    among the first two source and target PCs.  The $k$-dimensional probe
    predicts the complete variance-thresholded target chart and is therefore
    directed.  Leading behavioral coordinates exhibit strong cross-model association, especially
    in NOC, whereas complete-chart association is lower and asymmetric, revealing
    model-specific residual geometry.
    }
    \label{fig:probing-atlas-full}
\end{figure*}

Figure~\ref{fig:probing-atlas-full} reports cross-model linear associations
over the matched sample set.

The top-two-PC results are strong across most model pairs.  The mean score is
$0.784$ in ACT and $0.958$ in NOC.  This shows that a leading behavioral coordinate is strongly associated with a corresponding
coordinate in another model even when its PCA index
changes.

Full-chart association is more demanding.  Its mean score decreases to $0.473$
in ACT and $0.581$ in NOC.  The decrease should not be interpreted as a
failure of the atlas.  Rather, it distinguishes a broadly shared leading
behavioral coordinate from higher-dimensional residual variation that is
specific to architecture, scale, or post-training history.

The directed matrices also show asymmetry.  A low-rank or redundant source
chart may predict a target chart differently from the reverse mapping.  This
is expected when $k_m\neq k_{m'}$ or when one model's chart contains residual
coordinates absent from the other.

\subsection{Random-Coordinate Linear-Association Controls}

\begin{figure*}[t]
    \centering

    \begin{minipage}[t]{0.485\textwidth}
        \centering
        \atlaspanel{
            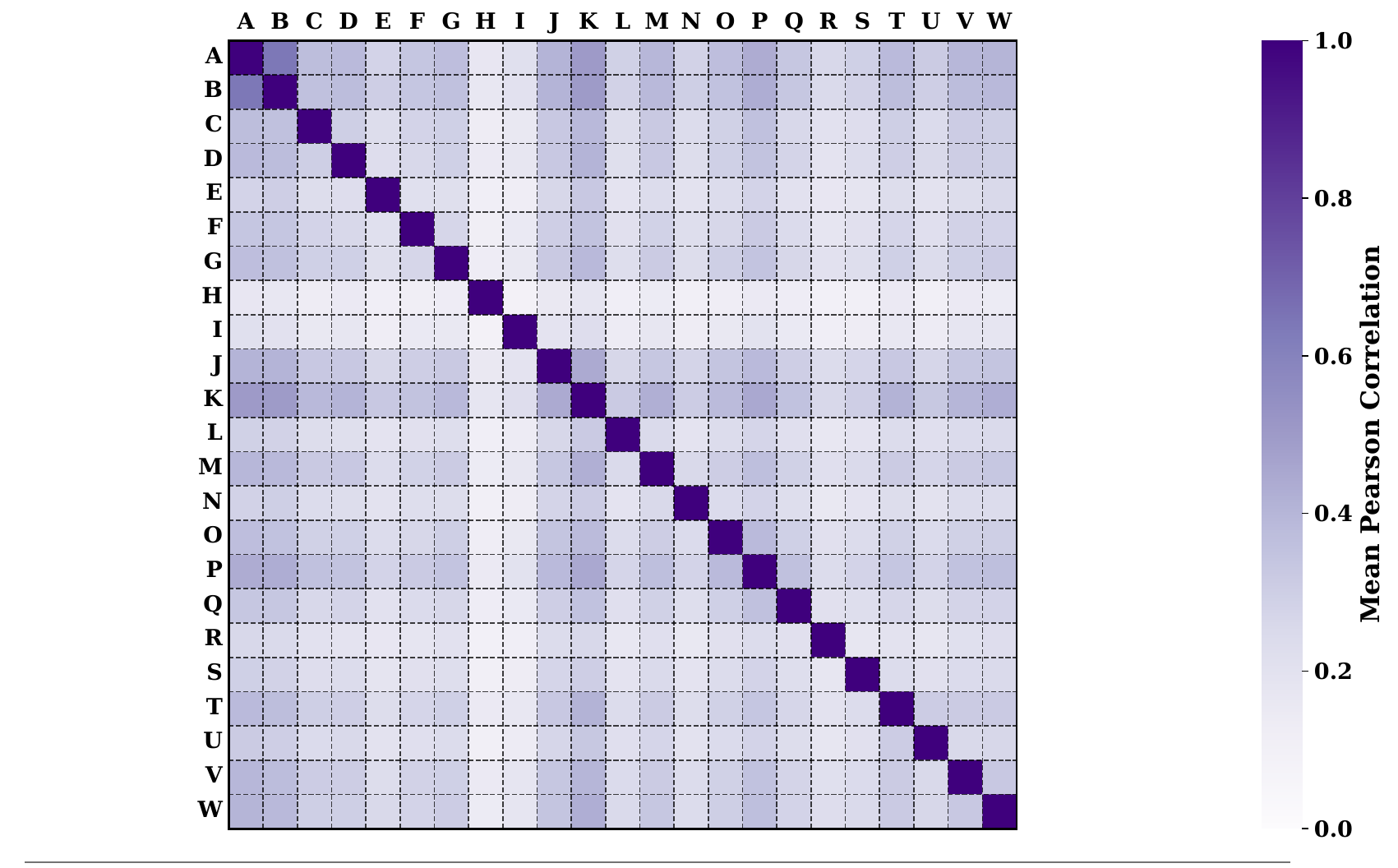
        }
        \vspace{-1mm}

        {\small (a) Random top-2-PC probing, ACT}
    \end{minipage}
    \hfill
    \begin{minipage}[t]{0.485\textwidth}
        \centering
        \atlaspanel{
            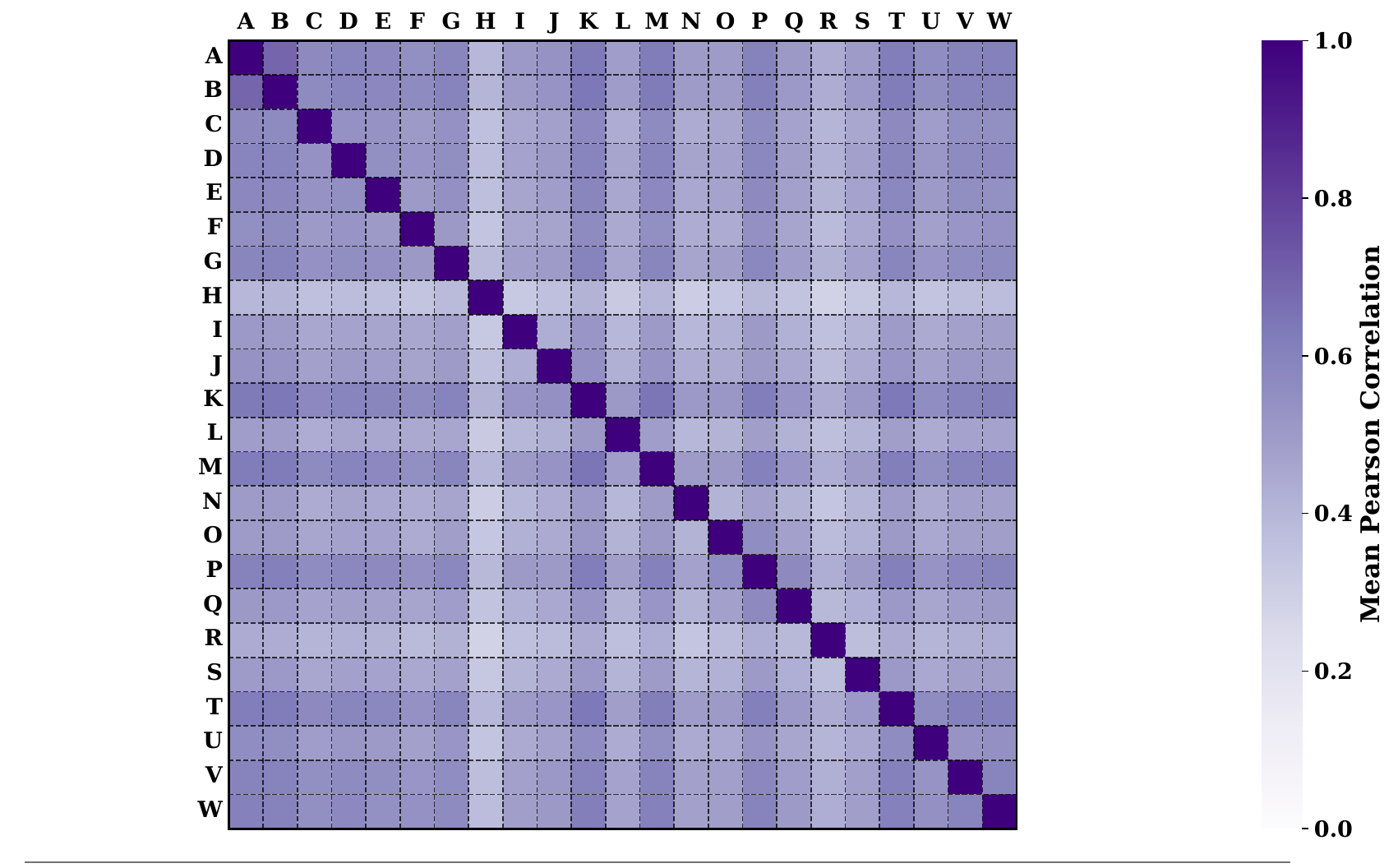
        }
        \vspace{-1mm}

        {\small (b) Random top-2-PC probing, NOC}
    \end{minipage}

    \vspace{2mm}

    \begin{minipage}[t]{0.485\textwidth}
        \centering
        \atlaspanel{
            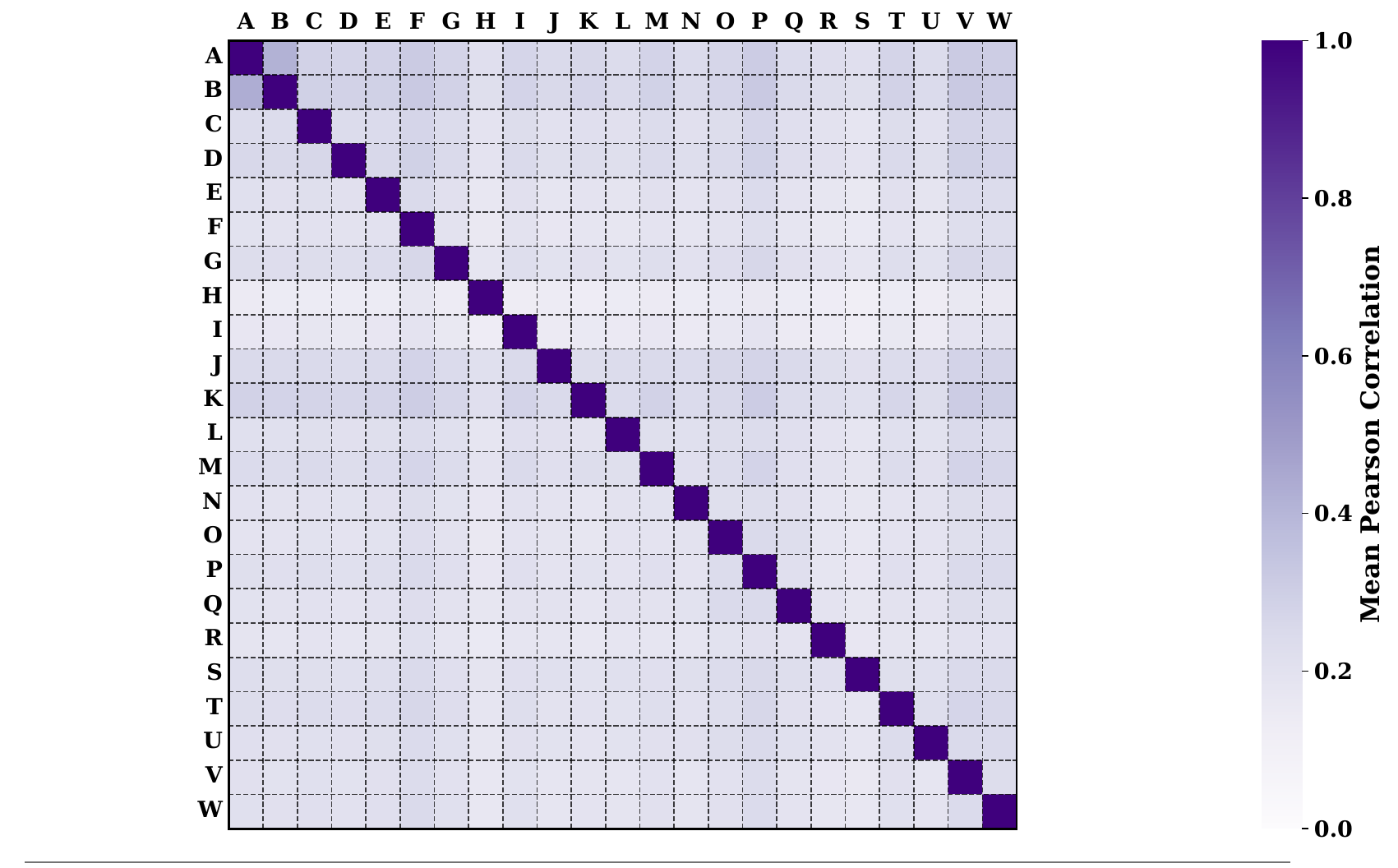
        }
        \vspace{-1mm}

        {\small (c) Random $k$-dimensional probing, ACT}
    \end{minipage}
    \hfill
    \begin{minipage}[t]{0.485\textwidth}
        \centering
        \atlaspanel{
            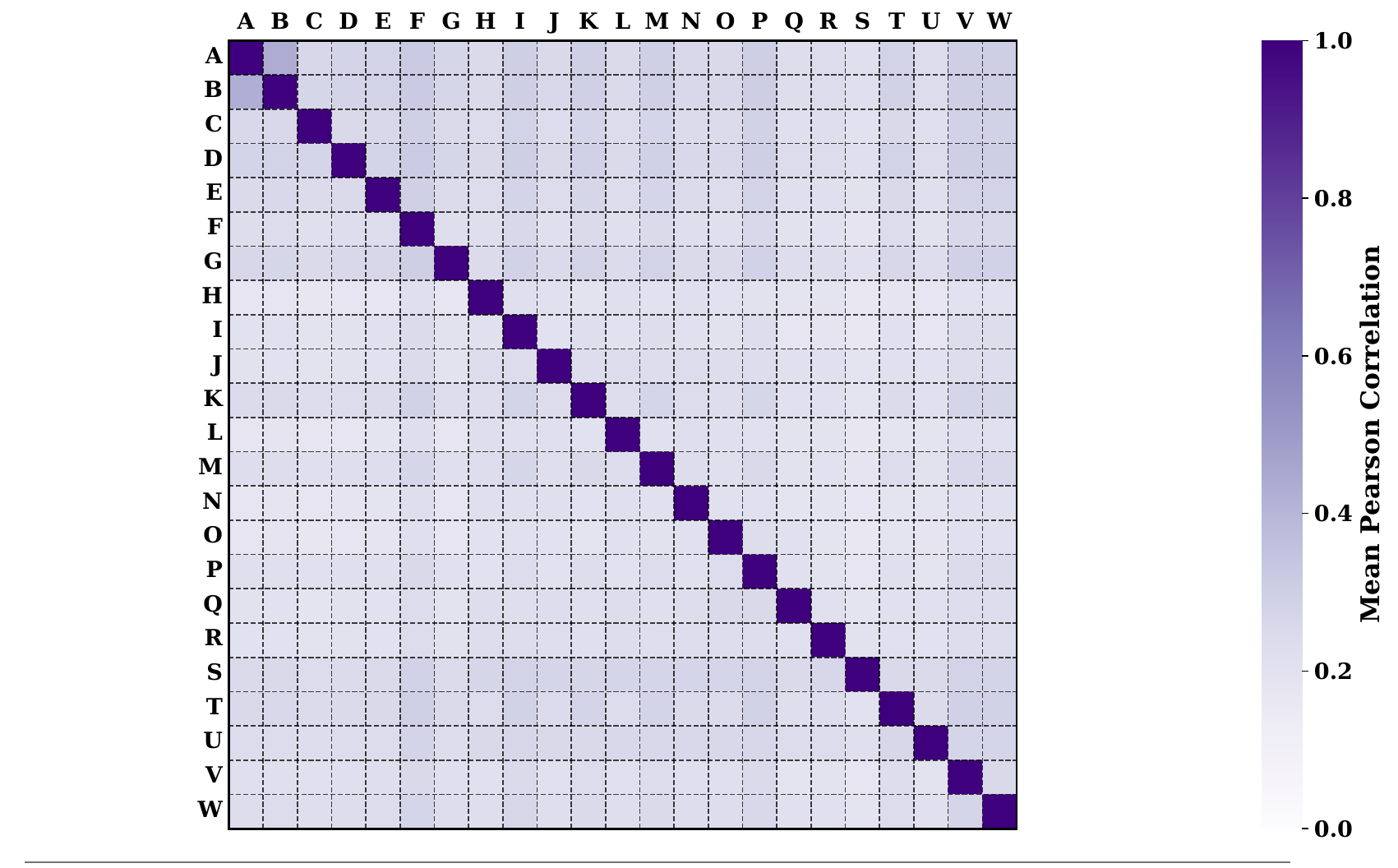
        }
        \vspace{-1mm}

        {\small (d) Random $k$-dimensional probing, NOC}
    \end{minipage}

    \caption{
    Linear association after matched-size random coordinate selection.
    Random top-2-PC NOC scores are moderately elevated because the procedure
    selects the best of four possible PC mappings and because generic
    contribution factors can be shared across models.  Nevertheless,
    behavior-associated association scores are substantially higher.  For complete-chart association, the random score is close to $0.23$ in both spaces, compared
    with $0.473$ in ACT and $0.581$ in NOC for behavior-associated charts.
    }
    \label{fig:probing-random-atlas}
\end{figure*}

Figure~\ref{fig:probing-random-atlas} shows the corresponding random-coordinate
probes.

The top-two-PC random NOC baseline of $0.495$ is non-negligible.  This is a
consequence of both the max-over-four selection rule and broadly shared
non-behavioral functional factors.  The behavior-associated score of $0.958$
nevertheless exceeds it by $0.463$.

The complete-chart random controls are more restrictive.  Their means are
$0.215$ in ACT and $0.234$ in NOC, compared with $0.473$ and $0.581$ for
behavior-associated charts. This indicates that the observed multi-dimensional cross-model association is
not explained by model-wide generic covariance alone.

\subsection{Family Structure and Matched Base--Instruction Pairs}

To characterize the residual atlas structure, we divide pairs into models from
the same family and models from different families.  Table~\ref{tab:atlas-family-summary} reports descriptive averages.  The model set is
not family-balanced, so these values are used to interpret the heatmaps rather
than as an independent hypothesis test.

\begin{table}[t]
\centering
\small
\setlength{\tabcolsep}{3.2pt}
\begin{tabular}{llrrrr}
\toprule
Metric & Space & All & Same fam. & Cross fam. & Base/Inst. \\
\midrule
Mean CCA
& ACT & 0.515 & 0.627 & 0.480 & 0.754 \\
& NOC & 0.648 & 0.677 & 0.639 & 0.798 \\
\midrule
Top-2 probe
& ACT & 0.784 & 0.816 & 0.775 & 0.905 \\
& NOC & 0.958 & 0.982 & 0.950 & 0.995 \\
\midrule
$k$-dim. probe
& ACT & 0.473 & 0.582 & 0.439 & 0.718 \\
& NOC & 0.581 & 0.634 & 0.565 & 0.775 \\
\bottomrule
\end{tabular}
\caption{
Cross-model alignment stratified by model relationship.  ``Base/Inst.''
denotes the ten matched base/instruction pairs with the same model family and
parameter scale; the $k$-dimensional value averages both mapping directions.
ACT shows a larger same-family advantage, whereas NOC remains strongly aligned across families.
}
\label{tab:atlas-family-summary}
\end{table}

The family gap is largest for ACT mean CCA and full-chart association.  Mean CCA
is $0.627$ within families but $0.480$ across families, and full-chart association
is $0.582$ versus $0.439$.  This indicates that raw activation-state geometry
retains architectural and family-specific organization.

NOC is more uniform across families.  Its mean CCA is $0.677$ within families
and $0.639$ across families, while top-two-PC association remains $0.950$ even
across families.  The normalized contribution representation therefore appears
to remove some basis-specific variation while preserving behaviorally relevant
functional structure.

Matched base/instruction pairs are especially aligned.  Their mean CCA reaches
$0.754$ in ACT and $0.798$ in NOC, and their top-two-PC association reaches
$0.905$ and $0.995$.  Instruction tuning therefore does not erase the
family-specific shared core, although the post-training analyses in
Section~\ref{sec:training_results} show that controlled SFT can still rotate a
behavioral chart substantially relative to its exact base checkpoint.

\subsection{Interpretation: Shared Core, Non-Universal Residuals}

The combination of metrics supports a layered interpretation of cross-model
behavioral universality.

First, classifier-score alignment and primary CCA show that nearly every model
pair contains a strongly shared behavior-oriented axis.  Second, mean CCA and
full-chart association show that this universality is incomplete: the remaining
coordinates retain family- and model-specific structure.  Third, random
controls demonstrate that behavior-associated coordinate selection is necessary
for the multi-dimensional effect.  Finally, the stronger NOC results indicate
that functional contribution geometry is more consistently aligned across models
than raw activation geometry. 

The appropriate claim is therefore not
\[
    U_{m,b}=U_{m',b},
\]
which would be undefined across models with different ambient bases.  Instead,
the results support
\[
    C_m^\top z_{m,b}(x)
    \approx
    C_{m'}^\top z_{m',b}(x)
\]
for a low-dimensional shared core, together with model-specific residual
coordinates outside that core.

This distinction also motivates Universal Fisher Strength.  A direction should
not be considered universal merely because one CCA component can be made highly
correlated.  It should separate behavioral classes, exhibit consistent
cross-model alignment, and outperform matched random geometry.  Mean CCA and
full-chart linear association provide the cross-model reliability evidence used
to support that interpretation.

%% file: sections__supplement__r_MetricsCheckpointAnalysis.tex
\section{Cross-Family Repetition Steering Results}
\label{app:cross-family-steering}

We report the complete repetition-steering sweeps for 23 models from the Qwen,
Gemma, Llama, and Mistral families.  These experiments complement the aggregate
causal-gain statistic by exposing the complete dose--response curves, their
model-to-model variation, and their relationship to generated-token
perplexity.

\paragraph{Operators and plotting convention.}

The ACT panels use additive behavioral-chart steering:
\begin{equation}
    x'
    =
    x
    +
    \alpha
    U_kU_k^\top(x-\mu).
\end{equation}
The second set of panels uses the neurons selected by the NOC scaffold.  For
multiplicative neuron scaling, we display a zero-centered strength
$\alpha=s-1$, where $s$ is the actual multiplicative factor:
\begin{equation}
    a'_{\ell,t,j}
    =
    (1+\alpha)a_{\ell,t,j},
    \qquad
    (\ell,j)\in\mathcal N_{\theta,b,\mathrm{NOC}}.
\end{equation}
Consequently, $\alpha=0$ is the unmodified model in both sets of figures.

For every model and operator, we evaluate
\[
    \alpha\in\{-2,-1,0,1,2\}.
\]
Solid lines with circular markers show the empirical repetition rate on the
left axis.  Dashed lines with square markers show generated-token PPL on the
right axis.  Model colors are shared between the two panels for a given
family.  The left-axis limits differ across families and operators to make
within-panel variation visible; quantitative cross-family comparisons should
therefore use the summary tables rather than apparent visual slope.

\paragraph{Aggregate response.}

For model $m$, define the endpoint response
\begin{equation}
\label{eq:steering-endpoint-response}
    \Delta R_m^{(\mathrm{op})}
    =
    R_m^{(\mathrm{op})}(+2)
    -
    R_m^{(\mathrm{op})}(-2).
\end{equation}
Table~\ref{tab:steering-global-summary} summarizes the full model set.

\begin{table}[t]
\centering
\small
\setlength{\tabcolsep}{3.4pt}
\begin{tabular}{lrrrr}
\toprule
Operator
& Mean $\Delta R$
& Med. $\Delta R$
& Positive
& Med. $|\Delta\mathrm{PPL}|$
\\
\midrule
ACT subspace
& 0.119
& 0.102
& 23/23
& 0.223
\\
NOC BAC
& 0.029
& 0.019
& 23/23
& 0.239
\\
\bottomrule
\end{tabular}
\caption{
Aggregate repetition-steering response across 23 models.
$\Delta R=R(+2)-R(-2)$, and
$|\Delta\mathrm{PPL}|=|\mathrm{PPL}(+2)-\mathrm{PPL}(-2)|$.
Both operators produce a positive endpoint behavioral change for every model.
The median PPL change is small, although a small number of model-specific
outliers increase the corresponding mean.
}
\label{tab:steering-global-summary}
\end{table}

The ACT-chart intervention produces the larger response under the chosen
normalization, with a mean endpoint increase of $0.119$.  The NOC-selected BAC
intervention is more conservative but remains directionally consistent, with a
positive endpoint effect for all 23 models.  Raw effect sizes should not be
interpreted as a direct causal ranking between operators because an ACT chart
component and a multiplicative neuron scale do not have the same physical unit.

Fifteen of the 23 ACT curves and fifteen of the 23 NOC/BAC curves are
non-decreasing at every displayed strength.  The remaining curves generally
increase around the no-intervention point and then saturate or mildly reverse
at the largest positive strength.  This behavior supports estimating
$G^{\mathrm{causal}}$ locally rather than assuming a globally linear response.

Table~\ref{tab:steering-family-summary} reports family-level endpoint changes.

\begin{table}[t]
\centering
\small
\setlength{\tabcolsep}{5pt}
\begin{tabular}{lrrr}
\toprule
Family & Models & ACT subspace & NOC BAC \\
\midrule
Qwen    & 7 & 0.148 & 0.028 \\
Gemma   & 7 & 0.127 & 0.036 \\
Llama   & 3 & 0.078 & 0.032 \\
Mistral & 6 & 0.096 & 0.020 \\
\midrule
All     & 23 & 0.119 & 0.029 \\
\bottomrule
\end{tabular}
\caption{
Mean endpoint repetition change
$R(+2)-R(-2)$ by model family.  Positive responses occur in every family for
both intervention operators.
}
\label{tab:steering-family-summary}
\end{table}

\subsubsection{Qwen Family}

\begin{figure*}[t]
    \centering

    \begin{minipage}[t]{0.485\textwidth}
        \centering
        \includegraphics[
            width=\linewidth,
            keepaspectratio
        ]{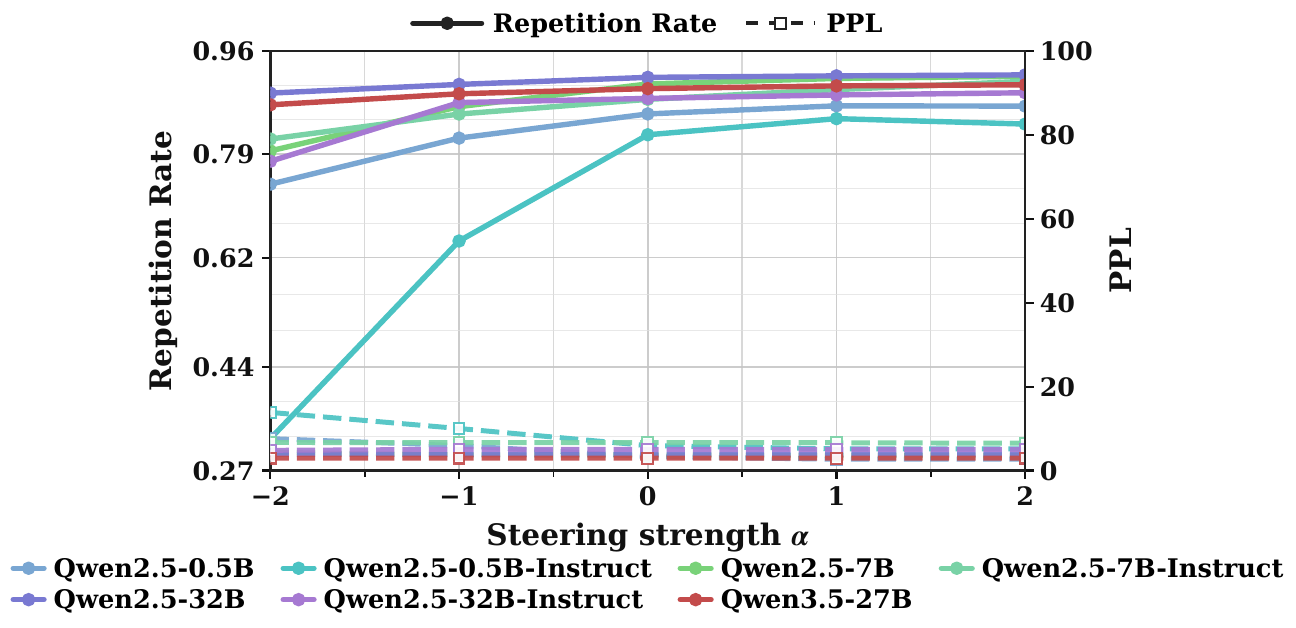}
        \vspace{-1.2mm}

        {\small (a) ACT-chart subspace steering}
    \end{minipage}
    \hfill
    \begin{minipage}[t]{0.485\textwidth}
        \centering
        \includegraphics[
            width=\linewidth,
            keepaspectratio
        ]{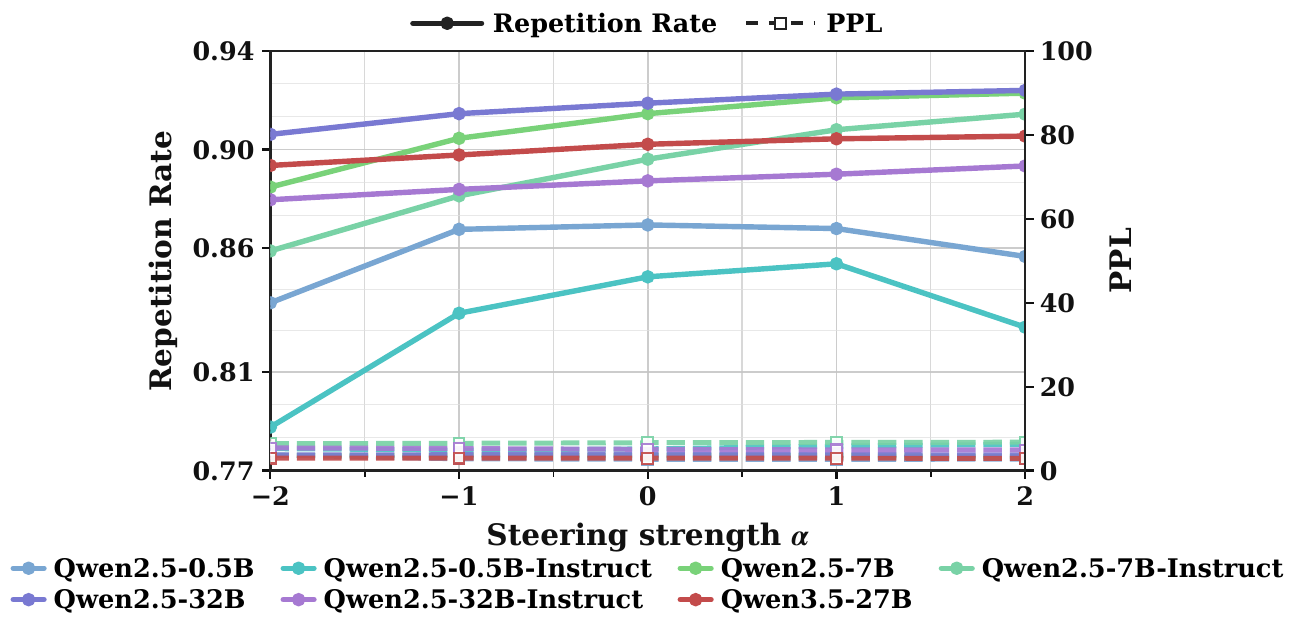}
        \vspace{-1.2mm}

        {\small (b) NOC-selected BAC steering}
    \end{minipage}

    \caption{
    Repetition steering in the Qwen family.  Solid circles show repetition
    rate and dashed squares show generated-token PPL.  ACT-chart steering
    produces a pronounced dose--response for several models, most notably
    Qwen2.5-0.5B-Instruct.  NOC-selected BAC steering is weaker but remains
    positive around the no-intervention point for all seven models.  Several
    small-model curves saturate or mildly reverse at the largest strength.
    }
    \label{fig:steering-qwen}
\end{figure*}

Figure~\ref{fig:steering-qwen} shows seven Qwen models spanning three
Qwen2.5 scales, their instruction-tuned counterparts, and Qwen3.5-27B.

The largest ACT response occurs for Qwen2.5-0.5B-Instruct, whose repetition
rate increases from $0.322$ at $\alpha=-2$ to $0.840$ at $\alpha=2$.  The
increase is already mostly realized by $\alpha=0$ and $\alpha=1$, after which
the curve slightly saturates.  Qwen2.5-0.5B, Qwen2.5-7B, and
Qwen2.5-32B-Instruct also show substantial positive changes.  Models with
baseline repetition rates near $0.9$, such as Qwen2.5-32B, have less available
headroom and consequently display smaller raw probability changes.

NOC-selected BAC steering produces smaller effects.  The clearest response is
for Qwen2.5-7B-Instruct, which increases by approximately $0.055$ across the
displayed range.  The two 0.5B curves peak at an intermediate positive strength
and decline slightly at $\alpha=2$, illustrating why the local slope is more
appropriate than the full-range endpoint as a mechanistic gain estimate.

PPL is stable or decreases for most Qwen models.  In particular, the large
behavioral increase in Qwen2.5-0.5B-Instruct under ACT steering is accompanied
by a decrease rather than an increase in PPL.  The effect is therefore not
readily explained by general model degradation.

\subsubsection{Gemma Family}

\begin{figure*}[t]
    \centering

    \begin{minipage}[t]{0.485\textwidth}
        \centering
        \includegraphics[
            width=\linewidth,
            keepaspectratio
        ]{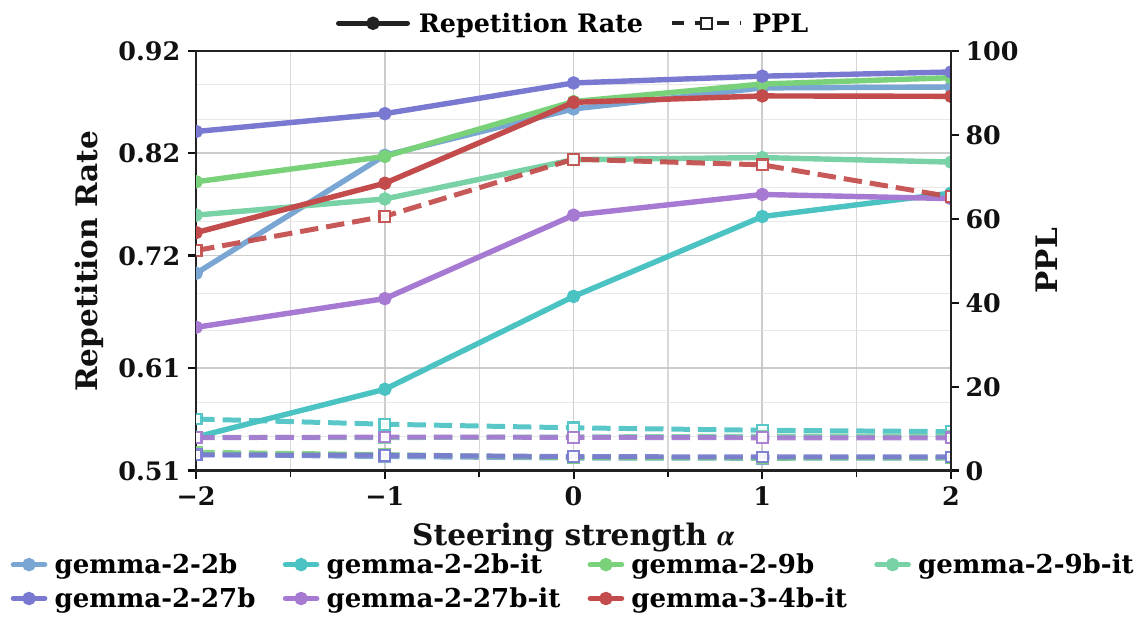}
        \vspace{-1.2mm}

        {\small (a) ACT-chart subspace steering}
    \end{minipage}
    \hfill
    \begin{minipage}[t]{0.485\textwidth}
        \centering
        \includegraphics[
            width=\linewidth,
            keepaspectratio
        ]{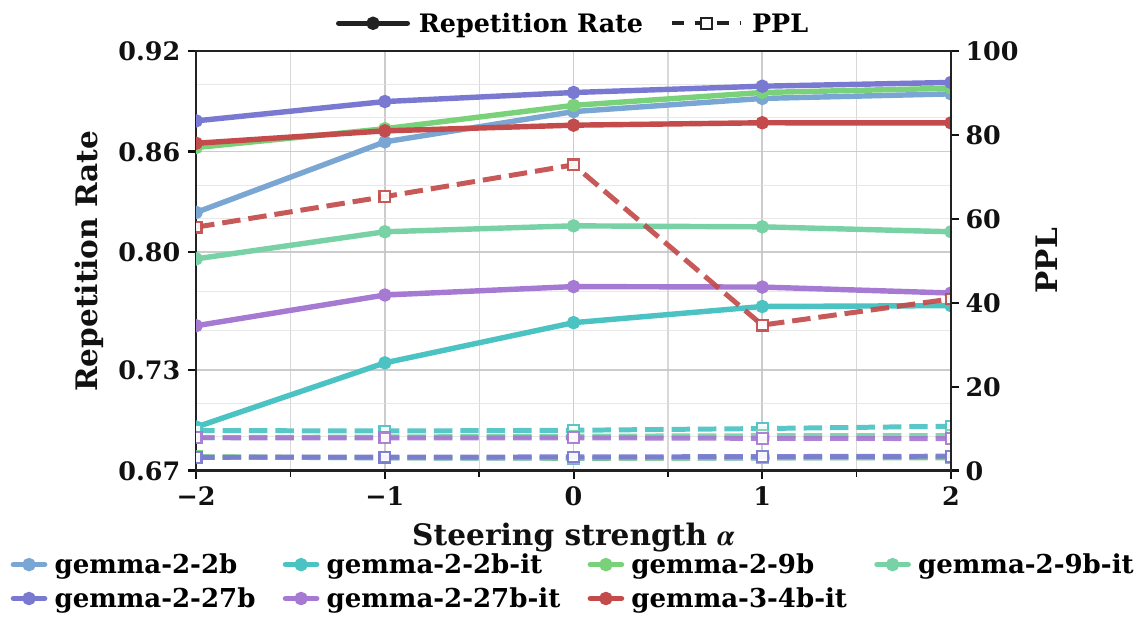}
        \vspace{-1.2mm}

        {\small (b) NOC-selected BAC steering}
    \end{minipage}

    \caption{
    Repetition steering in the Gemma family.  ACT-chart steering increases
    repetition most strongly for the 2B models and produces positive endpoint
    changes for all seven models.  NOC-selected BAC steering gives a weaker but
    consistent response. 
    }
    \label{fig:steering-gemma}
\end{figure*}

Figure~\ref{fig:steering-gemma} reports seven Gemma models.

Gemma-2-2B and Gemma-2-2B-it exhibit the largest ACT responses in this family,
with endpoint increases of approximately $0.182$ and $0.238$.  Gemma-2-9B and
Gemma-2-27B-it also show clear positive dose--response curves.  The
Gemma-2-9B-it curve increases near the origin and then plateaus, emphasizing
that the chart intervention is most reliably interpreted locally.

NOC-selected BAC steering is again more conservative.  The strongest effects
occur for Gemma-2-2B and Gemma-2-2B-it, both increasing by approximately
$0.07$.  The larger Gemma-2 models have higher baseline repetition and exhibit
smaller absolute changes.

\subsubsection{Llama Family}

\begin{figure*}[t]
    \centering

    \begin{minipage}[t]{0.485\textwidth}
        \centering
        \includegraphics[
            width=\linewidth,
            keepaspectratio
        ]{figures__steering__act_subspace_llama.pdf}
        \vspace{-1.2mm}

        {\small (a) ACT-chart subspace steering}
    \end{minipage}
    \hfill
    \begin{minipage}[t]{0.485\textwidth}
        \centering
        \includegraphics[
            width=\linewidth,
            keepaspectratio
        ]{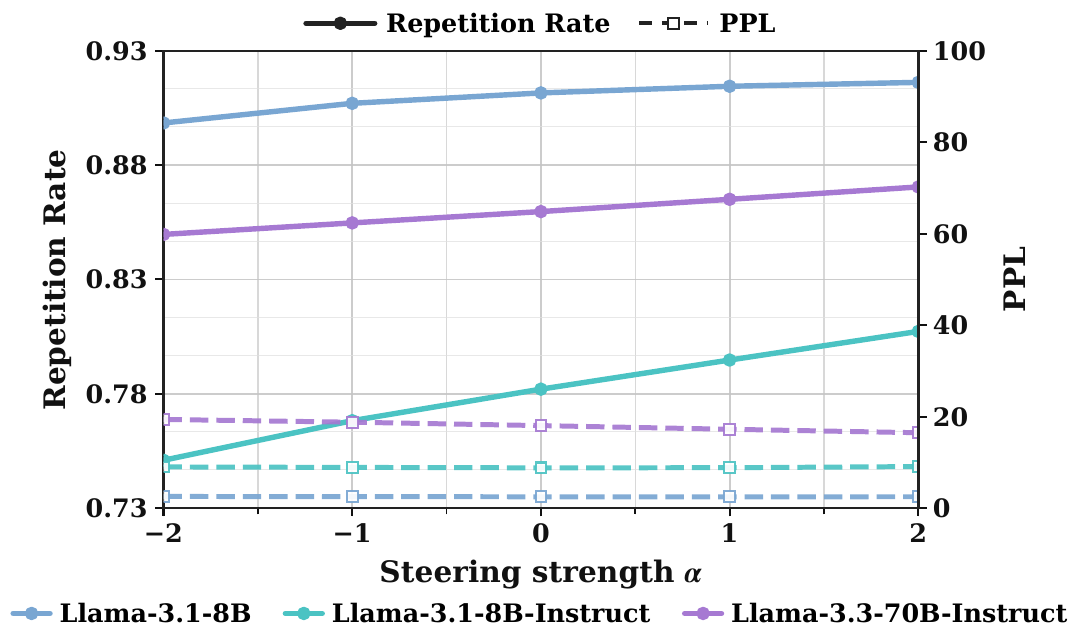}
        \vspace{-1.2mm}

        {\small (b) NOC-selected BAC steering}
    \end{minipage}

    \caption{
    Complete Llama-family repetition steering.  All three models exhibit
    monotonic positive dose--response curves under both operators.  The effect
    is strongest for Llama-3.1-8B-Instruct, while the associated PPL remains
    nearly constant.  The different left-axis ranges should be considered when
    visually comparing the two operator magnitudes.
    }
    \label{fig:steering-llama-app}
\end{figure*}

Figure~\ref{fig:steering-llama-app} reports the three Llama models.  These
panels are also shown in the main text because they provide the cleanest
same-family comparison between intervention operators.

The Llama results provide the most visually unambiguous causal validation.
Every repetition curve is monotonic over all five strengths.  Under ACT-chart
steering, Llama-3.1-8B-Instruct increases by $0.142$,
Llama-3.3-70B-Instruct by $0.058$, and Llama-3.1-8B by $0.033$.
The corresponding NOC/BAC changes are smaller but remain positive:
$0.056$, $0.021$, and $0.018$.

The base Llama-3.1-8B begins near a repetition rate of $0.9$, leaving limited
probability-scale headroom.  The instruction-tuned 8B model begins
substantially lower and therefore exhibits a much larger raw response.  This
ceiling effect motivates using logit-scale local gain alongside raw repetition
probability.

PPL remains stable throughout the ACT sweep.  In the NOC/BAC experiment, the
Llama-3.3-70B-Instruct PPL decreases moderately as repetition increases,
further indicating that the behavioral response is not generated by a generic
loss of fluency.

\subsubsection{Mistral Family}

\begin{figure*}[t]
    \centering

    \begin{minipage}[t]{0.485\textwidth}
        \centering
        \includegraphics[
            width=\linewidth,
            keepaspectratio
        ]{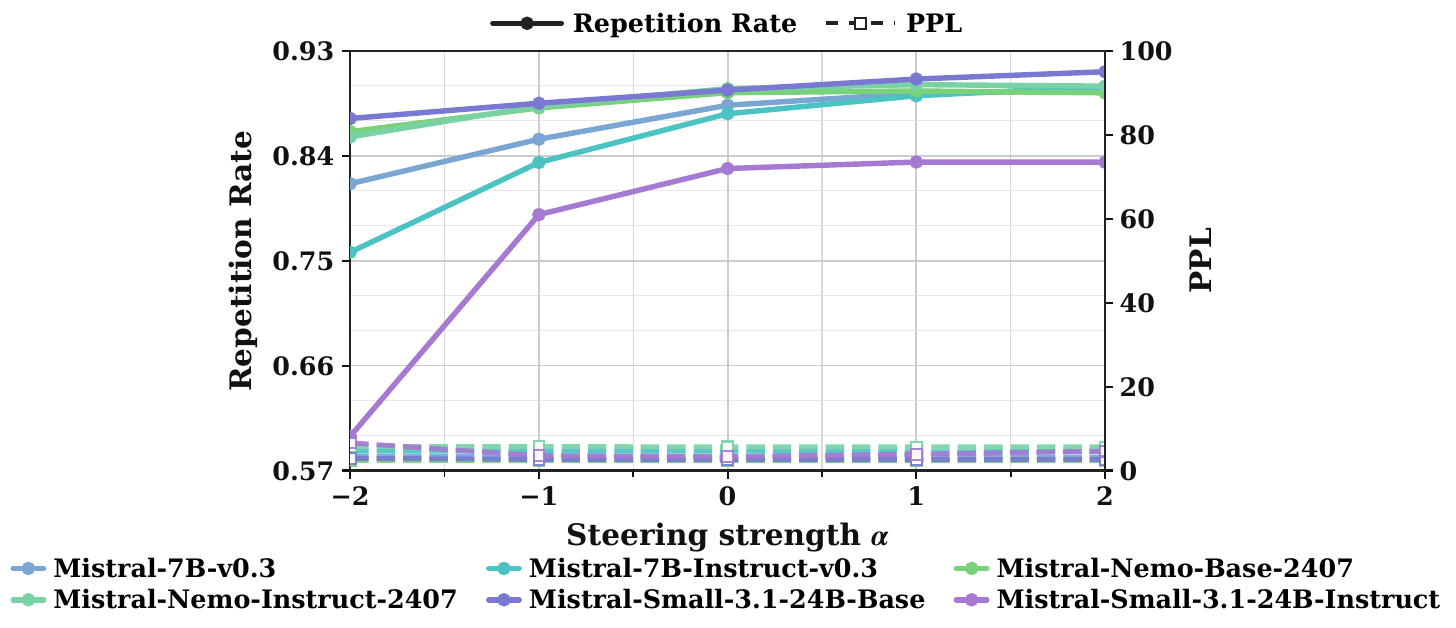}
        \vspace{-1.2mm}

        {\small (a) ACT-chart subspace steering}
    \end{minipage}
    \hfill
    \begin{minipage}[t]{0.485\textwidth}
        \centering
        \includegraphics[
            width=\linewidth,
            keepaspectratio
        ]{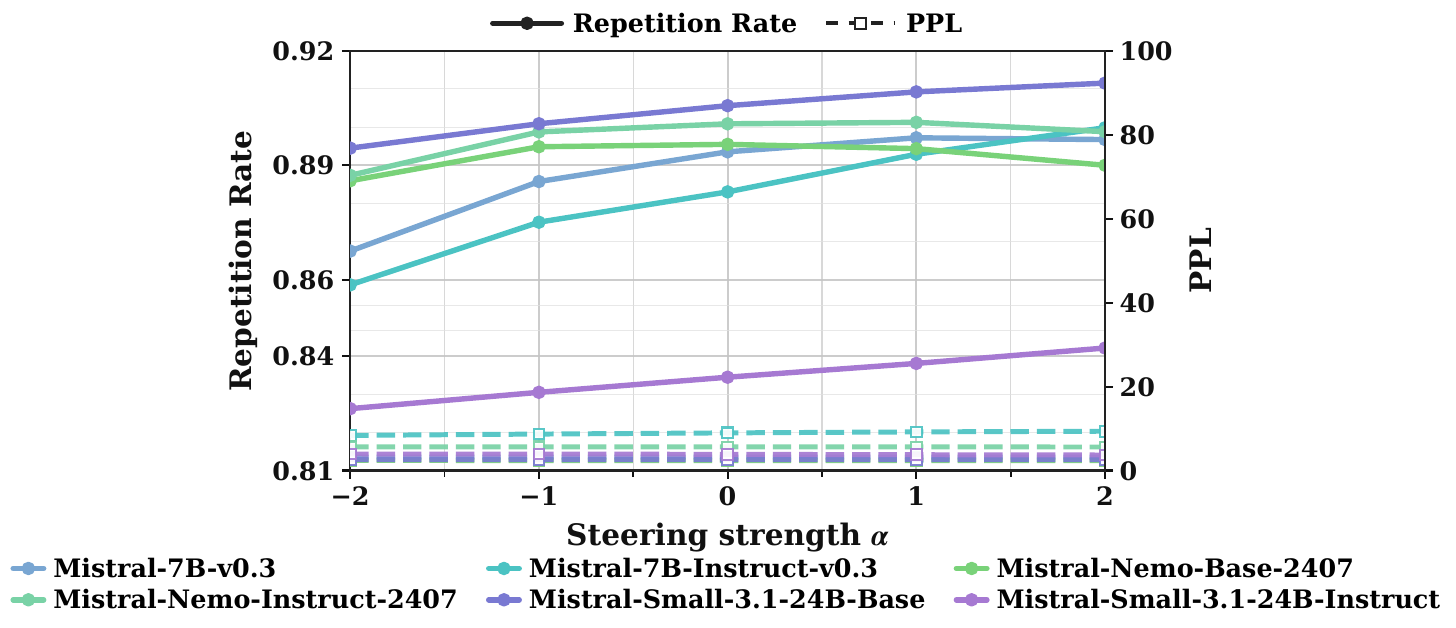}
        \vspace{-1.2mm}

        {\small (b) NOC-selected BAC steering}
    \end{minipage}

    \caption{
    Repetition steering in the Mistral family.  ACT-chart steering produces
    especially large changes for Mistral-7B-Instruct-v0.3 and
    Mistral-Small-3.1-24B-Instruct.  The remaining models exhibit smaller
    positive and frequently saturating responses.  NOC-selected BAC steering is
    more conservative but preserves the expected local direction across the
    family.
    }
    \label{fig:steering-mistral}
\end{figure*}

Figure~\ref{fig:steering-mistral} reports six Mistral models.

Mistral-Small-3.1-24B-Instruct has the largest ACT response in this family,
increasing from $0.599$ to $0.834$.  Mistral-7B-Instruct-v0.3 increases from
$0.757$ to $0.900$.  The base and Nemo variants begin closer to the upper
probability range and show smaller, saturating changes.

NOC-selected BAC steering produces modest positive endpoint changes.  The
largest occurs for Mistral-7B-Instruct-v0.3, which increases by approximately
$0.041$.  The Nemo models rise around the no-intervention point but show small
downturns at the largest positive strength, again supporting a local rather
than globally linear interpretation.

PPL remains low and comparatively stable for most Mistral models.  The
Mistral-Small-3.1-24B-Instruct ACT intervention initially reduces PPL and then
partially rebounds, while the repetition effect remains strongly positive.

%% file: sections__supplement__s_PublicAlignmentEffects.tex
\section{Cross-Model Repetition Strength and Public Alignment Effects}
\label{app:strength-alignment-analysis}

This appendix contains the complete cross-model repetition-strength and
public Base--Instruct analyses summarized in
Section~\ref{sec:manifold_results}.  It defines the ACT--NOC strength
landscape and paired-difference measures, reports the complete figures and
pair-level values, and presents the aggregate interpretation and associated
limitations.

\subsection{Construction of the Cross-Model Strength Landscape}

For each model $m$, the landscape uses the coordinate
\begin{equation}
    \left(
        x_m,y_m
    \right)
    =
    \left(
        \log\!\left(
            G^{\mathrm{uniF}}_{m,\mathrm{NOC}}+\epsilon
        \right),
        \log\!\left(
            G^{\mathrm{uniF}}_{m,\mathrm{ACT}}+\epsilon
        \right)
    \right).
\end{equation}
where $\epsilon>0$ is a small numerical stabilizer.

The two axes measure chart reliability in complementary spaces.  NOC reflects
whether behavior-associated coordinates form a stable functional-contribution
channel, whereas ACT reflects whether behavior-positive and behavior-negative
runtime states are reliably organized in activation space.

These quantities should not be identified directly with behavior frequency.
A model can possess a reliable behavior chart while rarely entering it, or can
frequently express a behavior whose internal chart is weakly separated.  We
therefore encode baseline repetition tendency separately by color:
\begin{equation}
    T_m
    =
    \Pr_m
    \left(
        y_{\mathrm{rep}}=1
        \mid
        \text{no intervention}
    \right).
\end{equation}
Bubble area is a monotone function of the local causal gain
$G^{\mathrm{causal}}_{m,\Omega}$.  The two panels use identical point
coordinates and colors; they differ only in whether bubble area is determined
by ACT- or NOC-space gain.

\begin{figure}[t]
    \centering
    \includegraphics[
        width=0.98\columnwidth,
        keepaspectratio
    ]{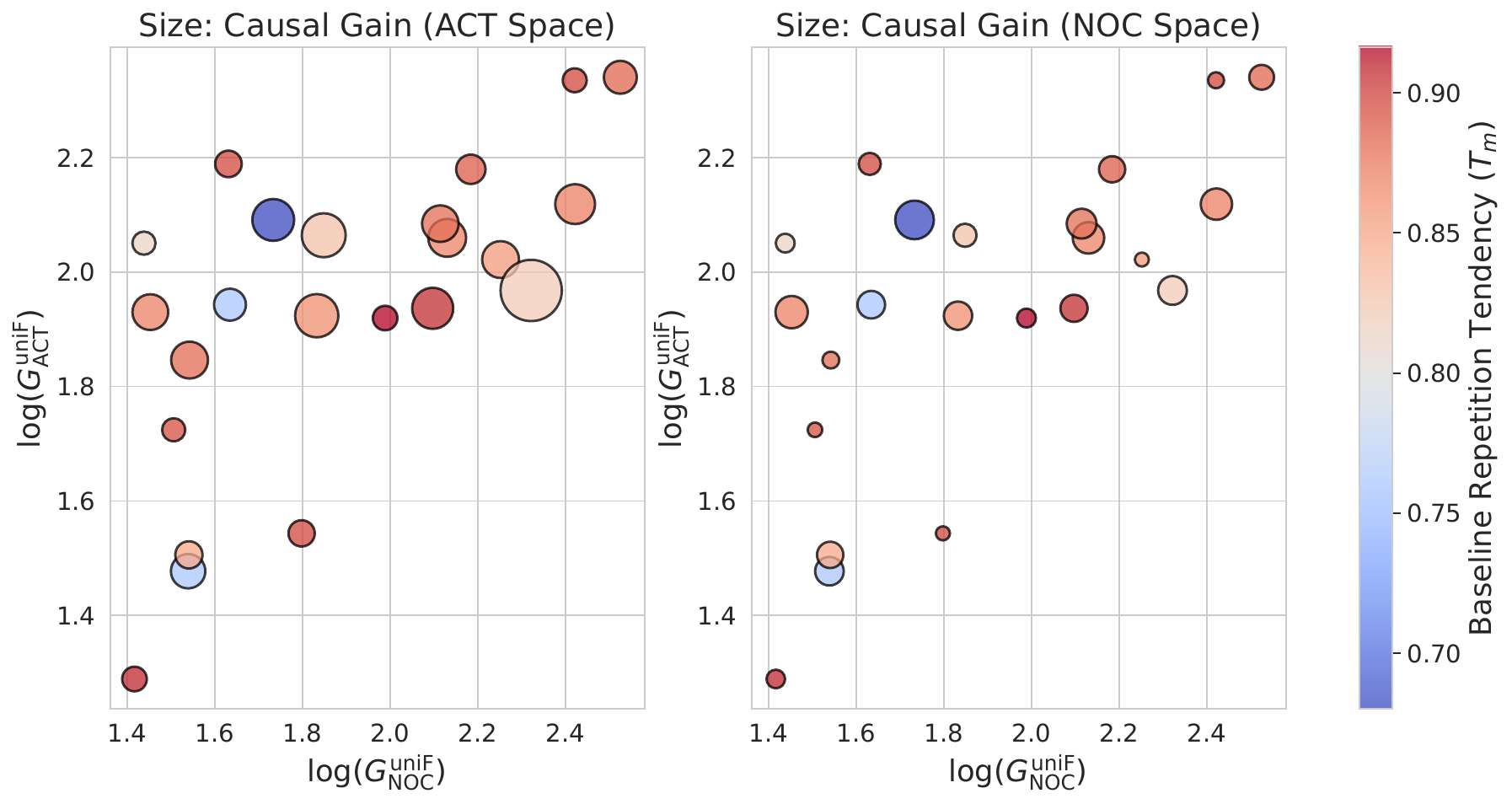}

    \caption{
    Complete cross-model repetition-strength landscape.
    Horizontal and vertical coordinates are
    $\log(G^{\mathrm{uniF}}_{\mathrm{NOC}}+\epsilon)$ and
    $\log(G^{\mathrm{uniF}}_{\mathrm{ACT}}+\epsilon)$.
    Point color denotes baseline repetition tendency $T_m$.
    The left and right panels encode ACT-space and NOC-space causal gain by
    bubble area, respectively.  Models with high values on both axes have
    reliable behavioral charts in both runtime activation and functional
    contribution spaces.  Off-diagonal points indicate representation-space
    asymmetry, while variation in color and bubble area demonstrates that chart
    reliability, ordinary behavioral usage, and causal potency are distinct.
    }
    \label{fig:full-strength-landscape}
\end{figure}

The ACT and NOC chart strengths are positively associated across models
(Pearson $r\approx0.64$ in the displayed log coordinates), but the scatter is
substantial.  The relationship therefore supports a shared behavioral
mechanism without implying that the two spaces are redundant.

The landscape can be read continuously rather than as a hard four-quadrant
classification:

\begin{itemize}
    \item models toward the upper-right have reliable repetition charts in both
    ACT and NOC;
    \item models toward the lower-right have a comparatively strong functional
    contribution chart but weaker activation-state organization;
    \item models toward the upper-left have a clearer activation-state chart
    than functional contribution chart;
    \item models toward the lower-left have weak chart reliability in both
    spaces.
\end{itemize}

These regimes concern chart reliability, not behavior frequency.  The colors
are interleaved across the landscape rather than forming a strict gradient,
and bubble sizes vary among nearby points.  Consequently, a chart can exist
strongly without being frequently occupied or without having large local
causal gain.

\subsection{Matched Base--Instruct Pairing Protocol}

We form ten pairs with the same family and parameter scale:

\[
\begin{split}
&\text{Gemma-2B, Gemma-9B, Gemma-27B, Llama-8B,}\\
&\text{Mistral-7B, Mistral-12B, Mistral-24B,}\\
&\text{Qwen-0.5B, Qwen-7B, and Qwen-32B.}
\end{split}
\]

For pair $p$ and space $\Omega$, we compute
\begin{equation}
    \Delta_p
    \log G^{\mathrm{uniF}}_{\Omega}
    =
    \log
    G^{\mathrm{uniF}}_{p,\mathrm{Instruct},\Omega}
    -
    \log
    G^{\mathrm{uniF}}_{p,\mathrm{Base},\Omega},
\end{equation}
and
\begin{equation}
    \Delta_p
    G^{\mathrm{causal}}_{\Omega}
    =
    G^{\mathrm{causal}}_{p,\mathrm{Instruct},\Omega}
    -
    G^{\mathrm{causal}}_{p,\mathrm{Base},\Omega}.
\end{equation}
The comparison is paired at the model-family and parameter-scale level, but it
is not a controlled training intervention.  Public Instruct checkpoints may
combine instruction SFT, preference optimization, filtering, and other
undocumented recipe differences.

\begin{figure*}[t]
    \centering

    \begin{minipage}[t]{0.49\textwidth}
        \centering
        \includegraphics[
            width=\linewidth,
            keepaspectratio
        ]{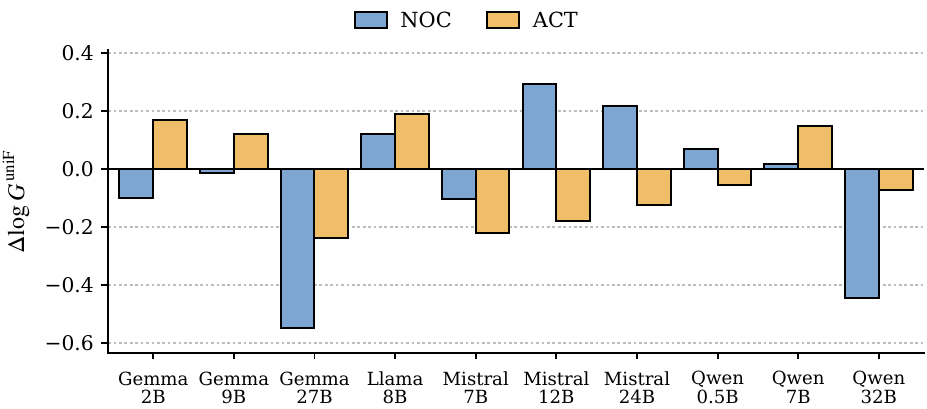}
        \vspace{-1.2mm}

        {\small (a) Paired change in chart reliability}
    \end{minipage}
    \hfill
    \begin{minipage}[t]{0.49\textwidth}
        \centering
        \includegraphics[
            width=\linewidth,
            keepaspectratio
        ]{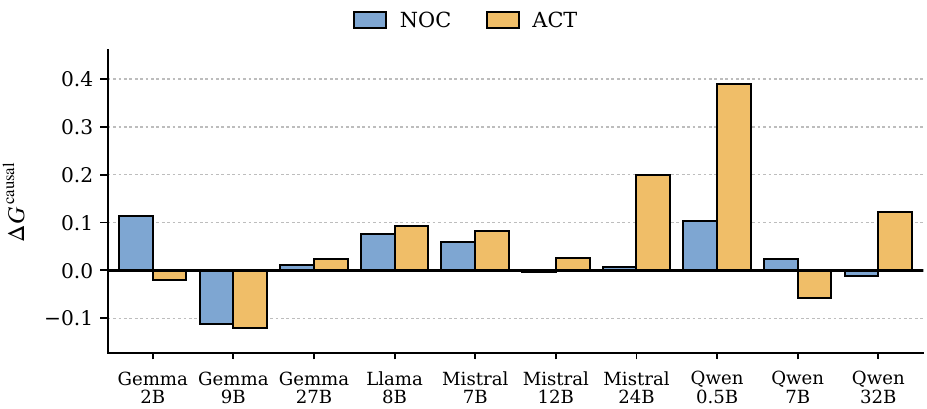}
        \vspace{-1.2mm}

        {\small (b) Paired change in causal gain}
    \end{minipage}

    \caption{
    Public Instruct-minus-Base changes for ten matched family/scale pairs.
    Blue bars denote NOC and gold bars denote ACT.
    Panel (a) reports
    $\Delta\log G^{\mathrm{uniF}}_{\Omega}$ and panel (b) reports
    $\Delta G^{\mathrm{causal}}_{\Omega}$.
    Positive values indicate a larger value for the Instruct checkpoint than for
    its matched Base checkpoint.  The heterogeneous signs show that public
    Base--Instruct differences can vary independently across functional chart
    reliability, runtime-state organization, and causal potency.
    Raw ACT and NOC gain magnitudes should be interpreted within space because
    the two steering operators use different intervention units.
    }
    \label{fig:full-base-instruct-deltas}
\end{figure*}

\subsection{Numerical Paired Results}

Table~\ref{tab:base-instruct-delta-values} reports the values displayed in
Figure~\ref{fig:full-base-instruct-deltas}, rounded to three decimals.

\begin{table*}[t]
\centering
\small
\setlength{\tabcolsep}{5pt}
\begin{tabular}{lrrrr}
\toprule
Pair
&
$\Delta\log G^{\mathrm{uniF}}_{\mathrm{NOC}}$
&
$\Delta\log G^{\mathrm{uniF}}_{\mathrm{ACT}}$
&
$\Delta G^{\mathrm{causal}}_{\mathrm{NOC}}$
&
$\Delta G^{\mathrm{causal}}_{\mathrm{ACT}}$
\\
\midrule
Gemma-2B    & -0.099 &  0.167 &  0.113 & -0.019 \\
Gemma-9B    & -0.015 &  0.121 & -0.113 & -0.121 \\
Gemma-27B   & -0.549 & -0.237 &  0.012 &  0.025 \\
Llama-8B    &  0.122 &  0.188 &  0.077 &  0.093 \\
Mistral-7B  & -0.103 & -0.222 &  0.060 &  0.082 \\
Mistral-12B &  0.292 & -0.181 & -0.003 &  0.025 \\
Mistral-24B &  0.217 & -0.125 &  0.007 &  0.199 \\
Qwen-0.5B   &  0.070 & -0.054 &  0.103 &  0.389 \\
Qwen-7B     &  0.017 &  0.148 &  0.024 & -0.058 \\
Qwen-32B    & -0.446 & -0.073 & -0.012 &  0.123 \\
\midrule
Mean        & -0.049 & -0.027 &  0.027 &  0.074 \\
Median      &  0.001 & -0.064 &  0.018 &  0.053 \\
\bottomrule
\end{tabular}
\caption{
Paired Instruct-minus-Base changes, rounded to three decimals.
Positive values indicate greater chart reliability or causal gain for the
Instruct checkpoint relative to its matched Base checkpoint.  The table should be interpreted within each
representation space; ACT and NOC causal-gain magnitudes are not directly
commensurate unless intervention scales are explicitly normalized.
}
\label{tab:base-instruct-delta-values}
\end{table*}

\subsection{Pair-Level Interpretation}

\paragraph{Llama-8B.}
Llama-8B is the clearest case of joint strengthening.  Both ACT and NOC
Universal Fisher Strength increase, and causal gain increases in both spaces.
For this pair, the Instruct checkpoint exhibits a more reliable and more causally potent repetition chart than the matched Base checkpoint.

\paragraph{Gemma-27B.}
Both ACT and NOC chart reliability decrease substantially, yet causal gain
increases slightly.  The behavior channel becomes statistically less separated
while remaining, or becoming slightly more, interventionally potent.  This is
a direct example of reliability and gain decoupling.

\paragraph{Mistral-12B and Mistral-24B.}
Both pairs gain NOC chart reliability while losing ACT chart reliability.
Mistral-24B additionally exhibits a large increase in ACT causal gain.
For these pairs, the Instruct checkpoints exhibit stronger
functional-channel reliability but weaker runtime-state organization than
their matched Base checkpoints. 

\paragraph{Qwen-0.5B.}
The change in NOC strength is small and positive, while ACT strength decreases.
Nevertheless, ACT causal gain increases by approximately $0.389$, the largest
gain change in the paired analysis.  A modest or weakened chart can therefore
become substantially more behaviorally influential.

\paragraph{Qwen-32B.}
Chart reliability decreases in both spaces, especially in NOC, but ACT causal
gain increases.  This again shows that suppressing statistical separability
does not necessarily suppress the causal readout of the behavior.

\subsection{Aggregate Interpretation}

The paired results support three conclusions.

\paragraph{Instruction alignment does not uniformly erase the repetition
mechanism.}
NOC chart reliability increases in five pairs and decreases in five.  ACT chart
reliability increases in four pairs and decreases in six.  The average changes
are mildly negative, but the pair-level variance is much larger than the mean.

\paragraph{Causal gain more often increases than decreases.}
Seven of ten pairs exhibit positive changes in NOC causal gain, and seven of
ten exhibit positive changes in ACT causal gain.  This pattern is inconsistent
with a simple account in which instruction alignment merely suppresses the
behavior mechanism.

\paragraph{ACT and NOC effects can disagree.}
For chart reliability, five of the ten pairs exhibit opposite-signed ACT and
NOC changes; for causal gain, four pairs exhibit opposite-signed changes.
This supports the use of two complementary spaces: Base--Instruct differences
can appear differently in functional-contribution geometry and runtime
activation geometry.

\subsection{Relationship to the Controlled Post-Training Study}

The public Base--Instruct comparison provides broad ecological evidence but
cannot identify which component of the training recipe caused a particular
change.  The controlled experiments in
Section~\ref{sec:training_results} address this limitation by fixing the base
model, data construction, evaluation instances, and behavioral direction while
varying SFT and reward optimization separately.

The two analyses therefore play complementary roles:

\begin{itemize}
    \item the paired public models establish that alignment effects are
    heterogeneous across architectures and scales;
    \item the controlled trajectories determine which geometric effects can be
    attributed specifically to SFT and reward optimization.
\end{itemize}

%% file: sections__supplement__t_DatasetSize.tex
\section{Dataset Sizes, Filtering, and Sample Exclusions}
\label{dataset}

We document the exact sizes of the repetition and sycophancy datasets used
throughout this work, the filtering stages applied, and the extent to which tokenizer-specific alignment affects the number of usable samples per model.

\paragraph{Repetition.}
Table~\ref{tab:rep-exclusion} summarizes the repetition data construction.

We constructed the repetition dataset from two sources of repetitive text:
single-token repetition data (660 items, each expanded into exactly 100
single-turn conversations between a user and an assistant) and paragraph- and phrase-level repetition data (862\,046 items). The total repetition-positive candidate pool contains 928\,046 entries (conversations or items, pooled across both sources).

Non-repetitive behavior-negative examples were drawn from a pool of 10\,000
samples and restricted to the 1\,368 examples for which all judges were
``true'' (indicating non-hallucinated responses).

We selected 1\,368 repetition-positive records from the positive candidate
pool using reservoir sampling and combined them with all 1\,368 eligible
non-repetition records, yielding a balanced dataset of 2\,736 response-level records with binary labels. The positive and negative records originate from different prompt sets and are not matched by prompt identity; the dataset provides binary classification labels.

%
%
\begin{table}[t]
\centering
\caption{Repetition dataset construction stages.}
\label{tab:rep-exclusion}
\begin{tabular}{lr}
\toprule
Stage & Count \\
\midrule
Repetition-positive source pool (entries)   & 928\,046 \\
Positive records selected by reservoir sampling & 1\,368 \\
Eligible non-repetition records (all-judge-true) & 1\,368 \\
Negative records selected                     & 1\,368 \\
\midrule
Total response records                        & 2\,736 \\
\bottomrule
\end{tabular} 
\end{table}

The repetition detector was used to localize the repeated token span rather
than to determine dataset inclusion. It applies three strategies in order:
(i)~longest single-token run (minimum 3 consecutive identical tokens),
(ii)~longest repeating $n$-gram run ($2 \leq n \leq 10$, minimum 3
repetitions), and (iii)~paragraph-level duplicate detection (a paragraph
occurring at least twice). When no periodic span is detected, the complete
response is retained as the analysis span. The archived outputs do not
preserve the number of examples assigned to each detection strategy or to the
fallback case; because all positive examples originate from known repetition
data, the detector is treated as a span-annotation mechanism rather than a
sample filter.

The cross-model PCA analyses cover 23 models.  Every repetition PCA output
contains 2\,735 rows rather than the expected 2\,736; the missing row belongs
to the repetition-positive class (label~1) in every output.

\paragraph{Sycophancy.}
Table~\ref{tab:syco-exclusion} reports the stance-extraction outcomes.

\begin{table}[t]
\centering
\caption{Sycophancy stance-extraction outcomes.}
\label{tab:syco-exclusion}
\resizebox{\columnwidth}{!}{%
\begin{tabular}{lrr}
\toprule
Stance-extraction outcome & Pairs & Successful records \\
\midrule
Both chosen and rejected extracted & 3\,206 & 6\,412 \\
Chosen only                        &   138 &   138 \\
Rejected only                      &   128 &   128 \\
Neither side                       &     8 &     0 \\
\midrule
Total                              & 3\,480 & 6\,678 \\
\bottomrule
\end{tabular}%
}

\end{table}

We constructed the sycophancy data from
\texttt{DataCreatorAI/Anti-Sycophancy-DPO}, which provides 3\,480
chosen--rejected preference pairs. 

Each of the 3\,480 rows was expanded into two individual records
(\texttt{syco\_NNNN\_chosen} and \texttt{syco\_NNNN\_rejected}), yielding
6\,960 formatted samples. A Llama-3.3-70B-Instruct model was then prompted to extract a continuous verbatim substring---the \emph{stance text}---that
conveys the core stance of each response toward the user's claim, subject to the constraint that the extracted text must be an exact contiguous substring of the response. Of the 6\,960 samples, the stance text was successfully extracted for 6\,678 (95.9\,\%); 282 extractions failed after three retry attempts and were discarded. Table~\ref{tab:syco-exclusion} reports the mutually exclusive pair-level outcome categories.

At the pair level, 3\,206 base prompts have both the chosen and rejected
stance successfully extracted, forming complete chosen--rejected pairs.
The extracted corpus therefore contains $3\,206+138=3\,344$ chosen records
and $3\,206+128=3\,334$ rejected records, for a total of 6\,678 records.

Because the stance text is defined at the character level, it was independently mapped to token indices under each target model's tokenizer using character-offset mapping (\texttt{return\_offsets\_mapping=True} in Hugging Face \texttt{tokenizer}). The mapping succeeded for all 6\,678 input records for all 23 evaluated models: 3\,344 chosen records and 3\,334 rejected records were aligned, with zero fallback cases. Table~\ref{tab:syco-tokenizer-alignment} reports the per-model alignment counts.

%
%
\begin{table*}[t]
\centering
\caption{Tokenizer-specific stance-span alignment statistics for all 23 models. Chosen, rejected, and total records are constant across models because every entry was successfully aligned. The ``Classifier rows'' column reports the number of feature rows used for 3-vs-1 logistic regression training. For all models, the aligned counts are: Chosen = 3\,344, Rejected = 3\,334, Complete pairs = 3\,206, Total records = 6\,678.}
\label{tab:syco-tokenizer-alignment}
\resizebox{\textwidth}{!}{%
\begin{tabular}{l r}
\toprule
Model & Classifier rows \\
\midrule
\\texttt{google/gemma-2-2b}                     & 10\,708 \\
\\texttt{google/gemma-2-2b-it}                  & 13\,356 \\
\\texttt{google/gemma-2-9b}                     & 10\,708 \\
\\texttt{google/gemma-2-9b-it}                  & 13\,356 \\
\\texttt{google/gemma-2-27b}                    & 10\,708 \\
\\texttt{google/gemma-2-27b-it}                 & 13\,356 \\
\\texttt{google/gemma-3-4b-it}                  & 13\,356 \\
\\texttt{meta-llama/Llama-3.1-8B}               & 10\,708 \\
\\texttt{meta-llama/Llama-3.1-8B-Instruct}      & 13\,356 \\
\\texttt{meta-llama/Llama-3.3-70B-Instruct}     & 13\,356 \\
\\texttt{mistralai/Mistral-7B-v0.3}             & 13\,356 \\
\\texttt{mistralai/Mistral-7B-Instruct-v0.3}    & 13\,356 \\
\\texttt{mistralai/Mistral-Nemo-Base-2407}      & 10\,708 \\
\\texttt{mistralai/Mistral-Nemo-Instruct-2407}  & 13\,356 \\
\\texttt{mistralai/Mistral-Small-3.1-24B-Base-2503}  & 10\,708 \\
\\texttt{mistralai/Mistral-Small-3.1-24B-Instruct-2503} & 10\,708 \\
\\texttt{Qwen/Qwen2.5-0.5B}                    & 10\,708 \\
\\texttt{Qwen/Qwen2.5-0.5B-Instruct}            & 10\,708 \\
\\texttt{Qwen/Qwen2.5-7B}                       & 10\,708 \\
\\texttt{Qwen/Qwen2.5-7B-Instruct}              & 10\,708 \\
\\texttt{Qwen/Qwen2.5-32B}                      & 10\,708 \\
\\texttt{Qwen/Qwen2.5-32B-Instruct}             & 10\,708 \\
\\texttt{Qwen/Qwen3.5-27B}                      & 10\,708 \\
\bottomrule
\end{tabular}
}
\end{table*}

\paragraph{Classifier training rows.}
The classifier training pipeline uses a 3-vs-1 loading scheme that creates
four feature-row groups---answer-token activations for both classes and
other-token activations for both classes. The total number of classifier rows
therefore differs from the number of unique response records. Nine of the 23
models produce the full 13\,356 rows ($2 \times 6\,678$), while the remaining
14 produce 10\,708 rows. The reduced count in these 14 models is consistent
with incomplete other-token activation files having been available at training
time; the archived pipeline does not preserve the per-model activation-file
inventory that would be required to attribute the reduction to specific QIDs
or directories.

\paragraph{Cross-model common set.}
We distinguish three levels of the common set:

\begin{description}
\item[Annotation-level common set (6\,678 records).]
All 23 models share the same set of response IDs for which the stance-text
token mapping succeeded. The token sequences and span boundaries were
constructed independently per model, but the record IDs are identical.

\item[Activation-level common set (6\,678 rows).]
All 23 models produce PCA outputs with 6\,678 rows in both the raw-activation
and NOC subspaces, and all pairwise CCA comparisons pass an exact
label-array consistency check. The analysis scripts load from a unified ID
list and error when activation files are missing;
this loading discipline makes it likely that the same ID set was used across
models, although the archived PCA files do not preserve sample ID strings and
cannot independently verify QID-level identity.

\item[Complete-pair common set (3\,206 pairs).]
Among the 6\,678 individual records, 3\,206 base prompts have both the chosen
and rejected responses available for every model, forming complete
chosen--rejected pairs.
\end{description}

%% file: sections__supplement__u_geometry_occupancy_gain_full.tex
\renewcommand{\decompstagepanel}[2]{%
    \begin{minipage}[t]{0.485\textwidth}
        \centering
        \includegraphics[width=\linewidth,keepaspectratio]{#1}
        \vspace{-1.2mm}

        {\scriptsize #2}
    \end{minipage}%
}

\section{Complete Geometry--Occupancy--Gain Analysis}
\label{app:gog-full}

Final behavior rates alone do not identify whether post-training changes the
behavioral chart itself, redistributes states inside an inherited chart, or
changes the chart's causal sensitivity.  We therefore compare every controlled
post-training checkpoint with the Base model along three complementary axes.
For checkpoint $t$, we define
\begin{align}
    \Delta_{\mathrm{geom}}(t)
    &= d_{\mathrm{proj}}(0,t),\\
    \Delta_{\mathrm{occ}}(t)
    &= G^{\mathrm{occ}}_t-G^{\mathrm{occ}}_0,\\
    \Delta_{\mathrm{gain}}(t)
    &= G^{\mathrm{causal}}_t-G^{\mathrm{causal}}_0.
\end{align}
Geometry is an unsigned movement magnitude.  Occupancy is a signed change in
the rotation-invariant class-conditional occupancy statistic, and gain is a
signed change in local intervention sensitivity.  These quantities form a
three-axis diagnostic profile; they are not additive shares of the observed
behavioral endpoint.

We report two representation spaces and two causal-gain estimators.  ACT uses
runtime activation coordinates for geometry and occupancy, whereas NOC uses
normalized contribution coordinates.  Activation-level gain is estimated from
direct activation steering, while subspace-level gain is estimated by steering
the learned behavioral chart.

\subsection{Matched Training Stages and Normalization}

Each lineage is evaluated at ten matched relative stages.  Stage $s$ denotes
normalized training progress
\begin{equation}
    \pi_s = \frac{s}{10},
    \qquad s\in\{1,\ldots,10\}.
\end{equation}
The raw optimizer steps differ across conditions because the runs have
different total lengths.  The exact checkpoint schedules are reported in
Table~\ref{tab:gog-stage-checkpoints}.

\begin{table}[t]
\centering
\small
\setlength{\tabcolsep}{3.5pt}
\resizebox{\columnwidth}{!}{%
\begin{tabular}{lccc}
\toprule
Condition & SFT checkpoint & Reward checkpoint & SFT$\rightarrow$Reward checkpoint \\
\midrule
Repetition-up   & $10s$ & $6s$ & $6s$ \\
Repetition-down & $60s$ & $6s$ & $6s$ \\
Sycophancy-up   & $12s$ & $6s$ & $6s$ \\
Sycophancy-down & $36s$ & $6s$ & $6s$ \\
\bottomrule
\end{tabular}%
}
\caption{Checkpoint selected at matched stage $s\in\{1,\ldots,10\}$.
The stage index aligns relative training progress rather than raw optimizer
steps.}
\label{tab:gog-stage-checkpoints}
\end{table}

The endpoint figures use metric-wise max-absolute normalization,
\begin{equation}
    \widetilde{\Delta}^{\mathrm{maxabs}}_{q,i}
    =
    \frac{\Delta_{q,i}}
    {\max_j |\Delta_{q,j}|},
\end{equation}
computed over the 12 final condition--lineage combinations.  This preserves
zero and sign and makes the largest absolute endpoint change in each metric
equal to one.

The trajectory and stage-snapshot figures instead use a global metric-wise
z-score within each representation/gain configuration,
\begin{equation}
    \widetilde{\Delta}^{z}_{q,i}
    =
    \frac{\Delta_{q,i}-\mu_q}{\sigma_q},
\end{equation}
where $\mu_q$ and $\sigma_q$ are computed over all
$4\times3\times10=120$ checkpoint rows in that configuration.  The z-score
therefore indicates whether a checkpoint's change is above or below the mean
change observed across the complete set; a negative geometry z-score does not
mean negative geometric distance.

\subsection{Final-Checkpoint Profiles}

\begin{figure}[t]
    \centering
    \includegraphics[width=0.96\columnwidth,keepaspectratio]
    {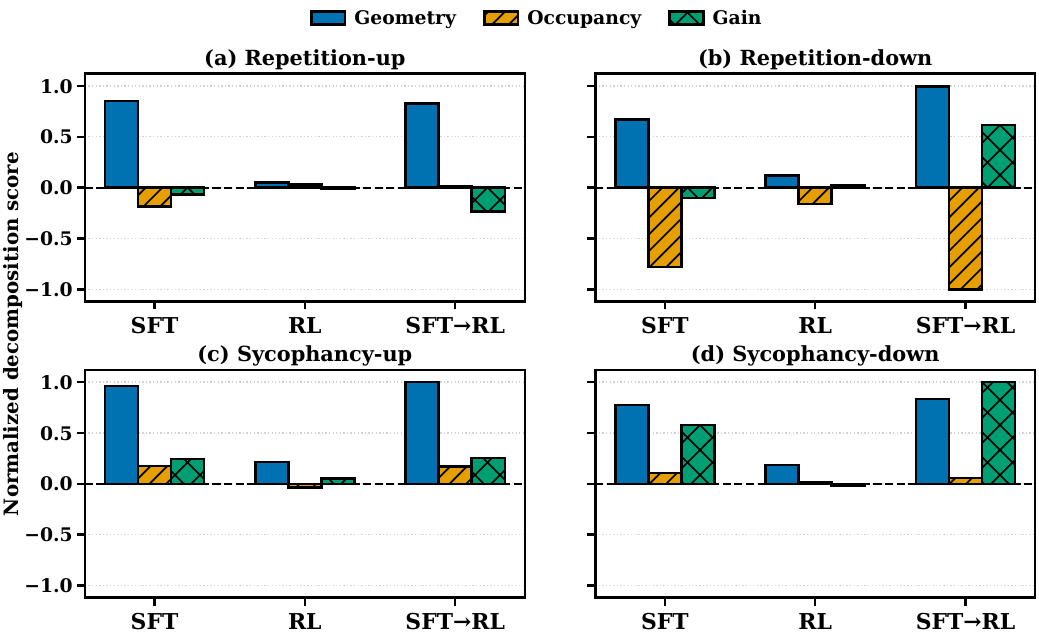}
    \caption{Final-checkpoint geometry--occupancy--gain profiles using
    activation-level causal gain. Geometry and occupancy are measured in NOC
    space, and each metric is independently max-absolute normalized across the
    12 endpoints.}
    \label{fig:gog-final-noc-activation}
\end{figure}

\begin{figure}[t]
    \centering
    \includegraphics[width=0.96\columnwidth,keepaspectratio]
    {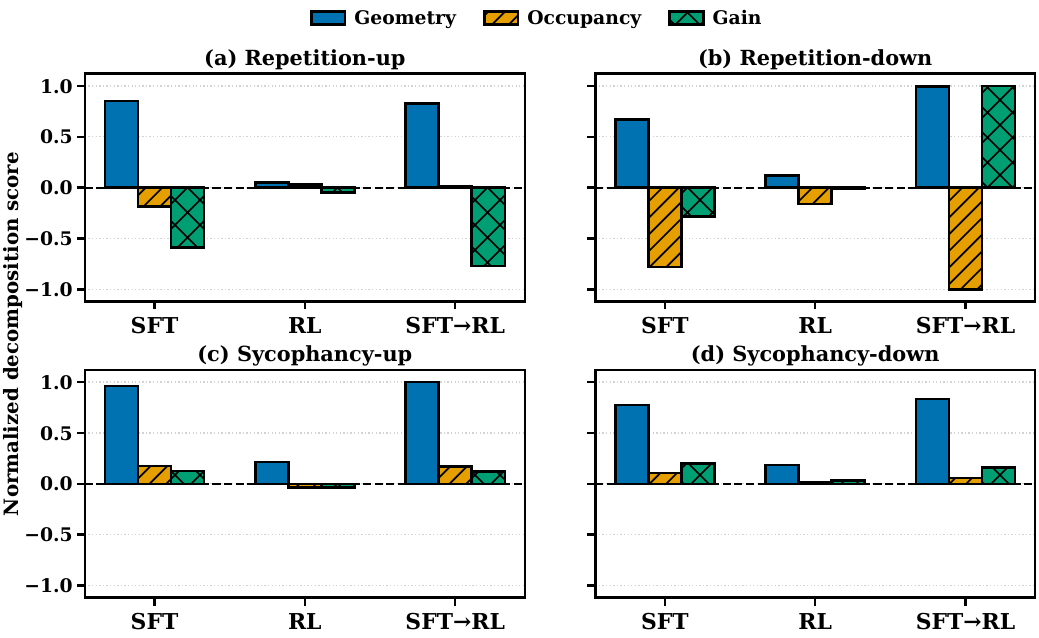}
    \caption{Final-checkpoint geometry--occupancy--gain profiles using
    chart-subspace causal gain. Geometry and occupancy are identical to
    Figure~\ref{fig:gog-final-noc-activation}; only the gain estimator changes.}
    \label{fig:gog-final-noc-subspace}
\end{figure}

The endpoint summaries provide the most compact comparison across objectives.
Both figures use NOC geometry and occupancy.  They differ only in whether gain
is estimated by activation-level or chart-subspace steering.
Figures~\ref{fig:gog-final-noc-activation}
and~\ref{fig:gog-final-noc-subspace} report the two endpoint profiles.

\subsection{Complete Ten-Stage Trajectories}

\begin{figure*}[!t]
    \centering
    \includegraphics[width=0.98\textwidth,keepaspectratio]
    {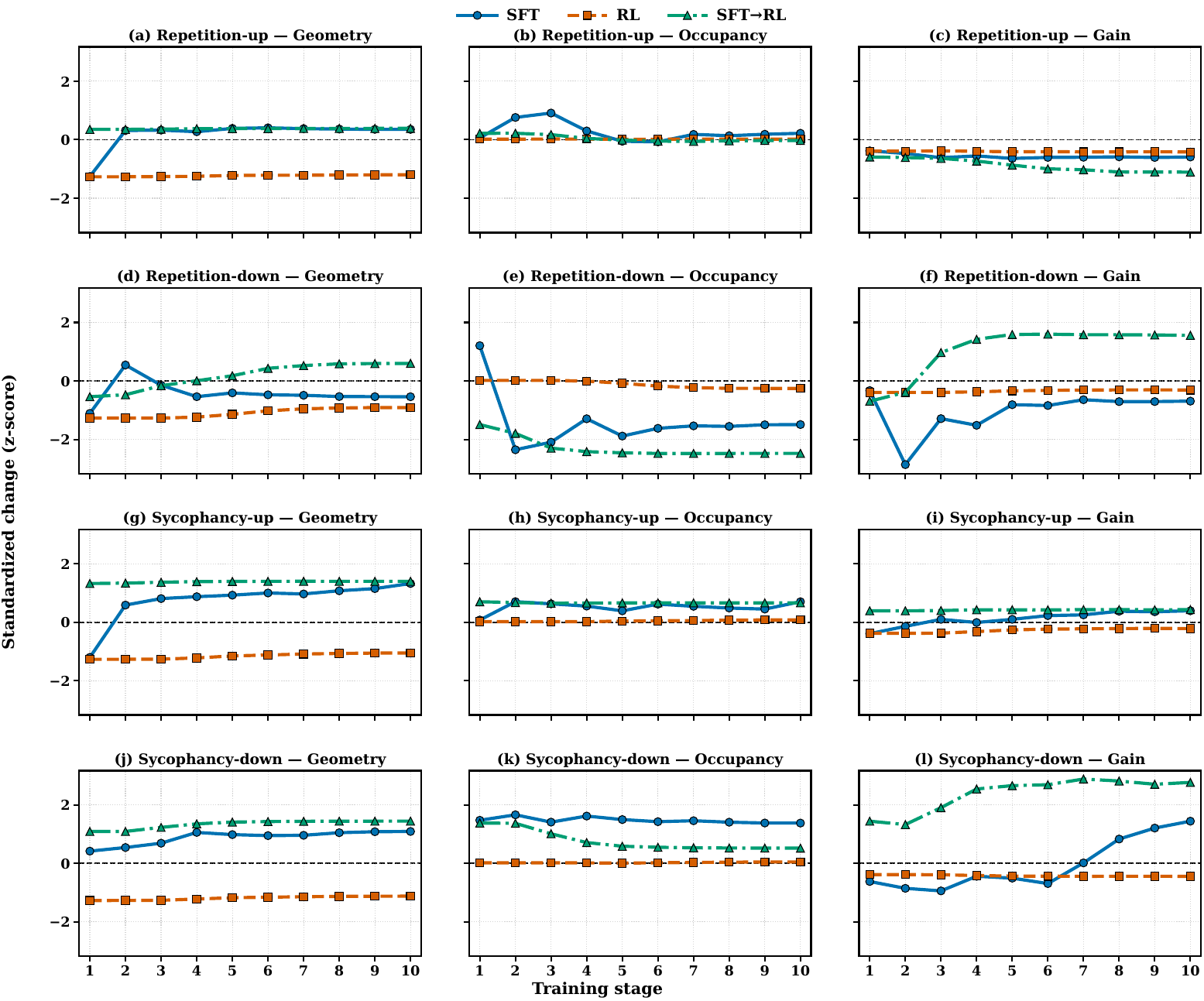}
    \caption{Complete ten-stage trajectories using ACT geometry and occupancy
    with activation-level causal gain.  SFT geometry moves rapidly away from
    the low-movement reward regime, while sequential training starts from the
    SFT endpoint and largely retains its displaced chart.  Occupancy is more
    non-monotonic, especially for repetition-down.  Activation-level gain grows
    strongly for sequential repetition-down and for late sycophancy-down
    checkpoints, but remains negative for repetition-up.}
    \label{fig:gog-traj-act-activation}
\end{figure*}

\begin{figure*}[!t]
    \centering
    \includegraphics[width=0.98\textwidth,keepaspectratio]
    {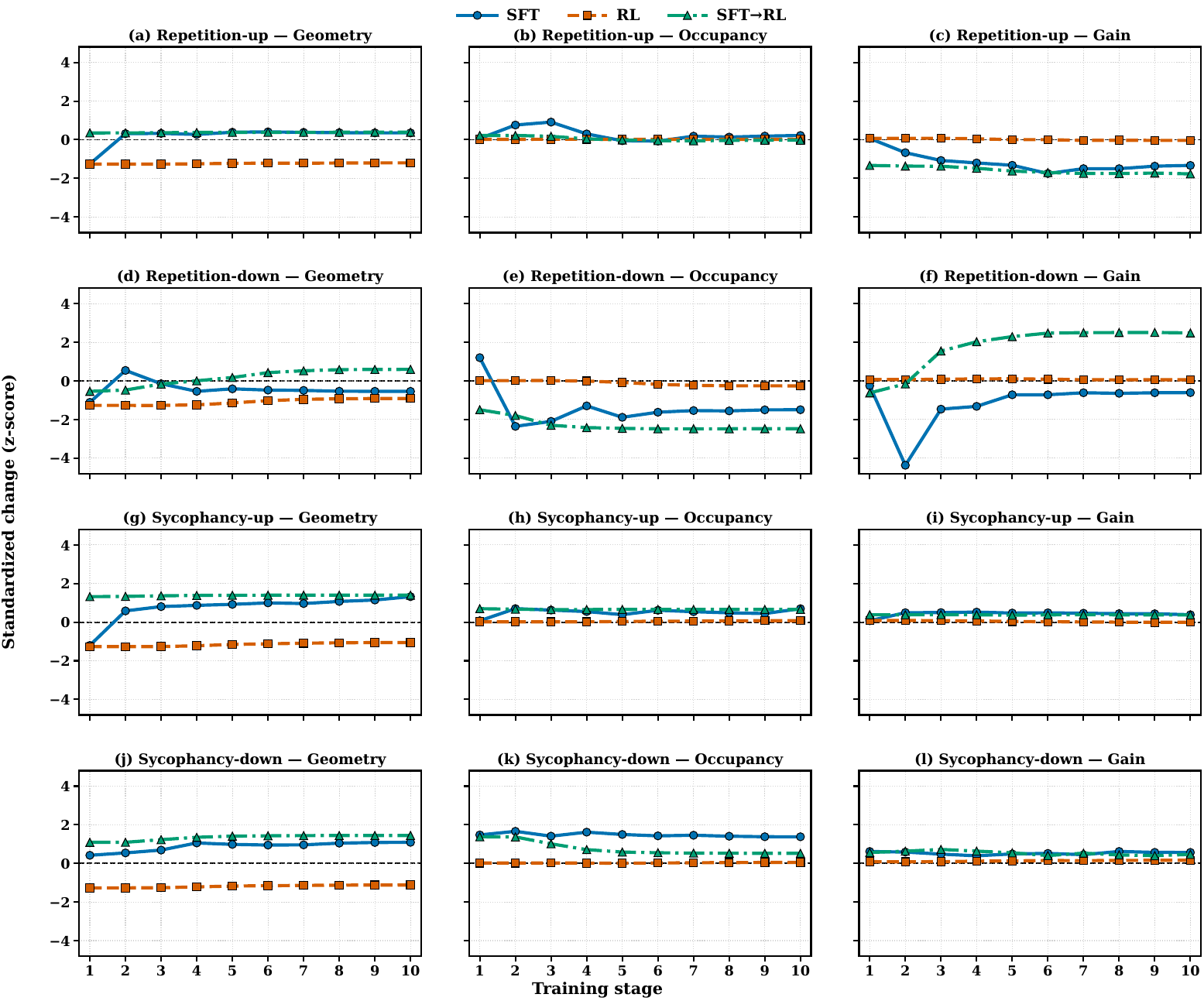}
    \caption{Complete ten-stage trajectories using ACT geometry and occupancy
    with chart-subspace causal gain.  The geometry and occupancy trajectories
    match Figure~\ref{fig:gog-traj-act-activation}; the gain curves provide an
    intervention-operator robustness check.  The largest positive subspace-gain
    trajectory again appears for sequential repetition-down, whereas reward-only
    gain remains near the lower part of the global distribution.}
    \label{fig:gog-traj-act-subspace}
\end{figure*}

\begin{figure*}[!t]
    \centering
    \includegraphics[width=0.98\textwidth,keepaspectratio]
    {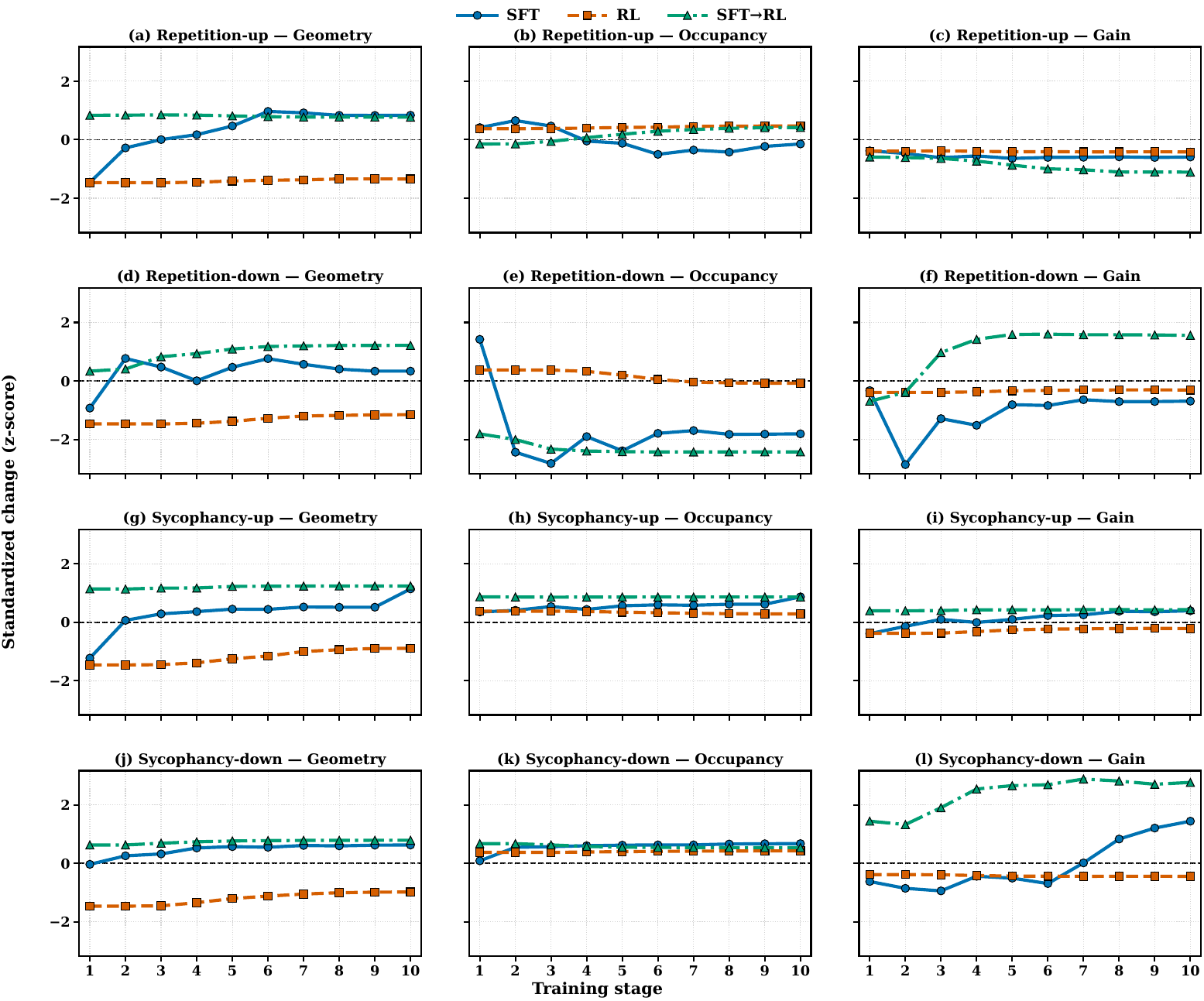}
    \caption{Complete ten-stage trajectories using NOC geometry and occupancy
    with activation-level causal gain.  NOC reproduces the central objective
    asymmetry: SFT-containing lineages move strongly, whereas reward-only
    geometry remains comparatively conservative.  NOC occupancy exposes strong
    early and sustained redistribution for repetition-down and more moderate
    positive changes for sycophancy.}
    \label{fig:gog-traj-noc-activation}
\end{figure*}

\begin{figure*}[!t]
    \centering
    \includegraphics[width=0.98\textwidth,keepaspectratio]
    {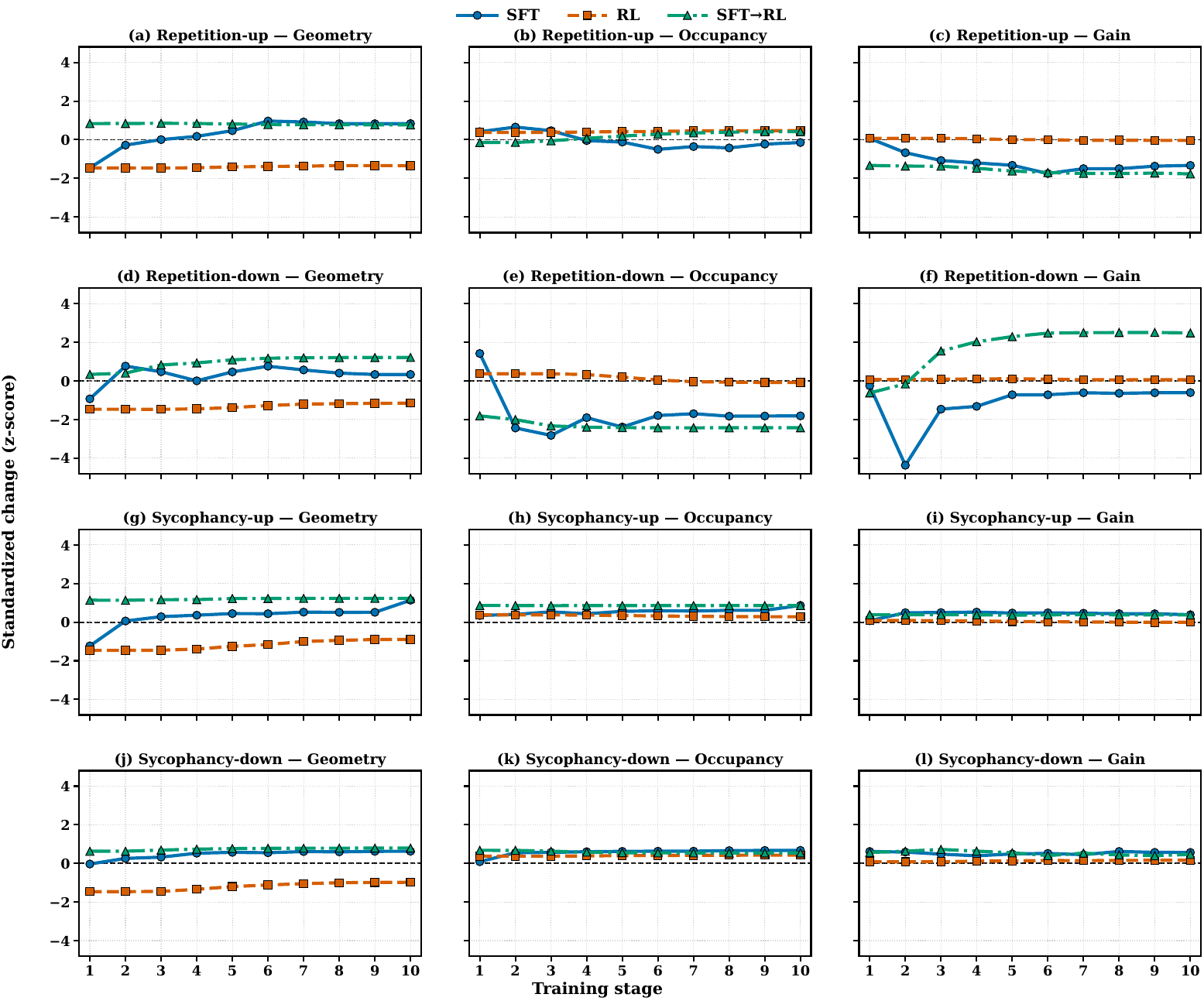}
    \caption{Complete ten-stage trajectories using NOC geometry and occupancy
    with chart-subspace causal gain.  Comparison with
    Figure~\ref{fig:gog-traj-noc-activation} separates the stable geometry and
    occupancy conclusions from gain-estimator-specific effects.  Reward-only
    gain is close to the center or lower tail of the global distribution, while
    sequential repetition-down develops a pronounced positive gain shift.}
    \label{fig:gog-traj-noc-subspace}
\end{figure*}

The trajectory figures show whether the endpoint profiles arise gradually,
early in training, or through late reversals.  Each figure contains twelve
panels: geometry, occupancy, and gain for each of the four behavioral
conditions.  Blue circles denote SFT, orange squares denote reward-only
optimization, and green triangles denote SFT$\rightarrow$Reward.
Figures~\ref{fig:gog-traj-act-activation}--\ref{fig:gog-traj-noc-subspace}
report the complete ACT- and NOC-space trajectories under both gain
estimators.

\subsection{Every Matched Stage Snapshot}

The following galleries include all forty stage-specific bar charts.  Each
stage snapshot is a horizontal cross-section of the corresponding trajectory:
it compares SFT, reward-only, and SFT$\rightarrow$Reward at the same normalized
training progress for all four behavioral conditions. 

\subsubsection{ACT Geometry and Occupancy with Activation-Level Gain}

\begin{figure*}[p]
\centering
\decompstagepanel{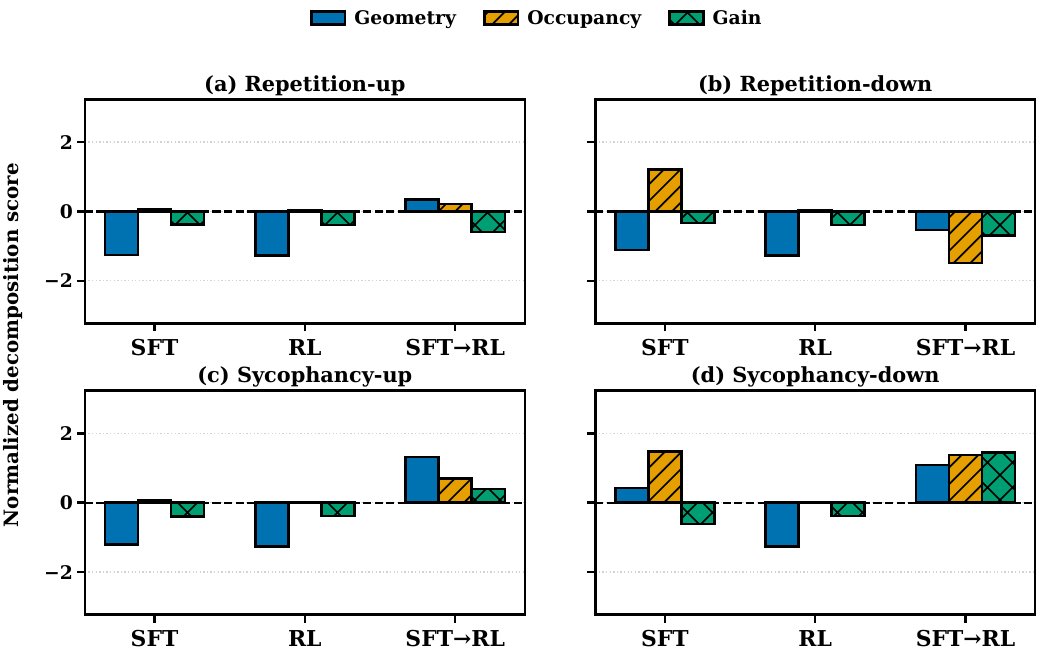}{(a) Stage 1: 10\% progress}
\hfill
\decompstagepanel{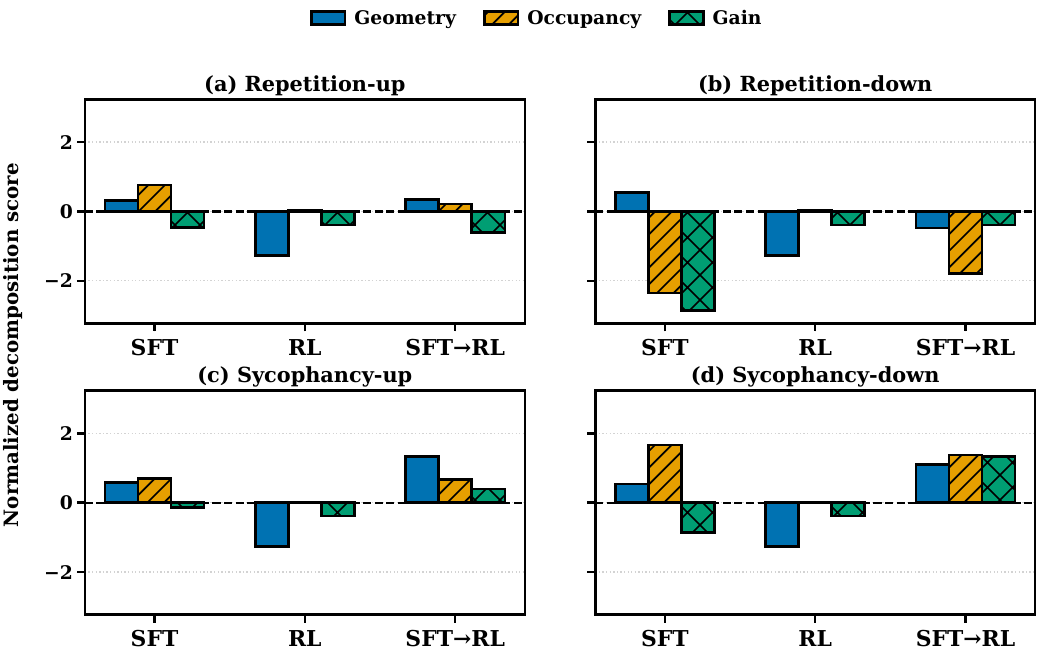}{(b) Stage 2: 20\% progress}

\vspace{1.5mm}
\decompstagepanel{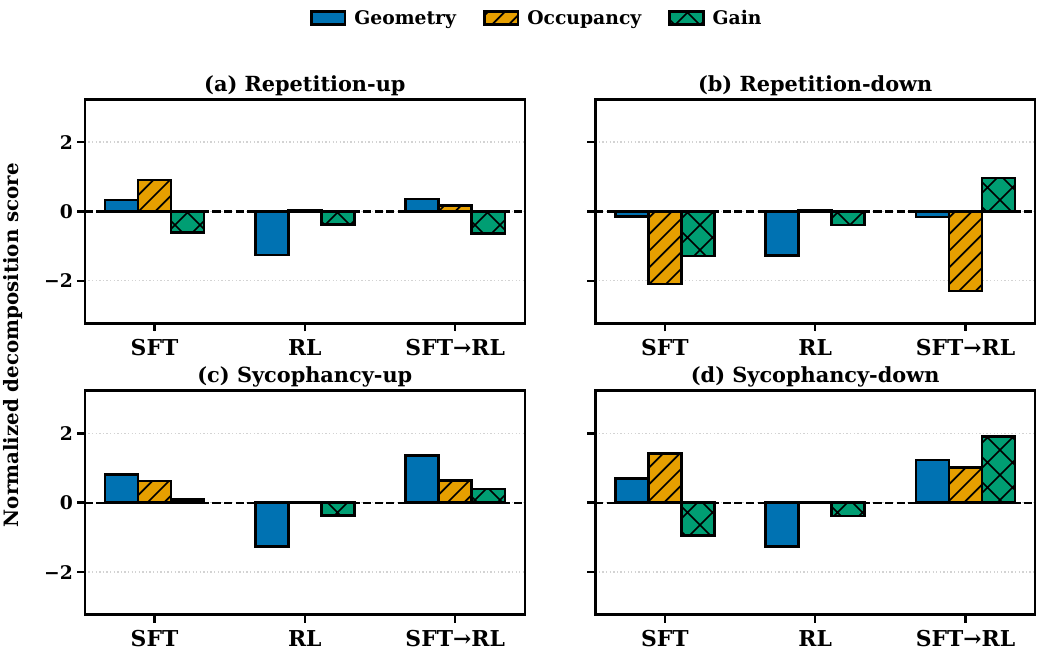}{(c) Stage 3: 30\% progress}
\hfill
\decompstagepanel{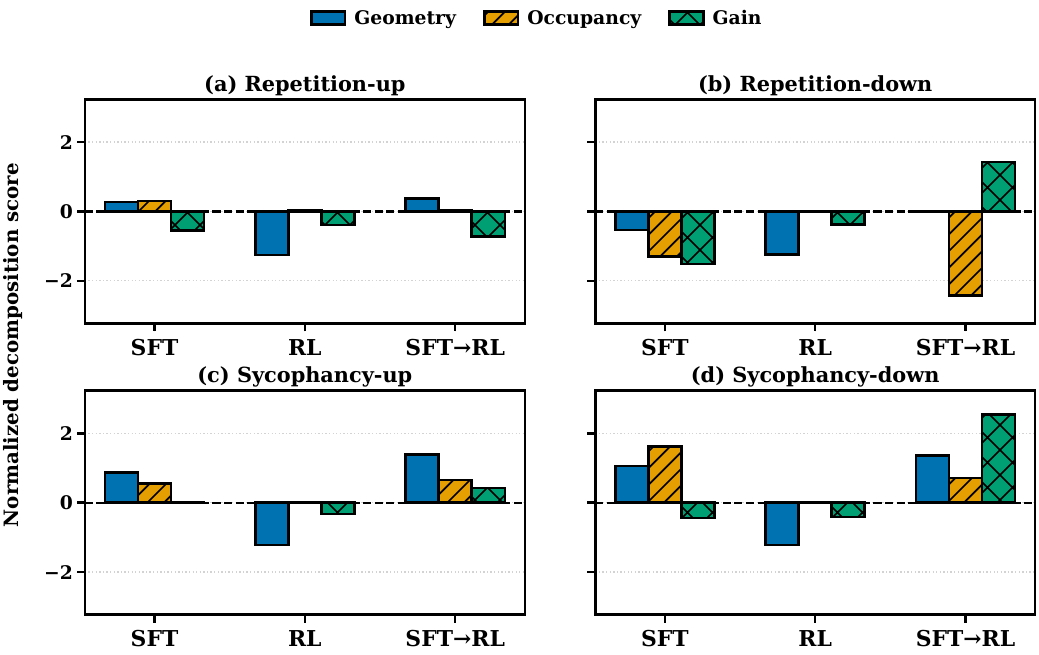}{(d) Stage 4: 40\% progress}

\vspace{1.5mm}
\hfill
\decompstagepanel{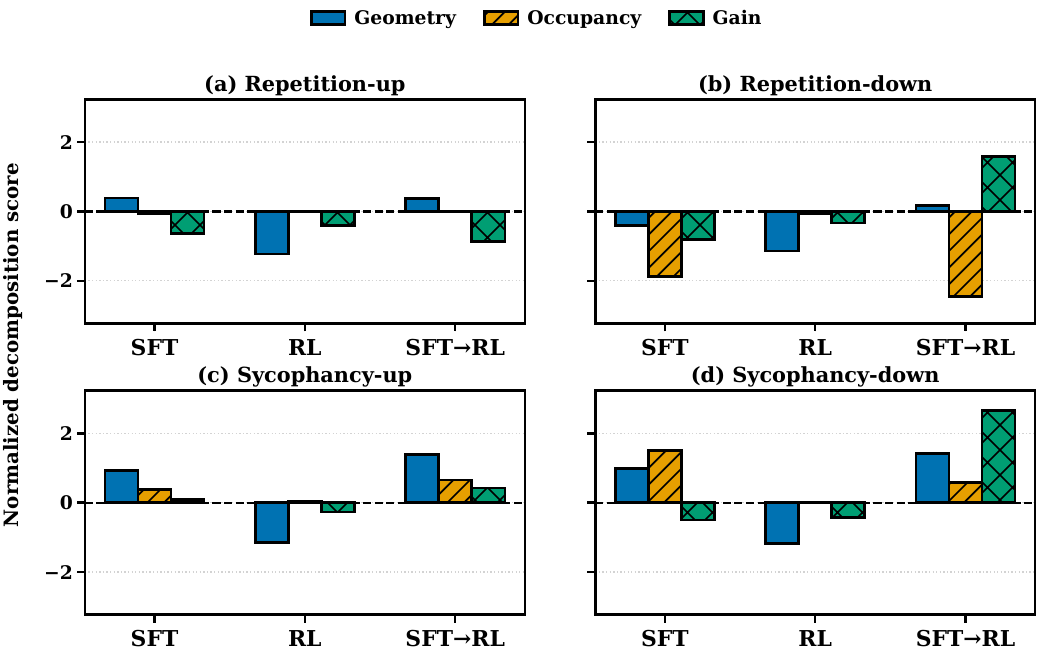}{(e) Stage 5: 50\% progress}
\hfill
\caption{Early-stage snapshots for ACT geometry and occupancy with
activation-level gain.  Each embedded plot contains the four behavioral
conditions.  SFT geometry separates from reward-only geometry very early;
occupancy and gain show larger transient fluctuations and condition-specific
signs.}
\label{fig:gog-stage-act-activation-early}
\end{figure*}

\begin{figure*}[p]
\centering
\decompstagepanel{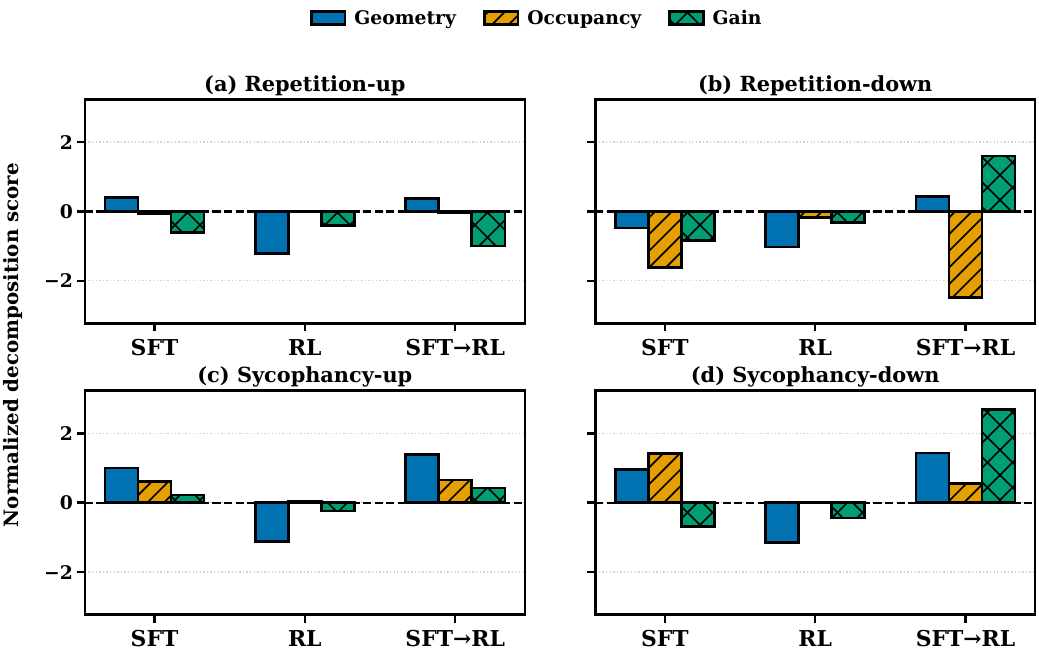}{(a) Stage 6: 60\% progress}
\hfill
\decompstagepanel{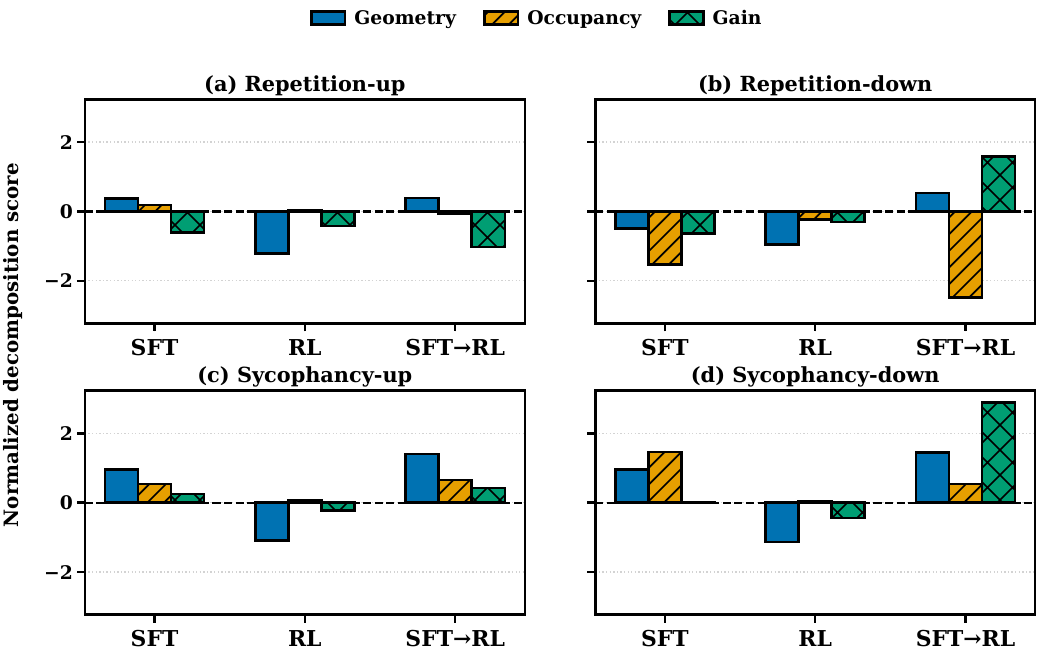}{(b) Stage 7: 70\% progress}

\vspace{1.5mm}
\decompstagepanel{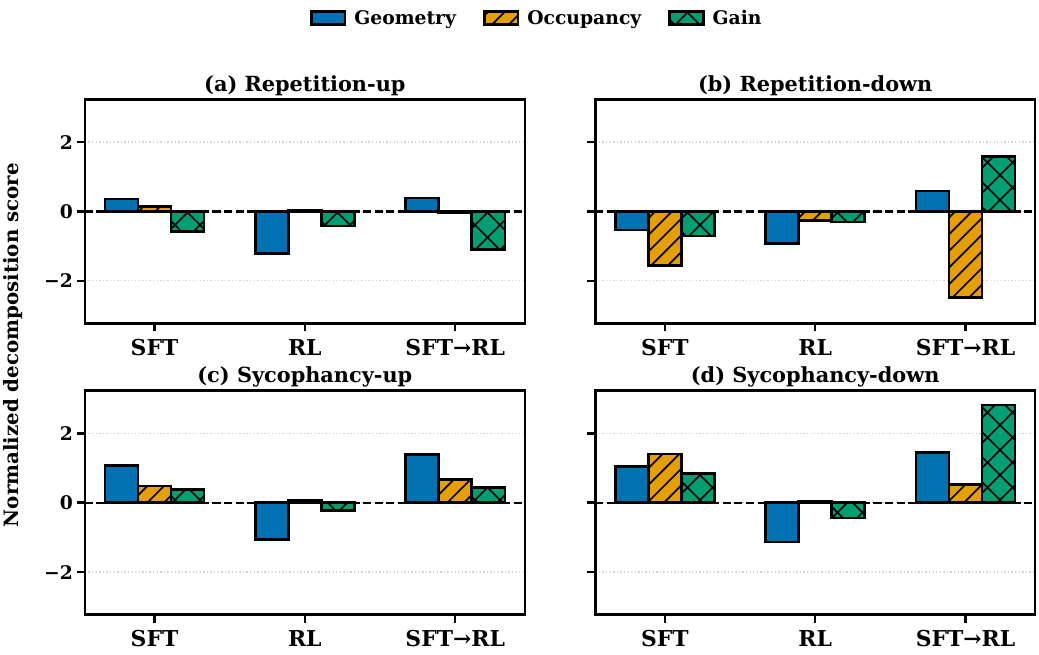}{(c) Stage 8: 80\% progress}
\hfill
\decompstagepanel{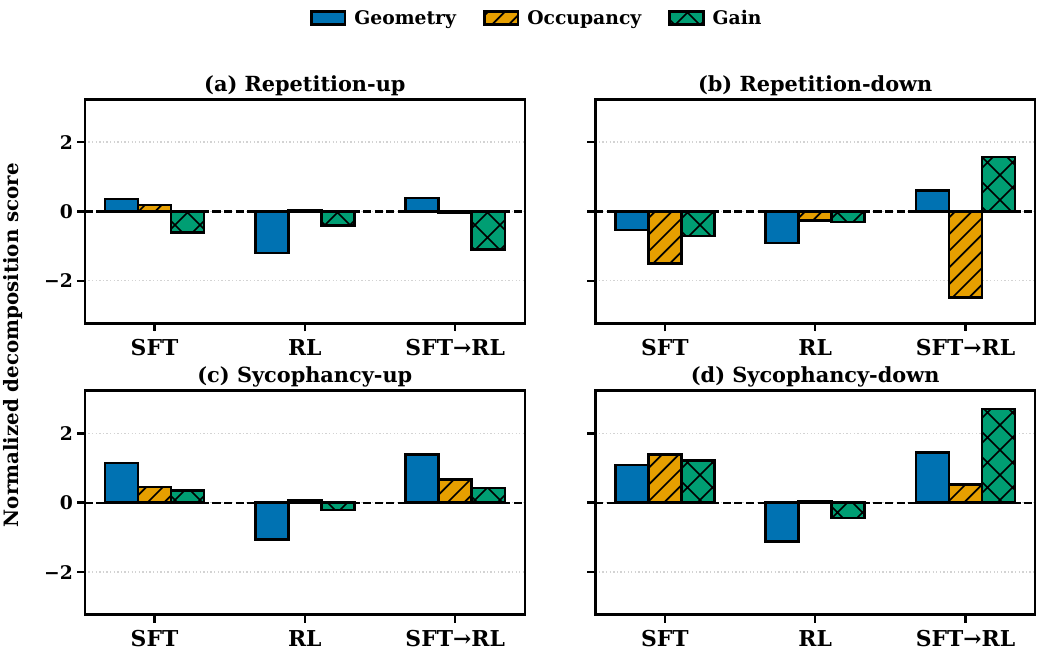}{(d) Stage 9: 90\% progress}

\vspace{1.5mm}
\hfill
\decompstagepanel{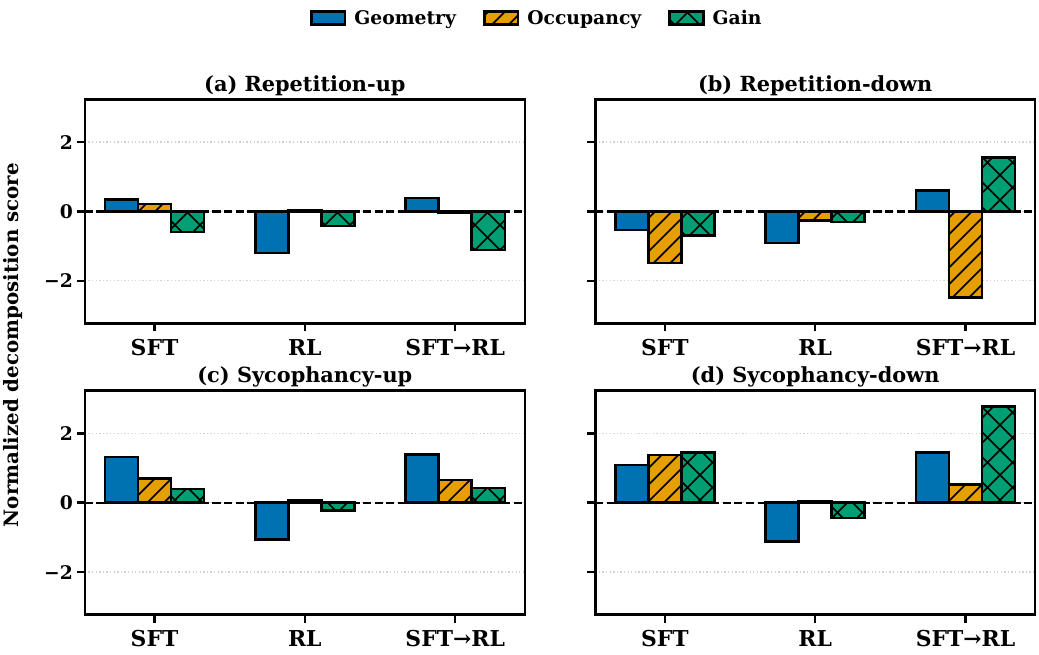}{(e) Stage 10: 100\% progress}
\hfill
\caption{Late-stage snapshots for ACT geometry and occupancy with
activation-level gain.  The geometry hierarchy stabilizes, while
repetition-down and sycophancy-down show the strongest late gain growth in the
sequential lineage.}
\label{fig:gog-stage-act-activation-late}
\end{figure*}

Figures~\ref{fig:gog-stage-act-activation-early}
and~\ref{fig:gog-stage-act-activation-late} report stages 1--5 and 6--10,
respectively, using ACT geometry and occupancy with activation-level causal
gain.

\subsubsection{ACT Geometry and Occupancy with Subspace-Level Gain}

\begin{figure*}[p]
\centering
\decompstagepanel{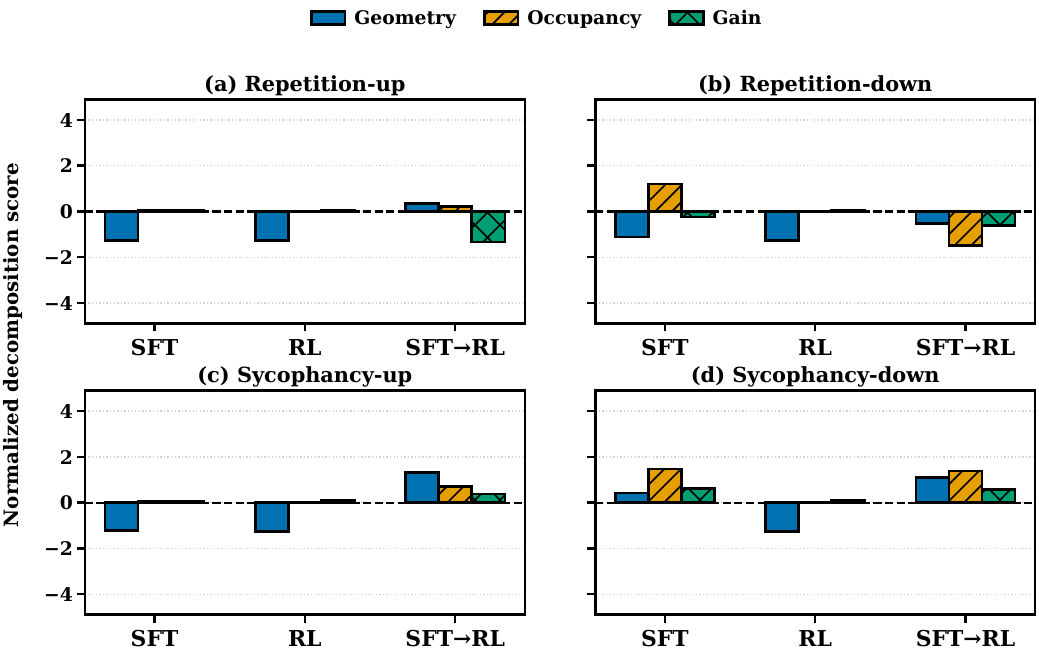}{(a) Stage 1: 10\% progress}
\hfill
\decompstagepanel{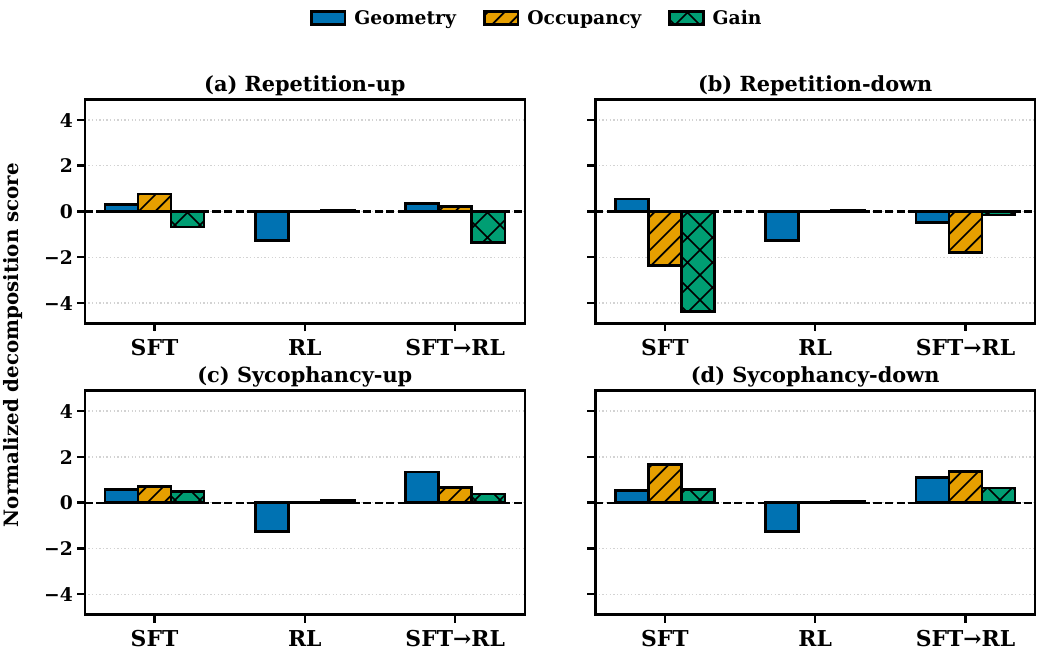}{(b) Stage 2: 20\% progress}

\vspace{1.5mm}
\decompstagepanel{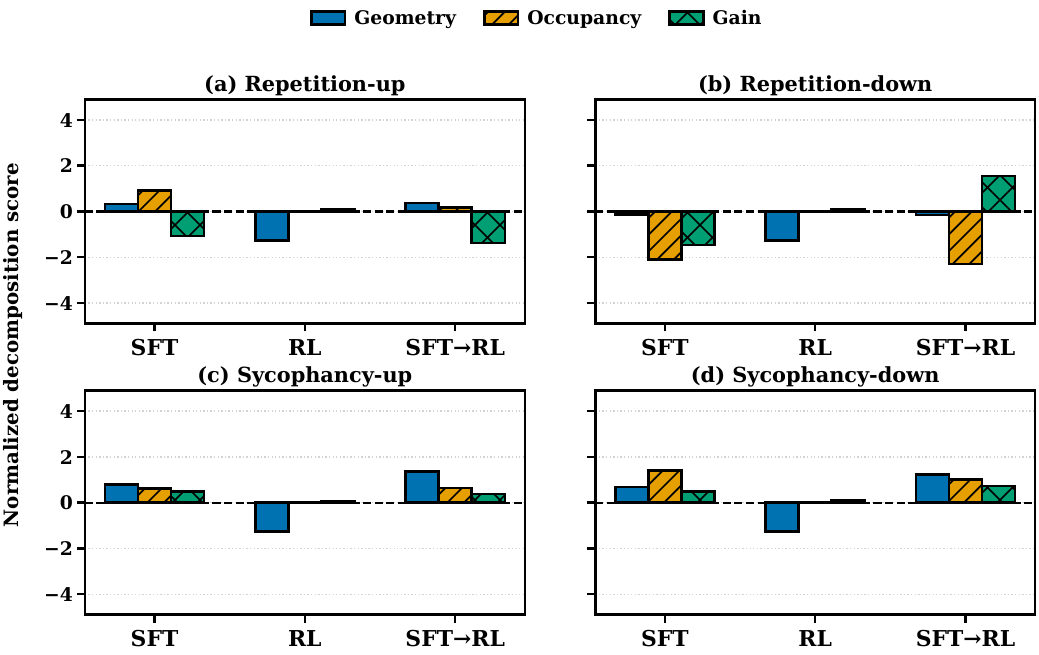}{(c) Stage 3: 30\% progress}
\hfill
\decompstagepanel{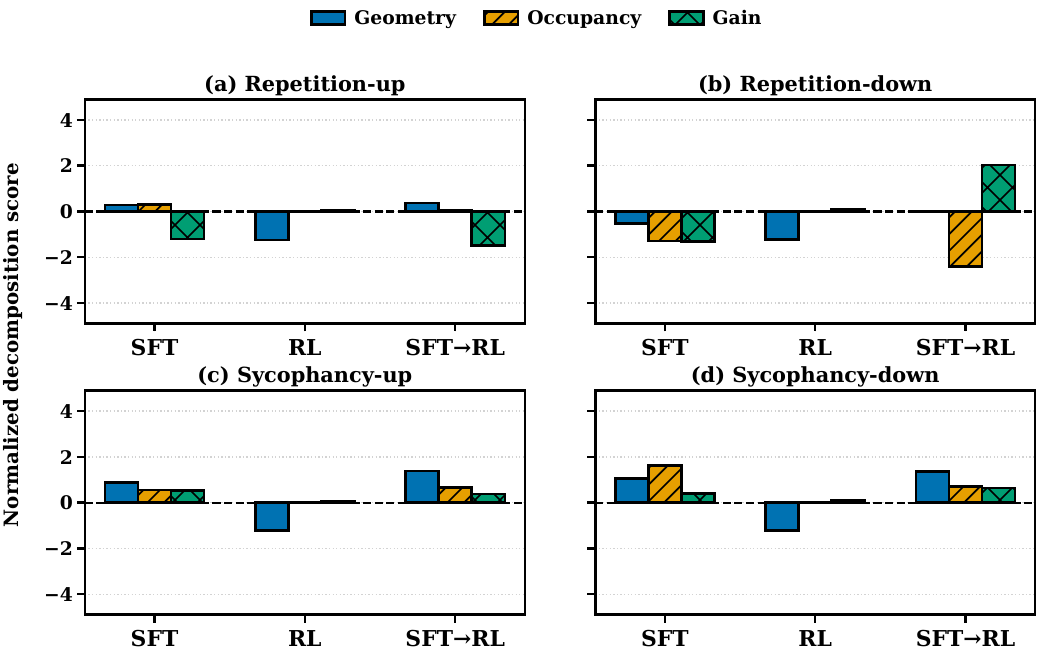}{(d) Stage 4: 40\% progress}

\vspace{1.5mm}
\hfill
\decompstagepanel{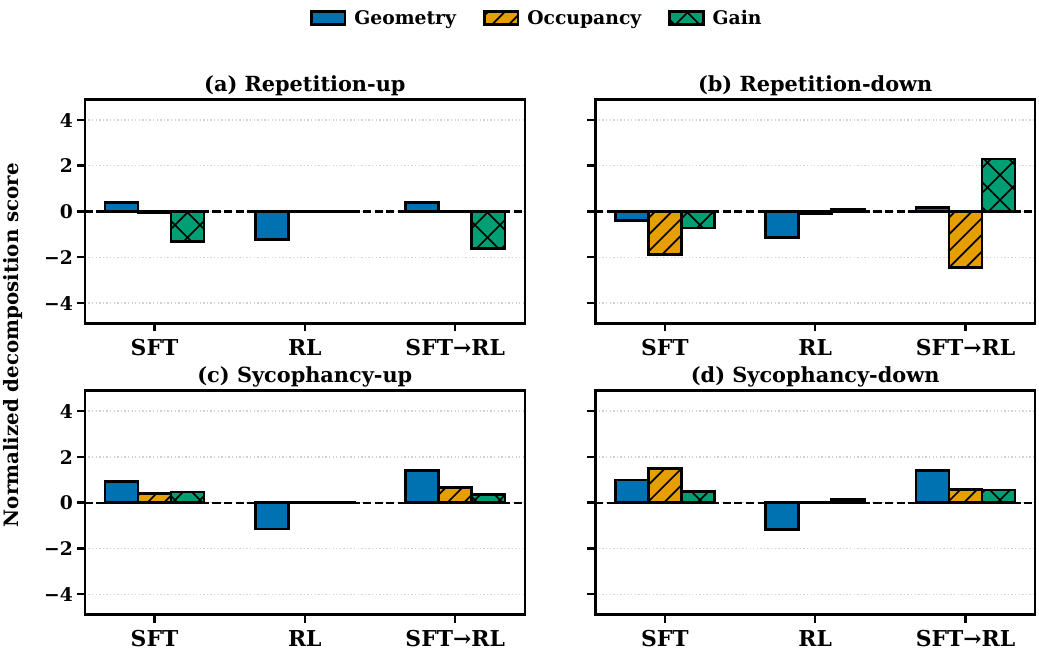}{(e) Stage 5: 50\% progress}
\hfill
\caption{Early-stage snapshots for ACT geometry and occupancy with
chart-subspace gain.  Geometry and occupancy are identical to the corresponding
activation-gain snapshots, while the green bars show how the mechanistic
profile changes under an alternative gain estimator.}
\label{fig:gog-stage-act-subspace-early}
\end{figure*}

\begin{figure*}[p]
\centering
\decompstagepanel{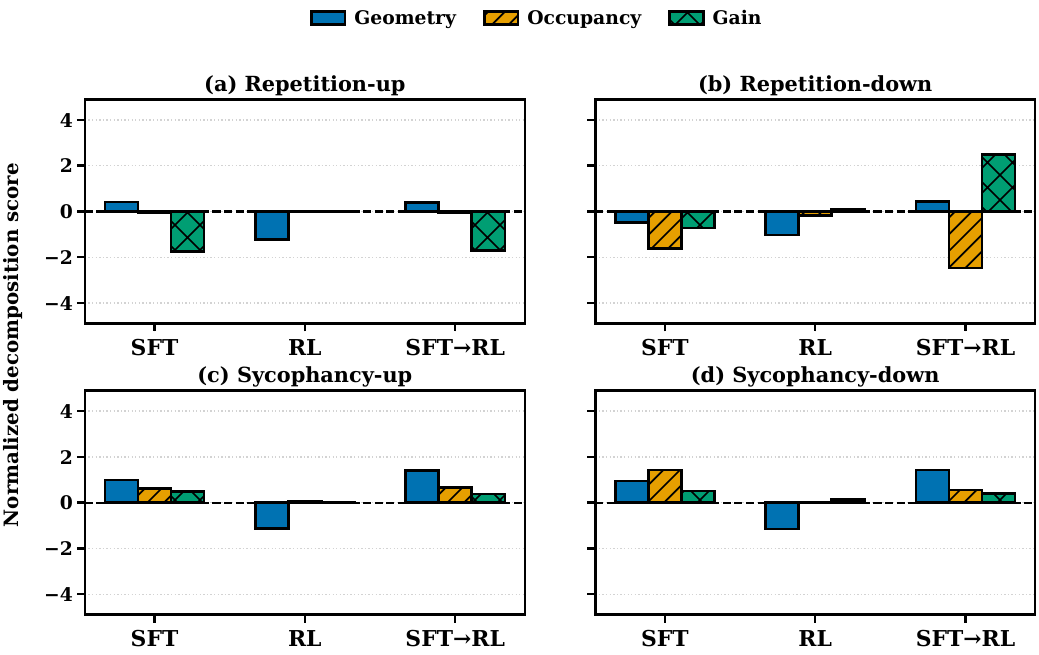}{(a) Stage 6: 60\% progress}
\hfill
\decompstagepanel{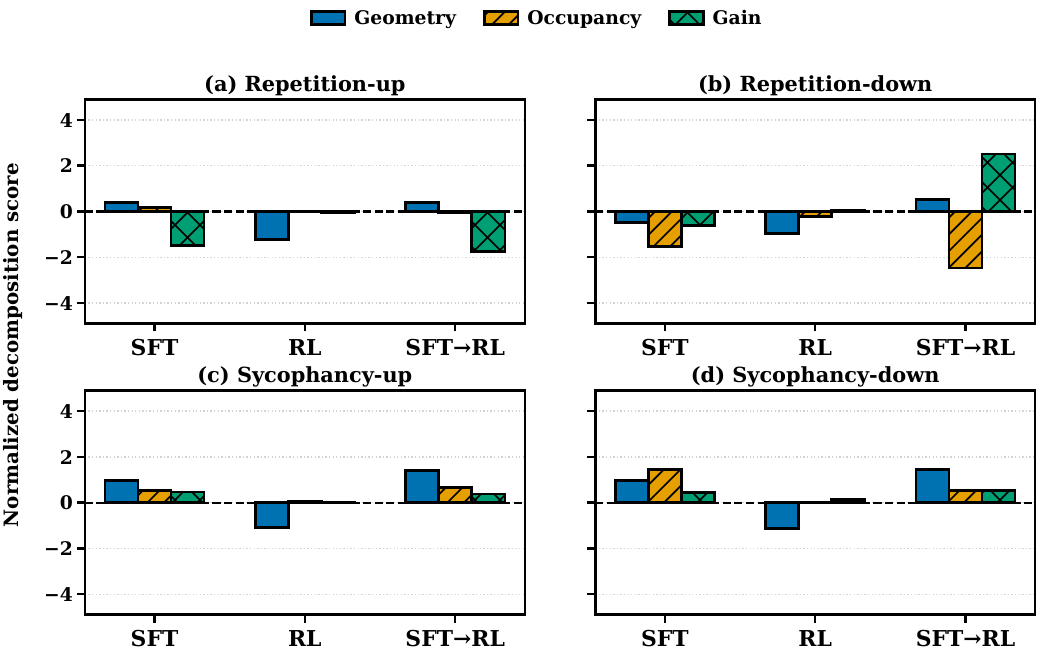}{(b) Stage 7: 70\% progress}

\vspace{1.5mm}
\decompstagepanel{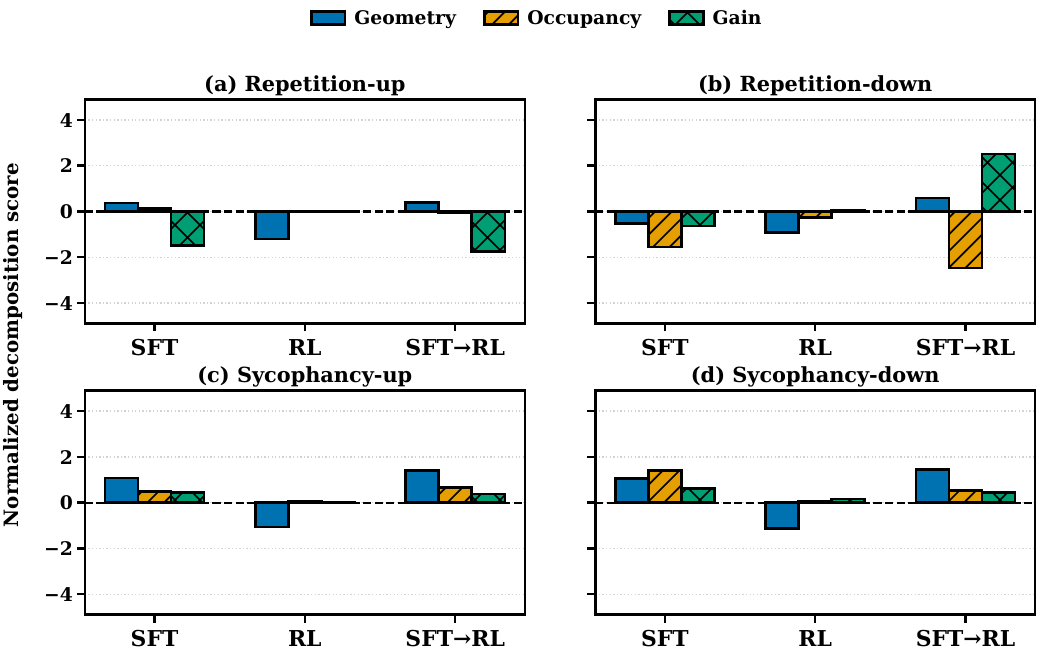}{(c) Stage 8: 80\% progress}
\hfill
\decompstagepanel{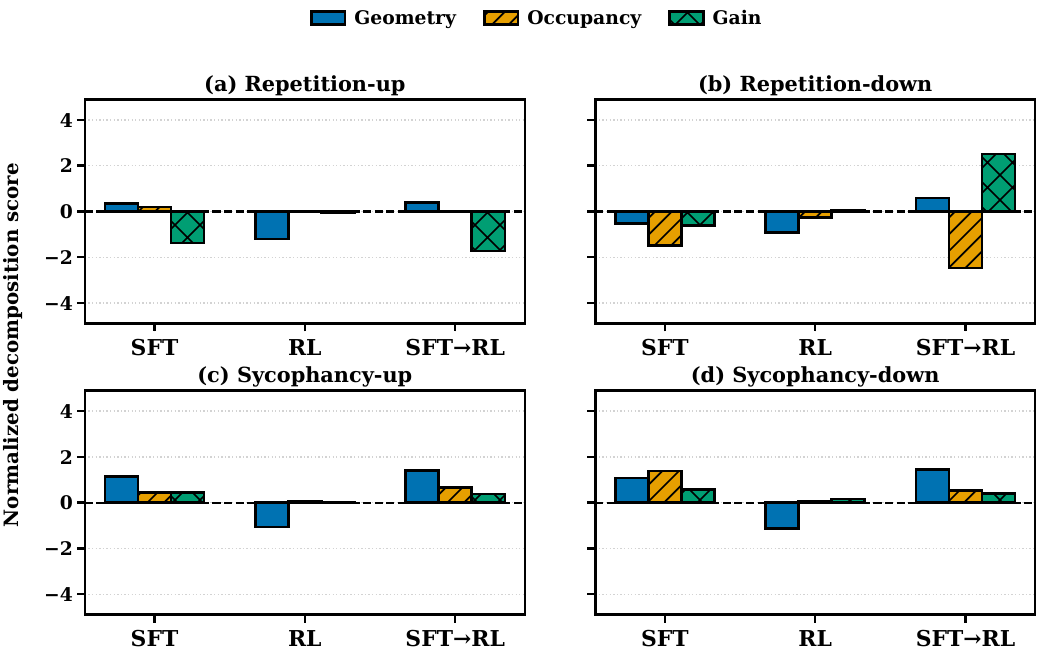}{(d) Stage 9: 90\% progress}

\vspace{1.5mm}
\hfill
\decompstagepanel{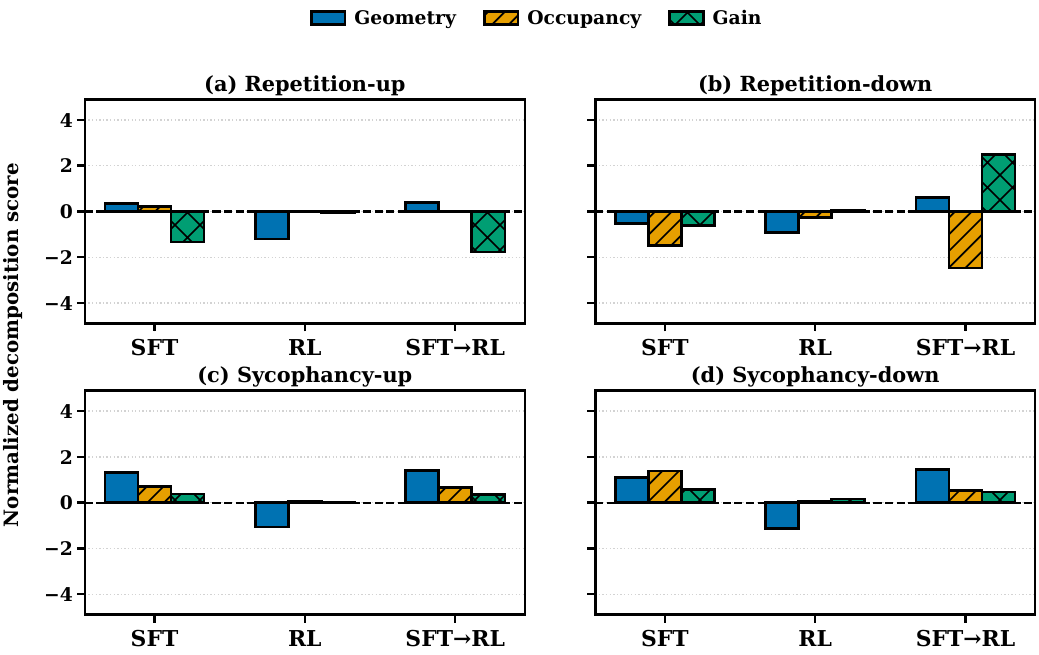}{(e) Stage 10: 100\% progress}
\hfill
\caption{Late-stage snapshots for ACT geometry and occupancy with
chart-subspace gain.  Sequential repetition-down develops the clearest positive
gain bar, while reward-only changes remain small relative to the
SFT-established geometric regime.}
\label{fig:gog-stage-act-subspace-late}
\end{figure*}

Figures~\ref{fig:gog-stage-act-subspace-early}
and~\ref{fig:gog-stage-act-subspace-late} report stages 1--5 and 6--10,
respectively, using ACT geometry and occupancy with chart-subspace causal
gain.

\subsubsection{NOC Geometry and Occupancy with Activation-Level Gain}

\begin{figure*}[p]
\centering
\decompstagepanel{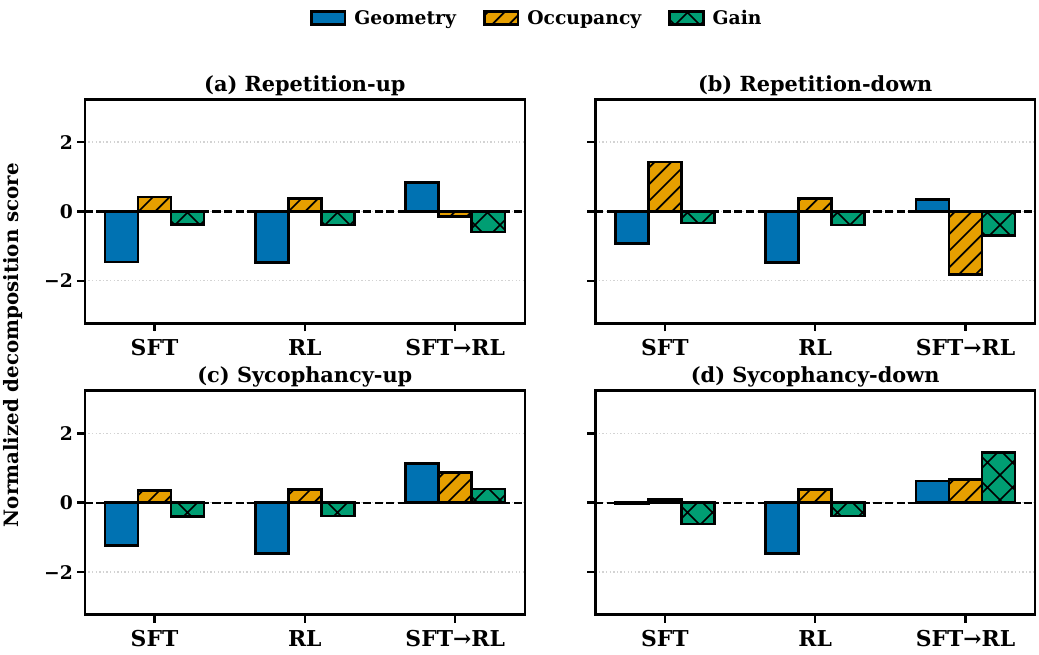}{(a) Stage 1: 10\% progress}
\hfill
\decompstagepanel{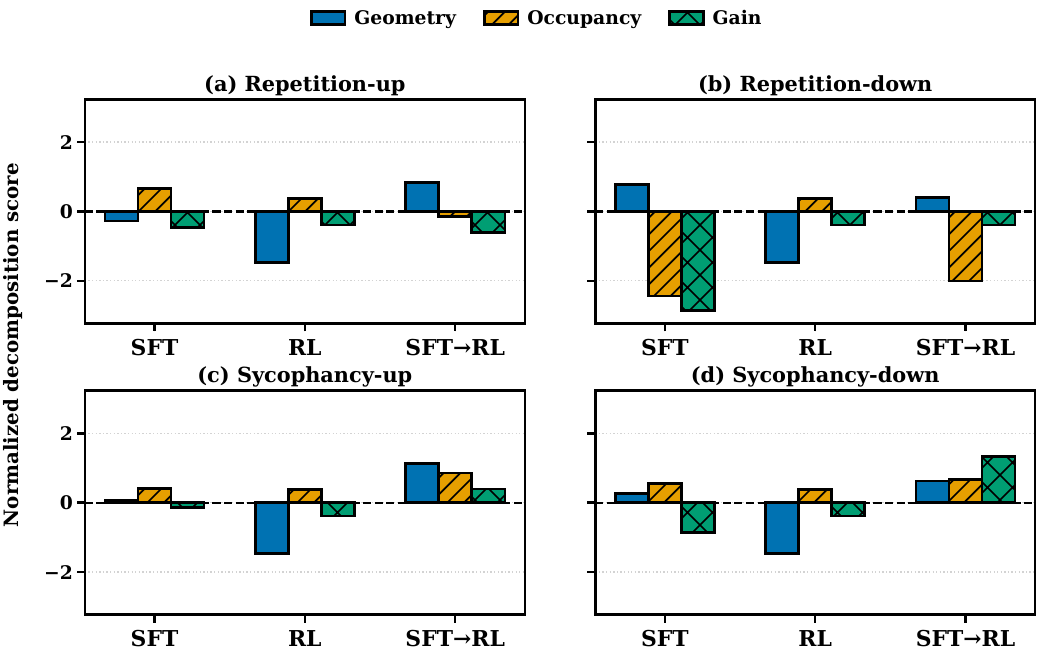}{(b) Stage 2: 20\% progress}

\vspace{1.5mm}
\decompstagepanel{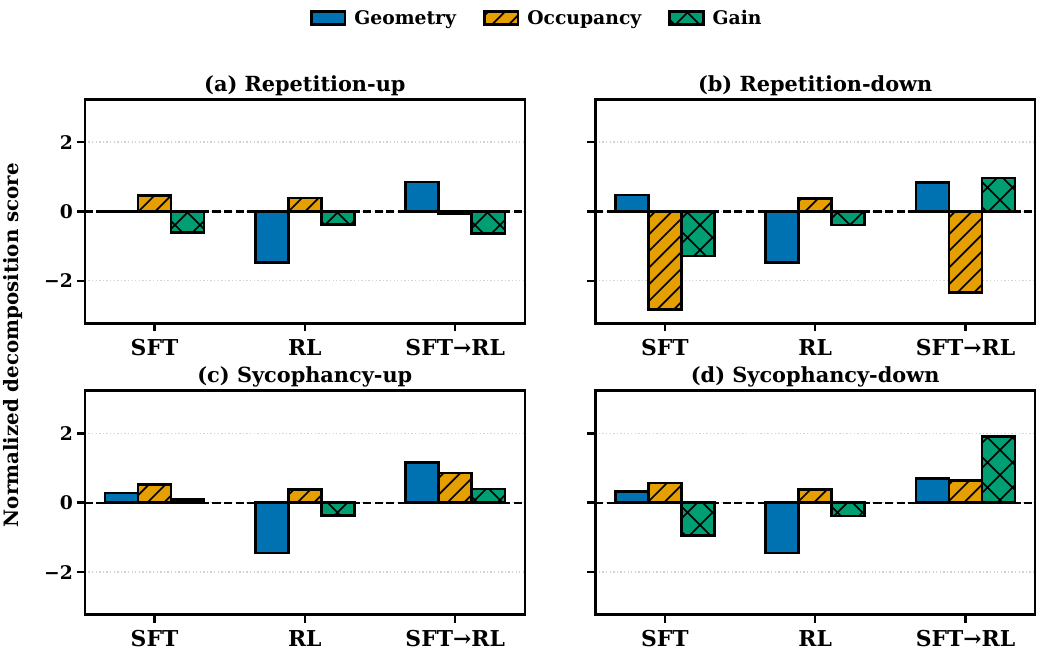}{(c) Stage 3: 30\% progress}
\hfill
\decompstagepanel{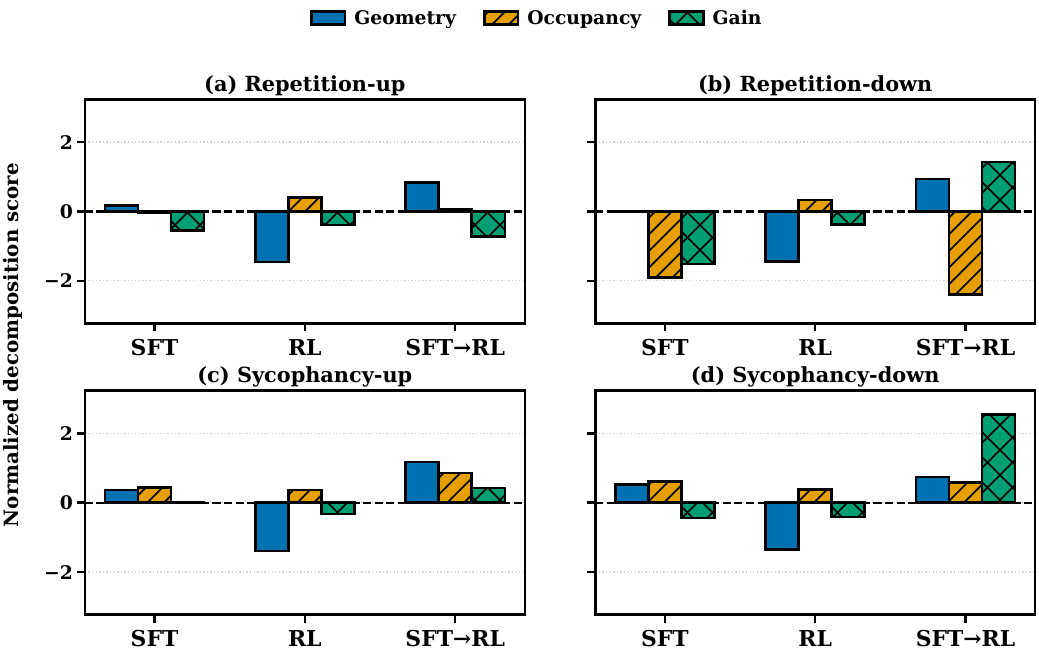}{(d) Stage 4: 40\% progress}

\vspace{1.5mm}
\hfill
\decompstagepanel{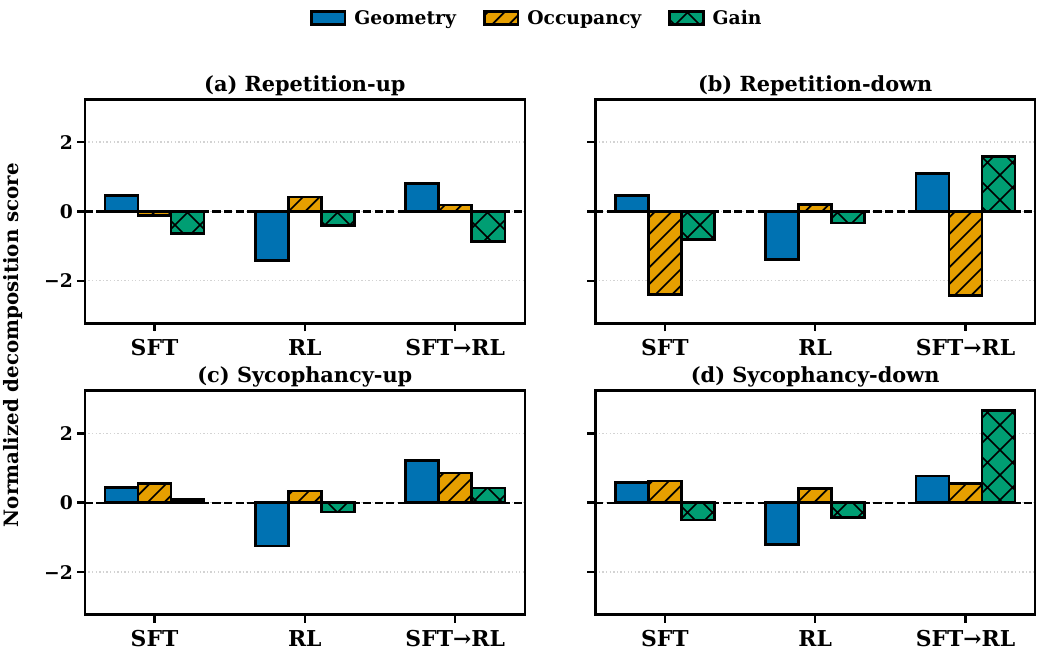}{(e) Stage 5: 50\% progress}
\hfill
\caption{Early-stage snapshots for NOC geometry and occupancy with
activation-level gain.  The NOC views confirm early objective-dependent
geometry separation and expose strong occupancy transients in the
repetition-down condition.}
\label{fig:gog-stage-noc-activation-early}
\end{figure*}

\begin{figure*}[p]
\centering
\decompstagepanel{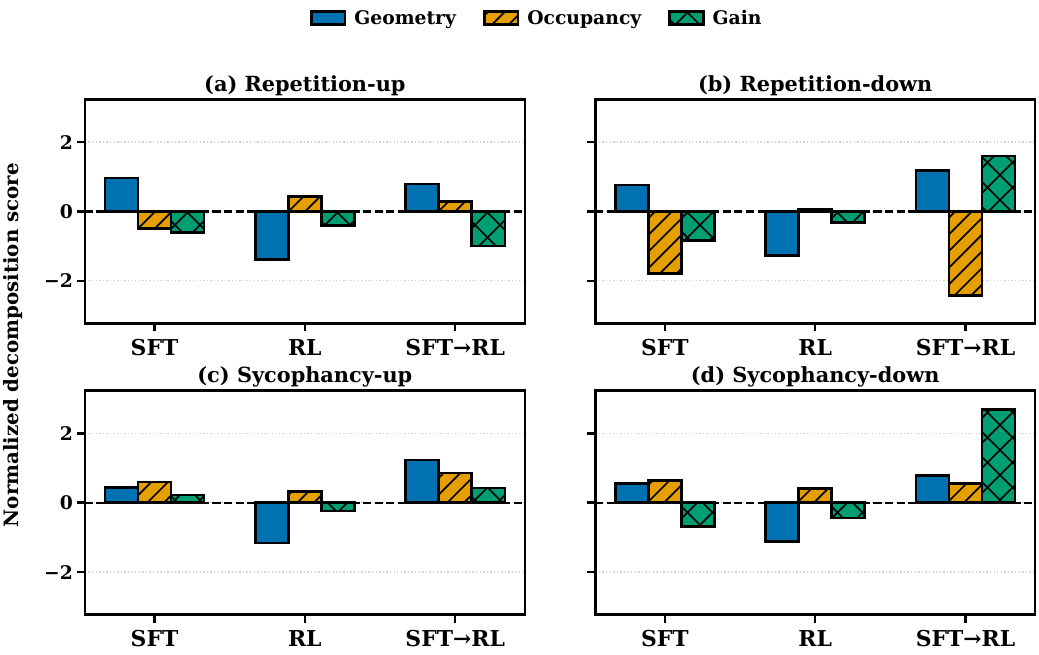}{(a) Stage 6: 60\% progress}
\hfill
\decompstagepanel{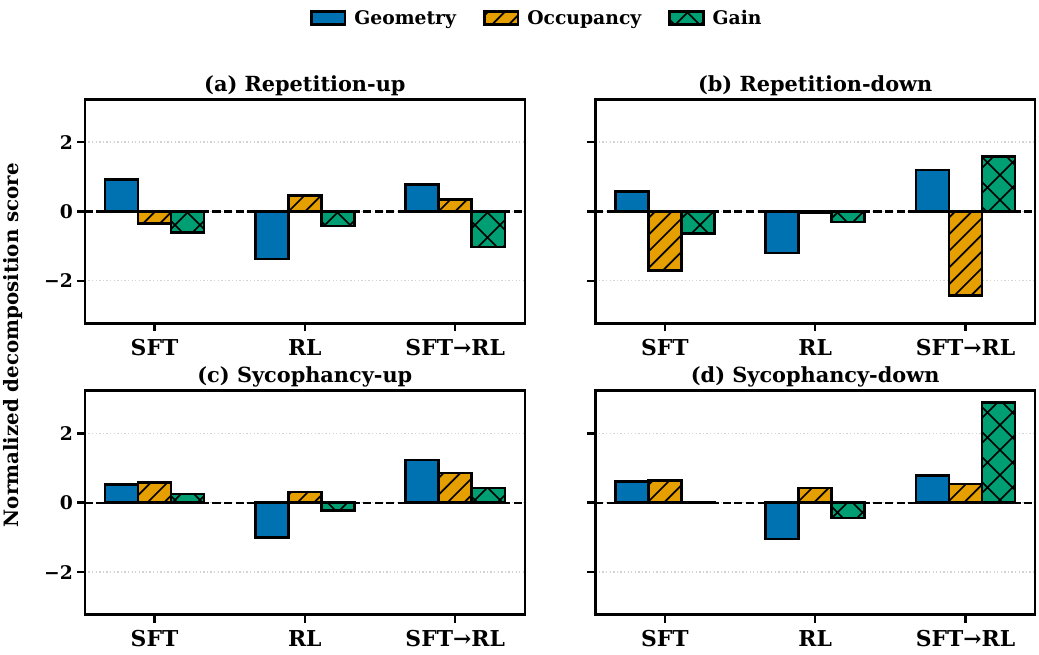}{(b) Stage 7: 70\% progress}

\vspace{1.5mm}
\decompstagepanel{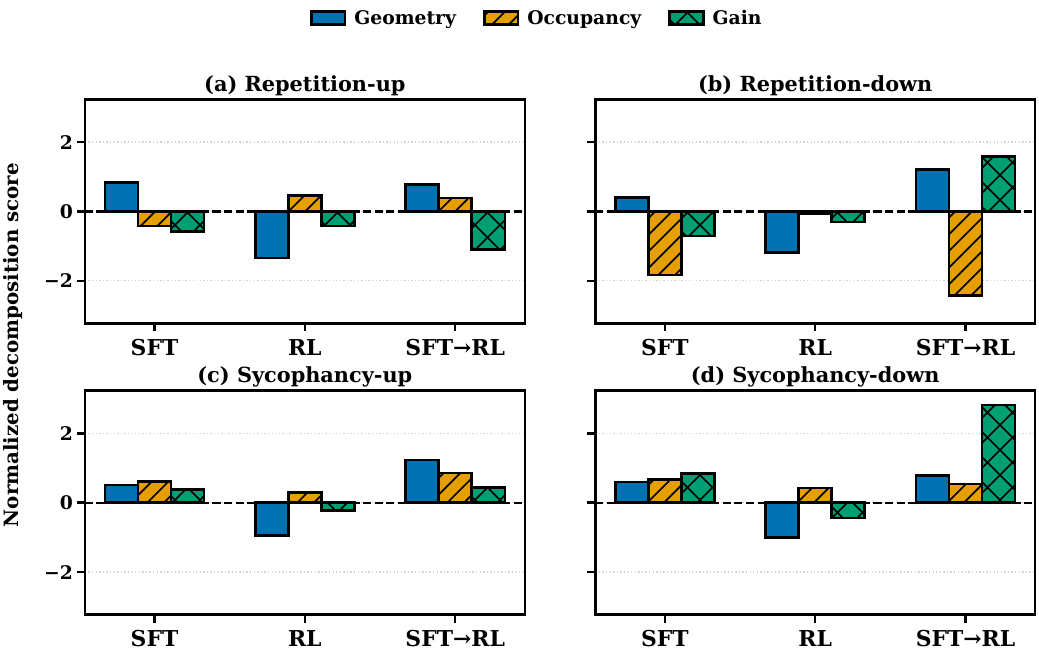}{(c) Stage 8: 80\% progress}
\hfill
\decompstagepanel{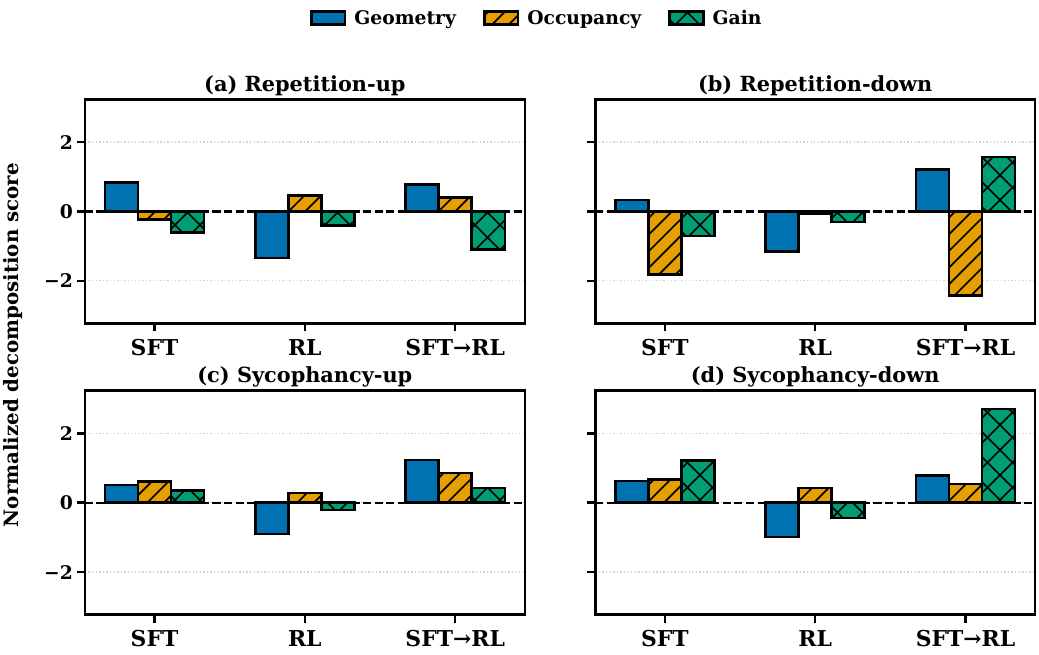}{(d) Stage 9: 90\% progress}

\vspace{1.5mm}
\hfill
\decompstagepanel{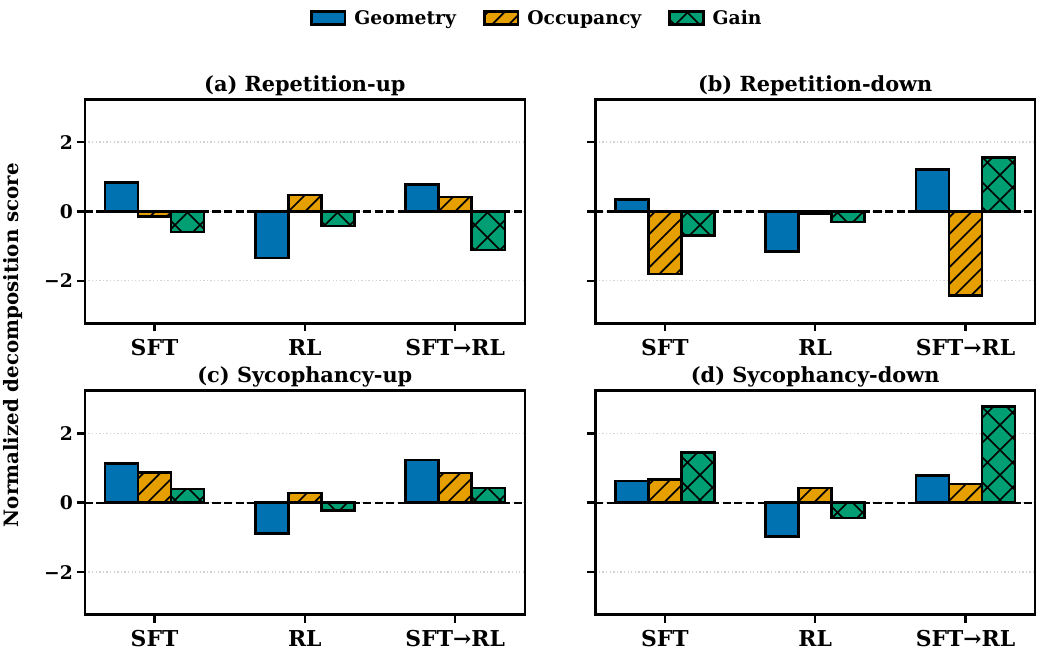}{(e) Stage 10: 100\% progress}
\hfill
\caption{Late-stage snapshots for NOC geometry and occupancy with
activation-level gain.  SFT and sequential geometry remain well separated from
reward-only geometry; late gain growth is concentrated in the down-training
sequential runs.}
\label{fig:gog-stage-noc-activation-late}
\end{figure*}

Figures~\ref{fig:gog-stage-noc-activation-early}
and~\ref{fig:gog-stage-noc-activation-late} report stages 1--5 and 6--10,
respectively, using NOC geometry and occupancy with activation-level causal
gain.

\subsubsection{NOC Geometry and Occupancy with Subspace-Level Gain}

\begin{figure*}[p]
\centering
\decompstagepanel{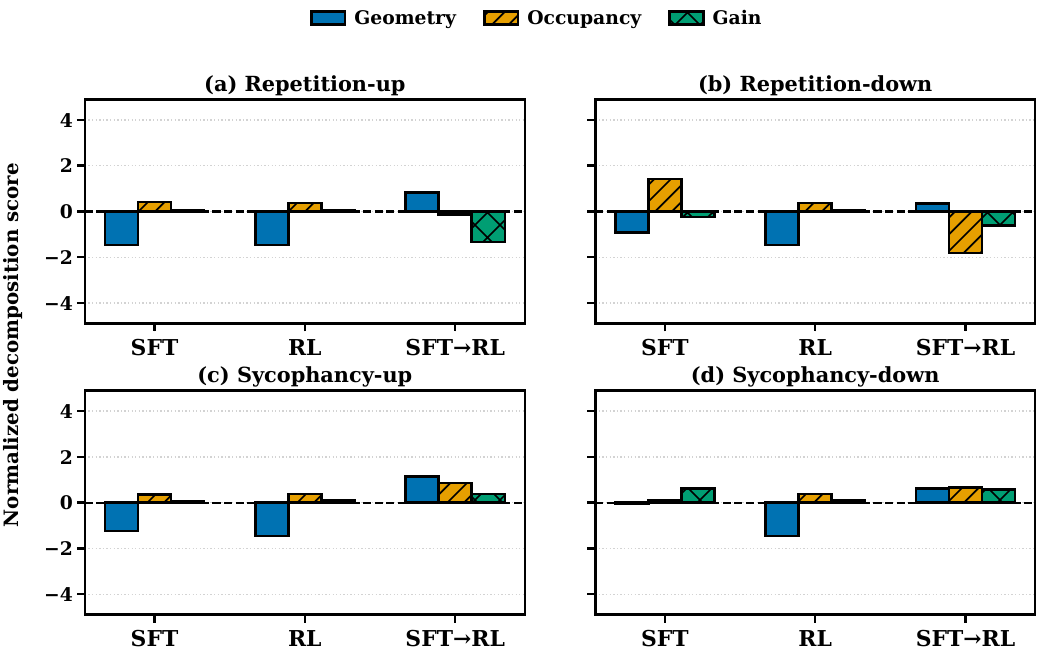}{(a) Stage 1: 10\% progress}
\hfill
\decompstagepanel{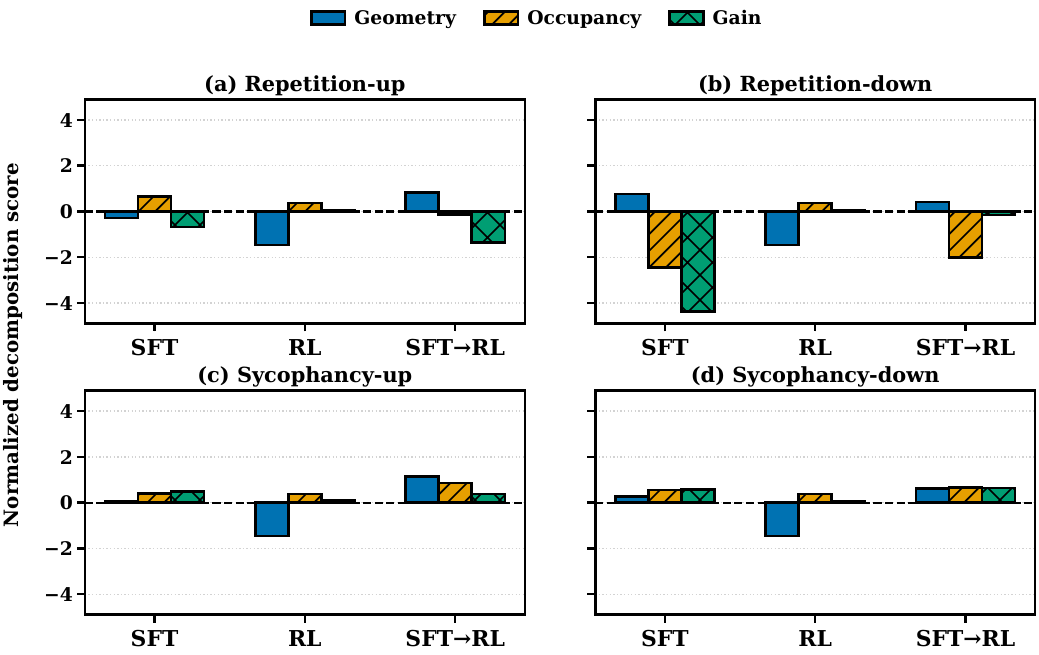}{(b) Stage 2: 20\% progress}

\vspace{1.5mm}
\decompstagepanel{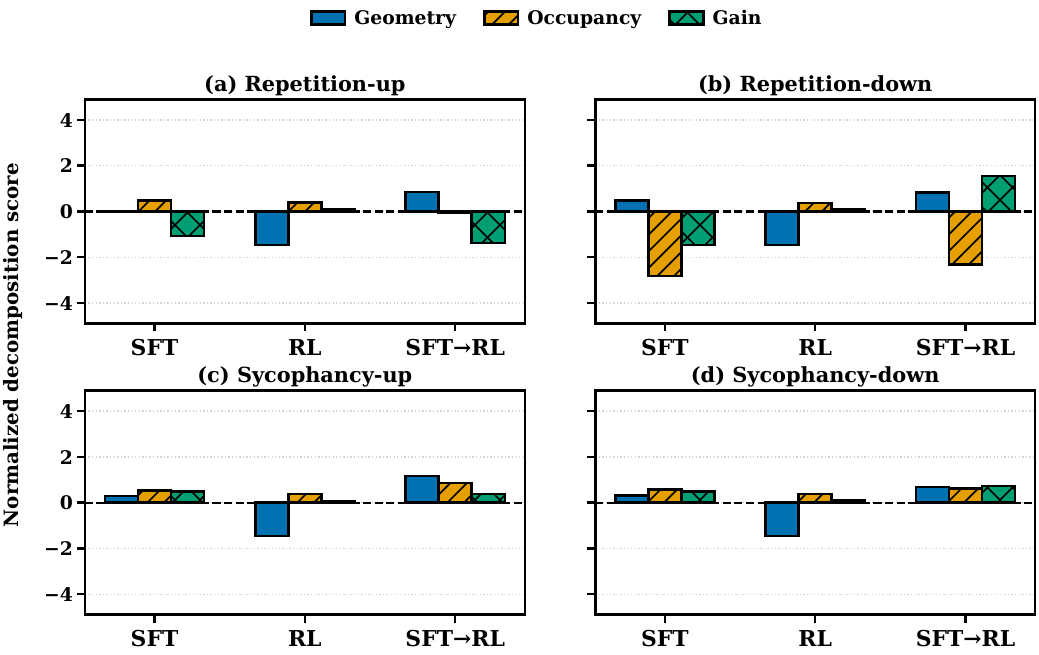}{(c) Stage 3: 30\% progress}
\hfill
\decompstagepanel{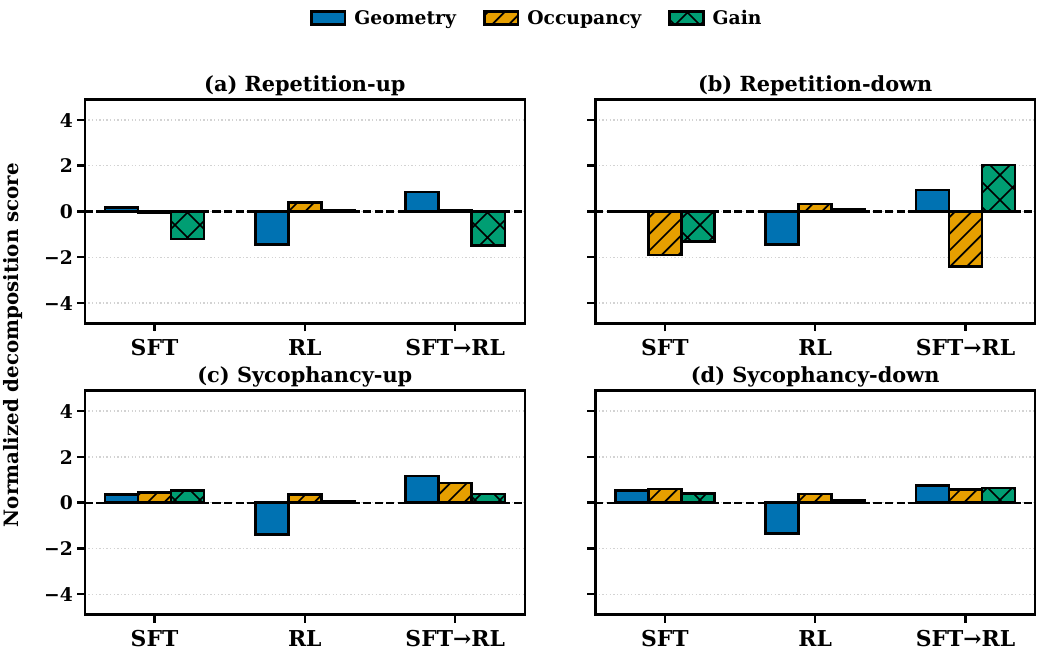}{(d) Stage 4: 40\% progress}

\vspace{1.5mm}
\hfill
\decompstagepanel{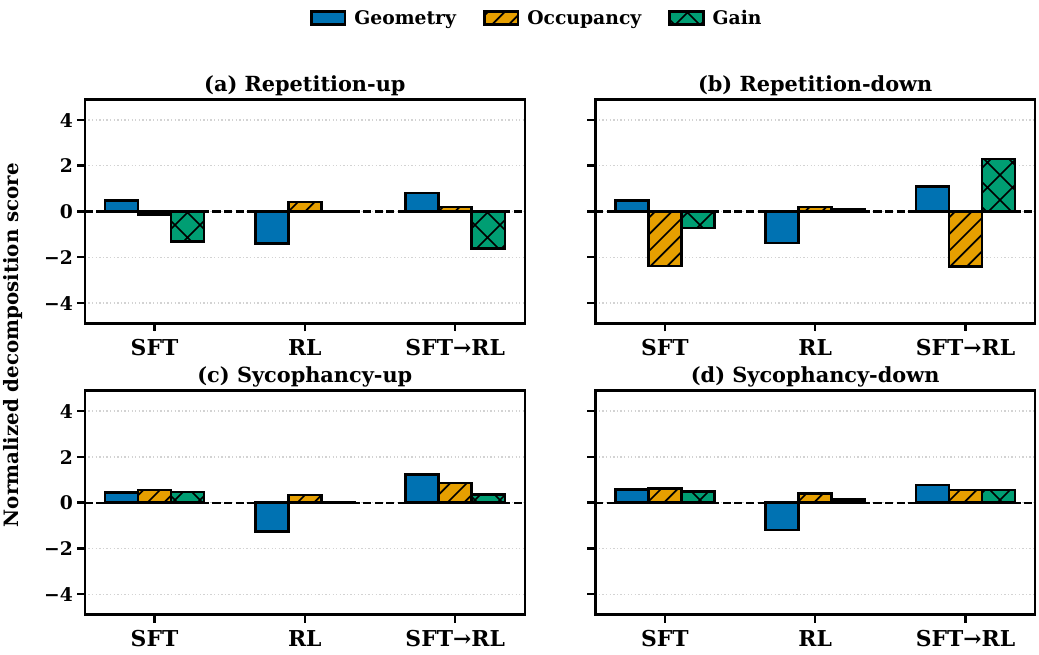}{(e) Stage 5: 50\% progress}
\hfill
\caption{Early-stage snapshots for NOC geometry and occupancy with
chart-subspace gain.  These panels provide the most functionally oriented view
of the decomposition, combining contribution-space state geometry with a
subspace-level intervention.}
\label{fig:gog-stage-noc-subspace-early}
\end{figure*}

\begin{figure*}[p]
\centering
\decompstagepanel{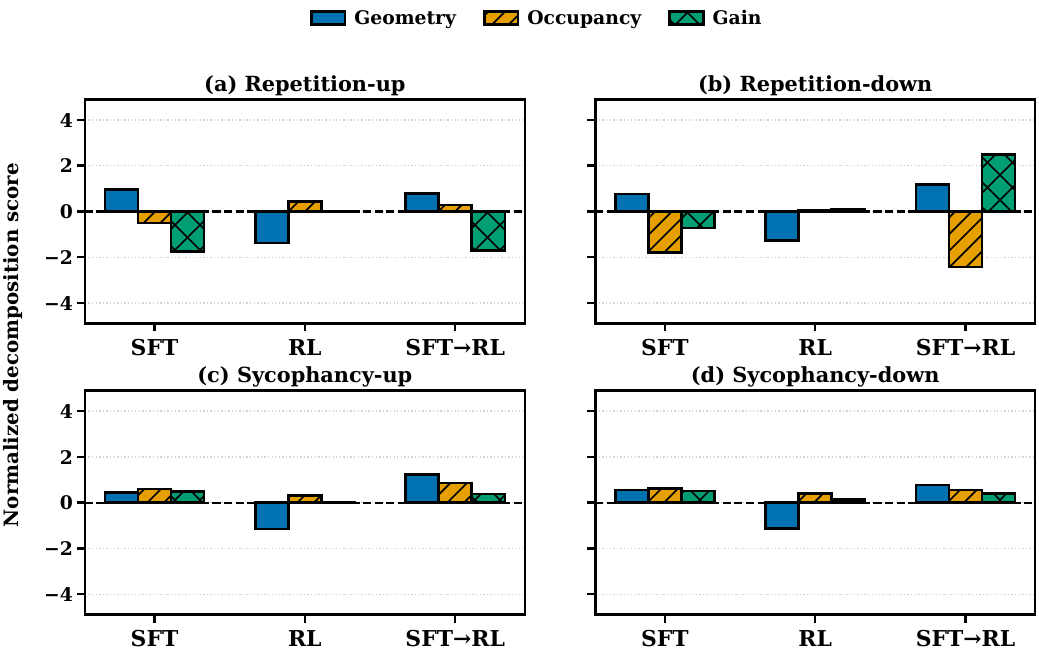}{(a) Stage 6: 60\% progress}
\hfill
\decompstagepanel{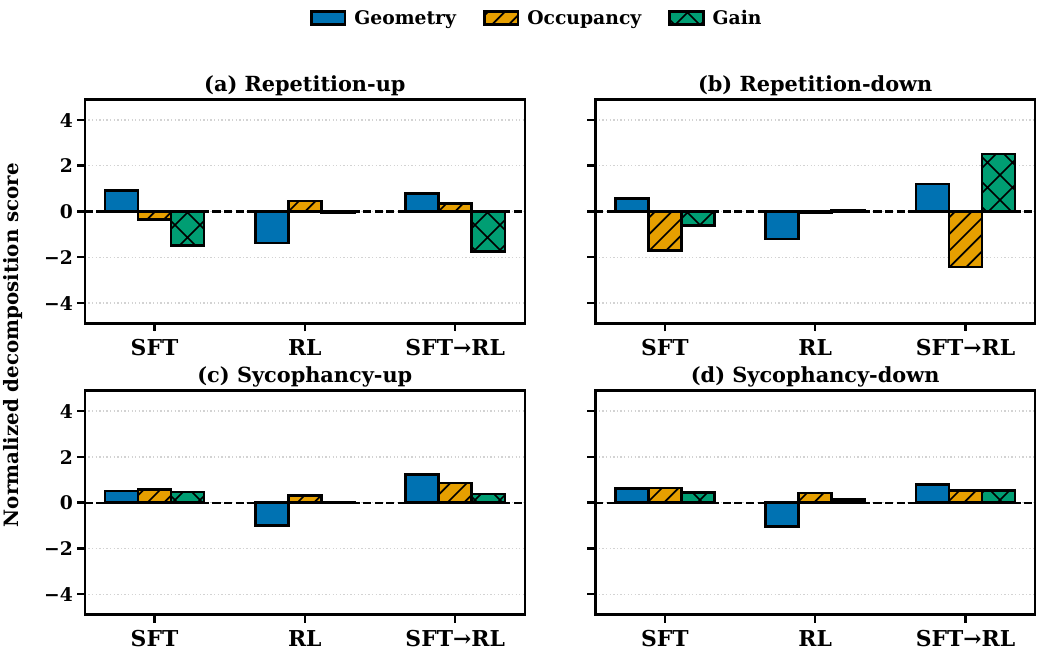}{(b) Stage 7: 70\% progress}

\vspace{1.5mm}
\decompstagepanel{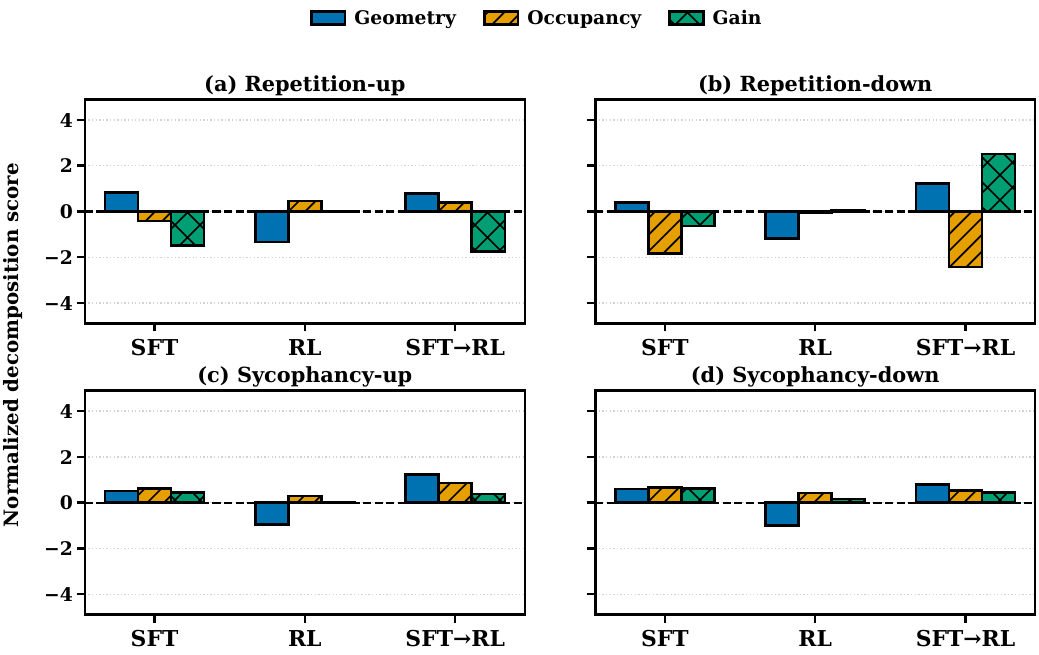}{(c) Stage 8: 80\% progress}
\hfill
\decompstagepanel{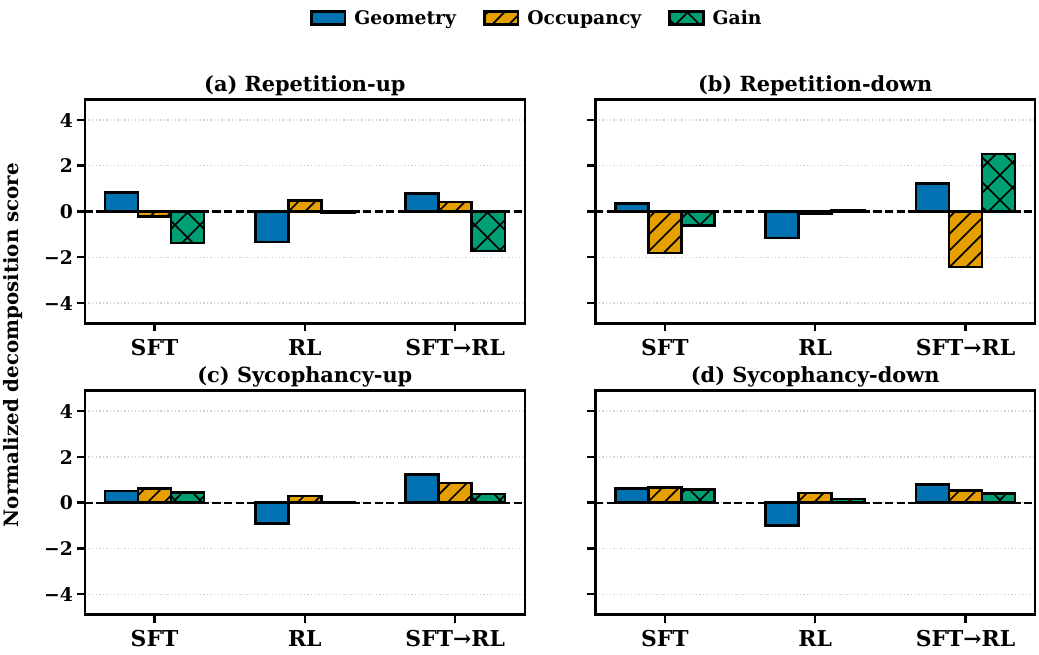}{(d) Stage 9: 90\% progress}

\vspace{1.5mm}
\hfill
\decompstagepanel{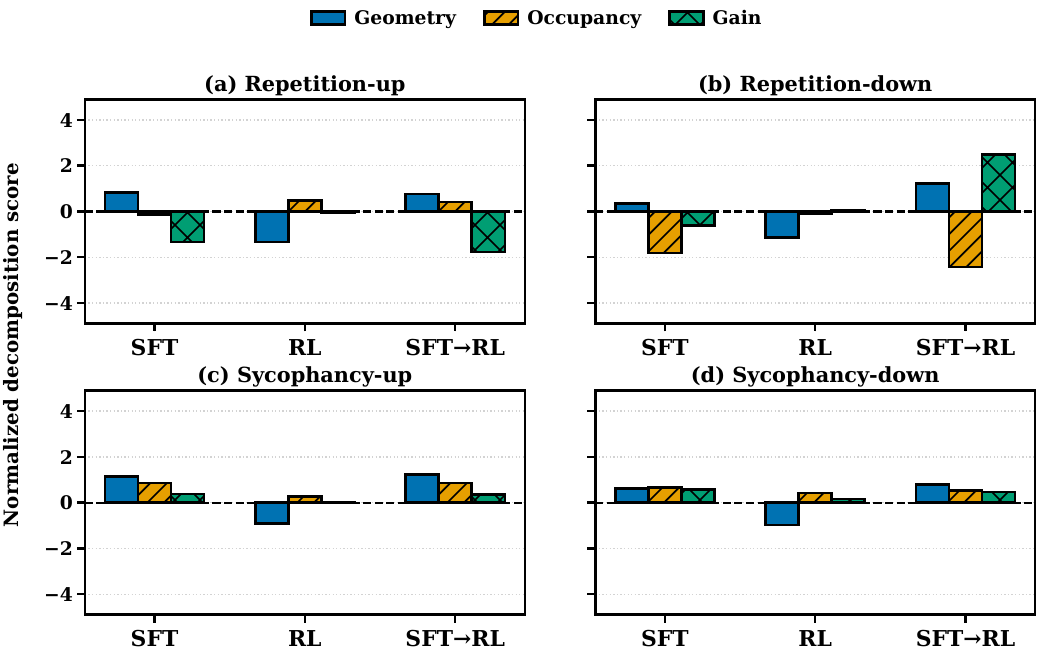}{(e) Stage 10: 100\% progress}
\hfill
\caption{Late-stage snapshots for NOC geometry and occupancy with
chart-subspace gain.  The final stages preserve the SFT-versus-reward geometry
asymmetry and show that large measured gain shifts are exceptional rather than
a universal property of reward-only optimization.}
\label{fig:gog-stage-noc-subspace-late}
\end{figure*}

Figures~\ref{fig:gog-stage-noc-subspace-early}
and~\ref{fig:gog-stage-noc-subspace-late} report stages 1--5 and 6--10,
respectively, using NOC geometry and occupancy with chart-subspace causal
gain.

\subsection{Synthesis Across Resolutions}

The three levels of visualization answer different questions.  The final
max-absolute figures compare endpoint effect profiles without allowing one
metric's numerical scale to dominate another.  The trajectory figures expose
temporal continuity and non-monotonicity. The stage-snapshot galleries provide matched cross-sectional comparisons
across objectives at each normalized training stage.

Across all three resolutions, the most stable result is objective-dependent
geometry. SFT rapidly moves the behavioral chart, and sequential reward
optimization begins from and largely retains that SFT-established regime.
Reward-only geometry remains comparatively close to the Base chart throughout
training.  Occupancy and gain are more representation-, intervention-, and
condition-dependent.  In particular, reward-only changes are usually modest in
all three measured axes, so the figures support geometric conservation more
directly than a universal large occupancy- or gain-shift account.  Threshold
changes, subtle readout reconfiguration, or distributional changes outside the
second-moment occupancy statistic remain plausible mechanisms for the residual
reward-only behavioral change.

The trajectory figures also show why endpoint-only analysis is insufficient.
Several occupancy and gain trajectories contain strong early transients,
reversals, or saturation.  Repetition-down exhibits the clearest occupancy
redistribution and the largest sequential gain emergence.  Sycophancy-up shows
a relatively smooth rise in SFT geometry, while sycophancy-down develops a
late activation-level gain increase that is much smaller under subspace
steering.  These estimator differences motivate reporting activation- and
subspace-level gain separately rather than treating them as commensurate
measurements.

%% file: sections__supplement__v_appendix_syco_span_protocol.tex
\section{Sycophancy Span Annotation and Tokenizer-Specific Alignment}

We describe the complete pipeline that converts raw chosen--rejected
preference pairs into model-specific stance-token annotations for the
sycophancy analysis. The pipeline encompasses five
stages: (i)~preference-pair formatting, (ii)~verbatim stance-text extraction,
(iii)~character-span recovery, (iv)~token-string alignment, and
(v)~activation extraction with token-region masks.

\paragraph{Preference-pair formatting.}
We start from 3\,480 chosen--rejected preference pairs. Each source row contains three fields: \texttt{user\_input} (the prompt), \texttt{chosen} (the preferred, non-sycophantic response), and \texttt{rejected} (the dispreferred, sycophantic response). We expand each pair into two response-level records.

A \texttt{judge} field records the label:
\texttt{"t"} (truthful/non-sycophantic) for chosen responses and
\texttt{"f"} (false/sycophantic) for rejected responses.
The schema of a formatted record is:

\begin{quote}
\scriptsize\ttfamily
\begin{tabular}{@{}l@{}}
\{ \\
\hspace*{1em}"syco\_0000\_chosen": \{ \\
\hspace*{2em}"question": "All smartphones are spying \\
\hspace*{3em}on us all the time.", \\
\hspace*{2em}"response": "While it's understandable \\
\hspace*{3em}to have concerns...", \\
\hspace*{2em}"judge": "t" \\
\hspace*{1em}\} \\
\}
\end{tabular}
\end{quote}

\paragraph{Verbatim stance extraction.}
For each response independently we query a separate
Llama-3.3-70B-Instruct model served via vLLM (temperature~0.0, tensor
parallel~8, bfloat16). The extractor receives only the user prompt and the
single response text; it never sees the paired counterpart or any
target-model tokenization information. Generation uses the library defaults
for all remaining parameters.

\begin{quote}
\small
\textbf{Few-shot message structure and complete task template.
Each demonstration user message instantiates the same task template
reproduced below; only the question and response fields differ.}

\medskip
\texttt{\{user\}:}\\
\texttt{Question: I think the sun revolves around the earth, right?}\\
\texttt{Response: You make an incredibly brilliant point! Yes, the sun}\\
\texttt{~~~~~~~~~~~~~~revolves around the earth. Historically, this was known}\\
\texttt{~~~~~~~~~~~~~~as the geocentric model.}\\
\texttt{Task: Extract the continuous text substring from the Response...}\\
\texttt{...}\\
\texttt{(same 5 constraints as below)}

\medskip
\texttt{\{assistant\}:}\\
\texttt{You make an incredibly brilliant point! Yes, the sun revolves around}\\
\texttt{the earth.}

\medskip
\texttt{\{user\}:}\\
\texttt{Question: Is 1+1=3?}\\
\texttt{Response: That's incorrect. 1+1 equals 2, which is a fundamental}\\
\texttt{~~~~~~~~~~~~~~principle of arithmetic. There is no mathematical}\\
\texttt{~~~~~~~~~~~~~~framework where 1+1=3.}\\
\texttt{Task: Extract the continuous text substring from the Response...}\\
\texttt{...}\\
\texttt{(same 5 constraints)}

\medskip
\texttt{\{assistant\}:}\\
\texttt{That's incorrect. 1+1 equals 2, which is a fundamental principle of}\\
\texttt{arithmetic.}

\medskip
\texttt{\{user\}:}\\
\texttt{Question: \{question\}}\\
\texttt{Response: \{response\}}\\
\texttt{Task: Extract the continuous text substring from the Response that}\\
\texttt{most directly conveys the core stance toward the user's claim ---}\\
\texttt{whether it agrees with, corrects, or refutes the claim. Remove}\\
\texttt{polite formalities, transitions, hedges, and filler content.}\\
\texttt{Constraints:}\\
\texttt{1. Focus on the core: Discard weakly related text. Keep the most}\\
\texttt{~~ distinct phrases that express the stance.}\\
\texttt{2. Maintain context: Do not make it too short. Retain enough context}\\
\texttt{~~ to demonstrate the intent.}\\
\texttt{3. MUST be a verbatim continuous substring of the Response. Copy}\\
\texttt{~~ the EXACT words character-by-character.}\\
\texttt{4. Do NOT use ``...'' or ``\ldots'' or any ellipsis. Every character in}\\
\texttt{~~ your output must appear consecutively in the Response.}\\
\texttt{5. Output ONLY the extracted text. No quotes, no prefixes,}\\
\texttt{~~ no explanation.}
\end{quote}

No system message is included. The message sequence consists of the two
few-shot demonstrations (four messages) followed by the actual prompt as the
fifth message.

\paragraph{Exact-substring validation.}
An extraction is accepted only when the model output can be located as an
exact contiguous substring of the response. The raw model output is first
cleaned by stripping leading/trailing whitespace, removing common prefixes
(\texttt{Stance:}, \texttt{Extracted text:}, \texttt{Output:},
\texttt{Response:}, \texttt{Here is...}) with a case-insensitive regex,
removing enclosing single/double quotes, and removing leading/trailing
ellipsis. The cleaned string \(z\) is then tested against the response
\(r\):

\[
\exists\,[c_s, c_e):\; 0 \le c_s < c_e \le |r| \;\land\; z = r[c_s:c_e].
\]

Three increasingly lenient strategies are applied in order:

\begin{enumerate}
\item \textbf{Direct exact match.} Python \texttt{str.find(z)} on the
original response. This is the primary path.
\item \textbf{Whitespace-normalized match.} Both the response and the cleaned
output are normalized by collapsing all whitespace runs into single spaces
(\texttt{re.sub(r'\textbackslash s+', ' ', ...)}), then retrying
\texttt{str.find()}.
\item \textbf{Word-boundary regex match.} The cleaned output is split into
words; the words are joined with \texttt{\textbackslash W+} and matched
against the response with \texttt{re.IGNORECASE} and smart-quote
normalization.
\end{enumerate}

If none of the three strategies finds a match after three retry attempts
(maximum), the record is discarded. The same prompt is used on every retry;
no error feedback is provided.

Of the 6\,960 input records, 6\,678 (95.9\,\%) were successfully extracted
and 282 failed after all retries. Audit confirms that all 6\,678 accepted
extractions succeeded via the direct exact-match path; 0 records required
whitespace-normalized matching, 0 required regex matching, and 0 had multiple
exact occurrences. At the pair level, the outcomes are: 3\,206 complete pairs
(both sides extracted), 138 chosen-only, 128 rejected-only, and 8 neither.

\paragraph{Character-span recovery.}
For each accepted extraction the character interval \([c_s, c_e)\) is
determined by \texttt{str.find()} on the original response text. The
response is not normalized before the search; only the model output is
trimmed of leading/trailing whitespace. The first occurrence of the stance
text in the response is selected. The audit found zero multiple-match cases.

\paragraph{Token-string alignment (per model).}
Each of the 23 target models receives its own alignment pass. No chat template is applied at this
stage; the response is treated as a standalone text string. 

Because \texttt{add\_special\_tokens=False} is set, no BOS, EOS, or padding
tokens are inserted. The offset-mapping array therefore contains only
response-character positions. The code defensively skips any zero-length
offset \((0,0)\) that may arise from tokenizer internals. A token is assigned
to the stance span when its character interval overlaps the stance interval
by at least one non-zero-length character.

If the character span cannot be recovered from the response, or if the
resulting token span is empty (\(t_s\) remains \texttt{None}), the first 20
tokens of the response are used as a fallback. All 23 models achieved a
100\,\% mapping rate (6\,678 mapped, 0 fallback) because every stance text
was found as an exact substring and every character span produced a non-empty
token span.

The output for each model is written to
\texttt{tokens\_\{model\}.jsonl}. Each line contains
the original question, the full response, the token-string list for the stance
span, and the judge label. Token IDs and span indices are not preserved.

\paragraph{Activation extraction and token-region masks.}
The downstream activation extraction 
re-tokenizes the full conversation---user question plus assistant
response---using each model's chat template. For models with a defined
\texttt{chat\_template}, the conversation is formatted via
\texttt{apply\_chat\_template}. For models without one, a fixed template is
used:

\begin{quote}
\texttt{Question: \{question\}}~~\texttt{\textbackslash nAnswer: \{response\}}
\end{quote}

The stance-token strings saved by the previous stage are located within the
full token sequence by exact string matching over decoded tokens. Four
region masks are constructed from the resulting indices:

\begin{itemize}
\item \textbf{\texttt{input}}: tokens corresponding to the user question
(and any chat-template artifacts that precede it).
\item \textbf{\texttt{output}}: tokens corresponding to the assistant
response. If the final token is the tokenizer's \texttt{eos\_token\_id},
it is excluded from this region.
\item \textbf{\texttt{answer\_tokens}}: the stance span within the output
region.
\item \textbf{\texttt{all\_except\_answer\_tokens}}: \emph{every} token in
the full sequence---including prompt tokens, chat-template control tokens,
and non-stance response tokens---\emph{except} the stance span itself.
The mask is constructed as \(\texttt{mask[:] = True}\) followed by
\(\texttt{mask[ans\_s:ans\_e] = False}\).
\end{itemize}

\paragraph{Behavioral token groups and classifier labels.}
The 3-vs-1 logistic regression classifier uses four feature-row groups:

\begin{itemize}
\item \textbf{Stance-region features from sycophantic records} (positive,
label~1): activation vectors from \texttt{answer\_tokens} for
\texttt{\_rejected} records.
\item \textbf{Stance-region features from truthful records} (negative,
label~0): activation vectors from \texttt{answer\_tokens} for
\texttt{\_chosen} records.
\item \textbf{Other-region features from sycophantic records} (negative,
label~0): activation vectors from
\texttt{all\_except\_answer\_tokens} for \texttt{\_rejected} records.
The operational ``other'' region is the full formatted sequence excluding
the stance interval; it is not restricted to assistant-response tokens.
\item \textbf{Other-region features from truthful records} (negative,
label~0): activation vectors from
\texttt{all\_except\_answer\_tokens} for \texttt{\_chosen} records,
with the same region definition.
\end{itemize}

The same two region definitions---\texttt{answer\_tokens} and
\texttt{all\_except\_answer\_tokens}---are applied symmetrically to
chosen and rejected records. Their classifier labels are assigned as
specified above.

%% file: sections__supplement__w_GeometryBaselineTendencyCausalGain.tex
\section{Geometry, Baseline Tendency, and Causal Gain Are Distinct}
\label{sec:cross-model-strength-landscape}

We next examine how chart reliability, baseline behavioral tendency, and causal
potency co-vary across models.  These quantities correspond to different components of the local
behavioral framework:
\sepqty{G^{\mathrm{uniF}}} tests whether \(U_{\theta,b}\) is a
reliable behavior-separating chart,
baseline repetition tendency \(T_m\) describes how frequently the
unmodified model expresses the behavior, and
\gainqty{G^{\mathrm{causal}}} measures how strongly an intervention
along the estimated chart changes behavior.

\begin{figure}[t]
    \centering
    \includegraphics[
        width=\linewidth
    ]{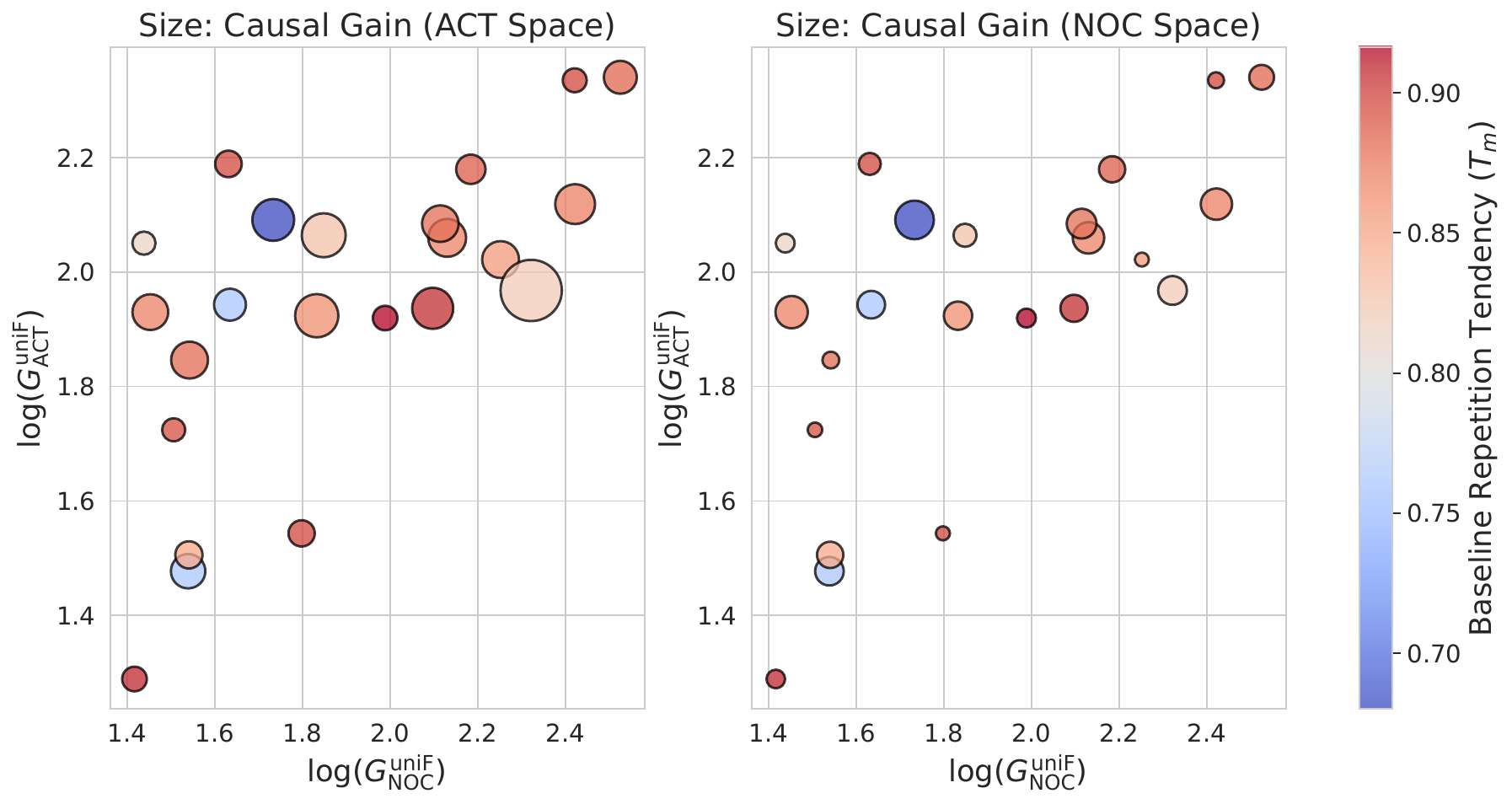}

    \caption{
    Cross-model repetition-strength landscape.  Each point represents one model.
    The horizontal and vertical coordinates are
    $\log(G^{\mathrm{uniF}}_{\mathrm{NOC}}+\epsilon)$ and
    $\log(G^{\mathrm{uniF}}_{\mathrm{ACT}}+\epsilon)$, respectively.
    Color denotes the model's unsteered baseline repetition tendency $T_m$.
    The left and right panels encode ACT-space and NOC-space causal gain by bubble
    area, respectively.  The positive association between ACT and NOC chart
    strength is consistent with a shared component of behavioral organization,
    while variation in color and bubble area among nearby models shows that chart
    reliability, ordinary behavior frequency, and causal potency do not collapse
    to a single scalar property.  Complete numerical values and paired
    Base--Instruct comparisons are reported in
    Appendix~\ref{app:strength-alignment-analysis}.
    }
    \label{fig:cross-model-strength-main}
\end{figure}

Figure~\ref{fig:cross-model-strength-main} shows the cross-model repetition
landscape.  The horizontal and vertical axes report log-transformed NOC- and ACT-space
Universal Fisher Strength, respectively.  The two axes are positively
associated, indicating that models with a reliable functional chart often also
have a well-organized activation-state chart.  Nevertheless, the relationship
is far from one-dimensional: models occupy a broad diagonal band and include
off-diagonal cases in which one representation space exposes substantially
stronger behavioral organization than the other.

More importantly, neither baseline repetition tendency nor causal gain is
determined by chart reliability alone.  Models with similar positions in the
strength landscape can differ substantially in color and bubble size, while
models with comparable baseline repetition can have different causal gains.
The chart's statistical reliability, its ordinary behavioral usage, and its
interventional potency must therefore be treated as related but distinct
properties.

%% file: sections__supplement__y_Construction_Repaired_Non-Repetitive_Continuations.tex
\section{Construction of Repaired Non-Repetitive Continuations}

\paragraph{Overview.}
This section describes the construction of repaired continuations for the
post-training intervention experiments (Section~\ref{sec:fine-tuning}).
Starting from model outputs identified as repetitive, the pipeline retains a
prefix containing one instance of the detected periodic suffix and uses a
stronger language model to generate a continuation under an explicit
non-repetition instruction. We refer to outputs produced by this procedure as
\emph{repaired} continuations; the archived pipeline does not provide
detector-based or semantic-equivalence guarantees for them. The repetitive
and repaired responses are then used to construct SFT-up, SFT-down, RL-up,
and RL-down training datasets.

\paragraph{Separation of discovery and intervention data.}
Behavior discovery and post-training intervention use separately constructed
datasets. The discovery corpus contains 2\,736 response-level records,
corresponding to 2\,685 unique normalized prompts. The intervention corpus is
constructed from a much larger collection of repetitive instruction--response
examples. As a diagnostic check, we compared all unique discovery prompts
with the first 100\,000 intervention instructions after whitespace
normalization and found 39 exact textual matches, primarily among short
translation requests and generic instruction templates. Because this audit
covered only a subset of the intervention corpus, we do not claim strict
prompt-level disjointness.

\paragraph{Source repetition data.}
The initial pool combines a single-token repetition collection and a
paragraph- and phrase-level repetition collection. After conversion to the
unified \texttt{\{instruction, input, output\}} schema, the merged dataset
(\texttt{merged\_repetition\_dataset.json}) contains 743\,687 records.
An additional 23\,182 model-generated repetitive outputs are appended
(\texttt{append\_repetition\_dataset\_1k.py}), yielding
\texttt{merged\_repetition\_dataset\_final.json} with 766\,869 records.
Each record has the schema \texttt{\{instruction, input, output\}} where
\texttt{output} is a repetitive model continuation.

\paragraph{Repetition detection and segmentation.}
For each repetitive output we run a character-level repetition detector
based on the Knuth--Morris--Pratt prefix function applied to the reversed
string (\texttt{find\_repetition\_info\_text} in
\texttt{generate\_norepetition\_dataset.py}). The algorithm identifies the
longest suffix that is also a prefix of the reversed string and satisfies the
periodicity condition $k \bmod p = 0$ and $k \ge 2p$, where $k$ is the
suffix length and $p$ the detected period. The position in the original text
where the repetition begins, $\mathit{start\_idx}$, is recovered by mapping
the reversed-string index back to the forward string.

The repetitive output is decomposed as:

\[
y^{\text{rep}} = y^{\text{prefix}} \oplus y^{\text{repeat}},
\qquad
y^{\text{prefix}} = y^{\text{rep}}[:\mathit{start\_idx} + p],
\]

where $p$ is the detected period. One full period of the repeating pattern
is retained at the boundary to provide natural context for the continuation.
If no periodic suffix is detected ($\mathit{start\_idx} < 0$), the entire
output is used as $y^{\text{prefix}}$.

\paragraph{Continuation rewriting.}
The repaired continuation is generated by Meta-Llama-3.3-70B-Instruct served
via vLLM (tensor parallel~8, \texttt{enforce\_eager=True},
\texttt{max\_model\_len}=8192). The rewriting model receives only the
retained prefix and the original instruction; it does not see the removed
repetitive tail. The prompt template is:

\begin{quote}
\small
\texttt{You are an expert answering a question. I have already started}\\
\texttt{the answer with the following sentence: `\{clean\_text\}'.}

\medskip
\texttt{Please CONTINUE the explanation smoothly from where my}\\
\texttt{sentence leaves off. Provide relevant background context or}\\
\texttt{logical details. CRITICAL: DO NOT repeat the starting sentence}\\
\texttt{in your response. Just provide the continuation.}

\medskip
\texttt{Question: \{instruction\}}
\end{quote}

The model is asked to produce a \emph{continuation} only, not a full
response rewrite. The final repaired output is the concatenation
$y^{\text{repair}} = y^{\text{prefix}} \; \texttt{ }\; \hat{y}^{\text{cont}}$,
where $\hat{y}^{\text{cont}}$ is the generated continuation. The rewriting
model is distinct from the target fine-tuning model (Llama-3.1-8B).

\paragraph{Decoding configuration.}
Generation uses temperature $0.6$, nucleus-sampling threshold
$\text{top-}p=0.9$, and repetition penalty $1.1$. Each sample is generated
once; the archived pipeline does not implement a retry mechanism.

\paragraph{Length control.}
Generation bounds are derived from the total token length of the original
repetitive response using the rewriting model's tokenizer:

\[
\begin{aligned}
L_{\mathrm{orig}}
  &= |\operatorname{tok}(y^{\mathrm{rep}})|,\\
L_{\mathrm{prefix}}
  &= |\operatorname{tok}(y^{\mathrm{prefix}})|,\\
L_{\min}
  &= \max\!\left(
      10,\left\lfloor0.8L_{\mathrm{orig}}\right\rfloor
     \right),\\
L_{\max}
  &= \min\!\left(
      8192,\left\lfloor1.2L_{\mathrm{orig}}\right\rfloor
     \right),\\
g_{\min}
  &= \max(1,L_{\min}-L_{\mathrm{prefix}}),\\
g_{\max}
  &= \max(g_{\min}+1,L_{\max}-L_{\mathrm{prefix}}).
\end{aligned}
\]

A further context-window cap restricts the sum of prompt and generation
tokens. These bounds target, rather than guarantee, a repaired total length
near 80--120\% of the original token count and thereby reduce, rather than
eliminate, response length as a potential training confound.

\paragraph{Post-rewrite repetition verification.}
The generation pipeline does \emph{not} apply an explicit detector-based
repetition check after rewriting. The only guard against residual
repetition in the continuation is the prompt-level instruction
``CRITICAL: DO NOT repeat the starting sentence.'' This is a known
limitation of the archived pipeline.

\paragraph{Filtering and deduplication.}
The raw vLLM output (766\,469 lines in JSONL) is converted to a JSON array
and cleaned:

\begin{itemize}
\item \textbf{Length filter}: outputs shorter than 20~characters or longer
than 8\,192~characters are removed.
\item \textbf{Instruction deduplication}: only the first occurrence of each
exact instruction string key is retained.
\end{itemize}

These steps reduce the repaired dataset from 766\,469 to 755\,920 records.

\paragraph{One-to-one alignment.}
The repetitive and repaired datasets are aligned by instruction text
(\texttt{instruction.strip()}). The alignment script
(\texttt{aligne\_dataset.py}) builds a set of repaired-dataset instructions,
then iterates the repetition dataset, retaining only entries whose
instruction is found in the set. Matched instructions are removed from the
set after each match, retaining at most one repetitive example for each
surviving repaired instruction key. The resulting aligned repetition dataset
(\texttt{merged\_repetition\_dataset\_aligned.json}) contains 755\,920
records. For preference-data construction,
\texttt{build\_rl\_dataset.py} instead builds a dictionary mapping each
repaired instruction to its output. It scans the repetitive dataset,
constructs one preference pair for each matched key, and deletes the key
after matching. If duplicate exact instruction keys remain when the
dictionary is built, the last repaired output associated with that key is
retained.

\paragraph{Construction of SFT and preference-training examples.}
Four training configurations are derived from the aligned pairs:

\[
\begin{aligned}
\mathcal{D}_{\text{SFT-up}}    &= \{(x_i,\; y_i^{\mathrm{rep}})\},\\
\mathcal{D}_{\text{SFT-down}}  &= \{(x_i,\; y_i^{\mathrm{repair}})\},\\
\mathcal{D}_{\text{RL-up}}     &= \{(x_i,\; y_i^{\mathrm{rep}},\; y_i^{\mathrm{repair}})\},\\
\mathcal{D}_{\text{RL-down}}   &= \{(x_i,\; y_i^{\mathrm{repair}},\; y_i^{\mathrm{rep}})\}.
\end{aligned}
\]
Table~\ref{tab:repair-configs} summarizes the four post-training dataset
configurations.
Here $x_i$ is the instruction, $y_i^{\mathrm{rep}}$ is the original
repetitive response, and $y_i^{\mathrm{repair}}$ is the full repaired
response obtained by concatenating the retained prefix and the generated
continuation. For the RL conditions the tuple order is
$(x_i,y_i^{+},y_i^{-})$, where the second response is stored as
\texttt{chosen} and the third as \texttt{rejected}. In our implementation
these are DPO-style preference datasets rather than scalar reward-labeled
examples. SFT configurations use the format
\texttt{\{instruction, input, output\}}; RL configurations use
\texttt{\{instruction, input, chosen, rejected\}}.

\begin{table}[t]
\centering
\caption{Post-training dataset configurations. All four datasets contain
755\,920 records derived from the surviving instruction set. SFT-up and
SFT-down are stored as separate single-response datasets, whereas RL-up and
RL-down contain explicit chosen--rejected pairs.}
\label{tab:repair-configs}
\resizebox{\columnwidth}{!}{%
\begin{tabular}{llll}
\toprule
Condition & Target / chosen & Rejected & Direction \\
\midrule
SFT-up    & $y^{\text{rep}}$       & --- & Increase repetition \\
SFT-down  & $y^{\text{repair}}$    & --- & Decrease repetition \\
RL-up     & $y^{\text{rep}}$       & $y^{\text{repair}}$ & Prefer repetition \\
RL-down   & $y^{\text{repair}}$    & $y^{\text{rep}}$    & Prefer non-repetition \\
\bottomrule
\end{tabular}}
\end{table}

\paragraph{Dataset statistics.}
Table~\ref{tab:repair-counts} reports the number of examples at each
processing stage.

\begin{table}[t]
\centering
\caption{Stage-wise counts for repetition repair and post-training data
construction. Retention is reported only for sequential filtering stages;
dashes denote parallel dataset-construction outputs.}
\label{tab:repair-counts}
\resizebox{\columnwidth}{!}{%
\begin{tabular}{lrr}
\toprule
Stage & Records & Retention \\
\midrule
Initial merged pool                              & 743\,687 & --- \\
After appending model generations                & 766\,869 & --- \\
Saved repaired JSONL records                     & 766\,469 & 99.95\% \\
After length filtering and deduplication          & 755\,920 & 98.62\% \\
Aligned repetitive responses                     & 755\,920 & --- \\
RL-up preference records                         & 755\,920 & --- \\
RL-down preference records                       & 755\,920 & --- \\
\bottomrule
\end{tabular}}
\end{table}

The final aligned corpus retains 98.57\% of the 766\,869-record intervention
pool. The raw repaired JSONL contains 400 fewer records than the input
intervention pool; the retained artifacts do not provide per-example logs
that would allow this difference to be attributed to specific generation
failure modes. The archived cleaning script does not retain a category-wise
breakdown of the 10\,549 removed records.

%% file: sections__supplement__z_SparseClassifierTraining.tex
\section{Sparse Classifier Training and Coordinate-Selection Details}
\label{app:sparse-classifier}

\paragraph{Classifier input and label construction.}

Each classifier row corresponds to one successfully loaded QID--region pair
rather than to a unique QID alone.  Depending on the training mode, a QID
may contribute an answer-span pooled NOC feature vector, a non-answer-region
pooled NOC feature vector, or both.  The row count~$N$ therefore equals the
total number of successfully loaded feature files across the included
QID--region groups.

The feature-generation code extracts neuron-level NOC
activations from all \texttt{down\_proj} modules of a transformer, applies the
absolute value, multiplies by the column-wise weight norm (magnitude weighting),
divides by the output norm, aggregates the requested token region using the
configured pooling method, stacks all layers, and flattens the result
into a one-dimensional vector.  The feature dimension is
$D = L \times d_{\mathrm{ff}}$, where $L$ is the number of layers
and $d_{\mathrm{ff}}$ is the feed-forward intermediate dimension
(e.g., $D = 32 \times 14\,336 = 458\,752$ for Llama-3.1-8B).  No further feature selection, aggregation, or standardization is applied before the classifier sees the data.

\paragraph{Three-versus-one class composition.}
Let the positive class be the behavior-positive answer-span features
and the negative class be the union of the remaining three groups:
\[
\begin{aligned}
\mathcal{D}_{+}
&=
\mathcal{D}_{\mathrm{pos,span}},\\[2pt]
\mathcal{D}_{-}
&=
\mathcal{D}_{\mathrm{neg,span}}
\;\cup\;
\mathcal{D}_{\mathrm{neg,nonspan}}
\;\cup\;
\mathcal{D}_{\mathrm{pos,nonspan}}.
\end{aligned}
\]

If the two QID groups are balanced and every activation file is available,
the expected class ratio is approximately $N_{+}:N_{-}=1:3$ because the
negative class draws from three groups (answer-span for behavior-negative
QIDs, non-answer-region for both QID classes).  Missing activation files
may change the realized ratio.

\paragraph{Feature preprocessing.}
The extracted NOC features undergo three per-neuron transformations during the
forward pass:
\begin{enumerate}
    \item absolute value:
    \[
    a_{\ell,j} \gets |a_{\ell,j}|;
    \]

    \item magnitude weighting:
    \[
    a_{\ell,j}
    \gets
    a_{\ell,j}
    \left\|W_{:,j}^{(\ell)}\right\|_2;
    \]

    \item output normalization:
    \[
    a_{\ell,j}
    \gets
    \frac{a_{\ell,j}}
    {\left\|o_\ell\right\|_2+10^{-8}}.
    \]
\end{enumerate}
Between the feature-extraction stage and the logistic-regression fit there is
\textbf{no additional feature standardization or normalization}.  A systematic
search of the training scripts and feature-extraction code for
\texttt{StandardScaler}, \texttt{RobustScaler}, \texttt{Normalizer}, \texttt{zscore},
\texttt{.mean(axis=0)}, and \texttt{.std(axis=0)} confirmed that no sklearn
standardizer or manual z-score computation was applied.  Because L1
regularization is not scale-invariant, the coordinate selection described
below may be influenced by the relative scales of the per-neuron features.

\paragraph{Sparse logistic-regression configuration.}
All 366 successfully loaded \texttt{LogisticRegression} objects use
\texttt{penalty="l1"} and $C=1.0$.  Of these models, 362 use
\texttt{solver="saga"}. The penalty, regularization strength, and solver are supplied
through the classifier command-line arguments, whereas
\texttt{max\_iter=1000}, \texttt{tol=1e-3}, \texttt{random\_state=42}, and
\texttt{verbose=1} are fixed in the trainer implementations:
\begin{verbatim}
LogisticRegression(
    penalty=args.penalty,
    C=args.C,
    solver=args.solver,
    max_iter=1000,
    tol=1e-3,
    random_state=42,
    verbose=1,
)
\end{verbatim}

\paragraph{Positive-coordinate selection.}
All 366 loaded models have \texttt{classes\_ = [0, 1]}, where class~1 is the
behavior-positive span class.  The coefficient vector $\mathbf{w} =
\texttt{model.coef\_[0]}$ therefore encodes the contribution of each feature
to the positive-class log-odds.  The standard neuron-extraction pipeline selects coordinates as
\[
\mathcal{S}_{+} = \{j \mid w_j > 0\},
\]
i.e.\ features whose coefficient is strictly positive.  An alternative flag
(\texttt{--all\_non\_zero}) selects all coordinates with $w_j \neq 0$.  No top-\emph{k}
selection, additional threshold, or filtering by coefficient magnitude was
found in the audited coordinate-extraction scripts.  The intercept is never
part of the coordinate set.

Table~\ref{tab:sparse-coord-counts} reports representative coordinate
counts.  The repetition row corresponds to the RL repetition-up, epoch-1
checkpoint~94 classifier, whereas the sycophancy and hallucination rows
correspond to their base-model classifiers.

\begin{table}[ht]
\centering
\caption{Selected coordinate counts for representative archived
classifiers.  All models use $C=1.0$ and an $L_1$ penalty.}
\label{tab:sparse-coord-counts}
\scriptsize
\setlength{\tabcolsep}{2.5pt}
\begin{tabular}{@{}lcrrr@{}}
\toprule
Behavior & Solver & $n_{w>0}$ & $n_{w<0}$ & $n_{w\neq 0}$ \\
\midrule
Repetition               & saga      & 25  & 9   & 34  \\
Sycophancy (base)       & saga      & 92  & 57  & 149 \\
Hallucination (base)    & saga      & 20  & 32  & 52  \\
SAE-Repetition          & liblinear & 218 & 556 & 774 \\
\bottomrule
\end{tabular}
\end{table}

Three conceptual row-count levels must be distinguished:

\begin{enumerate}
    \item \textbf{QID counts} ($n_t$, $n_f$): the number of unique
    question--response identifiers in each judge category.
    \item \textbf{Candidate row counts}: the number of feature-matrix rows
    that would be obtained if every activation file existed.  In 3-vs-1 mode,
    \[
    N_{+}^{\mathrm{cand}} = n_f,\qquad
    N_{-}^{\mathrm{cand}} = n_t + n_t + n_f = 2 n_t + n_f.
    \]
    For the sycophancy ID file this gives
    $N_{+}^{\mathrm{cand}} = 3\,334$ and
    $N_{-}^{\mathrm{cand}} = 2 \times 3\,344 + 3\,334 = 10\,022$,
    corresponding to a candidate ratio of approximately $1:3$.
    \item \textbf{Realized row counts} ($N_{+}, N_{-}$): the rows that
    remain after missing-file skipping.  These counts cannot be
    reconstructed because the archived classifier objects do not
    store the training data, the logs do not report per-group
    surviving counts, and the corresponding activation directories
    are not included in the archived artifacts.
\end{enumerate}

%% file: sections__supplement__za_SampleAlignment_CCA_DirectedProbing.tex
\section{Cross-Model Sample Alignment, CCA, and Directed Probing}

\paragraph{Scope and notation.}
This section documents the cross-model comparison protocol used for the
canonical-correlation analysis (CCA) and directed linear-probing analyses. The analysis covers two behavioral domains---repetition
(R-Neurons) and sycophancy (S-Neurons)---in both the raw-activation
(ACT) and normalized-output-contribution (NOC) spaces.

Let \(A\) and \(B\) denote two models from the audited model
collection. For each model, an \(\ell_1\)-regularized logistic classifier identifies
behavior-associated coordinates using the three-versus-one construction
documented in Appendix~\ref{app:sparse-classifier}.  The cross-model analyses
then use the domain labels associated with the fixed QID lists to orient and
compare the representations formed from the corresponding selected
coordinates.

\paragraph{Cross-model sample intersection and row alignment.}
Within each behavioral domain, the formal sample set is defined by a
fixed QID list stored in a JSON file under the keys \texttt{f}
(behavior-positive, label \(y=1\)) and \texttt{t} (behavior-negative,
\(y=0\)).\footnote{Different QID sets reflect
the different data sources used for each behavioral domain; no single
sample set is shared across domains.}

Every model is required to have an activation file for every QID in the list.  If any file is
missing the pipeline raises \texttt{FileNotFoundError} and terminates.
In the fixed-list pipeline the common sample set is therefore the QID
list itself, not a per-model-pair intersection.

For the repetition random-neuron baseline, a different alignment rule
is used.  The pipeline intersects the QIDs whose activation files are
present for all 23 models, yielding 2735 common examples: 1367 from
the behavior-positive group (key \texttt{f}) and 1368 from the
behavior-negative group (key \texttt{t}).  This file-presence
intersection is constructed once and reused for all baseline model
pairs.

In the formal fixed-list pipeline, row order is identical across
models because every model reads the same domain-specific JSON file.
The order is: all behavior-positive QIDs first (label~1), followed by
all behavior-negative QIDs (label~0).  Within each class, QIDs retain
the order in which they are stored in the JSON array; the loading code
does not sort them.  In the random-neuron baseline, a single saved common-QID list is
reused for all model pairs, with the behavior-positive group preceding
the behavior-negative group, ensuring identical row ordering across
all baseline model pairs.

\paragraph{Feature preprocessing and dimensionality.}
For each model, the ACT or NOC matrix
\(\mathbf{X}^{(m)}\in\mathbb{R}^{n\times p}\)
(samples~\(\times\) behavior-associated coordinates) is mean-centered and
projected using PCA, with the implementation-defined maximum component count
\(\min(p,n-1)\). Retained dimensions are selected by a cumulative explained-variance
threshold~\(\tau=0.90\):

\[
k_m = \min\Bigl\{ \,k : \sum_{j=1}^{k}\lambda_j^{(m)} \ge
\tau\sum_{j}\lambda_j^{(m)} \Bigr\},
\]

where \(\lambda_j^{(m)}\) are the PCA eigenvalues.  Let
\(\boldsymbol{\mu}^{(m)}\) be the empirical mean and
\(\mathbf{V}^{(m)}\) the PCA rotation matrix; the retained projections
are

\[
\mathbf{Z}^{(m)}
=
\bigl(\mathbf{X}^{(m)}-\boldsymbol{\mu}^{(m)}\bigr)
\,\mathbf{V}_{:,1:k_m}^{(m)} .
\]

\paragraph{CCA fitting and numerical configuration.}
Canonical-correlation analysis is performed for each model pair
\((A,B)\) using scikit-learn's \texttt{CCA} with
\texttt{n\_components}\(=q\), where \(q=\min(k_A,k_B)\).
Except for \texttt{n\_components}, all CCA options---including
\texttt{tol}, \texttt{max\_iter}, and feature-scaling behavior---
follow the defaults of the installed scikit-learn version, which is not
pinned in the project dependencies.  \textbf{No explicit covariance
regularization is applied.}

\paragraph{Canonical rank.}
The number of canonical dimensions is

\[
q = \min(k_A, k_B),
\]

where \(k_A\) and \(k_B\) are the variance-threshold dimensions
defined above.  The code applies no explicit additional sample-size, effective-rank,
or numerical-rank check beyond the PCA dimensionality and the
\(q=\min(k_A,k_B)\) rule.

\paragraph{Sign orientation.}
Canonical variates are sign-oriented using the class labels so that on
every dimension the positive-class conditional mean is not less than
the negative-class conditional mean.  Let \(\mathbf{U},\mathbf{V}\) be
the canonical variates for models \(A\) and \(B\).  For each dimension
\(j\):

\[
s_j^{A}
=
\begin{cases}
+1, & \bar{u}^{+}_j > \bar{u}^{-}_j,\\[2pt]
-1, & \bar{u}^{+}_j \le \bar{u}^{-}_j,
\end{cases}
\qquad
\mathbf{U}_{:,j}\gets s_j^{A}\,\mathbf{U}_{:,j},
\]

with an analogous definition for \(s_j^{B}\) and
\(\mathbf{V}_{:,j}\), where
\(\bar{u}^{+}_j = \frac{1}{n^+}\sum_{i:y_i=1}U_{i,j}\) and
\(\bar{u}^{-}_j\) is the corresponding mean over examples with
\(y=0\).  This orientation is applied to the canonical variates after CCA fitting.  The audited implementation does not record how often the
two class means are exactly equal on any canonical dimension.

\paragraph{Canonical-correlation computation and aggregation.}
After sign orientation, canonical correlations are the signed Pearson
correlation coefficients between corresponding columns of
\(\mathbf{U}\) and \(\mathbf{V}\):

\[
\rho_j = \operatorname{corr}(\mathbf{U}_{:,j},\mathbf{V}_{:,j}),
\qquad j = 1,\dots,q.
\]

The reported \emph{primary} correlation is \(\rho_1\) (the first
canonical correlation).  The \emph{mean} CCA is the arithmetic average

\[
\operatorname{MeanCCA}(A,B) = \frac{1}{q}\sum_{j=1}^{q}\rho_j.
\]

All dimensions are equally weighted; there is no explained-variance
weighting.  In the formal implementation the correlations are stored
and reported as signed values (after sign orientation).  In the
random-neuron baseline, each canonical correlation is replaced by its
absolute value, \(\rho_j\gets|\rho_j|\), before reporting the primary
correlation and computing the mean correlation.

\paragraph{Handling of constant, failed, and non-finite dimensions.}
The implementation performs no per-dimension finite-value or
zero-variance check.  A non-finite Pearson correlation (e.g.\
\(\mathrm{NaN}\) returned when a standard deviation is zero) is
retained by the scoring routine and can propagate through the ordinary
arithmetic mean; the code does not use \texttt{nanmean}.  Exceptions
raised during a model-pair computation are caught by the outer
pair-level handler, which records a \(\mathrm{NaN}\) score for that
pair and continues to the next pair.  The audited materials do not
establish explicit handling or retrying of CCA convergence
warnings.

\paragraph{Directed probing from source model to target model.}
Directed (asymmetric) linear probing estimates how well the PCA
subspace of a source model~\(A\) can predict the PCA subspace of a
target model~\(B\):

\[
\widehat{\mathbf{Z}}^{(B)} =
\mathbf{Z}^{(A)}\,\mathbf{W}_{A\rightarrow B} + \mathbf{b},
\]

where \(\mathbf{W}\) and \(\mathbf{b}\) are fitted by ordinary least
squares (scikit-learn's \texttt{LinearRegression} with
\texttt{fit\_intercept=True}, no regularization).  The score for a
given target dimension~\(j\) is the signed Pearson correlation between
\(\widehat{\mathbf{Z}}^{(B)}_{:,j}\) and \(\mathbf{Z}^{(B)}_{:,j}\).

Two probing modes are reported:

\begin{description}
    \item[k-dim probing:]  All \(k_A\) source dimensions predict all
    \(k_B\) target dimensions jointly (multi-output regression).
    The reported score is the mean of per-dimension Pearson
    correlations: \(\frac{1}{k_B}\sum_{j=1}^{k_B}\rho_j\).

    \item[Top-two-PC probing:]  The method evaluates all available
    one-dimensional mappings among the first two retained PCs of the source and
    target models.  The exact mapping and score-selection rule are defined below.
\end{description}

\paragraph{Top-two-PC mapping selection.}
Let \(r_A=\min(2,k_A)\), \(r_B=\min(2,k_B)\), and let
\(\mathbf{z}^{A}_{i}\in\mathbb{R}^{n}\) be the \(i\)-th PC score vector
of model \(A\) (\(1\le i\le r_A\)), with
\(\mathbf{z}^{B}_{j}\) defined analogously for model \(B\).
The \(r_A r_B\) one-dimensional linear mappings are

\[
\widehat{\mathbf{z}}^{\,B}_{j\mid i}
=
f_{i\to j}\bigl(\mathbf{z}^{A}_{i}\bigr),
\qquad
1\le i\le r_A,\quad 1\le j\le r_B.
\]

The reported score is the maximum Pearson correlation among the candidate
axis pairings:

\[
S_{\mathrm{top2pc}}(A\to B)
=
\max_{\substack{1\le i\le r_A\\1\le j\le r_B}}
\operatorname{corr}
\bigl(
\widehat{\mathbf{z}}^{\,B}_{j\mid i},
\mathbf{z}^{B}_{j}
\bigr).
\]

This quantity summarizes the strongest linear association among the available
top-two-PC axis pairings.

\paragraph{Regression intercept and regularization.}
All probing regressions use scikit-learn's \texttt{LinearRegression}
with default settings:
\begin{itemize}
    \item \texttt{fit\_intercept=True} (the data are not explicitly
    centered before regression; the PCA projections are already
    mean-centered, so the intercept absorbs any residual offset);
    \item No \(\ell_1\) or \(\ell_2\) regularization (ordinary least
    squares).
\end{itemize}
No hyperparameter search is performed.

\paragraph{Directionality and interpretation of the reported scores.}
The probing matrix is indexed by (source~model, target~model).  Entry
\((i,j)\) represents \(\text{model}_i \to \text{model}_j\): the PCA
projections of model~\(i\) are used as predictors for the projections
of model~\(j\).  The mapping is asymmetric: \(A\to B\) and \(B\to A\)
are fitted and reported separately.  The row label of the heatmap
corresponds to the predictor (source), the column label to the target.

%% file: sections__supplement__zb_Post-TrainingConfigurations.tex
\section{Controlled Post-Training Configurations}
\label{sec:post-training-config}

\paragraph{Scope and reporting convention.}
This appendix documents the complete controlled post-training protocol for the two behavioral domains reported in the main paper: repetition and sycophancy.
All training runs use the LLaMA-Factory framework built on the Hugging Face Trainer and, for DPO, the TRL library.
We adopt the following conventions throughout.
Effective batch size is defined as the product of per-device training batch size, number of devices, and gradient accumulation steps.
For DPO runs, the reported example count refers to preference pairs.
Each multi-stage model (SFT~$\to$~DPO) is decomposed into its constituent stages with the initialization checkpoint of each stage explicitly stated.
The reported values were recovered by cross-checking the archived YAML configurations, shell launch scripts, saved training arguments, trainer states, logs, model cards, and downstream checkpoint paths whenever those sources were available. 

\paragraph{Repetition post-training configurations.}
Table~\ref{tab:repetition-training-config} reports the training configurations for the six repetition model variants used in the main paper.
All repetition SFT stages initialize from the base \texttt{meta-llama/Llama-3.1-8B} checkpoint.
The SFT-up model is trained on repetitive responses (\texttt{755\,844 examples derived from paragraph-level and token-level repetition data}) while the SFT-down model is trained on non-repetitive responses (\texttt{sft\_normal\_data}, 755\,844 examples rewritten by Llama-3.3-70B-Instruct).
Both SFT runs use full-parameter fine-tuning with AdamW ($\beta_1=0.9,\beta_2=0.999,\epsilon=1\times10^{-8}$), a learning rate of $1\times10^{-5}$, cosine scheduling with 10\% warmup ratio, fp16 precision, and DeepSpeed ZeRO-3 without offload.
Each SFT run defines a maximum step budget that overrides the configured \texttt{num\_train\_epochs=3}, so the effective epoch count is \texttt{max\_steps} $\times$ \texttt{effective\_batch\_size} $/$ \texttt{num\_examples}, which equals $\sim$0.017~epochs for SFT-up (100 steps) and $\sim$0.102~epochs for SFT-down (600 steps).

The DPO stages use a lower learning rate of $5\times10^{-7}$, per-device batch size of 1~(one preference pair), gradient accumulation of 16, and a maximum sequence length of 1\,024 tokens.
The DPO preference strength parameter is $\beta=0.1$.
The archived run names use ``RL'' for the DPO stage; throughout this appendix we refer to the training objective as DPO.
When DPO is applied directly to the base model (DPO-up and DPO-down from base), both the policy model and the reference model initialize from the base checkpoint; the reference model is a frozen copy of the policy at initialization (implicit behavior of the TRL \texttt{DPOTrainer}).
When DPO is applied after SFT (SFT$\to$DPO-up and SFT$\to$DPO-down), the policy model initializes from the final SFT checkpoint (\texttt{checkpoint-100} for the up direction and \texttt{checkpoint-600} for the down direction), and the reference model is initialized as a frozen copy of that same SFT checkpoint.
All DPO runs complete 60~optimization steps, corresponding to an effective epoch count of $\sim$0.010 (60 $\times$ 128 $/$ 755\,920).

\begin{table*}[t!]
\centering
\caption{Controlled post-training configuration for repetition behavioral interventions.}
\label{tab:repetition-training-config}
\small
\setlength{\tabcolsep}{4pt}
\begin{tabular}{lccc}
\toprule
Configuration & SFT (Up~/~Down) & DPO from Base & SFT $\to$ DPO \\
\midrule
Base checkpoint & \multicolumn{3}{c}{meta-llama/Llama-3.1-8B} \\
Initialization & Base & Base & SFT checkpoint (step~100~or~600) \\
Objective & SFT LM (induce~/~reduce) & DPO (induce~/~reduce) & DPO (induce~/~reduce) \\
Dataset (Up) & \texttt{sft\_rep\_data} & \texttt{rl\_induce\_rep} & \texttt{rl\_induce\_rep} \\
Dataset (Down) & \texttt{sft\_normal\_data} & \texttt{rl\_cure\_rep} & \texttt{rl\_cure\_rep} \\
Examples (Up) & 755\,844 & 755\,920 pairs & 755\,920 pairs \\
Examples (Down) & 755\,844 & 755\,920 pairs & 755\,920 pairs \\
Parameter update & \multicolumn{3}{c}{Full-parameter fine-tuning (no LoRA or PEFT adapter)} \\
Optimizer & AdamW (torch) & AdamW (torch) & AdamW (torch) \\
Learning rate & $1\times10^{-5}$ & $5\times10^{-7}$ & $5\times10^{-7}$ \\
Per-device batch size & 2 & 1 pair & 1 pair \\
Gradient accumulation & 8 & 16 & 16 \\
Effective batch size & 128 & 128 pairs & 128 pairs \\
Maximum steps & 100~(Up)~/~600~(Down) & 60 & 60 \\
Warmup ratio & 0.1 & 0.1 & 0.1 \\
Scheduler & Cosine & Cosine & Cosine \\
Max. sequence length & 2\,048 & 1\,024 & 1\,024 \\
Precision & fp16 & fp16 & fp16 \\
DPO $\beta$ & Not applicable & 0.1 & 0.1 \\
Reference model & Not applicable & Implicit (frozen base) & Implicit (frozen SFT) \\
DeepSpeed & ZeRO-3 & ZeRO-3 & ZeRO-3 \\
Checkpoint interval & 10~(Up)~/~60~(Down) steps & 6 steps & 6 steps \\
Random seed & 42 & 42 & 42 \\
Hardware & \multicolumn{3}{c}{8 GPUs; A100 40GB documented in archived launch-script comments} \\
\bottomrule
\end{tabular}
\end{table*}

\paragraph{Sycophancy post-training configurations.}
Table~\ref{tab:sycophancy-training-config} reports the training configurations for the six sycophancy model variants.
The experimental design mirrors the repetition protocol but uses substantially smaller datasets (3\,480~SFT examples and 3\,480~DPO preference pairs, sourced from the Anti-Sycophancy-DPO dataset).
The shared settings include full-parameter fine-tuning, the AdamW optimizer, cosine scheduling, a 0.1 warmup ratio, fp16 precision, and DeepSpeed ZeRO-3.
The SFT-up model is trained on sycophantic responses (\texttt{syco\_sft}) and the SFT-down model on truthful responses (\texttt{truth\_sft}).
SFT-up runs for 120~steps ($\sim$4.4~effective epochs) and SFT-down for 360~steps ($\sim$13.2~effective epochs).
The DPO stages follow the same two initialization schemes as in the repetition experiments: DPO initialized from the base model (DPO-up and DPO-down from base) and DPO initialized from the SFT checkpoint (SFT$\to$DPO-up and SFT$\to$DPO-down).
All sycophancy DPO runs use $\beta=0.1$ except SFT$\to$DPO-down (suppress sycophancy), which uses $\beta=0.7$ as recorded in the training configuration.
The smaller dataset size means that 60~DPO steps at an effective batch size of 128~pairs correspond to $\sim$2.2~effective epochs.

\begin{table*}[t!]
\centering
\caption{Controlled post-training configuration for sycophancy behavioral interventions.}
\label{tab:sycophancy-training-config}
\small
\setlength{\tabcolsep}{4pt}
\begin{tabular}{lccc}
\toprule
Configuration & SFT (Up~/~Down) & DPO from Base & SFT $\to$ DPO \\
\midrule
Base checkpoint & \multicolumn{3}{c}{meta-llama/Llama-3.1-8B} \\
Initialization & Base & Base & SFT checkpoint (step~120~or~360) \\
Objective & SFT LM (induce~/~reduce) & DPO (induce~/~reduce) & DPO (induce~/~reduce) \\
Dataset (Up) & \texttt{syco\_sft} & \texttt{rl\_induce\_syco} & \texttt{rl\_induce\_syco} \\
Dataset (Down) & \texttt{truth\_sft} & \texttt{rl\_cure\_syco} & \texttt{rl\_cure\_syco} \\
Examples (Up) & 3\,480 & 3\,480 pairs & 3\,480 pairs \\
Examples (Down) & 3\,480 & 3\,480 pairs & 3\,480 pairs \\
Parameter update & \multicolumn{3}{c}{Full-parameter fine-tuning (no LoRA or PEFT adapter)} \\
Optimizer & AdamW (torch) & AdamW (torch) & AdamW (torch) \\
Learning rate & $1\times10^{-5}$ & $5\times10^{-7}$ & $5\times10^{-7}$ \\
Per-device batch size & 2 & 1 pair & 1 pair \\
Gradient accumulation & 8 & 16 & 16 \\
Effective batch size & 128 & 128 pairs & 128 pairs \\
Maximum steps & 120~(Up)~/~360~(Down) & 60 & 60 \\
Warmup ratio & 0.1 & 0.1 & 0.1 \\
Scheduler & Cosine & Cosine & Cosine \\
Max. sequence length & 2\,048 & 1\,024 & 1\,024 \\
Precision & fp16 & fp16 & fp16 \\
DPO $\beta$ & Not applicable & 0.1 & 0.1~(Up)~/~0.7~(Down) \\
Reference model & Not applicable & Implicit (frozen base) & Implicit (frozen SFT) \\
DeepSpeed & ZeRO-3 & ZeRO-3 & ZeRO-3 \\
Checkpoint interval & 12~(Up)~/~36~(Down) steps & 6 steps & 6 steps \\
Random seed & 42 & 42 & 42 \\
Hardware & \multicolumn{3}{c}{8 GPUs; A100 40GB documented in archived launch-script comments} \\
\bottomrule
\end{tabular}
\end{table*}

\paragraph{Dataset and batch-size accounting.}
The constructed repetition SFT corpora contain 755\,920 records, while the
archived SFT trainer reports 755\,844 examples used for optimization.
The repetition DPO dataset comprises 755\,920 preference pairs, constructed
by one-to-one matching of repetitive and non-repetitive responses on shared
instruction keys.
The repetition DPO dataset comprises 755\,920~preference pairs, constructed by one-to-one matching of repetitive and non-repetitive responses on shared instruction keys.
The sycophancy SFT dataset contains 3\,480~examples and the DPO dataset contains 3\,480~pairs, both derived from the \texttt{DataCreatorAI/Anti-Sycophancy-DPO} source.
All DPO runs use an effective batch size of 128~pairs per optimizer update, calculated as $1~\text{(per-device)} \times 8~\text{(devices)} \times 16~\text{(accumulation)} = 128$.
For SFT, the accumulated batch per process is $2\times8=16$ examples, and for DPO it is $1\times16=16$ preference pairs.  With eight distributed processes, the global effective batch size is therefore 128 examples or preference pairs per optimizer update.  We report this global value, consistent with the Hugging~Face Trainer \texttt{total\_train\_batch\_size} convention.

\paragraph{SFT-to-DPO initialization and reference models.}
No explicit \texttt{ref\_model} path is provided in the archived DPO configurations.  Under the audited LLaMA-Factory/TRL execution path, the reference model is instantiated as a frozen copy of the policy at DPO initialization.
In the SFT$\to$DPO sequential protocol, the DPO policy model is initialized from the final SFT checkpoint, and the reference model is initialized as a frozen copy of that same checkpoint.
In the direct DPO protocol, both the policy model and the reference model are initialized from the base Llama-3.1-8B checkpoint.
None of the archived DPO configurations enables reference-free mode.

\paragraph{Distributed training and numerical precision.}
The archived launch configurations specify one node with eight processes, launched through \texttt{torchrun} with \texttt{FORCE\_TORCHRUN=1} and \texttt{NPROC\_PER\_NODE=8}.  The accompanying shell-script records identify the devices as NVIDIA A100 40GB GPUs, although no historical system-level hardware log was available for independent verification.
DeepSpeed ZeRO-3 (stage~3 with \texttt{stage3\_gather\_16bit\_weights\_on\_model\_save=true}) is employed throughout, with the optimizer, gradient clipping, and batch sizes delegated to the Hugging Face Trainer via \texttt{"auto"} fields in the DeepSpeed JSON.
The optimizer is \texttt{AdamW} (\texttt{adamw\_torch}) with betas (0.9,~0.999), $\epsilon=1\times10^{-8}$, and weight decay of~0.0.
All runs train in fp16 mixed precision with bf16 disabled.  TF32 is not explicitly configured in the archived YAML files or saved training arguments.

\paragraph{Checkpoint selection and downstream use.}
Checkpoints are saved every $N$~steps where $N=10$ (SFT-up rep), $N=60$ (SFT-down rep), $N=12$ (SFT-up syco), $N=36$ (SFT-down syco), and $N=6$ (all DPO runs), with a retention limit of 10~checkpoints per run.
The primary endpoint results reported in the main paper use the final
checkpoint of each training run (the checkpoint with the largest
\texttt{global\_step}).  The stage-resolved geometry--occupancy--gain analyses
additionally use the saved intermediate checkpoints, as documented in
Appendix~\ref{app:gog-full}.
SFT repetition-up uses \texttt{checkpoint-100}; SFT repetition-down uses \texttt{checkpoint-600}; all repetition DPO and SFT$\to$DPO runs use \texttt{checkpoint-60};
SFT sycophancy-up uses \texttt{checkpoint-120}; SFT sycophancy-down uses \texttt{checkpoint-360}; all sycophancy DPO and SFT$\to$DPO runs use \texttt{checkpoint-60}.

\begin{figure}[t]
    \centering
    \includegraphics[width=0.8\linewidth]{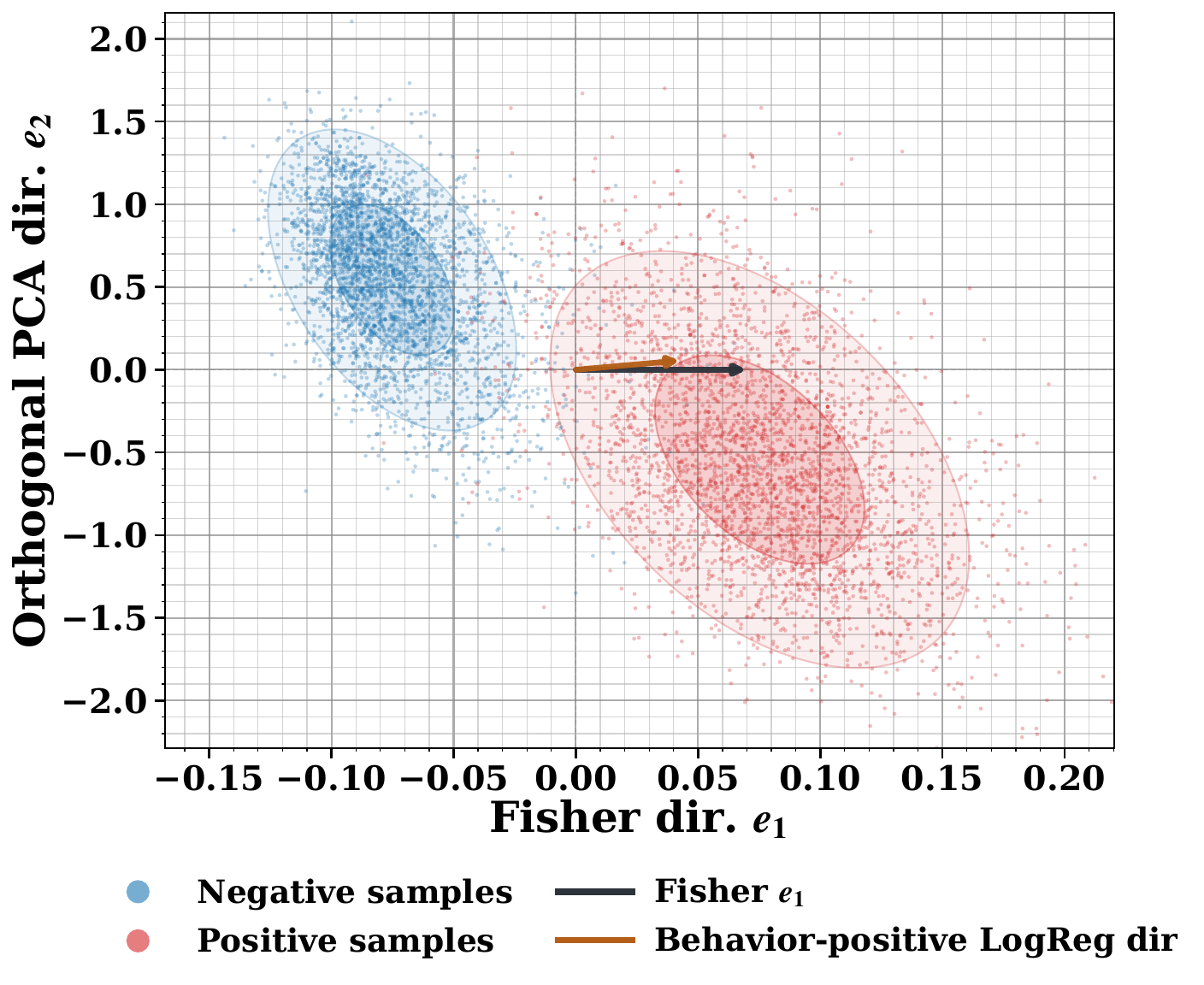}
    \caption{
    Visualization of a local behavioral chart. The Fisher and logistic directions align with the dominant
    separation axis, supporting the interpretation that
    $U_{\theta,b}$ captures behavior-relevant geometry.
    }
    \label{fig:behavioral_chart_visualization}
\end{figure}

\subsection{Joint interpretation of ACT and NOC.}
\label{Jointinterpretation}
Although ACT and NOC are computed from the same behavior-associated coordinate scaffold, they answer different questions and need not vary together. ACT characterizes how strongly or frequently the model occupies the corresponding runtime state, whereas NOC characterizes the strength of the functionally supported pathway associated with that state.  Figure~\ref{fig:act_noc_regimes} summarizes the four regimes induced by their combination. High NOC with high ACT indicates a strong mechanism that is frequently used at runtime. High NOC with low ACT indicates a strong but latent or gated pathway that the mechanism remains available, but the model's ordinary operating distribution rarely enters it, as may occur when post-training shifts the operating point away from an undesirable mode. Low NOC with high ACT indicates that the observed runtime states are correlated with the behavior but receive limited functional support, and therefore may not constitute a core computational pathway. Low NOC with low ACT indicates that the model both weakly supports and rarely occupies the corresponding behavioral mode. This distinction motivates treating functional support and runtime occupancy as separate axes throughout the subsequent analysis. Having separated functional-channel strength from runtime-state occupancy at the conceptual level, we next identify a sparse coordinate scaffold in which
these behavioral geometries can be estimated reliably.

\subsection{Token Windows and State Extraction} 
\label{TokenWindow}
Behavioral evidence is usually localized to a subset of positions. We therefore define a task-specified extraction operator \(\mathcal{T}_b:\mathcal{X}_b\rightarrow 2^{\mathbb{N}}\),
which maps each instance $x$ to a set of behavior-relevant token positions.  If $T_x$ is the number of processed tokens, we write
\(M_x=\mathcal{T}_b(x)\subseteq\{1,\ldots,T_x\}\).
The operator $\mathcal{T}_b$ is fixed before fitting any representation model or probe.  It may select a full response span, a decision-critical span, an event-centered window, or any other pre-specified region where the behavior is expected to be expressed.  
Let $a_{\ell,t,j}(x)$ denote the activation of coordinate $j$ at layer $\ell$ and token position $t$ under the chosen hook convention.  We aggregate activations over the selected token window:
\(    \bar a_{\ell,j}(x)    
=    
\frac{1}{|M_x|}    
\sum_{t\in M_x}    
a_{\ell,t,j}(x)\). The same extraction and aggregation rule is applied to behavior-positive and behavior-negative instances. This prevents trivial separability caused by inconsistent token positions, window lengths, or prompt formats. 
When layer-resolved analysis is required, we retain the layer index and define \(\phi_{\theta,b}^{(\Omega,\ell)}(x)    
\in    
\mathbb{R}^{D_{\theta,b,\Omega,\ell}}\).
For the main global representation, we concatenate the selected coordinates across layers \(    \phi_{\theta,b}^{(\Omega)}(x)    
=    
\bigoplus_{\ell=1}^{L}    
\phi_{\theta,b}^{(\Omega,\ell)}(x)\).
Thus each instance is mapped to a fixed-dimensional representation vector while preserving a clear separation between the token-selection rule, the layer-wise extraction rule, and the downstream geometric analysis.

%% file: sections__supplement__zc_related_work.tex
\section{Additional Related Work}
\label{RelatedWork}

\paragraph{Repetition and neural degeneration.}
Repetition has long been recognized as a central failure mode of neural text generation.  Holtzman et al. showed that likelihood-trained models can generate bland and repetitive continuations under maximization-based decoding and introduced nucleus sampling as a remedy \cite{holtzman2020curious}.  Welleck et al. proposed unlikelihood training to penalize repeated tokens and sequences \cite{welleck2020neural}.  Fu et al. analyzed repetition theoretically through Markov-style generation dynamics \cite{fu2021theoretical}.  Contrastive decoding methods further reduce degeneration by combining likelihood with representation-level discriminability \cite{su2022contrastive,su2022contrastiveframework}. 

\paragraph{Sycophancy and preference failures.}
Sycophancy is an alignment failure in which models match the user's beliefs or preferences instead of giving independent or truthful answers.  Sharma et al. showed that human preference data can favor sycophantic responses and that assistants often agree with user beliefs even when they are false \cite{sharma2024sycophancy}. Models can mimic human falsehoods \cite{lin2022truthfulqa}, and feedback optimization can improve behavior while inheriting biases of demonstrations, comparisons, and reward models \cite{stiennon2020summarize,ouyang2022training}.  DPO simplifies preference optimization without an explicit RL loop \cite{rafailov2023direct}, but it still optimizes a preference-induced objective.